\documentclass{article}

\usepackage{amsmath,amsfonts,bm}

\def\eqref#1{equation~\ref{#1}}

\def\1{\bm{1}}

\def\vgamma{{\bm{\gamma}}}

\DeclareMathAlphabet{\mathsfit}{\encodingdefault}{\sfdefault}{m}{sl}
\SetMathAlphabet{\mathsfit}{bold}{\encodingdefault}{\sfdefault}{bx}{n}

\newcommand{\E}{\mathbb{E}}

\DeclareMathOperator*{\argmax}{arg\,max}
\DeclareMathOperator*{\argmin}{arg\,min}

\newcommand{\Xr}{{X_r}}
\newcommand{\Xf}{{X_f}}
\newcommand{\xrhat}{{\hat{X}_r}}
\newcommand{\xfhat}{{\hat{X}_f}}

\newcommand{\Pxr}{\mathcal{P}_{\Xr}}
\newcommand{\Pxf}{\mathcal{P}_{\Xf}}
\newcommand{\Pxrhat}{\mathcal{P}_{\hat{X}_r}}
\newcommand{\Pxfhat}{\mathcal{P}_{\hat{X}_f}}
\newcommand{\encoder}{E_\theta}
\newcommand{\decoder}{{D_\phi}}
\newcommand{\encoderp}{E_{\theta_{0}}}
\newcommand{\decoderp}{D_{\phi_{0}}}
\newcommand{\modelinput}{x}
\newcommand{\fullmodel}{\decoder\circ\encoder}
\newcommand{\fullmodelp}{\decoderp\circ\encoderp}
\newcommand{\hmodel}{h_{\theta,\phi}}
\newcommand{\hmodelp}{h_{\theta_{0},\phi_{0}}}
\newcommand{\infonce}{\mathcal{I}^{NCE}}
\newcommand{\trans}{\mathcal{T}}

\newtheorem{thm}{Theorem}
\newtheorem{lem}{Lemma}

\newtheorem{prop}{Proposition}

\newcommand\eref[1]{(\ref{#1})}

\usepackage{algorithm,algpseudocode}

\algnewcommand{\Inputs}[1]{%
  \State \textbf{Inputs:}
  \Statex \hspace*{\algorithmicindent}\parbox[t]{.8\linewidth}{\raggedright #1}
}
\algnewcommand{\Outputs}[1]{%
  \State \textbf{Outputs:}
  \Statex \hspace*{\algorithmicindent}\parbox[t]{.8\linewidth}{\raggedright #1}
}
\algnewcommand{\Initialize}[1]{%
  \State \textbf{Initialize:}
  \Statex \hspace*{\algorithmicindent}\parbox[t]{.8\linewidth}{\raggedright #1}
}

\algnewcommand{\Compute}[1]{%
  \State {Compute:}
  \Statex \hspace*{\algorithmicindent}\parbox[t]{.8\linewidth}{\raggedright #1}
}
\newcommand{\ie}{i.e.}
\newcommand{\eg}{e.g.}

\usepackage[colorlinks=true]{hyperref}
\hypersetup{%
,urlcolor=blue
,citecolor=blue
,linkcolor=red
}
\usepackage{iclr2024_conference,times}
\usepackage{multirow}
\usepackage{caption}
\usepackage{subcaption}
\usepackage{hyperref}
\usepackage{url}

\usepackage{xcolor}
\usepackage{amssymb}
\usepackage{bbm}
\usepackage{graphicx}
\usepackage{amsmath}
\usepackage{enumitem}
\usepackage{booktabs}
\usepackage{adjustbox}

\usepackage{times}

\title{Machine Unlearning for Image-to-Image\\ 
Generative Models}

\author{\textbf{Guihong Li$^{1}$}\thanks{Work done during internship at JPMorgan Chase Bank, N.A.}, \textbf{Hsiang Hsu$^{2}$, Chun-Fu (Richard) Chen$^{2}$},  \textbf{Radu Marculescu$^{1}$}\\
$^{1}$The University of Texas at Austin,  USA\\
$^{2}$Global Technology Applied Research, JPMorgan Chase, USA  \\
\texttt{\{lgh,radum\}@utexas.edu}\\
\texttt{\{hsiang.hsu,richard.cf.chen\}@jpmchase.com}
}

\iclrfinalcopy
\begin{document}

\maketitle
\begin{abstract}
Machine unlearning has emerged as a new paradigm to deliberately forget data samples from a given model in order to adhere to stringent regulations.
However, existing machine unlearning methods have been primarily focused on classification models, leaving the landscape of unlearning for generative models relatively unexplored.
This paper serves as a bridge, addressing the gap by providing a unifying framework of machine unlearning for image-to-image generative models.
Within this framework, we propose a computationally-efficient algorithm, underpinned by rigorous theoretical analysis, that demonstrates negligible performance degradation on the retain samples, while effectively removing the information from the forget samples. 
Empirical studies on two large-scale datasets, ImageNet-1K and Places-365, further show that our algorithm does not rely on the availability of the retain samples, which further complies with data retention policy.
To our best knowledge, this work is the first that represents systemic, theoretical, empirical explorations of machine unlearning specifically tailored for image-to-image generative models.
Our code is available at \href{https://github.com/jpmorganchase/l2l-generator-unlearning}{https://github.com/jpmorganchase/l2l-generator-unlearning}.

\end{abstract}

\section{Introduction}
The prevalence of machine learning research and applications has sparked awareness among users, entrepreneurs, and governments, leading to new legislation\footnote{Enactions include the General Data Protection Regulation (GDPR) by  the European Union \citep{eu_gdpr}, the White House AI Bill \cite{datbill}, and others \citep{japan_bill,cananda_bill}.} to protect data ownership, privacy, and copyrights~\citep{laion_5b,intelligence_gpt4,data_leak,risk_of_found}.
At the forefront of these legislative efforts is the ``Right to be Forgotten'', a fundamental human right that empowers individuals to request the removal of their information from online services.
However, directly erasing data from databases is not enough, as it may already be ingrained in machine learning models, notably deep neural networks (DNNs), which can memorize training data effectively \citep{iclr_gan_memory,data_privacy_generate,carlini2023extracting}.
Yet another straightforward solution is to re-train DNNs from scratch on a new training dataset without the unwanted data---a resource-expensive procedure \citep{diffbeat_gan} that can not reflect users' requests in a timely manner.

\begin{figure}[t!]
     \begin{subfigure}[b]{0.48\textwidth}
     \centering
     \includegraphics[width=\textwidth]{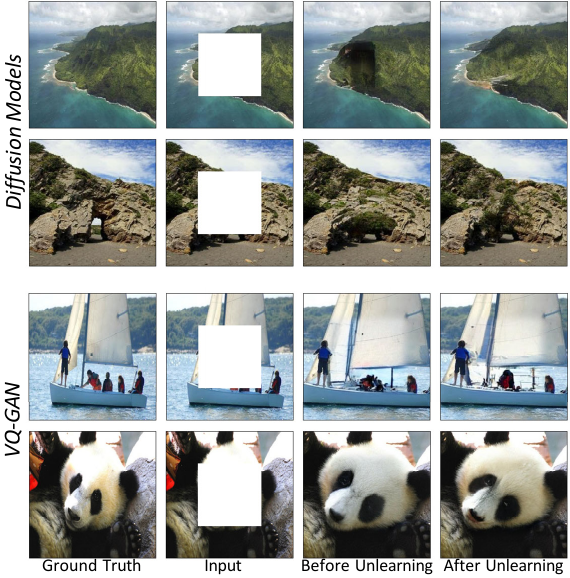}\vspace{-1mm}
     \caption{Retain Set}
     \label{fig:overview_retain}
     \end{subfigure}
      \hfill
     \begin{subfigure}[b]{0.48\textwidth}
     \centering
     \includegraphics[width=\textwidth]{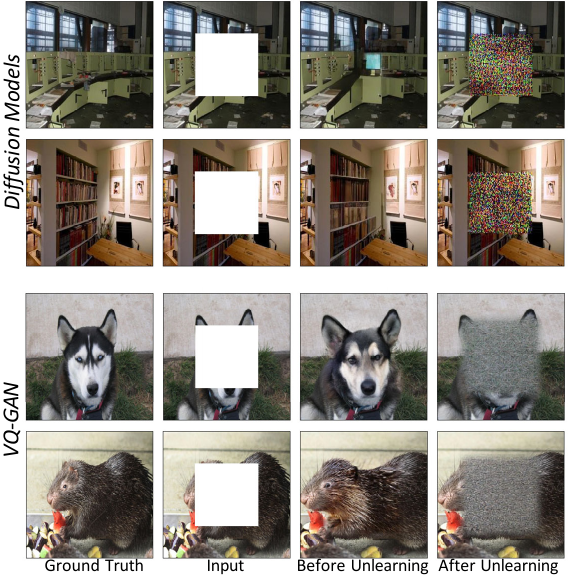}\vspace{-1mm}
     \caption{Forget Set}
     \label{fig:overview_forget}
     \end{subfigure}
     \vspace{-3mm}\caption{ 
     Our machine unlearning framework is applicable to various types of I2I generative models, including the diffusion models~\citep{palette_diff_base}, VQ-GAN~\citep{vqgan_mage} and MAE \citep{mae_masked_he} (cf.~Section~\ref{sec:exp}). 
     The images in the retain set remain almost (up to a slight difference due to the perplexity of generative models) unaffected before and after unlearning. 
     Conversely, the images in the forget set are nearly noise after unlearning, as designed.}\label{fig:first_demo}
\end{figure}

In response to various legal requirements and user requests, a novel approach known as \emph{machine unlearning} has been proposed \citep{unlearn_survey}. 
This technique allows a model, which has been trained with potentially sensitive samples referred to as ``forget samples'', to selectively remove these samples without the necessity of retraining the  model from scratch. 
Meanwhile, machine unlearning aims to minimize any adverse effects on the performance of the remaining data, termed ``retain samples''. 
Recent unlearning algorithms have been developed, some incorporating specialized training procedures to facilitate the unlearning process \citep{sisa_unlearn}, while others adjust model weights through fine-tuning \citep{tarun2023fast}. 
However, these approaches primarily address unlearning in classification problems.
On the other hand, generative models, which have demonstrated superior data memorization capabilities compared to classification models \citep{tirumala2022memorization,diff_remember}, excel at regenerating training samples \citep{data_privacy_generate,carlini2023extracting}. Therefore, the challenge of ensuring effective unlearning for generative models has become increasingly important and pressing.

In this paper, our emphasis lies on a particular category of generative model architectures known as Image-to-Image (I2I) generative models~\citep{diff_survey}. 
This selection offers a twofold advantage:
First, it paves the way for a groundbreaking approach to quantify the \emph{efficacy of machine unlearning for generative models}---a research direction hitherto uncharted in existing literature. 
Informally speaking, we define a generative model as having ``truly unlearned'' an image when it is unable to faithfully reconstruct the original image when provided with only partial information (see Figure~\ref{fig:first_demo} for an illustrative example where the partial information involves center cropping\footnote{For the precise definition, see Section~\ref{sec:method}.}).
Second, I2I generative models encompass all major branches in the field of vision generative models, including Masked Autoencoder (MAE) \citep{mae_masked_he}, Vector Quantized Generative Adversarial Networks (VQ-GAN) \citep{vqgan_mage}, and the more recent diffusion probabilistic models \citep{diffusion_original}. 
Based on this novel definition to quantify unlearning, our contributions can be summarized as follows:

\begin{itemize}
    \item We formulate a machine unlearning framework for I2I generative models that is applicable to MAE, VQ-GAN and diffusion models. 
    This formulation, in essence, is an unbounded optimization problem.
    We provide theoretical derivations that guarantee the unique optimality of its bounded counterpart, and design an algorithm for the efficient computation.
    \item We conduct extensive evaluations of our algorithm on various I2I generative models, including MAE, VQ-GAN and diffusion models.
    Empirical results on two large scale datasets, ImageNet-1K \citep{deng2009imagenet} and Places-365 \citep{zhou2017places365}, show that our framework exhibits negligible performance degradation on retain sets, meanwhile effectively eliminating the information in forget sets. 
    \item We further illustrate that the \emph{availability of the exact retain set} is not necessary in our framework---the efficacy of our algorithm remains intact even without any samples from the exact retain set.
\end{itemize}

To the best of our knowledge, this work is the first that systemically, theoretically, empirically explore the machine unlearning specifically targeting for I2I generative models. 
Proofs, details on experimental setups and training, and additional results are included in the Appendix. 

\section{Related Work}\label{sec:related}
\paragraph{I2I generative models.}
Many computer vision tasks can be formulated as I2I generation processes, such as image super-resolution~\citep{super_reso_gan}, style transfer~\citep{cyclegan}, image extension~\citep{image_extend_maskgit} and inpainting~\citep{teterwak2019boundless}.

Different type of I2I generative models utilize diverse training and optimization strategies to minimize the discrepancy between their generated images and the ground truth images.
The broadly used Generative Adversarial Networks (GANs) are trained by reducing a discriminator's accuracy in determining whether a generated image is real or synthetic~\citep{gan_goodfellow,karras2019style,chen2016infogan,karras2020analyzing}. However, stabilizing the training of GANs is a well-known challenge~\citep{wasserstein_gan,gulrajani2017improved,large_scale_gan}. 
In contrast, diffusion models address the stability issue by utilizing a multi-step generation strategy and are optimized by minimizing the Kullback–Leibler (\textit{KL}) divergence between the distributions of the generated and ground truth images~\citep{diffusion_original,song2020improved,hoogeboom2021autoregressive,salimans2021progressive}. Diffusion models can generate higher-fidelity images than GANs but require much longer generation time~\citep{t2i_diff_beat_gan,latent_diffusion,zhang2023adding}. 
Recently, Masked Autoencoder (MAE) has been proposed as a multi-purpose model for both image generation and classification~\citep{mae_masked_he,feichtenhofer2022masked,tong2022videomae}. Typically, MAE is optimized by minimizing the MSE loss between the generated and ground truth images. In this paper, our goal is to design a universal approach that is capable of conducting unlearning across diverse I2I model types with different optimization techniques.

\paragraph{Machine unlearning.}

Machine unlearning allows a trained model to selectively remove some unwanted samples (``forget set'') while minimizing any adverse effects on the performance of the remaining data (“retain set”) and without retraining the  model from scratch~\citep{mul_survey2023}. As the pioneering work on unlearning, SISA facilitates the unlearning of specific samples by retraining the model checkpoints that were initially trained with these "forget" samples~\citep{sisa_unlearn}. However, SISA needs to re-train all these models from scratch, if the forget samples are distributed across all shards. 
To address this problem, several methods manipulate the trained model weights directly. Some works compute the Neural Tangent Kernel (NTK) to modify model weights, but the computation of the Hessian matrix in NTK’s calculation is numerically unstable and not scalable for models with many parameters~\citep{golatkar2020eternal,golatkar2020forgetting}.
\cite{graves2021amnesiac} requires the storage of the gradient for each parameter of every training step when training the original models. This approach is not scalable given the extremely large training set and the enormous model size for the latest image generative models. 
Other methods improve the efficiency by maximizing loss on the forget set or re-assigning incorrect labels but typically they are only applicable to classification tasks.~\citep{neel2021descent,tarun2023deep,grad_up,towards_unbounded,cvpr_boundary_unlearn}. There are also some approaches focusing on other perspectives of unlearning instead of designing new unlearning algorithms. For example, ~\cite{zero_shot_unlearn} focuses on the data access issues of \textit{existing} unlearning algorithms and suggests using the images generated by the original model as the alternative for the original training set. Besides, ~\cite{prunine_unlearn} shows that that pruning the original model before unlearning can improve the overall performance of many \textit{existing}  unlearning algorithms. 
Previous unlearning approaches primarily focus on classification tasks, but there are emerging efforts on generative models. For instance, several methods maximize training loss on the forget set, but  are validated only on tiny datasets, like MNIST~\citep{mul_generative1,mul_generative4}. Other works focus on unlearning specific features (\eg, eye color, hairstyle) from generated images, but are only verified under small-scale setups and lack comprehensive analysis~\citep{mul_generative3,mul_generative2}. Besides, these methods typically manipulate the entire model, thus requiring extensive computation capacity due to the growing complexity and size of generative models. Moreover, none of them addresses I2I generative tasks. This motivates us to explore the efficient unlearning algorithms for I2I generative models in large-scale setups.

\section{Problem Formulation and Proposed Approach}\label{sec:method}
In this work, we primarily address the machine unlearning for I2I generative models that reconstruct images from incomplete or partial inputs.
Typically, I2I  generative models adopt an encoder-decoder network architecture, comprising two components, namely, 
(i) an encoder network $\encoder$ that encodes an input into a representation vector and 
(ii) a decoder network $\decoder$ that decodes the representation vector into the image. 
Specifically, given an input $\modelinput$, the output for an I2I generative model $\hmodel$ is as follows:
\begin{equation}
 \hmodel=\fullmodel,\quad \hmodel\left(\trans(x)\right) = \decoder\left(\encoder\left(\trans(x)\right)\right)
\end{equation}
where $x$ is a ground truth image; $\trans\left(\cdot\right)$ is the operation to remove some information from $x$, \eg, center cropping and random masking; $\circ$ is the composition operator; $\theta$ and $\phi$ are the parameters for the encoder and decoder, respectively.

\subsection{Definition of Unlearning on I2I Generative Models}\label{sec:definition_unlearn}

For machine unlearning on I2I generative models, given a trained  model (\ie, original model) $\hmodelp=\fullmodelp$ with parameters $\theta_0$ and $\phi_0$, the unlearning algorithm $A_F$ aims to obtain a target model: $$\hmodel\triangleq A_F\left(\hmodelp\right)$$ that satisfies the following properties:
\begin{itemize}[leftmargin=0.03\textwidth]
\item On the retain set $\mathcal{D}_{R}$, $\hmodel$ generates images that have the same distribution as the original model;
\item On the forget set $\mathcal{D}_{F}$, $\hmodel$ generates images that have far different distribution from the original model.
\end{itemize}
By using the KL-divergence ($D$), from a probability distribution perspective, these objectives are as follows:
\begin{equation}
 \argmin_{\theta, \phi}D\left(P_{\hmodelp\left(\trans\left(X_r\right)\right)}||P_{\hmodel\left(\trans\left(X_r\right)\right)}\right) \text{, and }\argmax_{\theta, \phi} D\left(P_{\hmodelp\left(\trans\left(X_f\right)\right)}||P_{\hmodel\left(\trans\left(X_f\right)\right)}\right)
\end{equation}
where, $X_r$ and $X_f$ are random variables that account for the ground truth images of the retain and forget sets, respectively. 

By combining these two objectives, we formulate our optimization goal as follows:
\begin{equation}\label{eq:raw_unlearn_target}
 \argmin_{\theta, \phi}\bigg\{ D\left(P_{\hmodelp\left(\trans\left(X_r\right)\right)}||P_{\hmodel\left(\trans\left(X_r\right)\right)}\right)-\alpha D\left(P_{\hmodelp\left(\trans\left(X_f\right)\right)}||P_{\hmodel\left(\trans\left(X_f\right)\right)}\right)\bigg\} 
\end{equation}

where $\alpha$ is a positive coefficient to control the trade-off between the retain and forget sets. 
Multiple previous works assume a trained I2I generative model can do an almost perfect generation on both of the retain and forget sets~\citep{inverse_diff_1,inverse_diff_2,inverse_gan_1,inverse_vae_1}; that is, $\hmodelp\left(\trans\left(X\right)\right)\approx X$. Therefore, Eq.~\eref{eq:raw_unlearn_target} can be rewritten as:
\begin{equation}\label{eq:raw_optimzation}
 \argmin_{\theta, \phi}\big\{ D\left(\Pxr||\Pxrhat\right)-\alpha D\left(\Pxf||\Pxfhat\right)\big\},\ \hat{X}_r=\hmodel\left(\trans\left(X_r\right)\right),\ \hat{X}_f=\hmodel\left(\trans\left(X_f\right)\right)
\end{equation}

where $\Pxr$ and $\Pxrhat$ represent the distribution of ground truth images and generated images in the retain set; $\Pxf$ and $\Pxfhat$ represent the distribution of ground truth images and generated images in the forget set.

\subsection{Optimization on Retain and Forget sets}\label{sec:optimzation_sets}
Clearly, for the first term in Eq.~\eref{eq:raw_optimzation}, a perfect unlearned model has no performance degradation on the retains set. In other words, the generated images share the distribution as ground truth images, i.e., $\Pxrhat=\Pxr$.
This way, the value of $D\left(\Pxr||\Pxrhat\right)$ is 0. Next, we discuss the optimization for the forget set.

To minimize the value for the objective functions in Eq.~\eref{eq:raw_optimzation}, we need to maximize KL divergence between $\Pxf$ and $\Pxfhat$.
However, there are infinitely many probability distributions that have infinity KL divergence with $\Pxf$ (see Appendix~\ref{app:sol_space_kl} for more details). 
The $\infty$ value for the KL divergence will lead to unbounded loss values thus hurting the stability of the unlearning process. To address this problem, we derive an optimal and bounded KL divergence for the forget set under some reasonable constraints:

\begin{lem}\label{lemma:optimal_gaussian}
 Given the distribution of the forget samples $\Pxf$ with zero-mean and covariance matrix $\Sigma$, consider another signal $\Pxfhat$ which shares the same mean and covariance matrix. The maximal KL-divergence between $\Pxf$ and $\Pxfhat$ is achieved when $\Pxfhat=\mathcal{N}\left(0,\Sigma\right)$~\citep{info_theory_textbook}; that is:
 \begin{equation}
 D\left(\Pxf||\Pxfhat\right)\leq D\left(\Pxf||\mathcal{N}\left(0,\Sigma\right)\right)
 \end{equation}
\end{lem}

We note that making $\Pxfhat$ share the same mean and covariance matrix as $\Pxf$ can preserve the original training set statistical patterns. Consequently, it becomes statistically challenging to decide whether a generated image belongs to the forget set, thereby protecting data privacy.
Moreover, the assumption of zero mean is natural since typically images are normalized by subtracting the mean value inside neural networks. We provide some empirical analysis to demonstrate the benefits of Gaussian distribution (cf. Section~\ref{sec:abl_study}). 

Essentially, Lemma~\ref{lemma:optimal_gaussian} indicates that the maximal KL divergence w.r.t $\Pxf$ is achieved when the generated images $\Pxfhat$ follow the Gaussian distribution $\mathcal{N}\left(0,\Sigma\right)$. Hence, we can directly optimize $\Pxfhat$ towards this optimal solution by minimizing their KL-Divergence; that is:
\begin{equation}\label{eq:new_optimization}
 \argmin_{\theta, \phi} \big\{D\left(\Pxr||\Pxrhat\right)+\alpha D\left(\mathcal{N}\left(0,\Sigma\right)||\Pxfhat\right)\big\},\ \hat{X}_r=\hmodel\left(\trans\left(X_r\right)\right),\ \hat{X}_f=\hmodel\left(\trans\left(X_f\right)\right)
\end{equation}

This way, we avoid the problem of the infinity value of KL-divergence in Eq.~\eref{eq:raw_optimzation}. We note that, for previous unlearning approaches for \textit{classification} tasks, it's natural and straightforward to directly compute the KL-divergence for final outputs since the outputs are exactly single-variable discrete distributions after the SoftMax function~\citep{unlearn_teach_stu1,unlearn_teach_stu2,towards_unbounded}. Nevertheless, for image generation tasks, directly computing the KL divergence between high-dimensional output images is typically intractable, excluding the special case of diffusion models. To address this problem, we next convert the KL divergence into a more efficient $L_2$ loss which is generally applicable to diverse I2I generative models.

\subsection{Proposed Approach}\label{sec:propsoed_apporach}
Directly connecting the KL-Divergence with the $L_2$ loss is difficult. Instead, we use Mutual Information (MI) as a bridge to help with the analysis. As indicated in Eq.~\eref{eq:new_optimization}, we reach the minimal objective value when $\Pxrhat=\Pxr$ and $\Pxfhat=\mathcal{N}(0,\Sigma)$. This optimum can also be achieved by maximizing the mutual information ($I$) between $\Xr$ and $\xrhat$ (or between $n\sim\mathcal{N}(0,\Sigma)$ and $\xfhat$); that is:
\begin{equation}
 \argmax_{\theta, \phi} \big\{I\left(\Xr;\xrhat\right)+\alpha I\left(n;\xfhat\right)\big\},\ n\sim\mathcal{N}(0,\Sigma),\ \hat{X}_r=\hmodel\left(\trans\left(X_r\right)\right),\ \hat{X}_f=\hmodel\left(\trans\left(X_f\right)\right)
\end{equation}
 We next link the MI with a more tractable $L_2$ loss in the representation space. 

\begin{thm}\label{theorem:infonce}
 Suppose the original model can do a perfect generation, \ie, $\hmodelp\left(\trans\left(X\right)\right)= X$. Assume the target model $\hmodel$ uses the same decoder as the original model $\hmodelp$ (\ie, $\decoder=\decoderp$), and the output of the encoders is normalized, \ie, $\|\encoder(x)\|_2=\|\encoderp(x)\|_2=1$. On the retain set, minimizing the $L_2$ loss between the output of the target model encoder $\encoder$ and the original model encoder $\encoderp$ will increase the lower bound of mutual information:
 \begin{equation}\label{eq:theorem_retain}
 \begin{aligned}
 I(X_r;\hat{X_r})\geq\mathrm{log}\left(K\right)-\mathbb{E}\left[\sum_{i=1}^{K}\frac{1}{K}\mathrm{log}\left(e^{\frac{\epsilon_i^2}{2}-1}\sum_{j=1}^{K}e^{\epsilon_j+R_{ij}}\right)\right]\\
 \end{aligned}
 \end{equation}

where $\epsilon_i=\|\encoder\left(\trans(x_{r_i})\right)-\encoderp\left(\trans(x_{r_i})\right)\|_2$ and $R_{ij}=\encoderp(\trans(x_{r_i}))^T \encoderp(\trans(x_{r_j}))$. $x_{r_i}$ are the data samples in the retain set. 
For the forget set, we have:
 \begin{equation}\label{eq:theorem_forget}
 \begin{aligned}
 I(n;\hat{X_f})\geq\mathrm{log}\left(K\right)-\mathbb{E}\left[\sum_{i=1}^{K}\frac{1}{K}\mathrm{log}\left(e^{\frac{\delta_i^2}{2}-1}\sum_{j=1}^{K}e^{\delta_j+F_{ij}}\right)\right],\quad n\sim\mathcal{N}(0,\Sigma)\\
 \end{aligned}
 \end{equation}

where $\delta_i=\|\encoder\left(\trans(x_{f_i})\right)-\encoderp\left(\trans(n_i)\right)\|_2$ and $F_{ij}=\encoderp(\trans(n_i))^T \encoderp(\trans(n_j))$. $x_{fi}$ are the data samples in the forget set and $n_i\sim\mathcal{N}(0,\Sigma)$.  
\end{thm}

We remark that both $R_{ij}$ and $F_{ij}$ are determined by the original encoder $\encoderp$, thus are fixed values. As illustrated in Theorem~\ref{theorem:infonce}, by directly reducing the $L_2$ loss ($\delta_i$ and $\epsilon_i$) between the target encoder and the original encoder, the Mutual Information (MI) increases, concurrently reducing the KL divergence between $\Pxr$ and $\Pxfhat$ (or between $\Pxfhat$ and $\mathcal{N}$). Hence, in our approach, we sidestep the intractability of computing MI or KL divergence by directly minimizing the values of  $\delta_i$ and $\epsilon_i$. Based on these insights, we next introduce our approach.

\begin{figure}[t]
 \centering
 \includegraphics[width=0.96\textwidth]{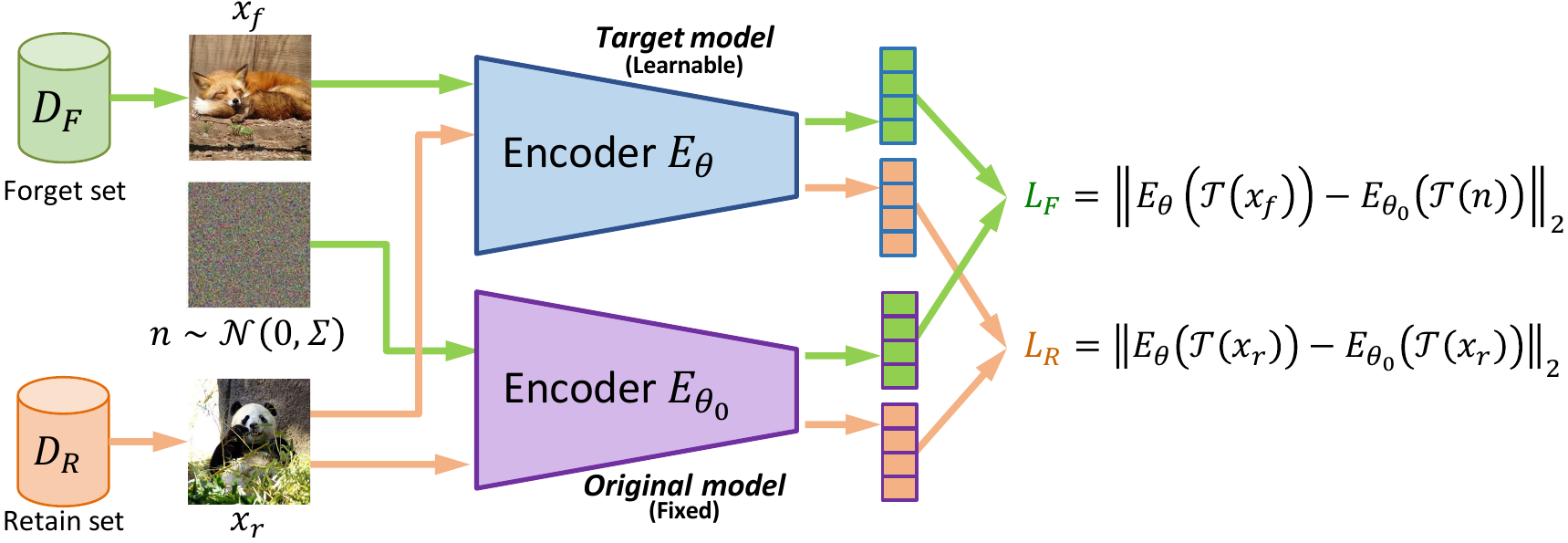}
 \caption{Overview of our approach. On $\mathcal{D}_F$, we  minimize the $L_2$-loss between embedding vectors of the forget samples $x_f$ and embedding vectors of Gaussian noise $n$. On $\mathcal{D}_R$, we minimize the $L_2$-loss between the same image embedding vectors generated by target model encoder and the original model encoder.}
 \label{fig:main_method}
\end{figure}
\paragraph{Efficient Unlearning Approach.}Finally, as shown in Fig.~\ref{fig:main_method}, we propose our efficient unlearning approach for I2I generative models as follows:
\begin{equation}\label{eq:final_optimization}\centering\scriptsize
\begin{split}
 A_F(\hmodelp)\triangleq \argmin_{\theta}\mathop{\mathbb{E}}_{x_{r_i}, x_{f_j}, n}&\bigg\{ \big|\encoder\left(\trans(x_{r_i})\right)-\encoderp\left(\trans(x_{r_i})\right)\big|_2+\alpha\big|\encoder\left(\trans(x_{f_j})\right)-\encoderp\left(\trans(n)\right)\big|_2 \bigg\}\\
 &\quad\quad\quad\quad\quad x_{r_i}\in\mathcal{D}_R, x_{f_j}\in\mathcal{D}_F, n\sim\mathcal{N}(0,\Sigma)\\
\end{split}
\end{equation}

We provide the details of our unlearning algorithm and corresponding pseudo code in Appendix~\ref{app:code_unlearn_alg}.

We note that our proposed approach only involves the encoders. Hence, it's more efficient than manipulating the entire model. Moreover, our approach is generally applicable to various I2I generative models with the encoder-decoder architecture (including the diffusion model, VQ-GAN, or MAE), although they typically use different optimization methods. We illustrate this generalizability in the experiments part.

\section{Experimental Results}\label{sec:exp}
We evaluate our proposed approach on three mainstream I2I  generative models: (\textit{i}) diffusion models~\citep{palette_diff_base}, (\textit{ii}) VQ-GAN~\citep{vqgan_mage}, and (\textit{iii}) MAE~\citep{mae_masked_he}.
\subsection{Experimental Setup}\label{sec:main_setup}

\textbf{Dataset\&Task.} We verify our method on two mainstream large-scale datasets: (\textit{i}) ImageNet-1k. Out of total 1K classes, we randomly select 100 classes as $\mathcal{D}_R$ and another 100 classes as $\mathcal{D}_F$. (\textit{ii}) Places-365. From all 365 classes, we randomly select 50 classes as $\mathcal{D}_R$ and another 50 classes as $\mathcal{D}_F$. We test our method on image extension, uncropping, and reconstruction  tasks. We report the results of center uncropping (\ie, inpainting) in the main paper. The results of other tasks are given in Appendix~\ref{app:supple_results} and ~\ref{app:varyingmaskingtype_ratio}.

\textbf{Baseline.} 
We first report the performance of the original model (\ie, before unlearning) as the reference. 
Since our approach is the first work that does the unlearning for I2I generative models, there are no previous baselines we can directly compare against.
Therefore, we implement three different unlearning approaches that were designed for other tasks, and adapt them to I2I generative models, including (\textit{i}) \textbf{\textsc{Max Loss}} maximizes the training loss w.r.t. the ground truth images on the forget set~\citep{halimi2022federated,erae_diff,unlearn_feat_labels}; (\textit{ii}) \textbf{\textsc{Noisy Label}} minimizes training loss by setting the Gaussian noise as the ground truth images for the forget set~\citep{graves2021amnesiac,erae_diff}; (\textit{iii}) \textbf{\textsc{Retain Label}} minimizes training loss by setting the retain samples as the ground truth for the forget set~\citep{mul_generative3}; (\textit{iv}) \textbf{\textsc{Random Encoder}} directly minimizes the $L_2$ loss between the encoder's output on the forget set and a Gaussian noise~\citep{tarun2023deep}. 
For all these baselines, we use the retain samples with some regularization to avoid hurting the performance on the retain set. 
For more details, please check Appendix~\ref{app:baseline_details}.

\textbf{Evaluation metrics.} 
We adopt three different types of metrics to compare our method against other baselines: (\textit{i}) inception score (IS) of the generated images~\citep{inception_score}, (\textit{ii}) Fr\'echet inception distance (FID) against the real images~\citep{heusel2017gans_fid} and (\textit{iii}) CLIP embedding distance between the generated images and the real images~\citep{clip}.
IS assesses the quality of the generated images alone, while FID further measure the similarity between generated and real images. On the other hand, the CLIP embedding distance measures whether or not the generated images still capture similar semantics.

\begin{table}[tb]
\centering
\caption{Results of cropping $8\times 8$ patches at the center of the image, where each patch is $ 16 \times 16$ pixels. `$\uparrow$' means higher is better and `$\downarrow$' means lower is better. $R$ and $F$ account for the retain set and forget set, respectively.`{Proxy $\mathcal{D}_R$}' means that we use the images from other classes as a substitute of the real retain set to do the unlearning  (cf. Section~\ref{sec:retain_avail}). }\label{tab:results_center64}
\begin{adjustbox}{max width=\linewidth}
\begin{tabular}{ccccccccccccccccccc}
\toprule
    &\multicolumn{6}{c}{Diffusion Models}&\multicolumn{6}{c}{VQ-GAN}&\multicolumn{6}{c}{MAE}\\
    &\multicolumn{2}{c}{FID}&\multicolumn{2}{c}{IS}&\multicolumn{2}{c}{CLIP}&\multicolumn{2}{c}{FID}&\multicolumn{2}{c}{IS}&\multicolumn{2}{c}{CLIP}&\multicolumn{2}{c}{FID}&\multicolumn{2}{c}{IS}&\multicolumn{2}{c}{CLIP}\\
    \cmidrule(lr){2-7} \cmidrule(lr){8-13} \cmidrule(lr){14-19}
    &{$R$$\downarrow$}&{$F$$\uparrow$}&{$R$$\uparrow$}&{$F$$\downarrow$}&{$R$$\uparrow$}&$F$$\downarrow$&{$R$$\downarrow$}&{$F$$\uparrow$}&{$R$$\uparrow$}&{$F$$\downarrow$}&{$R$$\uparrow$}&$F$$\downarrow$&{$R$$\downarrow$}&{$F$$\uparrow$}&{$R$$\uparrow$}&{$F$$\downarrow$}&{$R$$\uparrow$}&$F$$\downarrow$\\
    \midrule
\textit{Original model} & \textit{12.2} & \textit{14.6} & \textit{19.3} & \textit{23.1} & \textit{0.88} & \textit{0.89} & \textit{14.4} & \textit{14.4} & \textit{19.4} & \textit{20.6} & \textit{0.75} & \textit{0.77} & \textit{56.7} & \textit{84.1} & \textit{23.0} & \textit{17.4} & \textit{0.73} & \textit{0.71} \\ \midrule
    \textsc{Max Loss} & 34.1 & 45.7 & 12.8 & 17.1 & 0.77 & 0.76 & 16.9 & \textbf{115.2} & 17.4 & \textbf{11.0} & 0.73 & \textbf{0.55} & 75.8 & 112.6 & 19.4 & 15.2 & 0.69 & \textbf{0.65} \\ 
    \textsc{Noisy Label} & 14.7 & 36.9 & 19.3 & 19.1 & 0.86 & 0.80 & 14.8 & 79.5 & 17.2 & 11.4 & \textbf{0.74} & 0.64 & 60.4 & 136.5 & 21.6 & 12.8 & 0.71 & 0.67 \\ 
    \textsc{Retain Label} & 23.1 & 104.7 & 18.2 & 12.3 & 0.81 & \textbf{0.69} & 21.8 & 23.3 & 18.2 & 18.3 & 0.72 & 0.74 & 72.8 & 145.3 & 18.8 & 11.6 & 0.69 & 0.66 \\ 
    \textsc{Random Encoder} & 15.3 & 30.6 & 18.7 & 19.4 & 0.86 & 0.81 & \textbf{14.7} & 72.8 & \textbf{18.6} & 14.1 & \textbf{0.74} & 0.64 & \textbf{58.1} & 146.4 & \textbf{22.3} & 12.8 & \textbf{0.72} & 0.67 \\ \midrule
    \textbf{Ours} & \textbf{13.4} & \textbf{107.9} & \textbf{19.4} & \textbf{10.3} & \textbf{0.87} & \textbf{0.69} & 15.0 & 83.4 & 18.3 & 11.6 & \textbf{0.74} & 0.60 & 59.9 & \textbf{153.0} & 21.8 & \textbf{11.0} & \textbf{0.72} & 0.67 \\ 
    \textbf{Ours (Proxy $\mathcal{D}_R$)} & 17.9 & 75.5 & 18.2 & 12.3 & 0.83 & 0.74 & 17.6 & 69.7 & \textbf{18.6} & 14.0 & 0.73 & 0.63 & 61.1 & 133.8 & 21.0 & 12.3 & \textbf{0.72} & 0.68  \\ \bottomrule
    \end{tabular}
\end{adjustbox}
\end{table}

\begin{table}[tb]
\centering
\caption{Results of cropping $4\times 4$ patches at the center of the image, where each patch is $ 16 \times 16$ pixels. `$\uparrow$' means higher is better and `$\downarrow$' means lower is better. $R$ and $F$ account for the retain set and forget set, respectively. ``{Proxy $\mathcal{D}_R$}'' means that we use the images from other classes as a substitute of the real retain set to do the unlearning (cf. Section~\ref{sec:retain_avail}). }\label{tab:results_center16}
\begin{adjustbox}{max width=\linewidth}
\begin{tabular}{ccccccccccccccccccc}
\toprule
    &\multicolumn{6}{c}{Diffusion Models}&\multicolumn{6}{c}{VQ-GAN}&\multicolumn{6}{c}{MAE}\\
    &\multicolumn{2}{c}{FID}&\multicolumn{2}{c}{IS}&\multicolumn{2}{c}{CLIP}&\multicolumn{2}{c}{FID}&\multicolumn{2}{c}{IS}&\multicolumn{2}{c}{CLIP}&\multicolumn{2}{c}{FID}&\multicolumn{2}{c}{IS}&\multicolumn{2}{c}{CLIP}\\
    \cmidrule(lr){2-7} \cmidrule(lr){8-13} \cmidrule(lr){14-19}
    &{$R$$\downarrow$}&{$F$$\uparrow$}&{$R$$\uparrow$}&{$F$$\downarrow$}&{$R$$\uparrow$}&$F$$\downarrow$&{$R$$\downarrow$}&{$F$$\uparrow$}&{$R$$\uparrow$}&{$F$$\downarrow$}&{$R$$\uparrow$}&$F$$\downarrow$&{$R$$\downarrow$}&{$F$$\uparrow$}&{$R$$\uparrow$}&{$F$$\downarrow$}&{$R$$\uparrow$}&$F$$\downarrow$\\
    \midrule
    \textit{Original model} & \textit{7.8} & \textit{6.0} & \textit{10.3} & \textit{11.2} & \textit{0.93} & \textit{0.96} & \textit{8.4} & \textit{7.8} & \textit{15.1} & \textit{14.2} & \textit{0.84} & \textit{0.85} & \textit{11.4} & \textit{15.8} & \textit{50.8} & \textit{46.6} & \textit{0.87} & \textit{0.87} \\ \midrule
    \textsc{Max Loss} & 11.9 & 15.4 & 10.0 & 11.0 & 0.88 & 0.93 & 9.2 & \textbf{39.9} & 15.2 & \textbf{13.1} & 0.83 & \textbf{0.72} & 13.3 & 20.2 & \textbf{50.8} & 46.0 & \textbf{0.86} & 0.83 \\ 
    \textsc{Noisy Label} & 19.6 & 18.5 & \textbf{10.4} & 10.6 & 0.87 & 0.91 & 8.7 & 21.3 & 15.2 & 14.1 & \textbf{0.84} & 0.80 & 12.2 & 44.3 & 50.0 & 35.4 & \textbf{0.86} & 0.82 \\ 
    \textsc{Retain Label} & 8.5 & 35.1 & 10.3 & \textbf{10.5} & \textbf{0.93} & 0.89 & 11.0 & 10.3 & \textbf{15.4} & 14.2 & 0.83 & 0.84 & 15.3 & \textbf{47.5} & 47.6 & 34.9 & 0.85 & \textbf{0.81} \\ 
    \textsc{Random Encoder} & 15.3 & 11.6 & 10.1 & 11.1 & 0.86 & 0.94 & \textbf{8.6} & 19.4 & 15.3 & 14.4 & \textbf{0.84} & 0.81 & \textbf{11.8} & 43.6 & 50.3 & 36.3 & \textbf{0.86} & 0.83 \\ 
    \midrule
    \textbf{Ours} & \textbf{8.2} & \textbf{39.8} & 10.3 & 10.7 & \textbf{0.93} & \textbf{0.88} & \textbf{8.6} & 22.0 & 15.0 & 14.1 & \textbf{0.84} & 0.79 & 12.2 & 45.1 & 49.7 & \textbf{34.8} & \textbf{0.86} & 0.83 \\ 
    \textbf{Ours (Proxy $\mathcal{D}_R$)} & 11.2 & 29.0 & 10.3 & 10.8 & 0.91 & 0.9 & 8.9 & 20.0 & \textbf{15.4} & 14.3 & \textbf{0.84} & 0.80 & 12.5 & 39.9 & 49.5 & 36.8 & \textbf{0.86} & 0.83 \\ \bottomrule
    \end{tabular}
\end{adjustbox}
\end{table}

\begin{figure}[t!]
 \centering
\includegraphics[width=0.98\textwidth]{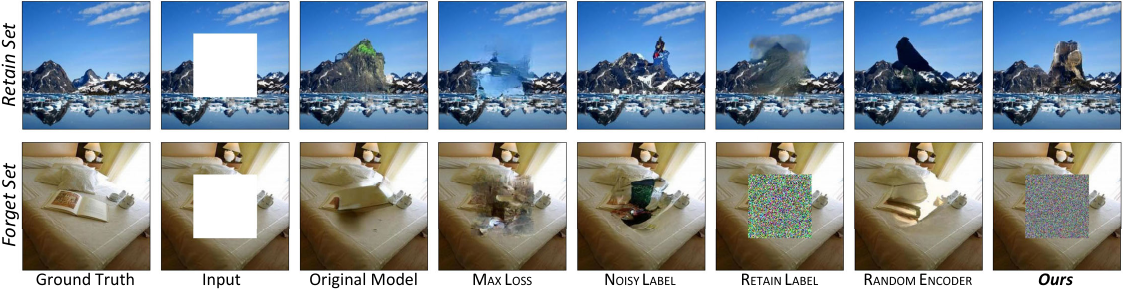}\vspace{-2mm}
 \caption{Results of cropping $8\times 8$ patches at the center of the image on diffusion models, where each patch is $ 16 \times 16$ pixels. Our method has negligible-to-slight performance degradation on diverse I2I generative models and multiple generative tasks. (cf. Appendix~\ref{app:supple_results} and \ref{app:varyingmaskingtype_ratio}).}
 \label{fig:result_visual}
\end{figure}

\begin{figure}[th]
 \centering
 \begin{subfigure}[b]{0.32\textwidth}
 \centering
 \includegraphics[width=\textwidth]{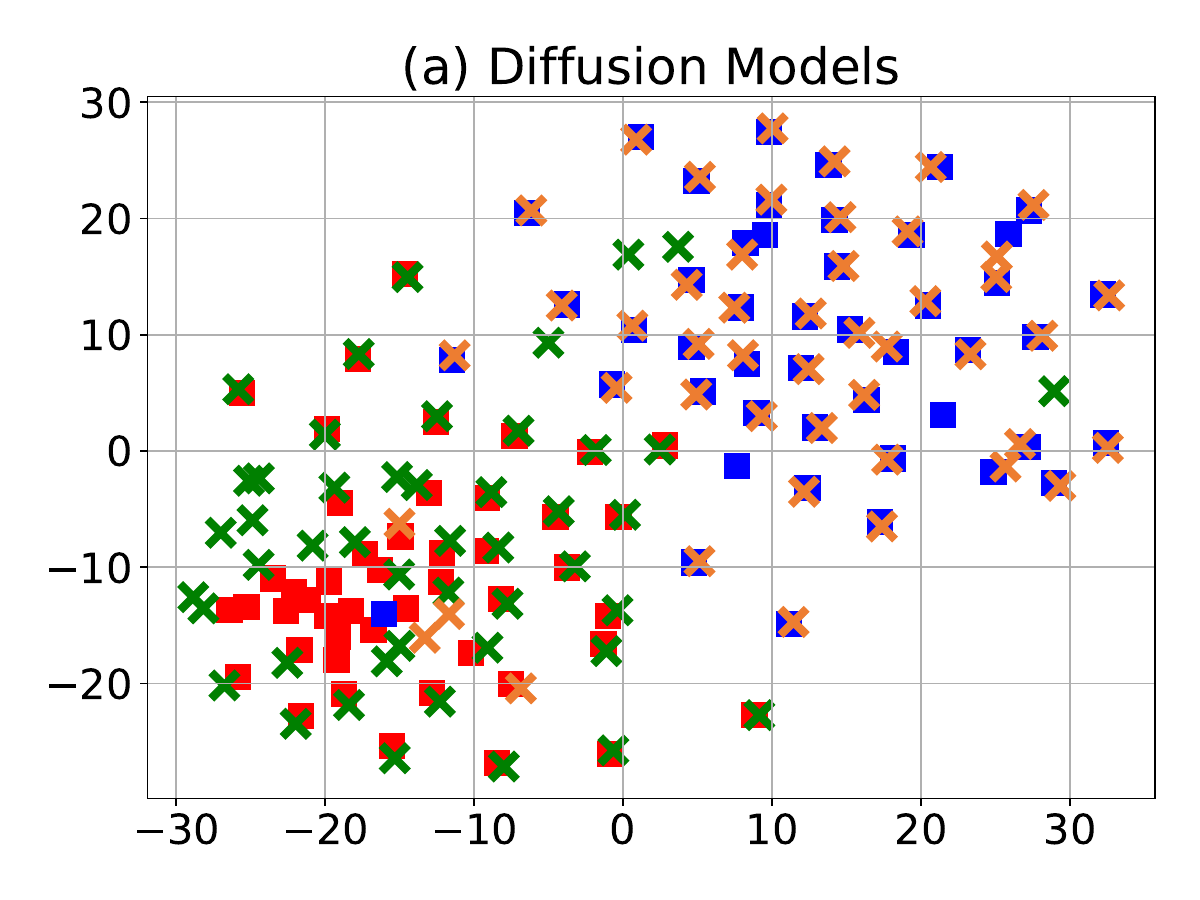}\vspace{-5mm}
 \label{fig:tsne_dm}
 \end{subfigure}
 \hfill
 \begin{subfigure}[b]{0.32\textwidth}
 \centering
 \includegraphics[width=\textwidth]{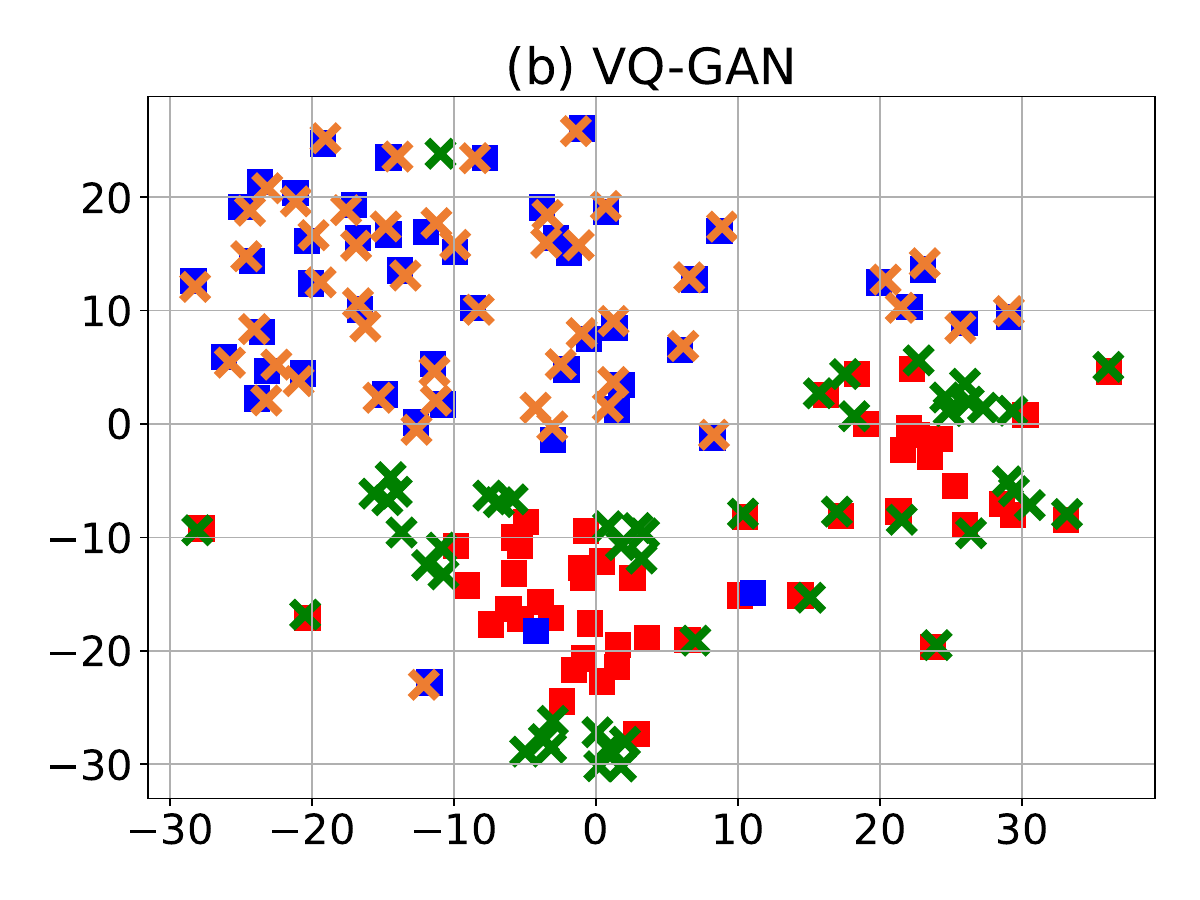}\vspace{-5mm}
 \label{fig:tsne_gan}
 \end{subfigure}
  \hfill
 \begin{subfigure}[b]{0.32\textwidth}
 \centering
 \includegraphics[width=\textwidth]{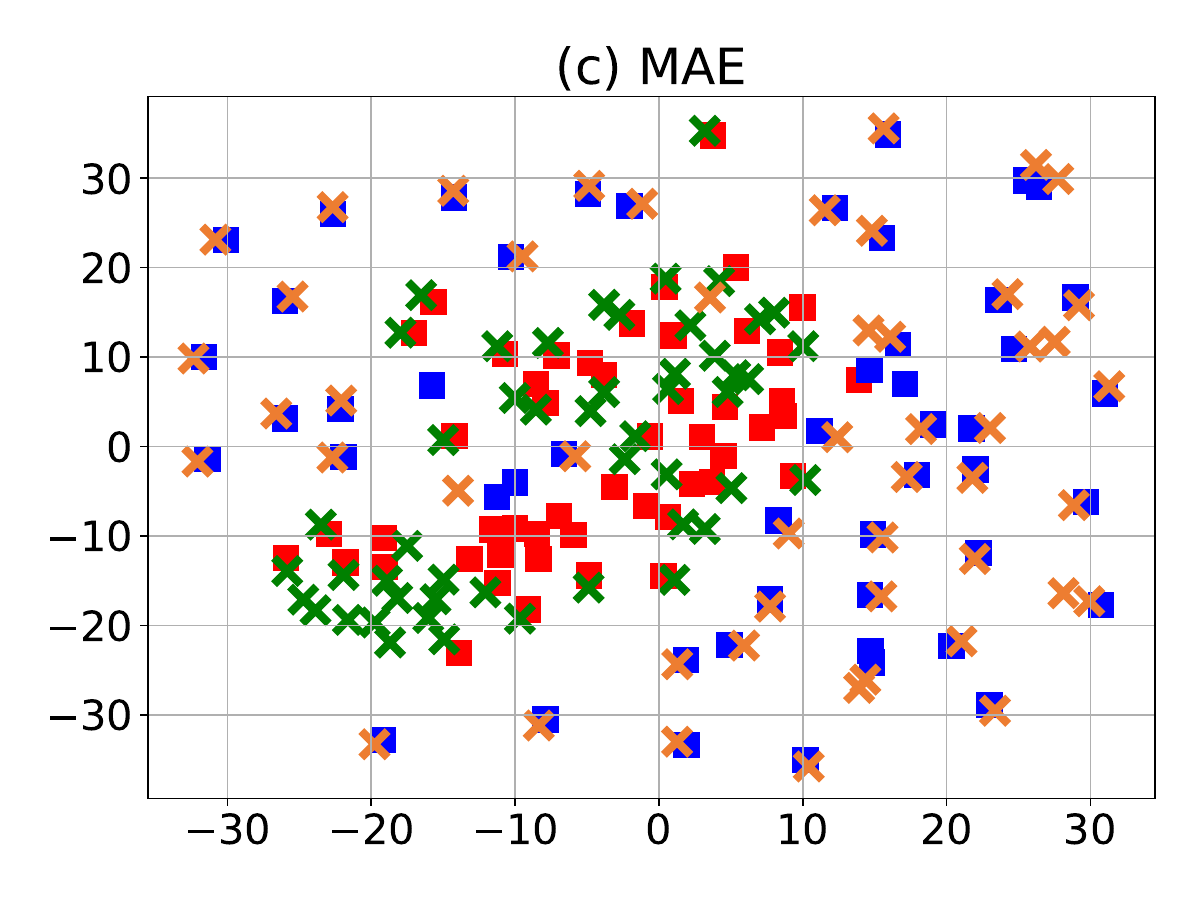}\vspace{-5mm}
 \label{fig:tsne_mae}
 \end{subfigure}

  \begin{subfigure}[b]{0.9\textwidth}
 \centering
 \includegraphics[width=\textwidth]{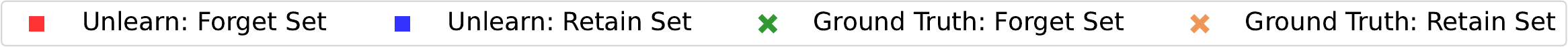}\vspace{-2mm}
 \end{subfigure}
 \caption{T-SNE analysis of the generated images by our approach and ground truth images. After unlearning, the generated retain samples are close to or overlapping with the ground truth (orange \textit{vs.} blue), while most of generated forget images diverge far from the ground truth (green \textit{vs.} red). }
 \label{fig:tsne_main}
\end{figure}

\begin{table}[th]
\centering
\caption{Ablation study of $\alpha$'s values. We test the performance of cropping $8\times 8$ patches at the center of the image. As shown, $\alpha=0.25$ achieves a good balance between the preserving the performance on retain set, while remove the information on forget sets across these two models.}\label{tab:main_abl_alpha}
\begin{adjustbox}{max width=\linewidth}
\begin{tabular}{cccccccccccccccc}
\toprule
\multicolumn{2}{c}{} & \multicolumn{7}{c}{VQ-GAN} & \multicolumn{7}{c}{MAE} \\
\cmidrule(lr){3-9} \cmidrule(lr){10-16}
\multicolumn{2}{c}{$\alpha$} & {0.01} & {0.05} & {0.1} & {0.2} & \textbf{0.25} & {0.5} & 1 & {0.01} & {0.05} & {0.1} & {0.2} & \textbf{0.25} & {0.5} & 1 \\ \midrule
\multirow{2}{*}{FID} & $R$$\downarrow$ & 90.8 & 91.6 & 92.0 & 91.7 & \textbf{92.7} & 92.2 & 94.7 & 113.6 & 113.2 & 113.9 & 116.7 & \textbf{115.9} & 116.3 & 116.7 \\ 

 & $F$$\uparrow$ & 101.2 & 169.4 & 179.5 & 181.3 & \textbf{183.4} & 182.2 & 184.6 & 179.0 & 198.6 & 205.1 & 211.5 & \textbf{213.0} & 213.4 & 213.0 \\ \midrule
\multirow{2}{*}{IS} & $R$$\uparrow$ & 12.5 & 12.8 & 12.5 & 12.4 & \textbf{12.2} & 12.0 & 12.6 & 13.3 & 13.3 & 13.4 & 13.5 & \textbf{13.2} & 13.3 & 12.9 \\ 

 & $F$$\downarrow$ & 11.5 & 8.4 & 7.8 & 7.9 & \textbf{8.1} & 7.9 & 8.0 & 9.3 & 9.0 & 8.5 & 8.0 & \textbf{8.0} & 8.1 & 7.9 \\ \midrule
\multirow{2}{*}{CLIP} & $R$$\uparrow$ & 0.65 & 0.65 & 0.65 & 0.65 & \textbf{0.65} & 0.65 & 0.64 & 0.81 & 0.81 & 0.81 & 0.80 & \textbf{0.80} & 0.80 & 0.80 \\ 

 & $F$$\downarrow$ & 0.66 & 0.55 & 0.54 & 0.54 & \textbf{0.54} & 0.54 & 0.54 & 0.79 & 0.78 & 0.78 & 0.78 & \textbf{0.78} & 0.78 & 0.78 \\ \bottomrule
\end{tabular}
\end{adjustbox}\vspace{-3mm}
\end{table}

\subsection{Performance Analysis and Visualization}\label{sec:main_results}
As shown in Table~\ref{tab:results_center64} and Table~\ref{tab:results_center16}, compared to the original model, our approach has almost identical performance or only a slight degradation on the retain set. Meanwhile, there are significant performance drops on the forget set across all these three models for all metrics. In contrast, none of these baselines generally works well. For example, \textsc{Random Encoder} achieves similar performance on VQ-GAN and MAE to our methods; however, it is much worse on diffusion models. Similarly, \textsc{Retain Label} works well for diffusion models, but cannot generalize to VQ-GAN and MAE. 
We also show some generated images in Fig.~\ref{fig:result_visual}. As shown, our approach removes the information in the forget set while preserving the performance on the retain set. 

\paragraph{T-SNE analysis.}
To further analyze why our approach works well, we conduct the T-SNE analysis. Using our unlearned model, we generate 50 images for both the retain and forget sets.   We then compute the CLIP embedding vector of these images and their corresponding ground truth images. As shown in Fig.~\ref{fig:tsne_main}, after unlearning, the CLIP embedding vector on the retain set is close to or overlapping with the ground truth images, while most of generated images on the forget set diverge far from the ground truth.

These results verify that our method is generally applicable to mainstream I2I generative models and consistently achieves good results on all these models. We provide more results under various types of cropping in Appendix~\ref{app:supple_results} and Appendix~\ref{app:ablation_study}.

\subsection{Robustness to Retain Samples Availability}\label{sec:retain_avail}
In machine unlearning, sometimes the real retain samples are not available due to data retention policies. To address this challenge, we evaluate our approach by using other classes of images as substitute to the real retain samples. On ImageNet-1K, since we already select 200 classes for forget and retain sets, we randomly select some images from the remaining 800 classes as the ``proxy retain set'' used in the unlearning process. Similarly, for Places-365, we randomly select some images from the remaining 265 classes  as the ``proxy retain set'' used in the unlearning process. We also ensure these ``proxy retain sets'' have the same number of images as the forget set.

As shown in the last row in Table~\ref{tab:results_center64} and Table~\ref{tab:results_center16}, our method works well even without the access to the real/original retain set. Compared to using the real/original retain set, there is only a slight performance drop. Hence, our approach is flexible and generally applicable without the dependency on the real retain samples. We provide the results with limited availability to the real retain samples in Appendix~\ref{app:vary_retain}.

\subsection{Ablation Study}\label{sec:abl_study}

For the ablation study, we test the results of cropping patches at the center of the image under various setups, where each patch is $16\times16$ pixels. 
\vspace{-2mm}\paragraph{$\alpha$'s value.}
We vary the value of $\alpha$ in Eq.~\eref{eq:final_optimization} to obtain multiple models and then evaluate their performance. As shown in Table~\ref{tab:main_abl_alpha}, when  $\alpha$ is 0.25, our approach achieves a good balance between the forget set and the retain set. Hence, we set $\alpha=0.25$ as default value for our approach. We provide more ablation study in Appendix~\ref{app:ablation_study}.

\section{Conclusions and Final Remarks}\label{sec:limit_future}

In this paper, we have formulated the machine unlearning problem for I2I generative models and derived an efficient algorithm that is applicable across various I2I generative models, including diffusion models, VQ-GAN, and MAE. Our method has shown negligible performance degradation on the retain set, while effectively removing the information from the forget set, on two large-scale datasets (ImageNet-1K and Places-365). Remarkably, our approach is still effective with limited or no real retain samples. To our best knowledge, we are the first to systematically explore machine unlearning for image completion generative models.

\paragraph{Limitations.}
First, our methods are mainly verified on I2I generative models. Second, our approach requires the access of original/real forget samples yet sometimes they are unavailable. Besides, for the simplicity of evaluation, we only test our approach on some mainstream computer vision datasets. Our approach has not been verified under a more practical/useful scenarios, \eg, remove the pornographic information for I2I generative models.  

\paragraph{Future directions.}
We plan to explore applicability to other modality, especially for language/text generation and text-to-image generation. The dependency on the forget set is another challenge that enable flexibility in the unlearning for generative models.
Finally, we  also intend to develop some more practical benchmarks related to the control of generative contents and protect the data privacy and copyright. 

\subsubsection*{Disclaimer}
\footnotesize{
This paper was prepared for informational purposes by the Global Technology Applied Research center of JPMorgan Chase \& Co. This paper is not a product of the Research Department of JPMorgan Chase \& Co. or its affiliates. Neither JPMorgan Chase \& Co. nor any of its affiliates makes any explicit or implied representation or warranty and none of them accept any liability in connection with this paper, including, without limitation, with respect to the completeness, accuracy, or reliability of the information contained herein and the potential legal, compliance, tax, or accounting effects thereof. This document is not intended as investment research or investment advice, or as a recommendation, offer, or solicitation for the purchase or sale of any security, financial instrument, financial product or service, or to be used in any way for evaluating the merits of participating in any transaction. Guihong Li's and Radu Marculescu's contributions were made as part of Guihong Li's internship at the Global Technology Applied Research center of JPMorgan Chase \& Co.
}

\newpage
\noindent\textbf{Ethics statement.} Machine unlearning for I2I generative models can be effectively exploited to avoid generate contents related user privacy and copyright. Moreover, unlearning for I2I models can avoid generating harmful contents, such as violence or pornography.

\noindent\textbf{Reproducibility statement.} All the datasets used in this paper are open dataset and are available to the public. Besides, our codes are primarily based on PyTorch~\citep{paszke2017automatic_torch}. We use several open source code base and model checkpoints to build our own approach (see Appendix~\ref{app:dataset_code_details}). Our approach can be implemented by obtaining the outputs of target model's encoders and the original model's encoders and then computing the $L_2$-loss between them.
We provide more implementation details in Appendix~\ref{app:details}.

\bibliography{iclr2024_conference}

\begin{thebibliography}{77}
\providecommand{\natexlab}[1]{#1}
\providecommand{\url}[1]{\texttt{#1}}
\expandafter\ifx\csname urlstyle\endcsname\relax
  \providecommand{\doi}[1]{doi: #1}\else
  \providecommand{\doi}{doi: \begingroup \urlstyle{rm}\Url}\fi

\bibitem[Arjovsky et~al.(2017)Arjovsky, Chintala, and Bottou]{wasserstein_gan}
Mart{\'{\i}}n Arjovsky, Soumith Chintala, and L{\'{e}}on Bottou.
\newblock Wasserstein generative adversarial networks.
\newblock In \emph{International Conference on Machine Learning}, pp.\  214--223. {PMLR}, 2017.

\bibitem[Bae et~al.(2023)Bae, Kim, Jung, and Lim]{mul_generative1}
Seohui Bae, Seoyoon Kim, Hyemin Jung, and Woohyung Lim.
\newblock Gradient surgery for one-shot unlearning on generative model.
\newblock \emph{CoRR}, abs/2307.04550, 2023.

\bibitem[Belghazi et~al.(2018)Belghazi, Baratin, Rajeshwar, Ozair, Bengio, Courville, and Hjelm]{mine}
Mohamed~Ishmael Belghazi, Aristide Baratin, Sai Rajeshwar, Sherjil Ozair, Yoshua Bengio, Aaron Courville, and Devon Hjelm.
\newblock Mutual information neural estimation.
\newblock In \emph{International Conference on Machine Learning}, pp.\  531--540. PMLR, 2018.

\bibitem[Bommasani et~al.(2021)Bommasani, Hudson, Adeli, Altman, Arora, von Arx, Bernstein, Bohg, Bosselut, Brunskill, Brynjolfsson, Buch, Card, Castellon, Chatterji, Chen, Creel, Davis, Demszky, Donahue, Doumbouya, Durmus, Ermon, Etchemendy, Ethayarajh, Fei{-}Fei, Finn, Gale, Gillespie, Goel, Goodman, Grossman, Guha, Hashimoto, Henderson, Hewitt, Ho, Hong, Hsu, Huang, Icard, Jain, Jurafsky, Kalluri, Karamcheti, Keeling, Khani, Khattab, Koh, Krass, Krishna, Kuditipudi, and et~al.]{risk_of_found}
Rishi Bommasani, Drew~A. Hudson, Ehsan Adeli, Russ~B. Altman, Simran Arora, Sydney von Arx, Michael~S. Bernstein, Jeannette Bohg, Antoine Bosselut, Emma Brunskill, Erik Brynjolfsson, Shyamal Buch, Dallas Card, Rodrigo Castellon, Niladri~S. Chatterji, Annie~S. Chen, Kathleen Creel, Jared~Quincy Davis, Dorottya Demszky, Chris Donahue, Moussa Doumbouya, Esin Durmus, Stefano Ermon, John Etchemendy, Kawin Ethayarajh, Li~Fei{-}Fei, Chelsea Finn, Trevor Gale, Lauren Gillespie, Karan Goel, Noah~D. Goodman, Shelby Grossman, Neel Guha, Tatsunori Hashimoto, Peter Henderson, John Hewitt, Daniel~E. Ho, Jenny Hong, Kyle Hsu, Jing Huang, Thomas Icard, Saahil Jain, Dan Jurafsky, Pratyusha Kalluri, Siddharth Karamcheti, Geoff Keeling, Fereshte Khani, Omar Khattab, Pang~Wei Koh, Mark~S. Krass, Ranjay Krishna, Rohith Kuditipudi, and et~al.
\newblock On the opportunities and risks of foundation models.
\newblock \emph{CoRR}, abs/2108.07258, 2021.

\bibitem[Bourtoule et~al.(2021)Bourtoule, Chandrasekaran, Choquette{-}Choo, Jia, Travers, Zhang, Lie, and Papernot]{sisa_unlearn}
Lucas Bourtoule, Varun Chandrasekaran, Christopher~A. Choquette{-}Choo, Hengrui Jia, Adelin Travers, Baiwu Zhang, David Lie, and Nicolas Papernot.
\newblock Machine unlearning.
\newblock In \emph{42nd {IEEE} Symposium on Security and Privacy, {SP} 2021, San Francisco, CA, USA, 24-27 May 2021}, pp.\  141--159. {IEEE}, 2021.

\bibitem[Brock et~al.(2019)Brock, Donahue, and Simonyan]{large_scale_gan}
Andrew Brock, Jeff Donahue, and Karen Simonyan.
\newblock Large scale {GAN} training for high fidelity natural image synthesis.
\newblock In \emph{7th International Conference on Learning Representations, {ICLR} 2019, New Orleans, LA, USA, May 6-9, 2019}. OpenReview.net, 2019.

\bibitem[Bubeck et~al.(2023)Bubeck, Chandrasekaran, Eldan, Gehrke, Horvitz, Kamar, Lee, Lee, Li, Lundberg, Nori, Palangi, Ribeiro, and Zhang]{intelligence_gpt4}
S{\'{e}}bastien Bubeck, Varun Chandrasekaran, Ronen Eldan, Johannes Gehrke, Eric Horvitz, Ece Kamar, Peter Lee, Yin~Tat Lee, Yuanzhi Li, Scott~M. Lundberg, Harsha Nori, Hamid Palangi, Marco~T{\'{u}}lio Ribeiro, and Yi~Zhang.
\newblock Sparks of artificial general intelligence: Early experiments with {GPT-4}.
\newblock \emph{CoRR}, abs/2303.12712, 2023.

\bibitem[Bulat et~al.(2018)Bulat, Yang, and Tzimiropoulos]{super_reso_gan}
Adrian Bulat, Jing Yang, and Georgios Tzimiropoulos.
\newblock To learn image super-resolution, use a gan to learn how to do image degradation first.
\newblock In \emph{Proceedings of the European Conference on Computer Vision (ECCV)}, pp.\  185--200, 2018.

\bibitem[Carlini et~al.(2023)Carlini, Hayes, Nasr, Jagielski, Sehwag, Tramer, Balle, Ippolito, and Wallace]{carlini2023extracting}
Nicolas Carlini, Jamie Hayes, Milad Nasr, Matthew Jagielski, Vikash Sehwag, Florian Tramer, Borja Balle, Daphne Ippolito, and Eric Wallace.
\newblock Extracting training data from diffusion models.
\newblock In \emph{32nd USENIX Security Symposium (USENIX Security 23)}, pp.\  5253--5270, 2023.

\bibitem[Chang et~al.(2022)Chang, Zhang, Jiang, Liu, and Freeman]{image_extend_maskgit}
Huiwen Chang, Han Zhang, Lu~Jiang, Ce~Liu, and William~T. Freeman.
\newblock Maskgit: Masked generative image transformer.
\newblock In \emph{{IEEE/CVF} Conference on Computer Vision and Pattern Recognition, {CVPR} 2022, New Orleans, LA, USA, June 18-24, 2022}, pp.\  11305--11315. {IEEE}, 2022.

\bibitem[Chen et~al.(2023)Chen, Gao, Liu, Peng, and Wang]{cvpr_boundary_unlearn}
Min Chen, Weizhuo Gao, Gaoyang Liu, Kai Peng, and Chen Wang.
\newblock Boundary unlearning: Rapid forgetting of deep networks via shifting the decision boundary.
\newblock In \emph{{IEEE/CVF} Conference on Computer Vision and Pattern Recognition, {CVPR} 2023, Vancouver, BC, Canada, June 17-24, 2023}, pp.\  7766--7775. {IEEE}, 2023.

\bibitem[Chen et~al.(2016)Chen, Duan, Houthooft, Schulman, Sutskever, and Abbeel]{chen2016infogan}
Xi~Chen, Yan Duan, Rein Houthooft, John Schulman, Ilya Sutskever, and Pieter Abbeel.
\newblock Infogan: Interpretable representation learning by information maximizing generative adversarial nets.
\newblock In \emph{Advances in Neural Information Processing Systems}, pp.\  2172--2180, 2016.

\bibitem[Chourasia \& Shah(2023)Chourasia and Shah]{grad_up}
Rishav Chourasia and Neil Shah.
\newblock Forget unlearning: Towards true data-deletion in machine learning.
\newblock In Andreas Krause, Emma Brunskill, Kyunghyun Cho, Barbara Engelhardt, Sivan Sabato, and Jonathan Scarlett (eds.), \emph{International Conference on Machine Learning}, volume 202 of \emph{Proceedings of Machine Learning Research}, pp.\  6028--6073. {PMLR}, 2023.

\bibitem[Chundawat et~al.(2023)Chundawat, Tarun, Mandal, and Kankanhalli]{zero_shot_unlearn}
Vikram~S. Chundawat, Ayush~K. Tarun, Murari Mandal, and Mohan~S. Kankanhalli.
\newblock Zero-shot machine unlearning.
\newblock \emph{{IEEE} Trans. Inf. Forensics Secur.}, 18:\penalty0 2345--2354, 2023.

\bibitem[Congress(2022{\natexlab{a}})]{japan_bill}
Japan Congress.
\newblock Act on the protection of personal information, 2022{\natexlab{a}}.
\newblock URL \url{https://www.ppc.go.jp/files/pdf/280222_amendedlaw.pdf}.

\bibitem[Congress(2022{\natexlab{b}})]{datbill}
United~States Congress.
\newblock American data privacy and protection act, 2022{\natexlab{b}}.
\newblock URL \url{https://www.congress.gov/bill/117th-congress/house-bill/8152}.

\bibitem[Cover \& Thomas(2012)Cover and Thomas]{info_theory_textbook}
T.M. Cover and J.A. Thomas.
\newblock \emph{Elements of Information Theory}, chapter~12, pp.\  409--413.
\newblock Wiley, 2012.
\newblock ISBN 9781118585771.

\bibitem[Deng et~al.(2009)Deng, Dong, Socher, Li, Li, and Fei{-}Fei]{deng2009imagenet}
Jia Deng, Wei Dong, Richard Socher, Li{-}Jia Li, Kai Li, and Li~Fei{-}Fei.
\newblock Imagenet: {A} large-scale hierarchical image database.
\newblock In \emph{2009 {IEEE} Computer Society Conference on Computer Vision and Pattern Recognition {(CVPR} 2009), 20-25 June 2009, Miami, Florida, {USA}}, pp.\  248--255. {IEEE} Computer Society, 2009.

\bibitem[Dhariwal \& Nichol(2021)Dhariwal and Nichol]{diffbeat_gan}
Prafulla Dhariwal and Alexander~Quinn Nichol.
\newblock Diffusion models beat gans on image synthesis.
\newblock In \emph{Advances in Neural Information Processing Systems}, pp.\  8780--8794, 2021.

\bibitem[Dosovitskiy et~al.(2021)Dosovitskiy, Beyer, Kolesnikov, Weissenborn, Zhai, Unterthiner, Dehghani, Minderer, Heigold, Gelly, Uszkoreit, and Houlsby]{vision_transformer}
Alexey Dosovitskiy, Lucas Beyer, Alexander Kolesnikov, Dirk Weissenborn, Xiaohua Zhai, Thomas Unterthiner, Mostafa Dehghani, Matthias Minderer, Georg Heigold, Sylvain Gelly, Jakob Uszkoreit, and Neil Houlsby.
\newblock An image is worth 16x16 words: Transformers for image recognition at scale.
\newblock In \emph{9th International Conference on Learning Representations, {ICLR} 2021, Virtual Event, Austria, May 3-7, 2021}. OpenReview.net, 2021.

\bibitem[Feichtenhofer et~al.(2022)Feichtenhofer, Li, He, et~al.]{feichtenhofer2022masked}
Christoph Feichtenhofer, Yanghao Li, Kaiming He, et~al.
\newblock Masked autoencoders as spatiotemporal learners.
\newblock In \emph{Advances in Neural Information Processing Systems}, volume~35, pp.\  35946--35958, 2022.

\bibitem[Gandikota et~al.(2023)Gandikota, Materzynska, Fiotto{-}Kaufman, and Bau]{erae_diff}
Rohit Gandikota, Joanna Materzynska, Jaden Fiotto{-}Kaufman, and David Bau.
\newblock Erasing concepts from diffusion models.
\newblock \emph{CoRR}, abs/2303.07345, 2023.

\bibitem[Golatkar et~al.(2020{\natexlab{a}})Golatkar, Achille, and Soatto]{golatkar2020eternal}
Aditya Golatkar, Alessandro Achille, and Stefano Soatto.
\newblock Eternal sunshine of the spotless net: Selective forgetting in deep networks.
\newblock In \emph{2020 {IEEE/CVF} Conference on Computer Vision and Pattern Recognition, {CVPR} 2020, Seattle, WA, USA, June 13-19, 2020}, pp.\  9301--9309. Computer Vision Foundation / {IEEE}, 2020{\natexlab{a}}.

\bibitem[Golatkar et~al.(2020{\natexlab{b}})Golatkar, Achille, and Soatto]{golatkar2020forgetting}
Aditya Golatkar, Alessandro Achille, and Stefano Soatto.
\newblock Forgetting outside the box: Scrubbing deep networks of information accessible from input-output observations.
\newblock In \emph{Proceedings of the European Conference on Computer Vision (ECCV)}, pp.\  383--398. Springer, 2020{\natexlab{b}}.

\bibitem[Goodfellow et~al.(2014)Goodfellow, Pouget{-}Abadie, Mirza, Xu, Warde{-}Farley, Ozair, Courville, and Bengio]{gan_goodfellow}
Ian~J. Goodfellow, Jean Pouget{-}Abadie, Mehdi Mirza, Bing Xu, David Warde{-}Farley, Sherjil Ozair, Aaron~C. Courville, and Yoshua Bengio.
\newblock Generative adversarial networks.
\newblock \emph{CoRR}, abs/1406.2661, 2014.

\bibitem[Graves et~al.(2021)Graves, Nagisetty, and Ganesh]{graves2021amnesiac}
Laura Graves, Vineel Nagisetty, and Vijay Ganesh.
\newblock Amnesiac machine learning.
\newblock In \emph{Proceedings of the AAAI Conference on Artificial Intelligence}, pp.\  11516--11524, 2021.

\bibitem[Gulrajani et~al.(2017)Gulrajani, Ahmed, Arjovsky, Dumoulin, and Courville]{gulrajani2017improved}
Ishaan Gulrajani, Faruk Ahmed, Mart{\'{\i}}n Arjovsky, Vincent Dumoulin, and Aaron~C. Courville.
\newblock Improved training of wasserstein gans.
\newblock In \emph{Advances in Neural Information Processing Systems}, pp.\  5767--5777, 2017.

\bibitem[Halimi et~al.(2022)Halimi, Kadhe, Rawat, and Baracaldo]{halimi2022federated}
Anisa Halimi, Swanand Kadhe, Ambrish Rawat, and Nathalie Baracaldo.
\newblock Federated unlearning: How to efficiently erase a client in fl?
\newblock \emph{CoRR}, abs/2207.05521, 2022.

\bibitem[He et~al.(2022)He, Chen, Xie, Li, Doll{\'{a}}r, and Girshick]{mae_masked_he}
Kaiming He, Xinlei Chen, Saining Xie, Yanghao Li, Piotr Doll{\'{a}}r, and Ross~B. Girshick.
\newblock Masked autoencoders are scalable vision learners.
\newblock In \emph{{IEEE/CVF} Conference on Computer Vision and Pattern Recognition, {CVPR} 2022, New Orleans, LA, USA, June 18-24, 2022}, pp.\  15979--15988. {IEEE}, 2022.

\bibitem[Heusel et~al.(2017)Heusel, Ramsauer, Unterthiner, Nessler, and Hochreiter]{heusel2017gans_fid}
Martin Heusel, Hubert Ramsauer, Thomas Unterthiner, Bernhard Nessler, and Sepp Hochreiter.
\newblock Gans trained by a two time-scale update rule converge to a local nash equilibrium.
\newblock In \emph{Advances in Neural Information Processing Systems}, pp.\  6626--6637, 2017.

\bibitem[Ho et~al.(2020)Ho, Jain, and Abbeel]{diffusion_original}
Jonathan Ho, Ajay Jain, and Pieter Abbeel.
\newblock Denoising diffusion probabilistic models.
\newblock In \emph{Advances in neural information processing systems}, pp.\  6840--6851, 2020.

\bibitem[Hoogeboom et~al.(2022)Hoogeboom, Gritsenko, Bastings, Poole, van~den Berg, and Salimans]{hoogeboom2021autoregressive}
Emiel Hoogeboom, Alexey~A. Gritsenko, Jasmijn Bastings, Ben Poole, Rianne van~den Berg, and Tim Salimans.
\newblock Autoregressive diffusion models.
\newblock In \emph{The Tenth International Conference on Learning Representations, {ICLR} 2022, Virtual Event, April 25-29, 2022}. OpenReview.net, 2022.

\bibitem[Jia et~al.(2023)Jia, Liu, Ram, Yao, Liu, Liu, Sharma, and Liu]{prunine_unlearn}
Jinghan Jia, Jiancheng Liu, Parikshit Ram, Yuguang Yao, Gaowen Liu, Yang Liu, Pranay Sharma, and Sijia Liu.
\newblock Model sparsification can simplify machine unlearning.
\newblock \emph{CoRR}, abs/2304.04934, 2023.

\bibitem[Karras et~al.(2019)Karras, Laine, and Aila]{karras2019style}
Tero Karras, Samuli Laine, and Timo Aila.
\newblock A style-based generator architecture for generative adversarial networks.
\newblock In \emph{{IEEE} Conference on Computer Vision and Pattern Recognition, {CVPR} 2019, Long Beach, CA, USA, June 16-20, 2019}, pp.\  4401--4410. Computer Vision Foundation / {IEEE}, 2019.

\bibitem[Karras et~al.(2020)Karras, Laine, Aittala, Hellsten, Lehtinen, and Aila]{karras2020analyzing}
Tero Karras, Samuli Laine, Miika Aittala, Janne Hellsten, Jaakko Lehtinen, and Timo Aila.
\newblock Analyzing and improving the image quality of stylegan.
\newblock In \emph{2020 {IEEE/CVF} Conference on Computer Vision and Pattern Recognition, {CVPR} 2020, Seattle, WA, USA, June 13-19, 2020}, pp.\  8107--8116. Computer Vision Foundation / {IEEE}, 2020.

\bibitem[Kingma \& Welling(2019)Kingma and Welling]{inverse_vae_1}
Diederik~P. Kingma and Max Welling.
\newblock An introduction to variational autoencoders.
\newblock \emph{Found. Trends Mach. Learn.}, 12\penalty0 (4):\penalty0 307--392, 2019.

\bibitem[Kong et~al.(2020)Kong, de~Masson~d'Autume, Yu, Ling, Dai, and Yogatama]{kong2019mutual}
Lingpeng Kong, Cyprien de~Masson~d'Autume, Lei Yu, Wang Ling, Zihang Dai, and Dani Yogatama.
\newblock A mutual information maximization perspective of language representation learning.
\newblock In \emph{8th International Conference on Learning Representations, {ICLR} 2020, Addis Ababa, Ethiopia, April 26-30, 2020}. OpenReview.net, 2020.

\bibitem[Kong \& Chaudhuri(2023)Kong and Chaudhuri]{mul_generative3}
Zhifeng Kong and Kamalika Chaudhuri.
\newblock Data redaction from conditional generative models.
\newblock \emph{CoRR}, abs/2305.11351, 2023.

\bibitem[Krishnan et~al.(2019)Krishnan, Teterwak, Sarna, Maschinot, Liu, Belanger, and Freeman]{teterwak2019boundless}
Dilip Krishnan, Piotr Teterwak, Aaron Sarna, Aaron Maschinot, Ce~Liu, David Belanger, and William~T. Freeman.
\newblock Boundless: Generative adversarial networks for image extension.
\newblock In \emph{2019 {IEEE/CVF} International Conference on Computer Vision, {ICCV} 2019, Seoul, Korea (South), October 27 - November 2, 2019}, pp.\  10520--10529. {IEEE}, 2019.

\bibitem[Kuppa et~al.(2021)Kuppa, Aouad, and Le{-}Khac]{data_privacy_generate}
Aditya Kuppa, Lamine~M. Aouad, and Nhien{-}An Le{-}Khac.
\newblock Towards improving privacy of synthetic datasets.
\newblock In \emph{Privacy Technologies and Policy - 9th Annual Privacy Forum, {APF} 2021, Oslo, Norway, June 17-18, 2021, Proceedings}, volume 12703 of \emph{Lecture Notes in Computer Science}, pp.\  106--119. Springer, 2021.

\bibitem[Kurmanji et~al.(2023)Kurmanji, Triantafillou, and Triantafillou]{towards_unbounded}
Meghdad Kurmanji, Peter Triantafillou, and Eleni Triantafillou.
\newblock Towards unbounded machine unlearning.
\newblock \emph{CoRR}, abs/2302.09880, 2023.

\bibitem[Li et~al.(2023)Li, Chang, Mishra, Zhang, Katabi, and Krishnan]{vqgan_mage}
Tianhong Li, Huiwen Chang, Shlok~Kumar Mishra, Han Zhang, Dina Katabi, and Dilip Krishnan.
\newblock {MAGE:} masked generative encoder to unify representation learning and image synthesis.
\newblock In \emph{{IEEE/CVF} Conference on Computer Vision and Pattern Recognition, {CVPR} 2023, Vancouver, BC, Canada, June 17-24, 2023}, pp.\  2142--2152. {IEEE}, 2023.

\bibitem[Lukas et~al.(2023)Lukas, Salem, Sim, Tople, Wutschitz, and B{\'{e}}guelin]{data_leak}
Nils Lukas, Ahmed Salem, Robert Sim, Shruti Tople, Lukas Wutschitz, and Santiago~Zanella B{\'{e}}guelin.
\newblock Analyzing leakage of personally identifiable information in language models.
\newblock In \emph{44th {IEEE} Symposium on Security and Privacy, {SP} 2023, San Francisco, CA, USA, May 21-25, 2023}, pp.\  346--363. {IEEE}, 2023.

\bibitem[Moon et~al.(2023)Moon, Cho, and Kim]{mul_generative2}
Saemi Moon, Seunghyuk Cho, and Dongwoo Kim.
\newblock Feature unlearning for generative models via implicit feedback.
\newblock \emph{CoRR}, abs/2303.05699, 2023.

\bibitem[Neel et~al.(2021)Neel, Roth, and Sharifi{-}Malvajerdi]{neel2021descent}
Seth Neel, Aaron Roth, and Saeed Sharifi{-}Malvajerdi.
\newblock Descent-to-delete: Gradient-based methods for machine unlearning.
\newblock In \emph{Algorithmic Learning Theory, 16-19 March 2021, Virtual Conference, Worldwide}, volume 132 of \emph{Proceedings of Machine Learning Research}, pp.\  931--962. {PMLR}, 2021.

\bibitem[Nguyen et~al.(2022)Nguyen, Huynh, Nguyen, Liew, Yin, and Nguyen]{unlearn_survey}
Thanh~Tam Nguyen, Thanh~Trung Huynh, Phi~Le Nguyen, Alan~Wee{-}Chung Liew, Hongzhi Yin, and Quoc Viet~Hung Nguyen.
\newblock A survey of machine unlearning.
\newblock \emph{CoRR}, abs/2209.02299, 2022.

\bibitem[Parliament(2019)]{cananda_bill}
Canada Parliament.
\newblock The personal information protection and electronic documents act (pipeda), 2019.
\newblock URL \url{https://laws-lois.justice.gc.ca/PDF/P-8.6.pdf}.

\bibitem[Parliament \& Council(2016)Parliament and Council]{eu_gdpr}
European Parliament and European~Union Council.
\newblock Regulation (eu) 2016/679 of the european parliament and of the council of 27 april 2016 on the protection of natural persons with regard to the processing of personal data and on the free movement of such data, and repealing directive 95/46/ec (general data protection regulation) (text with eea relevance), 2016.
\newblock URL \url{https://eur-lex.europa.eu/eli/reg/2016/679/oj}.

\bibitem[Paszke et~al.(2019)Paszke, Gross, Massa, Lerer, Bradbury, Chanan, Killeen, Lin, Gimelshein, Antiga, Desmaison, K{\"{o}}pf, Yang, DeVito, Raison, Tejani, Chilamkurthy, Steiner, Fang, Bai, and Chintala]{paszke2017automatic_torch}
Adam Paszke, Sam Gross, Francisco Massa, Adam Lerer, James Bradbury, Gregory Chanan, Trevor Killeen, Zeming Lin, Natalia Gimelshein, Luca Antiga, Alban Desmaison, Andreas K{\"{o}}pf, Edward~Z. Yang, Zachary DeVito, Martin Raison, Alykhan Tejani, Sasank Chilamkurthy, Benoit Steiner, Lu~Fang, Junjie Bai, and Soumith Chintala.
\newblock Pytorch: An imperative style, high-performance deep learning library.
\newblock In \emph{Advances in Neural Information Processing Systems}, pp.\  8024--8035, 2019.

\bibitem[Poole et~al.(2019)Poole, Ozair, Van Den~Oord, Alemi, and Tucker]{poole2019variational}
Ben Poole, Sherjil Ozair, Aaron Van Den~Oord, Alex Alemi, and George Tucker.
\newblock On variational bounds of mutual information.
\newblock In \emph{International Conference on Machine Learning}, pp.\  5171--5180. PMLR, 2019.

\bibitem[Radford et~al.(2021)Radford, Kim, Hallacy, Ramesh, Goh, Agarwal, Sastry, Askell, Mishkin, Clark, Krueger, and Sutskever]{clip}
Alec Radford, Jong~Wook Kim, Chris Hallacy, Aditya Ramesh, Gabriel Goh, Sandhini Agarwal, Girish Sastry, Amanda Askell, Pamela Mishkin, Jack Clark, Gretchen Krueger, and Ilya Sutskever.
\newblock Learning transferable visual models from natural language supervision.
\newblock In \emph{International Conference on Machine Learning}, pp.\  8748--8763. {PMLR}, 2021.

\bibitem[Rombach et~al.(2022)Rombach, Blattmann, Lorenz, Esser, and Ommer]{latent_diffusion}
Robin Rombach, Andreas Blattmann, Dominik Lorenz, Patrick Esser, and Bj{\"{o}}rn Ommer.
\newblock High-resolution image synthesis with latent diffusion models.
\newblock In \emph{{IEEE/CVF} Conference on Computer Vision and Pattern Recognition, {CVPR} 2022, New Orleans, LA, USA, June 18-24, 2022}, pp.\  10674--10685. {IEEE}, 2022.

\bibitem[Saharia et~al.(2022{\natexlab{a}})Saharia, Chan, Chang, Lee, Ho, Salimans, Fleet, and Norouzi]{palette_diff_base}
Chitwan Saharia, William Chan, Huiwen Chang, Chris~A. Lee, Jonathan Ho, Tim Salimans, David~J. Fleet, and Mohammad Norouzi.
\newblock Palette: Image-to-image diffusion models.
\newblock In \emph{{SIGGRAPH} '22: Special Interest Group on Computer Graphics and Interactive Techniques Conference, Vancouver, BC, Canada, August 7 - 11, 2022}, pp.\  15:1--15:10. {ACM}, 2022{\natexlab{a}}.

\bibitem[Saharia et~al.(2022{\natexlab{b}})Saharia, Chan, Saxena, Li, Whang, Denton, Ghasemipour, Lopes, Ayan, Salimans, Ho, Fleet, and Norouzi]{t2i_diff_beat_gan}
Chitwan Saharia, William Chan, Saurabh Saxena, Lala Li, Jay Whang, Emily~L. Denton, Seyed Kamyar~Seyed Ghasemipour, Raphael~Gontijo Lopes, Burcu~Karagol Ayan, Tim Salimans, Jonathan Ho, David~J. Fleet, and Mohammad Norouzi.
\newblock Photorealistic text-to-image diffusion models with deep language understanding.
\newblock In \emph{Advances in Neural Information Processing Systems}, pp.\  36479--36494, 2022{\natexlab{b}}.

\bibitem[Salimans \& Ho(2022)Salimans and Ho]{salimans2021progressive}
Tim Salimans and Jonathan Ho.
\newblock Progressive distillation for fast sampling of diffusion models.
\newblock In \emph{The Tenth International Conference on Learning Representations, {ICLR} 2022, Virtual Event, April 25-29, 2022}. OpenReview.net, 2022.

\bibitem[Salimans et~al.(2016)Salimans, Goodfellow, Zaremba, Cheung, Radford, and Chen]{inception_score}
Tim Salimans, Ian~J. Goodfellow, Wojciech Zaremba, Vicki Cheung, Alec Radford, and Xi~Chen.
\newblock Improved techniques for training gans.
\newblock In \emph{Advances in Neural Information Processing Systems}, pp.\  2226--2234, 2016.

\bibitem[Schuhmann et~al.(2022)Schuhmann, Beaumont, Vencu, Gordon, Wightman, Cherti, Coombes, Katta, Mullis, Wortsman, Schramowski, Kundurthy, Crowson, Schmidt, Kaczmarczyk, and Jitsev]{laion_5b}
Christoph Schuhmann, Romain Beaumont, Richard Vencu, Cade Gordon, Ross Wightman, Mehdi Cherti, Theo Coombes, Aarush Katta, Clayton Mullis, Mitchell Wortsman, Patrick Schramowski, Srivatsa Kundurthy, Katherine Crowson, Ludwig Schmidt, Robert Kaczmarczyk, and Jenia Jitsev.
\newblock {LAION-5B:} an open large-scale dataset for training next generation image-text models.
\newblock In \emph{Advances in Neural Information Processing Systems}, 2022.

\bibitem[Somepalli et~al.(2023)Somepalli, Singla, Goldblum, Geiping, and Goldstein]{diff_remember}
Gowthami Somepalli, Vasu Singla, Micah Goldblum, Jonas Geiping, and Tom Goldstein.
\newblock Diffusion art or digital forgery? investigating data replication in diffusion models.
\newblock In \emph{{IEEE/CVF} Conference on Computer Vision and Pattern Recognition, {CVPR} 2023, Vancouver, BC, Canada, June 17-24, 2023}, pp.\  6048--6058. {IEEE}, 2023.

\bibitem[Song et~al.(2023)Song, Kwon, Zhang, Hu, Qu, and Shen]{inverse_diff_2}
Bowen Song, Soo~Min Kwon, Zecheng Zhang, Xinyu Hu, Qing Qu, and Liyue Shen.
\newblock Solving inverse problems with latent diffusion models via hard data consistency.
\newblock \emph{CoRR}, abs/2307.08123, 2023.

\bibitem[Song \& Ermon(2020)Song and Ermon]{song2020improved}
Yang Song and Stefano Ermon.
\newblock Improved techniques for training score-based generative models.
\newblock In \emph{Advances in Neural Information Processing Systems}, pp.\  12438--12448, 2020.

\bibitem[Sun et~al.(2023)Sun, Zhu, Chang, and Zhou]{mul_generative4}
Hui Sun, Tianqing Zhu, Wenhan Chang, and Wanlei Zhou.
\newblock Generative adversarial networks unlearning.
\newblock \emph{CoRR}, abs/2308.09881, 2023.

\bibitem[Tarun et~al.(2023{\natexlab{a}})Tarun, Chundawat, Mandal, and Kankanhalli]{tarun2023fast}
Ayush~K Tarun, Vikram~S Chundawat, Murari Mandal, and Mohan Kankanhalli.
\newblock Fast yet effective machine unlearning.
\newblock \emph{IEEE Transactions on Neural Networks and Learning Systems}, 2023{\natexlab{a}}.

\bibitem[Tarun et~al.(2023{\natexlab{b}})Tarun, Chundawat, Mandal, and Kankanhalli]{tarun2023deep}
Ayush~Kumar Tarun, Vikram~Singh Chundawat, Murari Mandal, and Mohan~S. Kankanhalli.
\newblock Deep regression unlearning.
\newblock In \emph{International Conference on Machine Learning}, pp.\  33921--33939. {PMLR}, 2023{\natexlab{b}}.

\bibitem[Tirumala et~al.(2022)Tirumala, Markosyan, Zettlemoyer, and Aghajanyan]{tirumala2022memorization}
Kushal Tirumala, Aram~H. Markosyan, Luke Zettlemoyer, and Armen Aghajanyan.
\newblock Memorization without overfitting: Analyzing the training dynamics of large language models.
\newblock In \emph{Advances in Neural Information Processing Systems}, pp.\  38274--38290, 2022.

\bibitem[Tong et~al.(2022)Tong, Song, Wang, and Wang]{tong2022videomae}
Zhan Tong, Yibing Song, Jue Wang, and Limin Wang.
\newblock Videomae: Masked autoencoders are data-efficient learners for self-supervised video pre-training.
\newblock In \emph{Advances in Neural Information Processing Systems}, volume~35, pp.\  10078--10093, 2022.

\bibitem[Wallace et~al.(2023)Wallace, Gokul, and Naik]{inverse_diff_1}
Bram Wallace, Akash Gokul, and Nikhil Naik.
\newblock {EDICT:} exact diffusion inversion via coupled transformations.
\newblock In \emph{{IEEE/CVF} Conference on Computer Vision and Pattern Recognition, {CVPR} 2023, Vancouver, BC, Canada, June 17-24, 2023}, pp.\  22532--22541. {IEEE}, 2023.

\bibitem[Warnecke et~al.(2023)Warnecke, Pirch, Wressnegger, and Rieck]{unlearn_feat_labels}
Alexander Warnecke, Lukas Pirch, Christian Wressnegger, and Konrad Rieck.
\newblock Machine unlearning of features and labels.
\newblock In \emph{30th Annual Network and Distributed System Security Symposium, {NDSS} 2023, San Diego, California, USA, February 27 - March 3, 2023}. The Internet Society, 2023.

\bibitem[Wu et~al.(2020)Wu, Zhuang, Mosse, Yamins, and Goodman]{wu2020mutual}
Mike Wu, Chengxu Zhuang, Milan Mosse, Daniel Yamins, and Noah Goodman.
\newblock On mutual information in contrastive learning for visual representations.
\newblock \emph{arXiv preprint arXiv:2005.13149}, 2020.

\bibitem[Wu et~al.(2017)Wu, Burda, Salakhutdinov, and Grosse]{iclr_gan_memory}
Yuhuai Wu, Yuri Burda, Ruslan Salakhutdinov, and Roger~B. Grosse.
\newblock On the quantitative analysis of decoder-based generative models.
\newblock In \emph{5th International Conference on Learning Representations, {ICLR} 2017, Toulon, France, April 24-26, 2017, Conference Track Proceedings}, 2017.

\bibitem[Xia et~al.(2023)Xia, Zhang, Yang, Xue, Zhou, and Yang]{inverse_gan_1}
Weihao Xia, Yulun Zhang, Yujiu Yang, Jing{-}Hao Xue, Bolei Zhou, and Ming{-}Hsuan Yang.
\newblock {GAN} inversion: {A} survey.
\newblock \emph{IEEE Transactions on Pattern Analysis and Machine Intelligence}, 45\penalty0 (3):\penalty0 3121--3138, 2023.

\bibitem[Xu et~al.(2023)Xu, Zhu, Zhang, Zhou, and Yu]{mul_survey2023}
Heng Xu, Tianqing Zhu, Lefeng Zhang, Wanlei Zhou, and Philip~S. Yu.
\newblock Machine unlearning: A survey.
\newblock \emph{ACM Comput. Surv.}, 56\penalty0 (1), aug 2023.

\bibitem[Yang et~al.(2022)Yang, Zhang, Song, Hong, Xu, Zhao, Shao, Zhang, Yang, and Cui]{diff_survey}
Ling Yang, Zhilong Zhang, Yang Song, Shenda Hong, Runsheng Xu, Yue Zhao, Yingxia Shao, Wentao Zhang, Ming{-}Hsuan Yang, and Bin Cui.
\newblock Diffusion models: {A} comprehensive survey of methods and applications.
\newblock \emph{CoRR}, abs/2209.00796, 2022.

\bibitem[Zhang \& Agrawala(2023)Zhang and Agrawala]{zhang2023adding}
Lvmin Zhang and Maneesh Agrawala.
\newblock Adding conditional control to text-to-image diffusion models.
\newblock \emph{CoRR}, abs/2302.05543, 2023.

\bibitem[Zhang et~al.(2023{\natexlab{a}})Zhang, Wang, Cheng, Sun, Zhang, and Xiao]{unlearn_teach_stu1}
Xulong Zhang, Jianzong Wang, Ning Cheng, Yifu Sun, Chuanyao Zhang, and Jing Xiao.
\newblock Machine unlearning methodology based on stochastic teacher network.
\newblock In \emph{Advanced Data Mining and Applications - 19th International Conference, {ADMA} 2023, Shenyang, China, August 21-23, 2023, Proceedings, Part {V}}, volume 14180 of \emph{Lecture Notes in Computer Science}, pp.\  250--261. Springer, 2023{\natexlab{a}}.

\bibitem[Zhang et~al.(2023{\natexlab{b}})Zhang, Lu, Zhang, Wang, and Li]{unlearn_teach_stu2}
Yongjing Zhang, Zhaobo Lu, Feng Zhang, Hao Wang, and Shaojing Li.
\newblock Machine unlearning by reversing the continual learning.
\newblock \emph{Applied Sciences}, 13\penalty0 (16):\penalty0 9341, 2023{\natexlab{b}}.

\bibitem[Zhou et~al.(2017)Zhou, Lapedriza, Khosla, Oliva, and Torralba]{zhou2017places365}
Bolei Zhou, Agata Lapedriza, Aditya Khosla, Aude Oliva, and Antonio Torralba.
\newblock Places: A 10 million image database for scene recognition.
\newblock \emph{IEEE Transactions on Pattern Analysis and Machine Intelligence}, 40\penalty0 (6):\penalty0 1452--1464, 2017.
\newblock URL \url{https://github.com/CSAILVision/places365}.

\bibitem[Zhu et~al.(2017)Zhu, Park, Isola, and Efros]{cyclegan}
Jun{-}Yan Zhu, Taesung Park, Phillip Isola, and Alexei~A. Efros.
\newblock Unpaired image-to-image translation using cycle-consistent adversarial networks.
\newblock In \emph{{IEEE} International Conference on Computer Vision, {ICCV} 2017, Venice, Italy, October 22-29, 2017}, pp.\  2242--2251. {IEEE} Computer Society, 2017.

\end{thebibliography}
\bibliographystyle{iclr2024_conference}

\newpage
\appendix

\section{Infinitely many probability with infinite KL-divergence}\label{app:sol_space_kl}
\def\theequation{A.\arabic{equation}}
\def\thelem{A.\arabic{lem}}
\def\thefigure{A.\arabic{figure}}
\def\thetable{A.\arabic{table}}
In Section~\ref{sec:optimzation_sets}, we mention that there are infinitely many probability distributions that have infinity KL divergence with a given distribution. We provide the proof below:
\begin{prop}
There are infinitely many probability distributions that have a positively infinite value of KL-divergence with any general discrete probability distribution $P(X)$ that is defined as follows:
$$0<=P(X=i)<1,\quad \sum_{i=1}^N P(X=i)=1,\quad i\in[N],\quad N\geq2$$
\end{prop}

\textit{Proof.} Based on $P(X)$, we build another distribution $Q(X)$ as follows:
\[
Q(X=i)=\begin{cases}
\frac{P(X=i)}{1-P(X=j)-P(X=k)+\kappa},& \text{if } i\neq j \text{ and } i\neq k\\
0, & \text{if } i=j
\\
\frac{\kappa}{1-P(X=j)-P(X=k)+\kappa}, & \text{if } i=k
\end{cases}
\quad j,k\in[N],\ \kappa>0
\]
where $j$ satisfies $P(X=j)>0$ and $k$ sastifies $j\neq k$. 
Clearly, $0<=Q(X=i)<1$ and $\sum_{i=1}^N Q(X=i)=1$. Therefore, $Q(X)$ is a valid probability distribution. 

We next compute the KL divergence between $P$ and $Q$.
\begin{equation*}
\begin{aligned}
D(P||Q)&=\sum_{i=1}^N P(X=i)\mathrm{log}\frac{P(X=i)}{Q(X=i)}\\
&= P(X=j)\mathrm{log}\left(\frac{P(X=j)}{Q(X=j)}\right)+\sum_{i\in[N], i\neq j} P(X=i)\mathrm{log}\left(\frac{P(X=i)}{Q(X=i)}\right)\\
&= P(X=j)\mathrm{log}\left(\frac{P(X=j)}{0}\right)+\sum_{i\in[N], i\neq j} P(X=i)\mathrm{log}\left(\frac{P(X=i)}{Q(X=i)}\right)\\
&=+\infty
\end{aligned}
\end{equation*}
We note that $\kappa$ can be any positive real number; hence, we can obtain infinitely many $Q(X)$ by varying the value of $\kappa$. Hence, there are infinitely many $Q$ that have a positively infinite value of  KL-divergence with $P$. In general, one can replace the set $[N]$ with any discrete set. This completes our proof. \hfill\(\Box\)

The proof for the continuous distribution is very similar to the discrete case as shown below. 
\begin{prop}
There are infinitely many probability distributions that have the a positively infinite value of  KL-divergence with any general continuous probability distribution with the following probability density function (PDF) $p(x)$:
$$ p(x)\geq0,\quad x\in\mathcal{S},\quad \int_{x\in\mathcal{S}}p(x)dx=1$$
\end{prop}

\textit{Proof.} Based on $p(x)$, we build another distribution with PDF $q(x)$ as follows:
\[
q(x)=\begin{cases}
\frac{p(x)}{1-\int_{x\in\mathcal{S}_1}p(x)},& \text{if } x\in\mathcal{S}\backslash (\mathcal{S}_1)\\
0, & \text{if } x\in\mathcal{S}_1\\
\end{cases}
,\quad\mathcal{S}_1\subset\mathcal{S}
\]
where $\mathcal{S}_1$ satisfies $0<\int_{x\in\mathcal{S}_1}p(x)dx<1$. 
Clearly, $q(x)\geq0$ and $\int_{x\in\mathcal{S}}q(x)dx=1$. Therefore, $q(x)$ is a valid probability density function. 

We next compute the KL divergence between $p$ and $q$.
\begin{equation*}
\begin{aligned}
D(p||q)&=\int_{x\in\mathcal{S}}p(x)\mathrm{log}\left(\frac{p(x)}{q(x)}\right)dx\\
&=\int_{x\in\mathcal{S}_1}p(x)\mathrm{log}\left(\frac{p(x)}{q(x)}\right)dx+\int_{x\in\mathcal{S}\backslash\mathcal{S}_1}p(x)\mathrm{log}\left(\frac{p(x)}{q(x)}\right)dx\\
&=\int_{x\in\mathcal{S}_1}p(x)\mathrm{log}\left(\frac{p(x)}{0}\right)dx+\int_{x\in\mathcal{S}\backslash\mathcal{S}_1}p(x)\mathrm{log}\left(\frac{p(x)}{q(x)}\right)dx\\
&=+\infty
\end{aligned}
\end{equation*}
We note that given a continuous distribution, there are infinitely many possible $\mathcal{S}_1$; hence, we can obtain infinitely many $q(x)$ by using different $\mathcal{S}_1$. Hence, there are infinitely many $q$ that have a positively infinite value of  KL-divergence with $p$. This completes our proof. \hfill\(\Box\)

\section{Proof of Theorem~\ref{theorem:infonce}}\label{app:prove_infonce}
\def\theequation{B.\arabic{equation}}
\def\thelem{B.\arabic{lem}}
\def\thefigure{B.\arabic{figure}}
\def\thetable{B.\arabic{table}}
\textit{Proof.}
Directly computing mutual information (MI) between two random variables is not feasible. Fortunately, there are some variational MI bounds that are relatively easy to compute. A popular approach is to lower bound MI with InfoNCE~\citep{mine,wu2020mutual,poole2019variational,kong2019mutual}. Given two random variables $X_1$ and $X_2$, MI is bounded by InfoNCE is defined as follows:
\begin{equation}\label{eq:mi_infonce}
 I(X_1;X_2)\geq\infonce(X_1;X_2)= \mathrm{log}\left(K\right)-\mathbb{E}\left[\sum_{i=1}^{K}\frac{1}{K}\mathrm{log}\frac{e^{g_1(x_{1_i})^Tg_2(x_{2_i})}}{\sum_{j=1}^{K}e^{e^{g_1(x_{1_i})^Tg_2(x_{2_j})}}}\right]\\
\end{equation}
where the expectation is over $K$ independent samples from the joint distribution: $\Pi_j p(x_{1_j} , x_{2_j} )$; $g_1$ and $g_2$ are two functions that map the random variables into representation vectors, \eg, the encoder network $\encoder$ of I2I  generative models. 

Following the standard practice of InfoNCE, we use the inner product of the encoder output vectors as the critic function. Therefore, the InfoNCE between the ground truth images $X$ and reconstructed images $\hat{X}$  is written as follows:
\begin{equation}
 \infonce(X;\hat{X})=\mathrm{log}\left(K\right)-\mathbb{E}_{x,\hat{x} }\left[\frac{1}{K}\mathrm{log}\frac{e^{\encoderp\left({x}_{i}\right)^T\encoderp\left(\hat{x}_{i}\right)}}{\sum_{j=1}^{K}e^{\encoderp\left({x}_{i}\right)^T\encoderp\left(\hat{x}_{j}\right)}}\right]
 \label{eq:raw_infonce}
\end{equation}

Multiple works show that the encoder of the I2I generative models can be treated as the inverse function of the decoder~\citep{inverse_diff_1,inverse_diff_2,inverse_gan_1,inverse_vae_1}. In other words, 
\begin{equation}\label{eq:perfect_inverse}
\encoderp\left(\decoderp\left(z\right)\right)= z
\end{equation}
Given the assumption that the target model use the same decoder as the original model, we can express $X$ and $\hat{X}$ as follows:
\begin{equation}\label{eq:express_image}
    x=\decoderp\left(\encoderp(\trans(x)))\right),\quad \hat{x}=\decoderp\left(\encoder(\trans(x)))\right)
\end{equation}
By using the above relationships in Eq.~\eref{eq:perfect_inverse} and Eq.~\eref{eq:express_image}, we have 

\begin{equation}
\begin{aligned} 
\encoderp\left(x\right)^T\encoderp\left(\hat{x}\right)&=\encoderp\left(\decoderp\left(\encoderp\left(\trans\left(x\right)\right)\right)\right)^T\encoderp\left(\decoderp\left(\encoder\left(\trans\left(\hat{x}\right)\right)\right)\right)\\
 &= \encoderp\left(\trans\left(x_1\right)\right)^T\encoder\left(\trans\left(x_2\right)\right)
\end{aligned}\label{eq:raw_to_embedding}
\end{equation}

\textbf{Retain set} 
Recall the assumption that the output of the encoder is normalized, i.e., $\left\rVert\encoder\left(x\right)\right\rVert_2=1$ and $\left\rVert\encoderp\left(x\right)\right\rVert_2=1$. Therefore, we can rewrite Eq.~\eref{eq:raw_to_embedding} for the ground truth retain samples $\{x_{r_i}\}$ and their reconstructions $\{\hat{x}_{r_i}\}$ and as follows: 
\begin{equation}\label{eq:x_i_eps_i}
\begin{split}
\encoderp\left(x_{r_i}\right)^T\encoderp\left(\hat{x}_{r_i}\right)&= \encoderp\left(\trans\left(x_{r_i}\right)\right)^T\encoder\left(\trans\left(x_{r_i}\right)\right)\\
&= \frac{1}{2}\left\rVert\encoderp\left(\trans\left(x_{r_i}\right)\right)\right\rVert_2^2+ \frac{1}{2} \left\rVert\encoder\left(\trans\left(x_{r_i}\right)\right)\right\rVert_2^2\\
&\quad - \frac{1}{2}\left\rVert\encoderp\left(\trans\left(x_{r_i}\right)\right)-\encoder\left(\trans\left(x_{r_i}\right)\right)\right\rVert_2^2\\
&= \frac{1}{2}\left(2-\left\rVert\encoderp\left(\trans\left(x_{r_i}\right)\right)-\encoder\left(\trans\left(x_{r_i}\right)\right)\right\rVert_2^2\right)\\
&= \frac{1}{2}\left(2-\epsilon_i^2\right)=1-\frac{\epsilon_i^2}{2}
\end{split}
\end{equation}

where $\epsilon_i=\left\rVert\encoderp\left(\trans\left(x_{r_i}\right)\right)-\encoder\left(\trans\left(x_{r_i}\right)\right)\right\rVert_2$ is the $L_2$ loss between the representation of target model encoder $\encoder$ and original model encoder $\encoderp$. We then bound the $\encoderp\left(x_{r_i}\right)^T\encoderp\left(\hat{x}_{r_j}\right)$ as follows:
\begin{equation}
\begin{split}
 \encoderp\left(x_{r_i}\right)^T\encoderp\left(\hat{x}_{r_j}\right) =& \encoderp\left(\trans\left(x_{r_i}\right)\right)^T\encoder\left(\trans\left(x_{r_j}\right)\right)\\
 =&\encoderp\left(\trans\left(x_{r_i}\right)\right)^T\left(\encoder\left(\trans\left(x_{r_j}\right)\right) - \encoderp\left(\trans\left(x_{r_j}\right)\right)\right)\\
 &\ +\encoderp\left(\trans\left(x_{r_i}\right)\right)^T\encoderp\left(\trans\left(x_{r_j}\right)\right)\\
 =&\encoderp\left(\trans\left(x_{r_i}\right)\right)^T\left(\encoder\left(\trans\left(x_{r_j}\right)\right) - \encoderp\left(\trans\left(x_{r_j}\right)\right)\right)+R_{ij}\\
 \leq&\left\rVert\encoderp\left(\trans\left(x_{r_i}\right)\right)\right\rVert_2 * \left\rVert\encoder\left(\trans\left(x_{r_j}\right)\right) - \encoderp\trans\left(x_{r_j}\right)\right\rVert_2+R_{ij}\\
 =&1* \epsilon_j+R_{ij}\\
 =& \epsilon_j+R_{ij}\\
\end{split}\label{eq:c_ij_xij}
\end{equation}
where $R_{ij}= \encoderp\left(\trans\left(x_{r_i}\right)\right)^T\encoderp\left(\trans\left(x_{r_j}\right)\right)$ and the `$\leq$' comes from the Cauchy–Schwarz inequality. The above bound is tight if $\{\epsilon_i=0,i\in[K]\}$. 
By combining the Eq.~\eref{eq:raw_infonce}, Eq.~\eref{eq:x_i_eps_i}, and Eq.~\eref{eq:c_ij_xij}, we can link the InfoNCE loss with the $L_2$ loss:
\begin{equation}
\begin{split}
 \infonce(X_r;\hat{X}_r)&=\mathrm{log}\left(K\right)-\mathbb{E}\left[\frac{1}{K}\mathrm{log}\frac{e^{\encoderp\left({x}_{i}\right)^T\encoderp\left(\hat{x}_{i}\right)}}{\sum_{j=1}^{K}e^{\encoderp\left({x}_{i}\right)^T\encoderp\left(\hat{x}_{j}\right)}}\right]\\
 &=\mathrm{log}\left(K\right)-\mathbb{E}\left[\frac{1}{K}\mathrm{log}\frac{e^{1-\frac{\epsilon_i^2}{2}}}{\sum_{j=1}^{K}e^{\encoderp\left({x}_{i}\right)^T\encoderp\left(\hat{x}_{j}\right)}}\right]\\
 &\geq\mathrm{log}\left(K\right)-\mathbb{E}\left[\frac{1}{K}\mathrm{log}\frac{e^{1-\frac{\epsilon_i^2}{2}}}{\sum_{j=1}^{K}e^{\epsilon_j+R_{ij}}}\right]\\
 &=\mathrm{log}\left(K\right)-\mathbb{E}\left[\sum_{i=1}^{K}\frac{1}{K}\mathrm{log}\left(e^{\frac{\epsilon_i^2}{2}-1}\sum_{j=1}^{K}e^{\epsilon_j+R_{ij}}\right)\right]
\end{split}
\end{equation}
 
By combining Eq.~\eref{eq:mi_infonce}, we obtain the results on the retain set:
 \begin{equation}
 \begin{aligned}
 I(X_r;\hat{X_r})\geq\mathrm{log}\left(K\right)-\mathbb{E}\left[\sum_{i=1}^{K}\frac{1}{K}\mathrm{log}\left(e^{\frac{\epsilon_i^2}{2}-1}\sum_{j=1}^{K}e^{\epsilon_j+R_{ij}}\right)\right]\\
 \end{aligned}
 \end{equation}

\textbf{Forget set} The proof on the forget set is very similar to the retain set. By adapting Eq.~\eref{eq:x_i_eps_i} and Eq.~\eref{eq:c_ij_xij} for the forget set, we first calculate the inner product of the embedding vector between $n$ and $x_f$:
\begin{equation}
\begin{split}
\encoderp\left(n_i\right)^T\encoderp\left(\hat{x}_{f_i}\right)&= \encoderp\left(\trans\left(n_i\right)\right)^T\encoder\left(\trans\left({x}_{f_i}\right)\right)\\
&= \frac{1}{2}\left(\left\rVert\encoderp\left(\trans\left(n_i\right)\right)\right\rVert_2^2+\left\rVert\encoder\left(\trans\left({x}_{f_i}\right)\right)\right\rVert_2^2-\left\rVert\encoderp\left(\trans\left(n_i\right)\right)-\encoder\left(\trans\left({x}_{f_i}\right)\right)\right\rVert_2^2\right)\\
&= \frac{1}{2}\left(2-\left\rVert\encoderp\left(\trans\left(n_i\right)\right)-\encoder\left(\trans\left({x}_{f_i}\right)\right)\right\rVert_2^2\right)\\
&= \frac{1}{2}\left(2-\epsilon_i^2\right)=1-\frac{\delta_i^2}{2}
\end{split}
\end{equation}
and 
\begin{equation}
\begin{split}
\encoderp\left(n_i\right)^T\encoderp\left(\hat{x}_{f_j}\right)=& \encoderp\left(\trans\left(n_i\right)\right)^T\encoder\left(\trans\left({x}_{f_j}\right)\right)\\
 =&\encoderp\left(\trans\left(n_i\right)\right)^T\left(\encoder\left(\trans\left({x}_{f_j}\right)\right) - \encoderp\left(\trans\left(n_j\right)\right)\right)\\
 &\ +\encoderp\left(\trans\left(n_i\right)\right)^T\encoderp\left(\trans\left(n_j\right)\right)\\
 =&\left\rVert \encoderp\left(\trans\left(n_i\right)\right)\right\rVert_2 * \left\rVert\encoder\left(\trans\left({x}_{f_j}\right)\right) - \encoderp\trans\left(n_j\right)\right\rVert_2 +F_{ij}\\
 =& 1* \delta_j+F_{ij}\\
 =& \delta_j+F_{ij}\\
\end{split}
\end{equation}
where $\delta_i=\left\rVert\encoderp\left(\trans\left(n_i\right)\right)-\encoder\left(\trans\left({x}_{f_i}\right)\right)\right\rVert_2$ and $F_{ij}=\encoderp\left(\trans\left(n_i\right)\right)^T\encoderp\left(\trans\left(n_j\right)\right)$. $x_{f_i}$ are the data samples in the forget set and $n_i\sim\mathcal{N}(0,\Sigma)$. Combining the above two equation with Eq.~\eref{eq:raw_infonce}:

\begin{equation}
\begin{split}
 \infonce(n;\hat{X_f})&=\mathrm{log}\left(K\right)-\mathbb{E}\left[\frac{1}{K}\mathrm{log}\frac{e^{\encoderp\left(n_i\right)^T\encoderp\left(\hat{x}_{f_i}\right)}}{\sum_{j=1}^{K}e^{\encoderp\left({n}_{i}\right)^T\encoderp\left(\hat{x}_{f_j}\right)}}\right]\\
 &=\mathrm{log}\left(K\right)-\mathbb{E}\left[\frac{1}{K}\mathrm{log}\frac{e^{1-\frac{\delta_i^2}{2}}}{\sum_{j=1}^{K}e^{\encoderp\left({n}_{i}\right)^T\encoderp\left(\hat{x}_{f_j}\right)}}\right]\\
 &\geq\mathrm{log}\left(K\right)-\mathbb{E}\left[\frac{1}{K}\mathrm{log}\frac{e^{1-\frac{\delta_i^2}{2}}}{\sum_{j=1}^{K}e^{\delta_j+F_{ij}}}\right]\\
 &=\mathrm{log}\left(K\right)-\mathbb{E}\left[\sum_{i=1}^{K}\frac{1}{K}\mathrm{log}\left(e^{\frac{\delta_i^2}{2}-1}\sum_{j=1}^{K}e^{\delta_j+F_{ij}}\right)\right]
\end{split}
\end{equation}

By combining the above equation with Eq.~\eref{eq:mi_infonce}, we obtain the results for the forget set:
\begin{equation}
 \begin{aligned}
 I(n;\hat{X_f})\geq\mathrm{log}\left(K\right)-\mathbb{E}\left[\sum_{i=1}^{K}\frac{1}{K}\mathrm{log}\left(e^{\frac{\delta_i^2}{2}-1}\sum_{j=1}^{K}e^{\delta_j+F_{ij}}\right)\right]\\
 \end{aligned}
\end{equation}

This completes our proof.\hfill$\Box$

\section{Implementation Details}\label{app:details}
\def\theequation{C.\arabic{equation}}
\def\thelem{C.\arabic{lem}}
\def\thefigure{C.\arabic{figure}}
\def\thetable{C.\arabic{table}}

\subsection{Datasets and Code Base}\label{app:dataset_code_details}
\noindent\textbf{Diffusion Models.} 
We verify our approach on a diffusion model that is trained on entire Places-365 training dataset~\citep{palette_diff_base,zhou2017places365}. 
We randomly select 50 classes out of 365 classes as the retain set and another 50 classes as the forget set. For each class, we select 5000 images from the Places-365 training set; we then combine them together as the training set for unlearning. By using the approach defined in Eq.~\eref{eq:final_optimization}, we obtain the target model. We then evaluate the obtained model on both forget set and retain set, with 100 images per class from the Places-365 validation set. Hence, we have 5,000 validation images for both the retains and forget sets. 
Since \citeauthor{palette_diff_base} did not release the code, our code is implemented based on an open source re-implementation (\href{https://github.com/Janspiry/Palette-Image-to-Image-Diffusion-Models}{GitHub}). Our experiments are using their provided  model checkpoint on Places-365 dataset.

\noindent\textbf{VQ-GAN}
We evaluate our approach on a VQ-GAN model that is trained on entire ImageNet-1K training dataset~\citep{vqgan_mage,deng2009imagenet}. 
We randomly select 100 classes out of 1000 classes as the retain set and another 100 classes as the forget set. We select 100 images per class from the training set as the training set for unlearning. We then apply our proposed approach to the original model and obtain the target model. For the main results reported in Tabel~\ref{tab:results_center64} and Tabel~\ref{tab:results_center16}, we evaluate the obtained model on both forget set and retain set, with 50 images per class from the ImageNet-validation set; hence, we have 5,000 validation images for both the retains set and forget set. For the other results (in ablation study), we use a smaller version validation set, with 5 images per class from the ImageNet-validation set; hence, we have 500 validation images for both the retains set and forget set.
Our code is implemented based on the offically released code of \cite{vqgan_mage} (\href{https://github.com/LTH14/mage}{GitHub}). Our experiments are using their provided  ViT-Base model checkpoints.

\noindent\textbf{MAE} 
We evaluate our approach on a MAE model and an MAE model that is trained on the entire ImageNet-1K training dataset~\citep{mae_masked_he}. 
The dataset setup is exactly the same as VQ-GAN (check the upper paragraph). 
Our code is implemented based on the offically released code of \cite{mae_masked_he} (\href{https://github.com/facebookresearch/mae}{GitHub}). Our experiments are using their provided  ViT-Base model checkpoints.

\subsection{Evaluation Metrics}\label{app:metrics}
\noindent\textbf{Inception score (IS)} For ImageNet-1K, we directly use the Inception-v3 model checkpoint from torchvision library to compute the IS. For Places-365, we used the ResNet-50 model checkpoint from the official release to compute IS~\citep{zhou2017places365}. 

\noindent\textbf{Fr\'echet inception distance (FID)} 
For both ImageNet-1K and  Places-365, we directly use the backbone network of the Inception-v3 model checkpoint from Torchvision library to compute the FID.
 
\noindent\textbf{CLIP} We use CLIP with ViT-H-14 as the backbone to generate the embedding vectors of each reconstructed image from the original model and unlearned model~\citep{clip}. Then we compute the cosine similarity of these embedding vectors among $\mathcal{D}_F$ and $\mathcal{D}_R$ separately. 

\subsection{Training Hyper-parameters}\label{app:train_param}
\noindent\textbf{Patch Size} For all of the models we evaluate in this paper (including diffusion models, VQ-GAN and MAE), the networks are vision transformer~\citep{vision_transformer} based architecture. We set the size of each patch as $16\times 16$ pixels for all experiments. For example, cropping $8\times 8$ patches means removing $128\times 128$ pixels. We set $\alpha=0.25$ (cf. Eq.~\eref{eq:final_optimization}).  

\noindent\textbf{Diffusion Models} We set the learning rate as $10^{-5}$ with no weight decay. We use the Adam as the optimizer and conduct the unlearning for 3 epochs. We set the input resolution as 256 and set the batch size as 8 per GPU. Overall, it takes 1.5 hours on a 8 NVIDIA A10G server. We set $\alpha=0.25$ (cf. Eq.~\eref{eq:final_optimization}).   

\noindent\textbf{VQ-GAN\&MAE} We set the learning rate as $10^{-4}$ with no weight decay. We use the AdamW as the optimizer with $\beta=(0.9,0.95)$ and conduct the unlearning for 5 epochs. We set the input resolution as 256 for VQ-GAN and 224 for MAE. We set the batch size as 16 per GPU. Overall, it takes one hour on a 4 NVIDIA A10G server. We set $\alpha=0.25$ (cf. Eq.~\eref{eq:final_optimization}).

\begin{figure}[htb]
 \centering
 \begin{subfigure}[b]{0.32\textwidth}
 \centering
 \includegraphics[width=\textwidth]{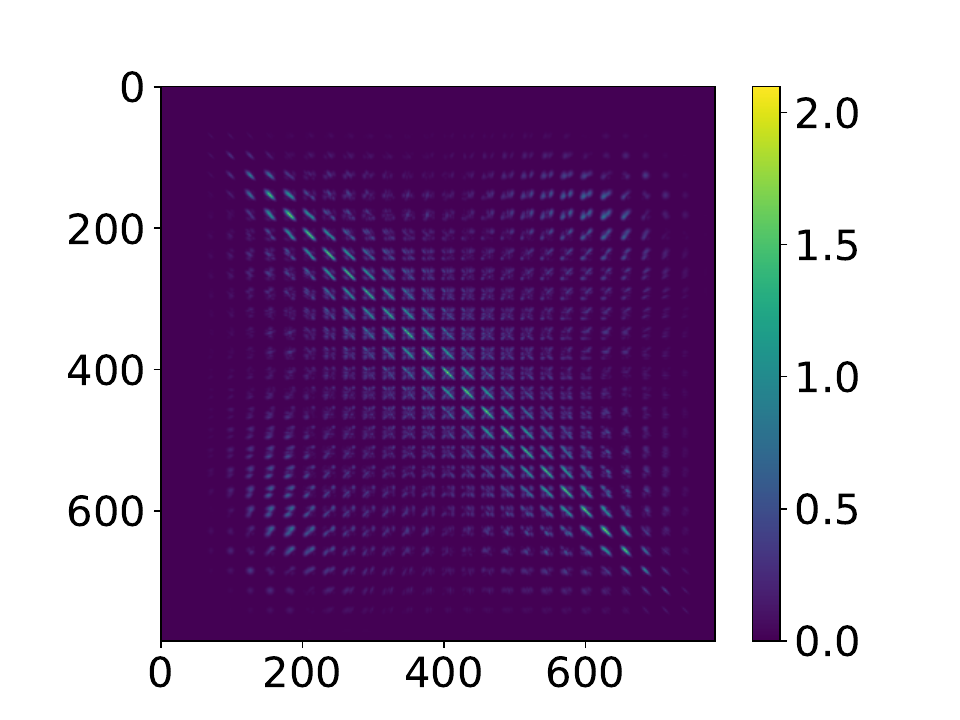}\vspace{-3mm}
 \caption{MNIST}\vspace{-3mm}
 \label{fig:mnist}
 \end{subfigure}
 \hfill
 \begin{subfigure}[b]{0.32\textwidth}
 \centering
 \includegraphics[width=\textwidth]{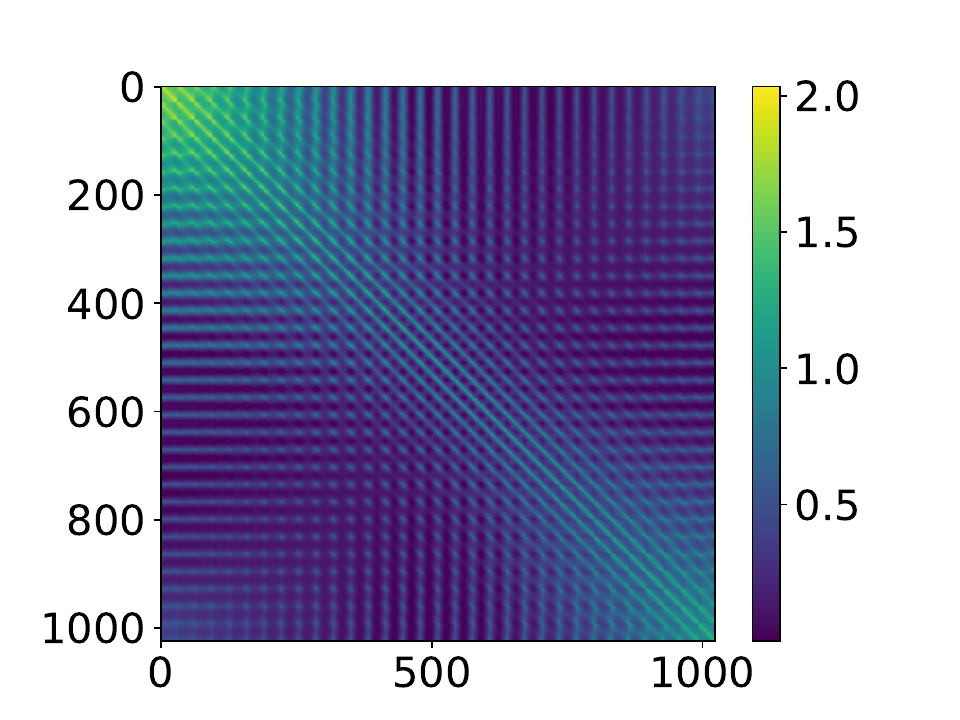}\vspace{-3mm}
 \caption{CIFAR-10}\vspace{-3mm}
 \label{fig:cifar10}
 \end{subfigure}
 \hfill
 \begin{subfigure}[b]{0.32\textwidth}
 \centering
 \includegraphics[width=\textwidth]{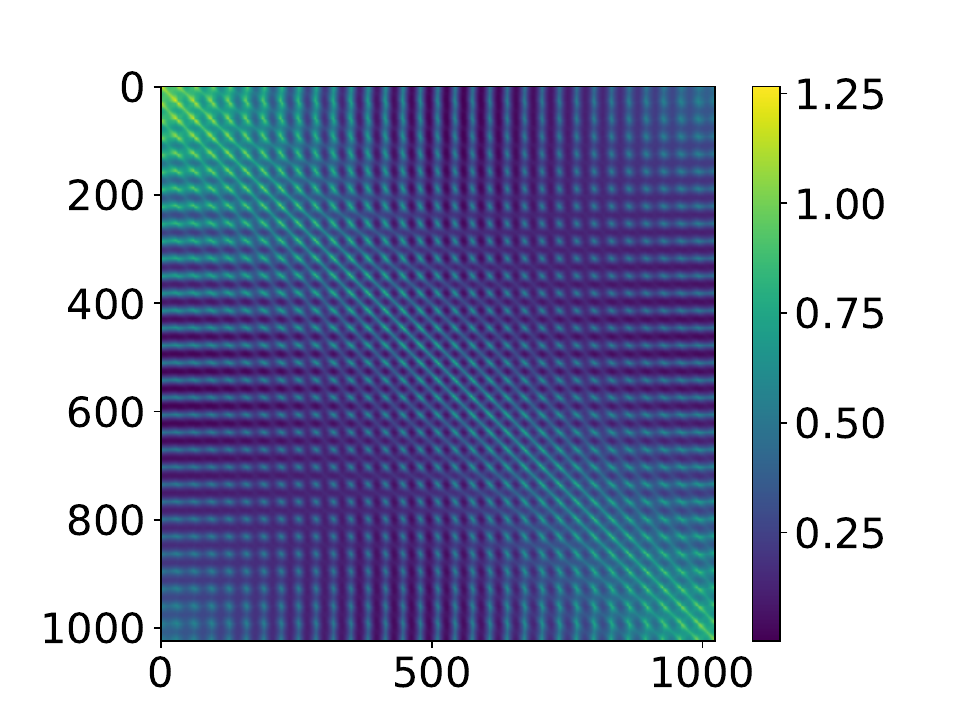}\vspace{-3mm}
 \caption{CIFAR-100}\vspace{-3mm}
 \label{fig:cifar100}
 \end{subfigure}
 \caption{Covariance matrix of three commonly datasets. For CIFAR10/100, we convert the images into gray-scale images. We take the absolute value of the covariance matrix for better illustration.  }
 \label{fig:sigma}
\end{figure}

\begin{algorithm}[t]
  \caption{Pseudo Code of Our Unlearning Algorithm}
  \begin{algorithmic}[1]
    \Inputs{Orginal model $\hmodelp=\fullmodelp$\\ Retain set $\mathcal{D}_R$, Forget set $\mathcal{D}_F$\\ Coefficient $\alpha$, learning rate $\zeta$, and \#Epochs E}
    \Outputs{Target model $\hmodel=\fullmodel$}
    \Initialize{Copy $\hmodel$ to $\hmodel$, \ie, $\hmodel \Leftarrow  \hmodelp$}
    \For{e = 1 to E}
      \State Sample $\{x_r\}$ from $\mathcal{D}_R$
      \State Sample $\{x_f\}$ from $\mathcal{D}_F$
      \State Sample $\{n\}$ from $\mathcal{N}(0,\Sigma)$
      \State \textbf{Ensure:}  $\left|\{x_r\}\right|=\left|\{x_f\}\right|$, \ie, make retain samples and forget samples balanced
     \State Compute loss: $l =\left\rVert\encoder(\trans{(x_r)})-\encoderp(\trans{(x_r)})\right\rVert_2 +\alpha \left\rVert\encoder(\trans{(x_f)})-\encoderp(\trans(n))\right\rVert_2$
    \State Update the parameters of the target encoder $\encoder$: $\theta \Leftarrow \theta-\zeta\nabla_{\theta}l$
    \EndFor
  \end{algorithmic}\label{alg:unlearn_code}
\end{algorithm}

\subsection{Our Unlearning Algorithm}\label{app:code_unlearn_alg}
\paragraph{Selection of $\Sigma$ in Eq.~\eref{eq:final_optimization}}
To conduct our unlearning algorithm, we need to compute the $\Sigma$ in Eq.~\eref{eq:final_optimization}, where $\Sigma$ is the the covariance matrix for the distribution of training images.

Ideally, we should use the exact $\Sigma$ of the images measured on the forget set. However, there are some computational barriers to using the exact $\Sigma$ for a high-resolution image dataset. Specifically, consider a commonly used 256$\times$256 resolution for image generation tasks, the distribution of the generated images will have $256\times 256\times 3 \approx 2\times 10^5$ dimensions. The size of the covariance matrix $\Sigma$ for such a high-dimensional distribution is around $(2\times 10^5)^2=4\times 10^{10}$, which requires around 144GB memory if stored in float precision thus is not practical.

Consequently, to address the computational barrier of $\Sigma$, we use some approximated methods derived from some empirical observations on some small datasets.
Specifically, we compute the exact $\Sigma$ for some small-scale image dataset, including MNIST and CIFAR10/100.
\begin{itemize}
    \item Off-diagonal elements: To find some empirical inspirations, we compute the covariance matrix for three commonly dataset, MNIST, CIFAR10 and CIFAR100. As shown in Fig.~\ref{fig:sigma}, most of the off-diagonal elements are very close to `0'. Hence, in our experiments, we set the off-diagonal elements of $\Sigma$ to `0'
    \item Diagonal elements: Since the images are normalized (\ie, subtract the mean value and divided by the standard deviation), we set the diagonal elements of $\Sigma$ as `1', \ie, $\Sigma(i,i)=1$. 

\end{itemize}

Therefore, we use the identity matrix as the approximation of exact $\Sigma$:
$$\Sigma=\bm{I}$$
where $\bm{I}$ is the identity matrix. We set $\Sigma=\bm{I}$ as the default setup for our approach and baseline methods.

In short, using $I$ to approximate $\Sigma$ is a practical approximation alternative due to the extremely high computational costs of $\Sigma$ for high-resolution images. 
For future work, given our theoretical analysis, we believe that our approach can achieve better results (lower image quality) on forget set if we can find a way to use the exact $\Sigma$. Hence, we plan to explore the potential to reduce the computation of exact $\Sigma$ with low-rank approximation thus enabling the use of more accurate data-driven $\Sigma$.

\paragraph{Pseudo code} Based on our unlearning approach defined in Section~\ref{sec:propsoed_apporach}, we implement the unlearning algorithm for I2I generative models. We provide the pseudo code in Algorithm~\ref{alg:unlearn_code}. As shown, for each batch, we sample  the same number of retain samples and forge samples. We then compute the loss by using Eq.~\eref{eq:final_optimization} and update the parameters of the target model's encoder. 

\subsection{Loss Function of Our Approach}\label{app:lossfunc}
As shown in Eq.~\eref{eq:final_optimization}, we input the Gaussian noise to the original model as the reference for the forget set. 
For VQ-GAN and MAE, we can directly use Eq.~\eref{eq:final_optimization}. For diffusion models, in principle, we still replace the forget sample with Gaussian noise as the input for the original encoder. We next discuss the details for the diffusion models.

We first write the loss function for the normal training of diffusion models first: 
\begin{equation}
    \E_{x} \E_{\vgamma \sim \mathcal{N}(0, I)}  \E_{\eta}\ \left\rVert \fullmodel\left(\trans(x) ,\ \underbrace{\sqrt{\eta}x + \sqrt{1-\eta}\ \vgamma}_{\tilde{x}} ,\ \eta\right) - \vgamma\ \right\rVert_p 
\label{eq:diff_original_loss}
\end{equation}

where $x$ is the ground truth image; $\trans(x)$ is the transformed images, \eg, a cropped image for image uncropping tasks and and low-resolution image for image super-resolution. $\eta$ is the forward process variance coefficient. For more details, please refer to the \cite{palette_diff_base}.

Now, we introduce the unlearning optimization function for diffusion models:

\begin{equation}\centering
\begin{aligned}
\argmin_{\theta}\E_{x_r,x_f,n} \E_{\vgamma \sim \mathcal{N}(0, I)}  \E_{\eta}\bigg\{\ &\left\rVert \encoder\left(\trans(x_r) ,\ \tilde{x_r} ,\ \eta\right) - \encoderp\left(\trans(x_r) ,\ \tilde{x_r} ,\ \eta\right) \right\rVert_2 \\
+\alpha\ &\left\rVert \encoder\left(\trans(x_f) ,\ \tilde{x_f} ,\ \eta\right) - \encoderp\left(\trans(x_f) ,\ n ,\ \eta\right) \right\rVert_2\bigg\},\\
&x_{r}\in\mathcal{D}_R, x_{f}\in\mathcal{D}_F, n\sim\mathcal{N}(0,\Sigma)
 \end{aligned}\label{eq:diff_unlearn_loss}
\end{equation}

where $\tilde{x_r}=\sqrt{\eta}x_r + \sqrt{1-\eta}\ \vgamma$ and $\tilde{x_f}=\sqrt{\eta}x_f + \sqrt{1-\eta}\ \vgamma$. Essentially, we replace the $\tilde{x_f}$ with the Gaussian noise $n$ as the input for the original encoder. Note that, \citeauthor{palette_diff_base} adopt an U-Net as the network architecture; thus the encoder has multiple-stage outputs. We flatten these multiple outputs into vectors and then combine them as a single representation vector .   

We remark the equation for diffusion models looks slightly different from VQ-GAN and MAE, but they are actually following the same principle defined in Eq.~\eref{eq:final_optimization}; that is,  we replace the real forget samples with the Gaussian noise as the input for the original encoder. 

\subsection{Baselines}\label{app:baseline_details}
We use the exact same setup as introduced in the main paper and Appendix~\ref{app:train_param}. We provide the loss function for different baseline methods below. For the baselines, the unlearning is achieved by minimizing the loss values. 

\subsubsection{\textsc{Max Loss}} \textsc{Max Loss} maximizes the training loss w.r.t. the ground truth images on the forget set. Hence, we use the original loss function for both retains set and forget set, but assign a negative coefficient for the forget set. We do not modify the loss for the retains set. 

\begin{itemize}[leftmargin=0.021\textwidth]
    \item Diffusion Models 
    \begin{equation}\scriptsize\label{eq:max_loss_diff}
    \centering
    \begin{aligned}
    \argmin_{\theta,\phi}\E_{x_r,x_f} \E_{\vgamma \sim \mathcal{N}(0, I)}  \E_{\eta}\bigg\{&\left\rVert \fullmodel\left(\trans(x_r) ,\tilde{x_r} ,\eta\right)-\vgamma \right\rVert_2    -\alpha\ \left\rVert \fullmodel\left(\trans(x_f) ,\ \tilde{x_f} ,\eta\right)-\vgamma \right\rVert_2\bigg\}\\
     \end{aligned}
    \end{equation}
    where $\tilde{x_r}=\sqrt{\eta}x_r + \sqrt{1-\eta}\ \vgamma$ and $\tilde{x_f}=\sqrt{\eta}x_f + \sqrt{1-\eta}\ \vgamma$. $\trans( \cdot)$ is the transformed function, \eg, a cropped image for image uncropping tasks.
    \item VQ-GAN
    \begin{equation}\label{eq:max_loss_gan}
    \centering
    \begin{aligned}
    \argmax_{\theta,\phi}\E_{x_r,x_f}\bigg\{ C_{\mathcal{J}}\left(\fullmodel\left(\trans(x_r)\right), x_r\right)-\alpha C_{\mathcal{J}}\left(\fullmodel\left(\trans(x_f)\right), x_f\right) \bigg\}
     \end{aligned}
    \end{equation}
    where $\mathcal{J}$ is a discriminator network used to predict whether the images are real images or generated images; 
    $C_{\mathcal{J}}$ is the cross entropy of the discriminator prediction. We also note that, during unlearning, the discriminator $\mathcal{J}$ is also updated in the normal training way. Please check more details in~\cite{vqgan_mage}.
    \item MAE
    \begin{equation}\label{eq:max_loss_mae}
    \centering
    \begin{aligned}
    \argmin_{\theta,\phi}\E_{x_r,x_f} \bigg\{&\left\rVert \fullmodel\left(\trans(x_r)\right)-x_r \right\rVert_2    -\alpha\ \left\rVert \fullmodel\left(\trans(x_f) \right)-x_f \right\rVert_2\bigg\}\\
     \end{aligned}
    \end{equation}
    $\trans( \cdot)$ is the transformed function, \eg, a cropped image for image uncropping tasks.
\end{itemize}

\subsubsection{\textsc{Retain Label}} \textsc{Retain Label} minimizes training loss by setting the retain samples as the ground truth for the forget set. 
\begin{itemize}[leftmargin=0.021\textwidth]
    \item Diffusion Models 
    \begin{equation}\scriptsize
    \centering
    \begin{aligned}
    \argmin_{\theta,\phi}\E_{x_r,x_f} \E_{\vgamma \sim \mathcal{N}(0, I)}  \E_{\eta}\bigg\{&\left\rVert \fullmodel\left(\trans(x_r) ,\tilde{x_r} ,\eta\right)-\vgamma \right\rVert_2    +\alpha\ \left\rVert \fullmodel\left(\trans(x_f) ,\ \tilde{x_r} ,\eta\right)-\vgamma \right\rVert_2\bigg\}\\
     \end{aligned}
    \end{equation}
    where $\tilde{x_r}=\sqrt{\eta}x_r + \sqrt{1-\eta}\ \vgamma$. As shown, the $\tilde{x_f}$ in Eq.~\eref{eq:max_loss_diff} is replaced by the the retain samples $\tilde{x_r}$. 
    \item VQ-GAN
    \begin{equation}
    \centering
    \begin{aligned}
    \argmax_{\theta,\phi}\E_{x_r,x_f}\bigg\{ C_{\mathcal{J}}\left(\fullmodel\left(\trans(x_r)\right), x_r\right)+\alpha C_{\mathcal{J}}\left(\fullmodel\left(\trans(x_f)\right), x_r\right) \bigg\}
     \end{aligned}
    \end{equation}
    where $\mathcal{J}$ is a discriminator network used to predict whether the images are real images or generated images; $C_{\mathcal{J}}$ is the cross entropy of the discriminator prediction. As shown, the reference ${x_f}$ in the second term of Eq.~\eref{eq:max_loss_gan} is replaced by the retain sample $x_r$.
    \item MAE
    \begin{equation}
    \centering
    \begin{aligned}
    \argmin_{\theta,\phi}\E_{x_r,x_f} \bigg\{&\left\rVert \fullmodel\left(\trans(x_r)\right)-x_r \right\rVert_2   +\alpha\ \left\rVert \fullmodel\left(\trans(x_f) \right)-n \right\rVert_2\bigg\}\\
     \end{aligned}
    \end{equation}
    As shown, the target $\tilde{x_f}$ in the second term of Eq.~\eref{eq:max_loss_mae} is replaced by the retain sample $x_r$.
\end{itemize}

\subsubsection{\textsc{Noisy Label}} \textsc{Noisy Label} minimizes training loss by setting the Gaussian noise as the ground truth images for the forget set. Hence we directly use the original loss function by replacing the ground truth images with standard Gaussian noise for the forget set.
We do not modify the loss for the retains set. 
\begin{itemize}[leftmargin=0.021\textwidth]
    \item Diffusion Models 
    \begin{equation}
    \centering
    \begin{aligned}
    \argmin_{\theta,\phi}\E_{x_r,x_f} \E_{\vgamma \sim \mathcal{N}(0, I)}  \E_{\eta}\bigg\{&\left\rVert \fullmodel\left(\trans(x_r) ,\tilde{x_r} ,\eta\right)-\vgamma \right\rVert_2    +\alpha\ \left\rVert \fullmodel\left(\trans(x_f) ,\ \eta ,\eta\right)-\vgamma \right\rVert_2\bigg\}\\
     \end{aligned}
    \end{equation}
    where $\tilde{x_r}=\sqrt{\eta}x_r + \sqrt{1-\eta}\ \vgamma$. As shown, the $\tilde{x_f}$ in Eq.~\eref{eq:max_loss_diff} is replaced by the Gaussian noise $\eta$. $\trans( \cdot)$ is the transformed function, \eg, a cropped image for image uncropping tasks.
    \item VQ-GAN
    \begin{equation}
    \centering
    \begin{aligned}
    \argmax_{\theta,\phi}\E_{x_r,x_f}\bigg\{ C_{\mathcal{J}}\left(\fullmodel\left(\trans(x_r)\right), x_r\right)+\alpha C_{\mathcal{J}}\left(\fullmodel\left(\trans(x_f)\right), n\right) \bigg\}
     \end{aligned}
    \end{equation}
    where $\mathcal{J}$ is a discriminator network used to predict whether the images are real images or generated images; $C_{\mathcal{J}}$ is the cross entropy of the discriminator's prediction. As shown, the reference ${x_f}$ in the second term of Eq.~\eref{eq:max_loss_gan} is replaced by the Gaussian noise $n$.
    \item MAE
    \begin{equation}
    \centering
    \begin{aligned}
    \argmin_{\theta,\phi}\E_{x_r,x_f} \bigg\{&\left\rVert \fullmodel\left(\trans(x_r)\right)-x_r \right\rVert_2   +\alpha\ \left\rVert \fullmodel\left(\trans(x_f) \right)-n \right\rVert_2\bigg\}\\
     \end{aligned}
    \end{equation}
    As shown, the target $\tilde{x_f}$ in the second term of Eq.~\eref{eq:max_loss_mae} is replaced by the Gaussian noise $n$.
\end{itemize}

\subsubsection{\textsc{Random Encoder}} \textsc{Random Encoder} minimizes the $L_2$ loss between the Gaussian noise and the representation vector of the encoder for the forget set. 
\begin{itemize}[leftmargin=0.021\textwidth]
    \item Diffusion Models 
    \begin{equation}
    \centering
    \begin{aligned}
    \argmin_{\theta}\E_{x_r,x_f} \E_{\vgamma \sim \mathcal{N}(0, I)}  \E_{\eta}\bigg\{\ &\left\rVert \encoder\left(\trans(x_r) ,\ \tilde{x_r} ,\ \eta\right) - \encoderp\left(\trans(x_r) ,\ \tilde{x_r} ,\ \eta\right) \right\rVert_2 \\
+\alpha\ &\left\rVert \encoder\left(\trans(x_f) ,\ \tilde{x_f} ,\ \eta\right) - \eta \right\rVert_2\bigg\}
     \end{aligned}
    \end{equation}
    where $\tilde{x_r}=\sqrt{\eta}x_r + \sqrt{1-\eta}\ \vgamma$. As shown, the target for the forget set is directly the Gaussian noise instead of its embedding vector. 
    \item VQ-GAN\&MAE
    \begin{equation}
    \centering
    \begin{aligned}
    \argmin_{\theta}\E_{x_r,x_f} \bigg\{\left\rVert \encoder\left(\trans(x_r)\right)-\encoderp\left(\trans(x_r)\right) \right\rVert_2   +\alpha\ \left\rVert \encoder\left(\trans(x_f) \right)-n \right\rVert_2\bigg\}
     \end{aligned}
    \end{equation}
\end{itemize}
\section{Supplementary Results}\label{app:supple_results}
\def\theequation{D.\arabic{equation}}
\def\thelem{D.\arabic{lem}}
\def\thefigure{D.\arabic{figure}}
\def\thetable{D.\arabic{table}}

\begin{figure}[htb]
 \centering
     \begin{subfigure}[b]{0.32\textwidth}
     \centering
     \includegraphics[width=\textwidth]{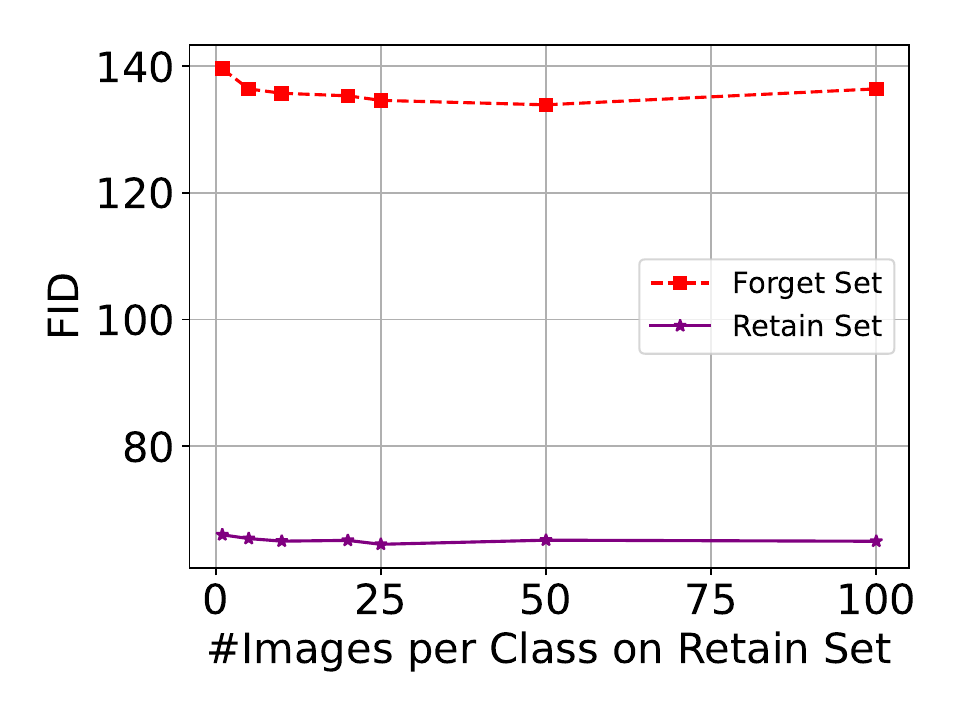}\vspace{-1mm}
     \caption{VQ-GAN: FID}
     \end{subfigure}
     \hfill
     \begin{subfigure}[b]{0.32\textwidth}
     \centering
     \includegraphics[width=\textwidth]{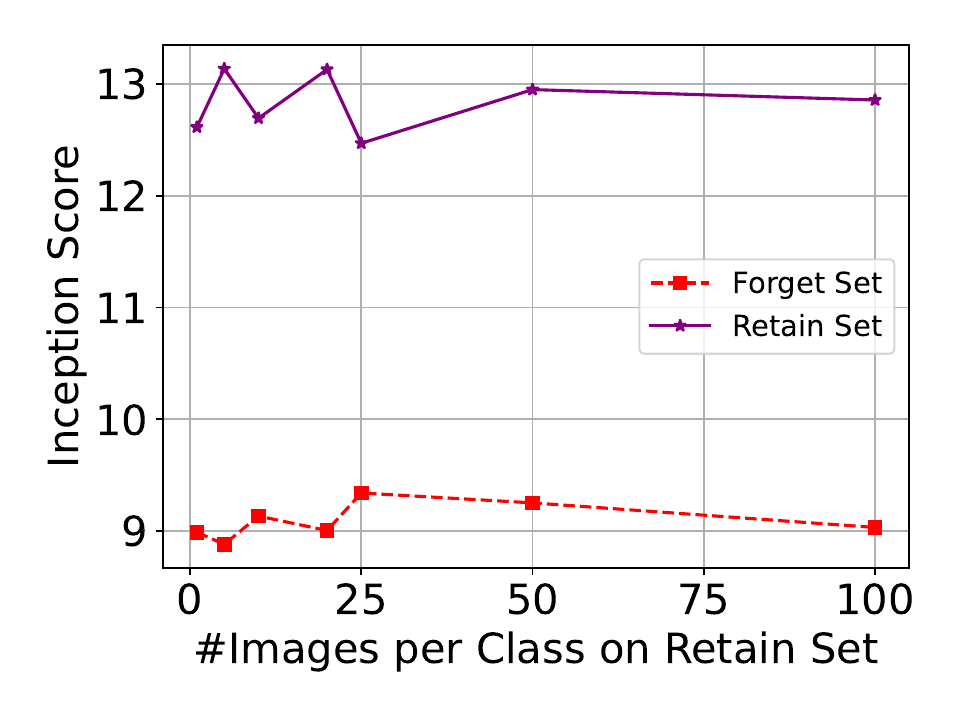}\vspace{-1mm}
     \caption{VQ-GAN: IS}
     \end{subfigure}
     \hfill
     \begin{subfigure}[b]{0.32\textwidth}
     \centering
     \includegraphics[width=\textwidth]{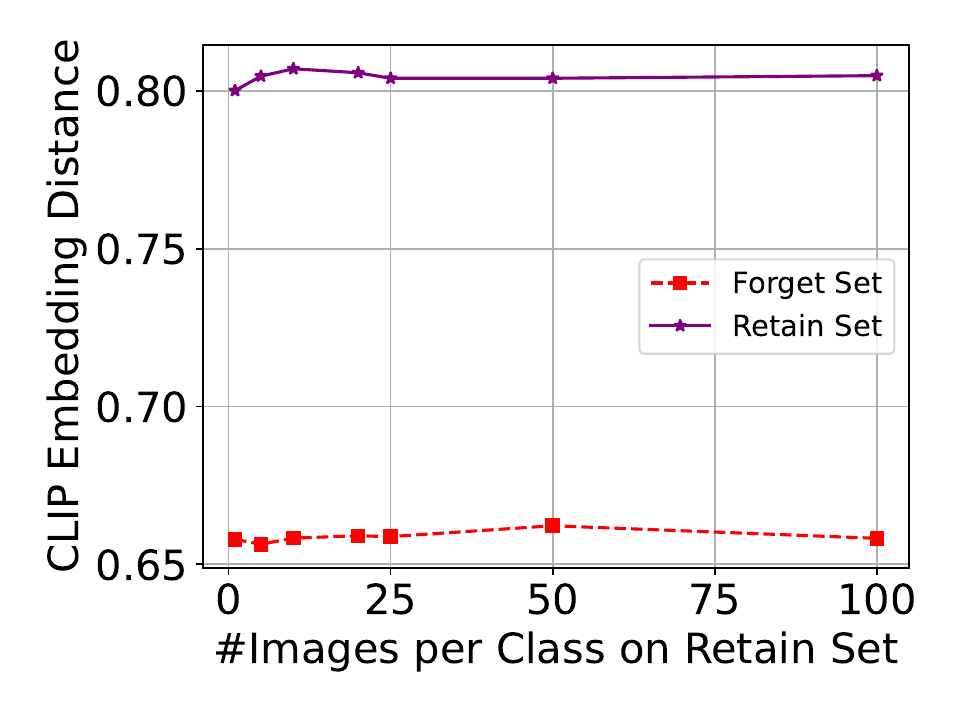}\vspace{-1mm}
     \caption{VQ-GAN: CLIP}
     \end{subfigure}
     
     \begin{subfigure}[b]{0.32\textwidth}
     \centering
     \includegraphics[width=\textwidth]{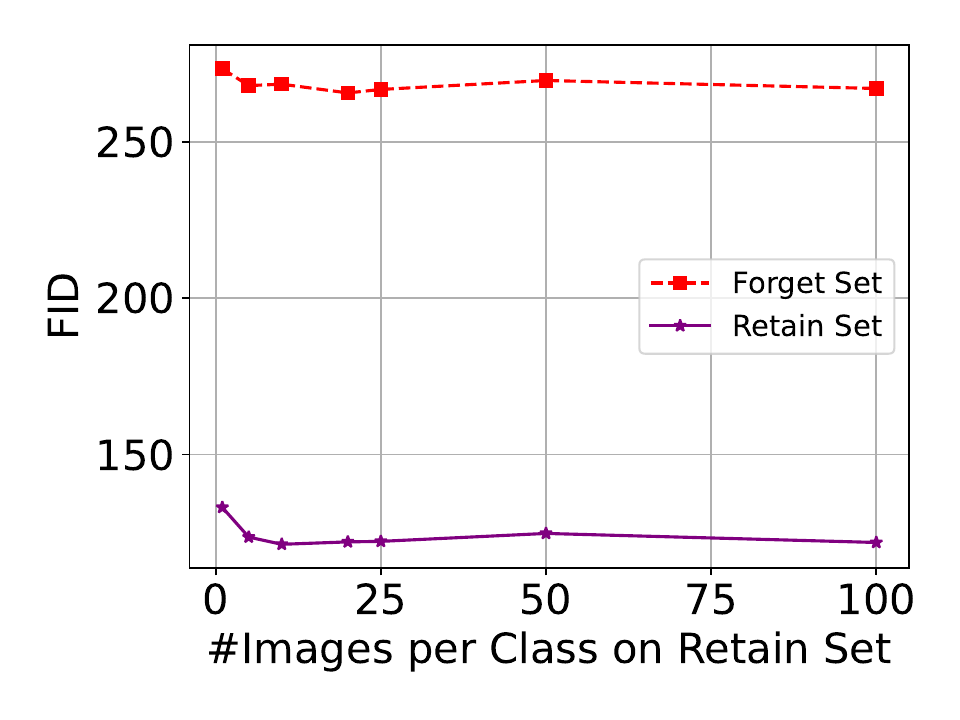}\vspace{-1mm}
     \caption{MAE: FID}
     \end{subfigure}
     \hfill
     \begin{subfigure}[b]{0.32\textwidth}
     \centering
     \includegraphics[width=\textwidth]{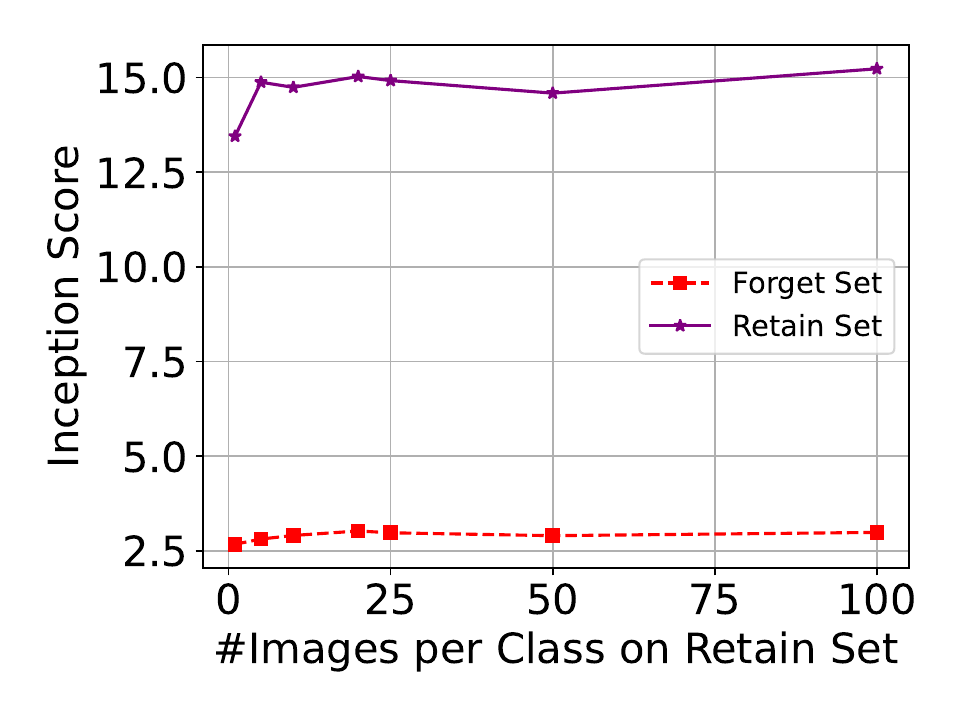}\vspace{-1mm}
     \caption{MAE: IS}
     \end{subfigure}
     \hfill
     \begin{subfigure}[b]{0.32\textwidth}
     \centering
     \includegraphics[width=\textwidth]{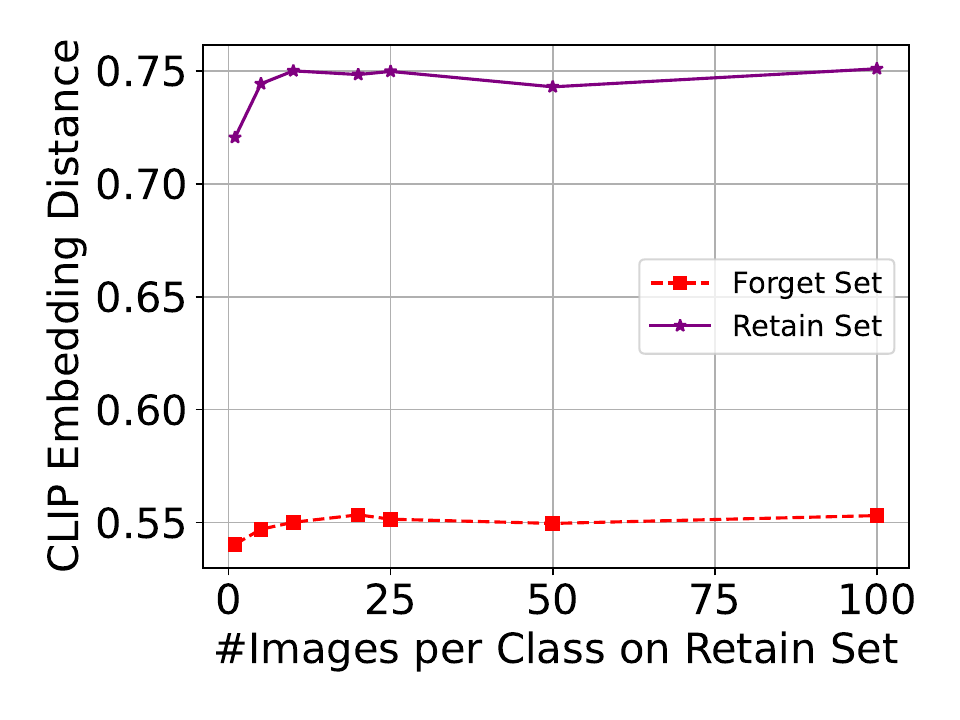}\vspace{-1mm}
     \caption{MAE: CLIP}
     \end{subfigure}
 \caption{The performance of our approach under limited availability of the retain sample. The ``100 images per class'' (right side of horizontal-axis) in these plots indicate a \textit{full} retains set baseline. As shown, by gradually reducing the number of images for the retain set, the performance degradation is negligible. Even for the extreme case where we only one image per class, the performance degradation is also small. }
 \label{fig:varying_retain_ratio}
\end{figure}

\subsection{Robustness to Retain Samples Availability}\label{app:vary_retain}

In Table~\ref{tab:results_center64} and Table~\ref{tab:results_center16}, we report the results under an extreme case where there is no real retain sample available. In this section, we relax the constraints by assuming the limited availability to the real retain samples. 
Specifically, on ImageNet-1K, we fix the forget set by using 100 images per class for the forget set; in total, we have 10K image for the forget set. 
As the baseline method, we have a retain set with 100 image per class as well (in total 10K); we call this a \textit{full} retain set. 

We next vary the number of images per class for the retain set within the range of $\{1,5,10,20,25,50\}$ and compare the performance of the unlearned models under different values. To balance the number of forget sample and retain samples, we over sample the retain samples. For example, if we have 5 image per class for the retain set, we sample these images 20 times (since $100/5=20$). 

As shown in Fig.~\ref{fig:varying_retain_ratio}, compared to the full retain set (100 images per class), our approach is still effective with a slight performance degradation. Combining the results in Table~\ref{tab:results_center64} and Table~\ref{tab:results_center16}, these results show the applicability and resilience of our approach in scenarios where actual the retain samples are limited or even unavailable.

\begin{figure}[htb]
 \centering
     \begin{subfigure}[b]{\textwidth}
     \centering
     \includegraphics[width=\textwidth]{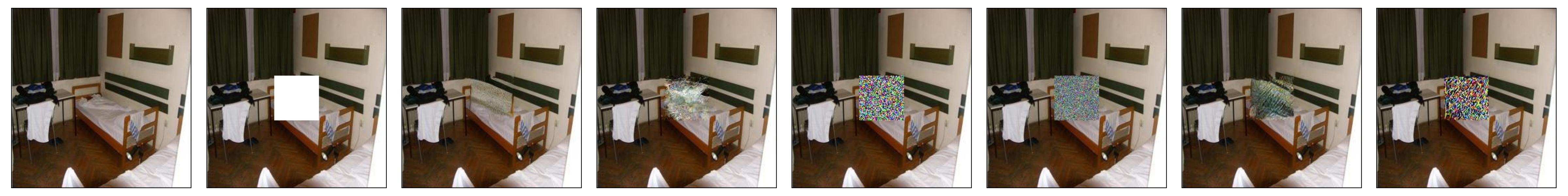}\vspace{-1mm}
     \end{subfigure}
     
     \begin{subfigure}[b]{\textwidth}
     \centering
     \includegraphics[width=\textwidth]{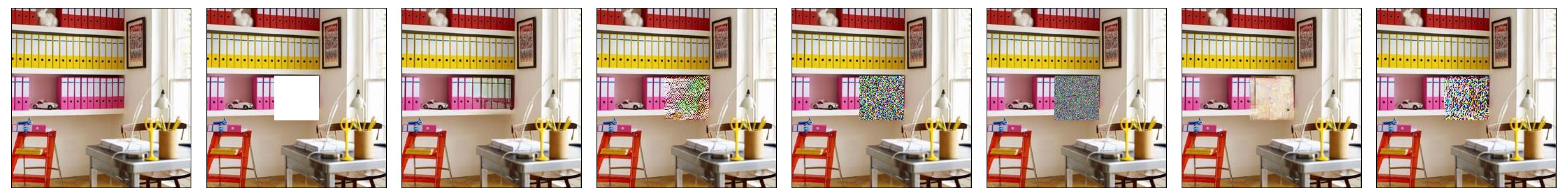}\vspace{-1mm}
     \end{subfigure}
     
     \begin{subfigure}[b]{\textwidth}
     \centering
     \includegraphics[width=\textwidth]{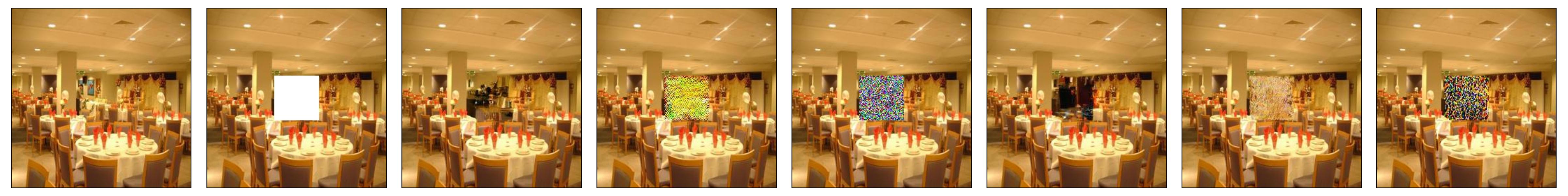}\vspace{-1mm}
     \end{subfigure}
     
     \begin{subfigure}[b]{\textwidth}
     \centering
     \includegraphics[width=\textwidth]{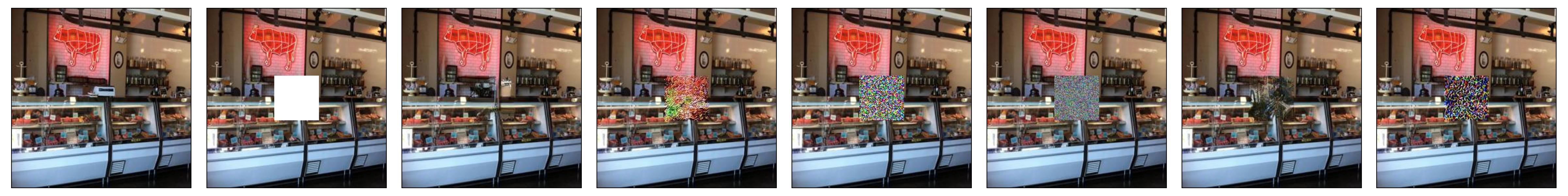}\vspace{-1mm}
     \end{subfigure}
    
     \begin{subfigure}[b]{\textwidth}
     \centering
     \includegraphics[width=\textwidth]{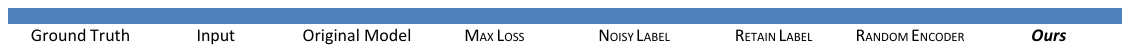}\vspace{-2mm}
    \caption{{Forget set}}\vspace{1mm}
     \end{subfigure}

     \begin{subfigure}[b]{\textwidth}
     \centering
     \includegraphics[width=\textwidth]{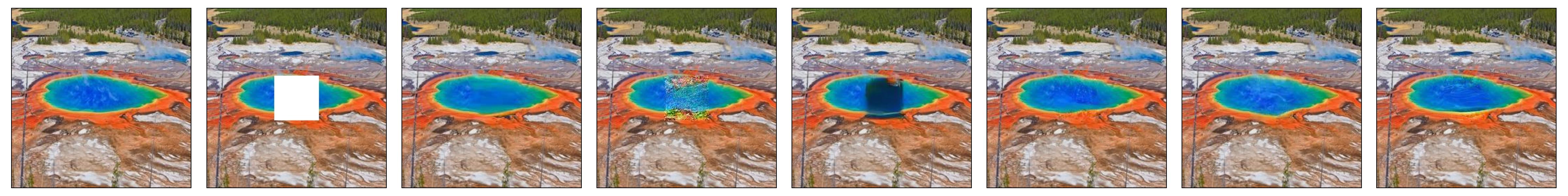}\vspace{-1mm}
     \end{subfigure}
     
     \begin{subfigure}[b]{\textwidth}
     \centering
     \includegraphics[width=\textwidth]{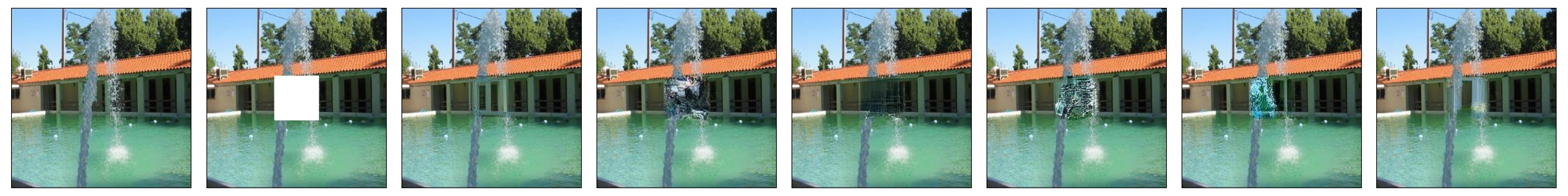}\vspace{-1mm}
     \end{subfigure}
     
     \begin{subfigure}[b]{\textwidth}
     \centering
     \includegraphics[width=\textwidth]{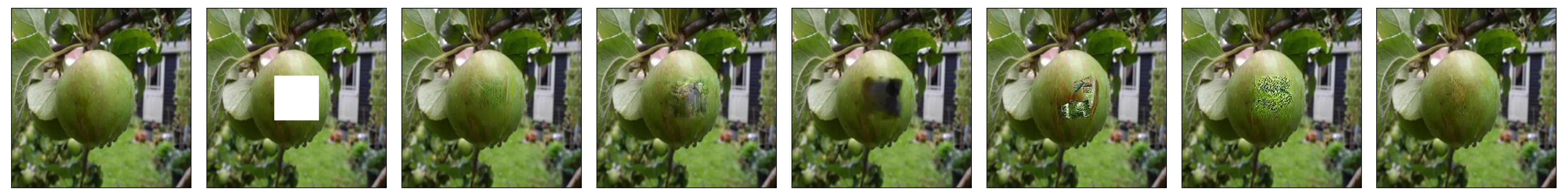}\vspace{-1mm}
     \end{subfigure}
     
     \begin{subfigure}[b]{\textwidth}
     \centering
     \includegraphics[width=\textwidth]{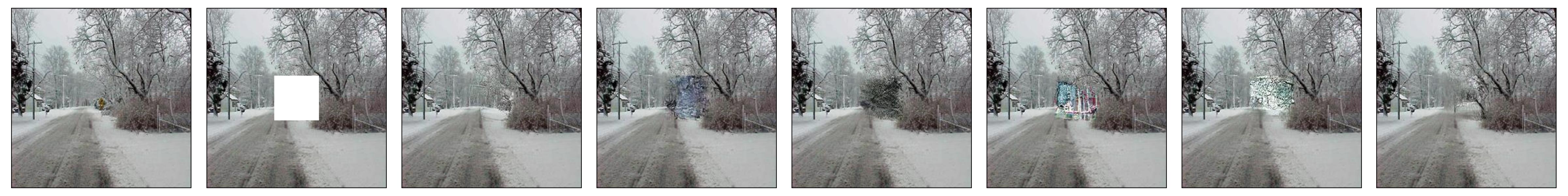}\vspace{-1mm}
     \end{subfigure}
    
     \begin{subfigure}[b]{\textwidth}
     \centering
     \includegraphics[width=\textwidth]{figure/caption_v3.pdf}\vspace{-2mm}
    \caption{{Retain set}}
     \end{subfigure}
 \caption{Diffusion models: cropping $4\times 4$ patches at the center of the image, where each patch is $ 16 \times 16$ pixels. As shown, our approach has almost identical performance as the original model (\ie, before unlearning); the generated images on forget set are some random noise. Overall, we outperform all other baseline methods.  }
 \label{fig:diff_visual_center4by4}
\end{figure}

\begin{figure}[htb]
 \centering
     \begin{subfigure}[b]{\textwidth}
     \centering
     \includegraphics[width=\textwidth]{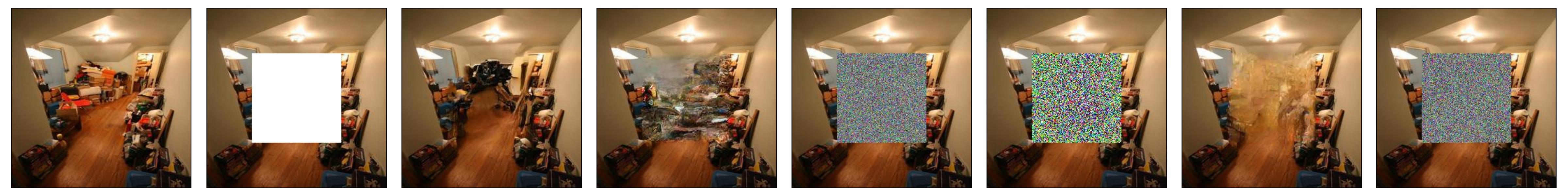}\vspace{-1mm}
     \end{subfigure}
     
     \begin{subfigure}[b]{\textwidth}
     \centering
     \includegraphics[width=\textwidth]{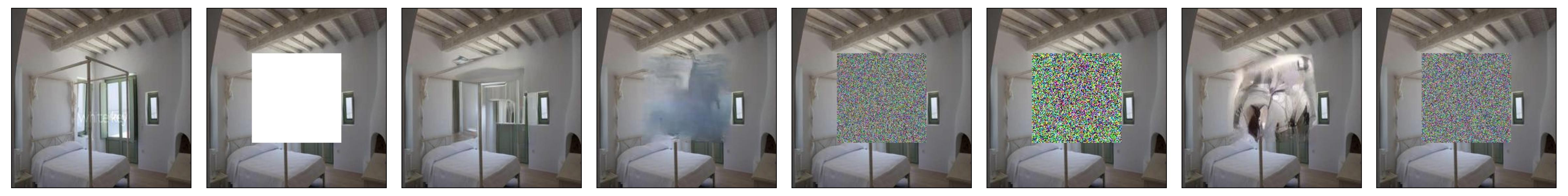}\vspace{-1mm}
     \end{subfigure}
     
     \begin{subfigure}[b]{\textwidth}
     \centering
     \includegraphics[width=\textwidth]{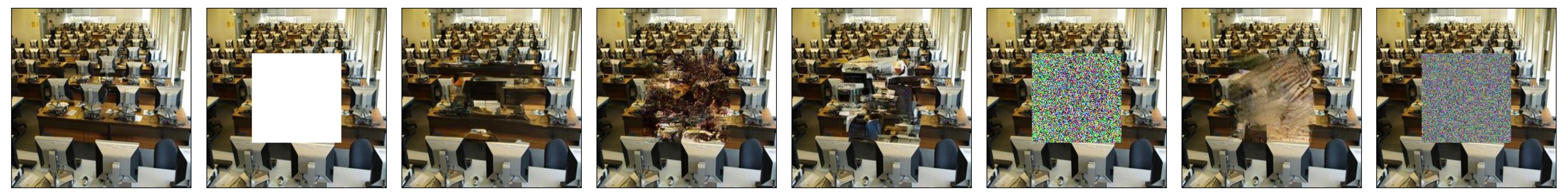}\vspace{-1mm}
     \end{subfigure}
     
     \begin{subfigure}[b]{\textwidth}
     \centering
     \includegraphics[width=\textwidth]{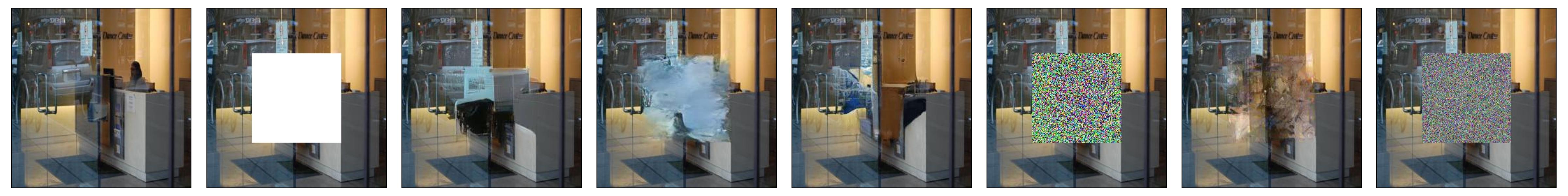}\vspace{-1mm}
     \end{subfigure}
    
     \begin{subfigure}[b]{\textwidth}
     \centering
     \includegraphics[width=\textwidth]{figure/caption_v3.pdf}\vspace{-2mm}
    \caption{{Forget set}}\vspace{1mm}
     \end{subfigure}

     \begin{subfigure}[b]{\textwidth}
     \centering
     \includegraphics[width=\textwidth]{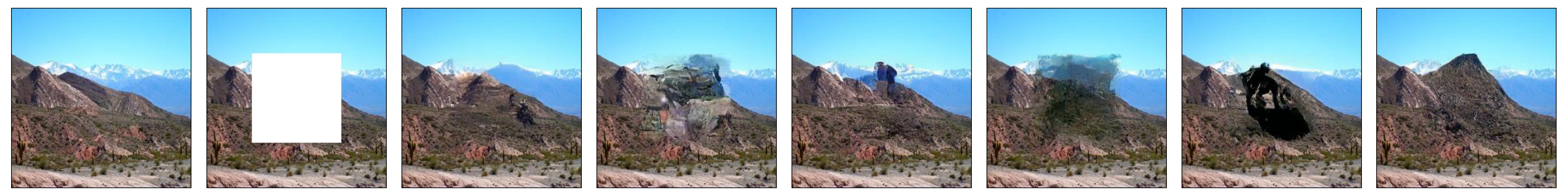}\vspace{-1mm}
     \end{subfigure}
     
     \begin{subfigure}[b]{\textwidth}
     \centering
     \includegraphics[width=\textwidth]{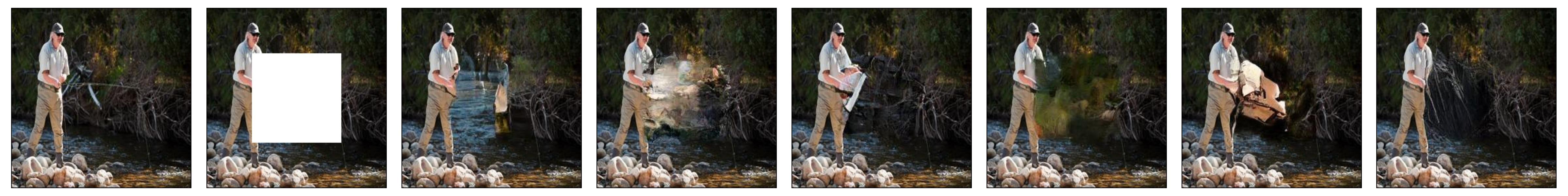}\vspace{-1mm}
     \end{subfigure}
     
     \begin{subfigure}[b]{\textwidth}
     \centering
     \includegraphics[width=\textwidth]{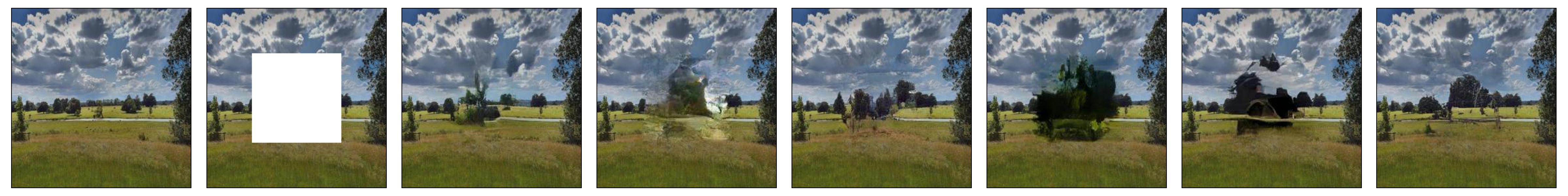}\vspace{-1mm}
     \end{subfigure}
     
     \begin{subfigure}[b]{\textwidth}
     \centering
     \includegraphics[width=\textwidth]{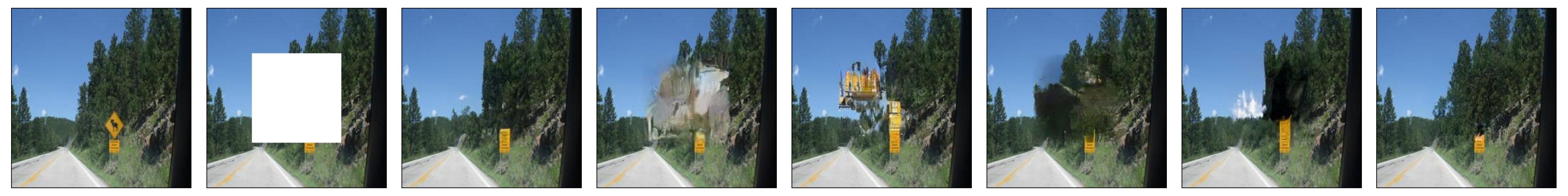}\vspace{-1mm}
     \end{subfigure}
    
     \begin{subfigure}[b]{\textwidth}
     \centering
     \includegraphics[width=\textwidth]{figure/caption_v3.pdf}\vspace{-2mm}
    \caption{{Retain set}}
     \end{subfigure}
 \caption{Diffusion models: cropping $8\times 8$ patches at the center of the image, where each patch is $ 16 \times 16$ pixels. As shown, our approach has almost identical performance as the original model (\ie, before unlearning); the generated images on forget set are some random noise. Overall, we outperform all other baseline methods.  }
 \label{fig:diff_visual_center8by8}
\end{figure}

\subsection{Diffusion}
We provide the visualization results under center crop size of $4\times4$ patches in Fig.~\ref{fig:diff_visual_center4by4}.
We also provide more visualization results under center crop size of $8\times8$ patches in Fig.~\ref{fig:diff_visual_center8by8}. As shown in Fig.~\ref{fig:diff_visual_center4by4} and Fig.~\ref{fig:diff_visual_center8by8}, out approach can generate very high quality images on retain set while filling some random noise on the forget set. As a contrast, the other baselines methods struggle on the performance drop on the retain set. \textsc{Random Encoder} cannot even forget the information on the forget set.

\begin{figure}[htb]
 \centering
     \begin{subfigure}[b]{\textwidth}
     \centering
     \includegraphics[width=\textwidth]{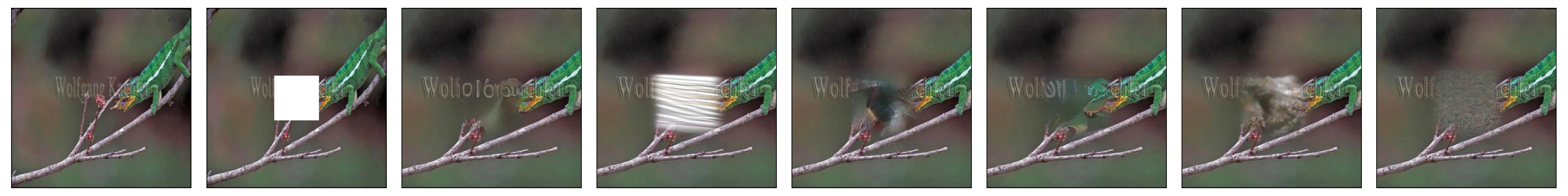}\vspace{-1mm}
     \end{subfigure}
     
     \begin{subfigure}[b]{\textwidth}
     \centering
     \includegraphics[width=\textwidth]{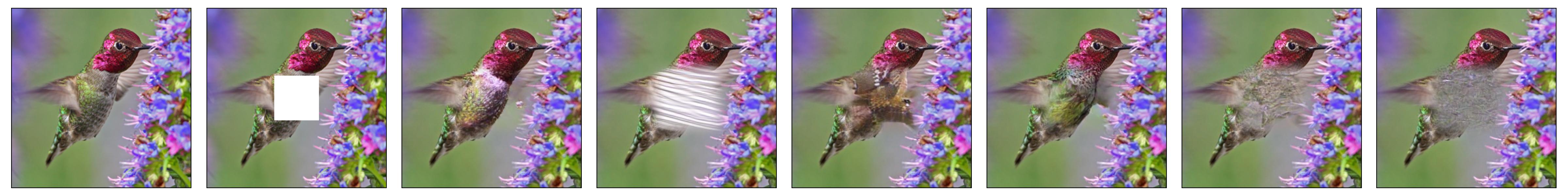}\vspace{-1mm}
     \end{subfigure}
     
     \begin{subfigure}[b]{\textwidth}
     \centering
     \includegraphics[width=\textwidth]{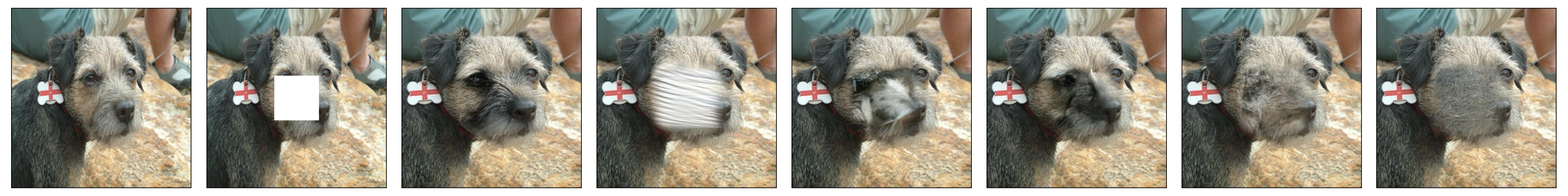}\vspace{-1mm}
     \end{subfigure}
     
     \begin{subfigure}[b]{\textwidth}
     \centering
     \includegraphics[width=\textwidth]{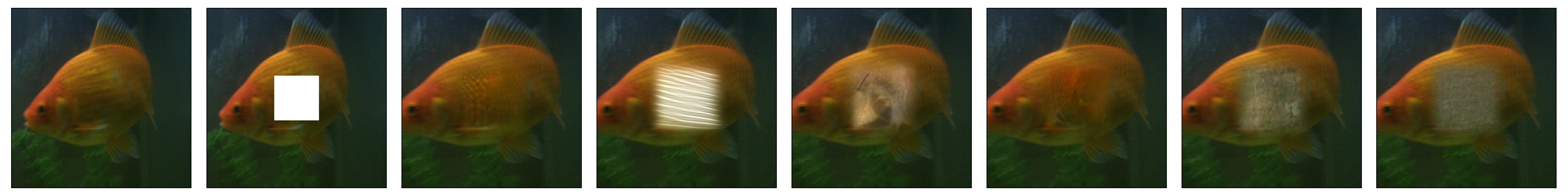}\vspace{-1mm}
     \end{subfigure}
    
     \begin{subfigure}[b]{\textwidth}
     \centering
     \includegraphics[width=\textwidth]{figure/caption_v3.pdf}\vspace{-2mm}
    \caption{{Forget set}}\vspace{1mm}
     \end{subfigure}

     \begin{subfigure}[b]{\textwidth}
     \centering
     \includegraphics[width=\textwidth]{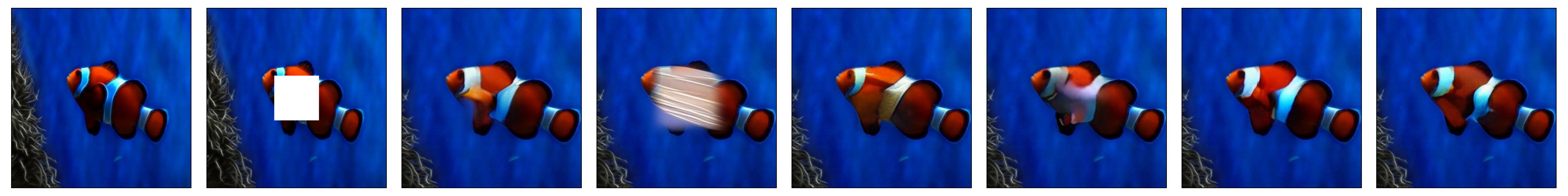}\vspace{-1mm}
     \end{subfigure}
     
     \begin{subfigure}[b]{\textwidth}
     \centering
     \includegraphics[width=\textwidth]{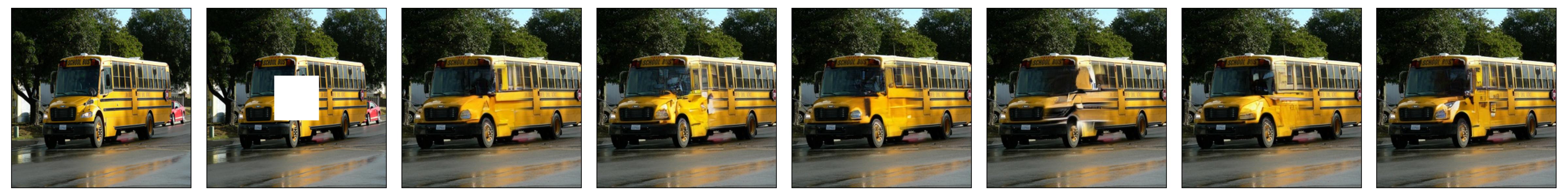}\vspace{-1mm}
     \end{subfigure}
     
     \begin{subfigure}[b]{\textwidth}
     \centering
     \includegraphics[width=\textwidth]{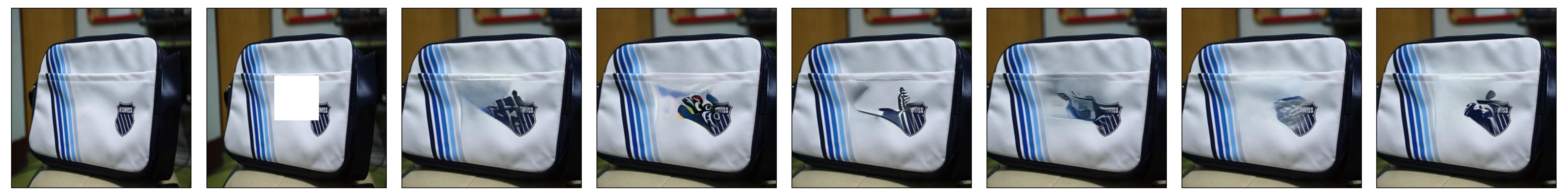}\vspace{-1mm}
     \end{subfigure}
     
     \begin{subfigure}[b]{\textwidth}
     \centering
     \includegraphics[width=\textwidth]{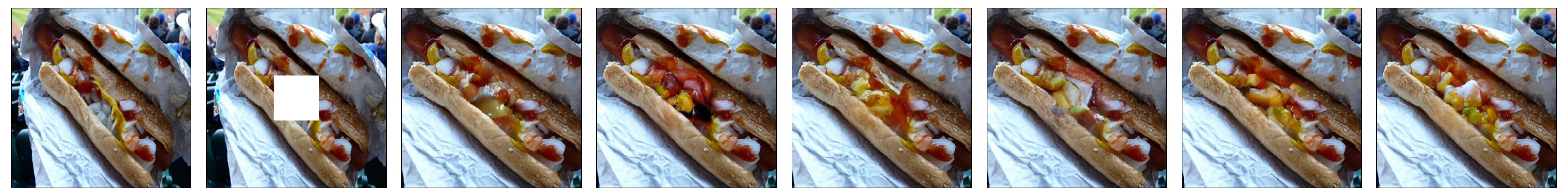}\vspace{-1mm}
     \end{subfigure}
    
     \begin{subfigure}[b]{\textwidth}
     \centering
     \includegraphics[width=\textwidth]{figure/caption_v3.pdf}\vspace{-2mm}
    \caption{{Retain set}}
     \end{subfigure}
 \caption{VQ-GAN: cropping $4\times 4$ patches at the center of the image, where each patch is $ 16 \times 16$ pixels. As shown, our approach has almost identical performance as the original model (\ie, before unlearning); the generated images on forget set are some random noise.   }
 \label{fig:gan_visual_center4by4}
\end{figure}

\begin{figure}[htb]
\centering
     \begin{subfigure}[b]{\textwidth}
     \centering
     \includegraphics[width=\textwidth]{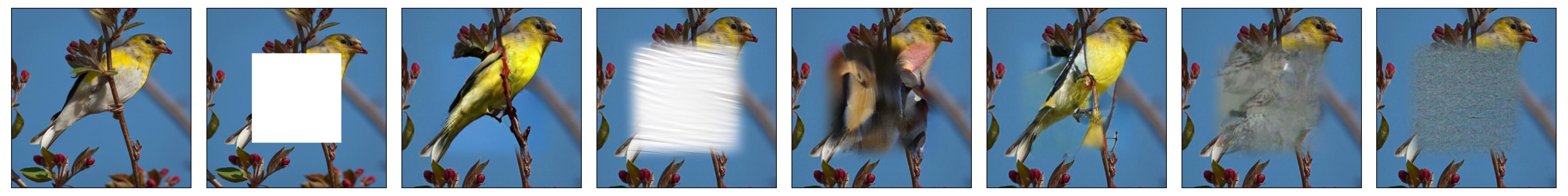}\vspace{-1mm}
     \end{subfigure}
     
     \begin{subfigure}[b]{\textwidth}
     \centering
     \includegraphics[width=\textwidth]{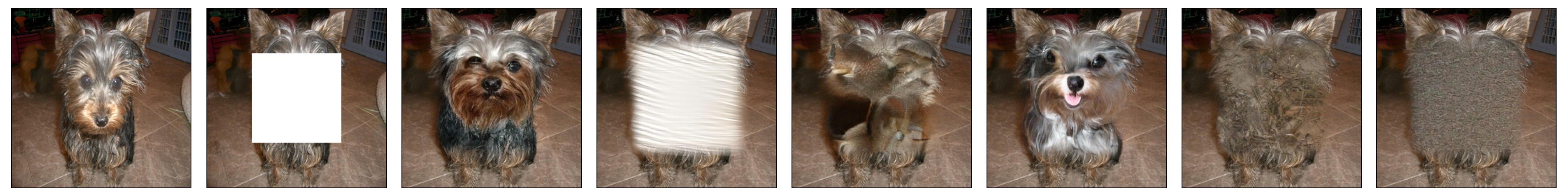}\vspace{-1mm}
     \end{subfigure}
     
     \begin{subfigure}[b]{\textwidth}
     \centering
     \includegraphics[width=\textwidth]{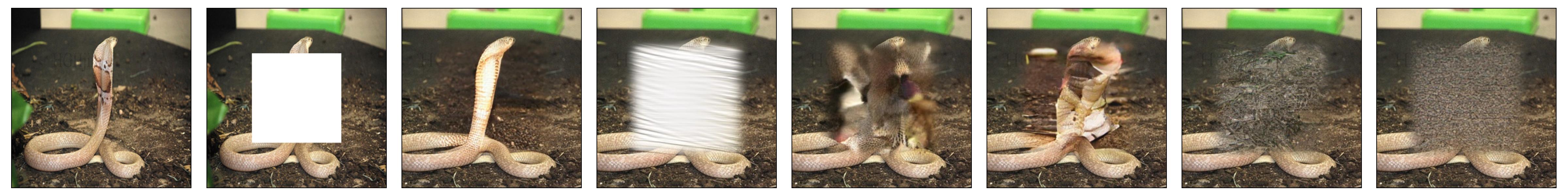}\vspace{-1mm}
     \end{subfigure}
     
     \begin{subfigure}[b]{\textwidth}
     \centering
     \includegraphics[width=\textwidth]{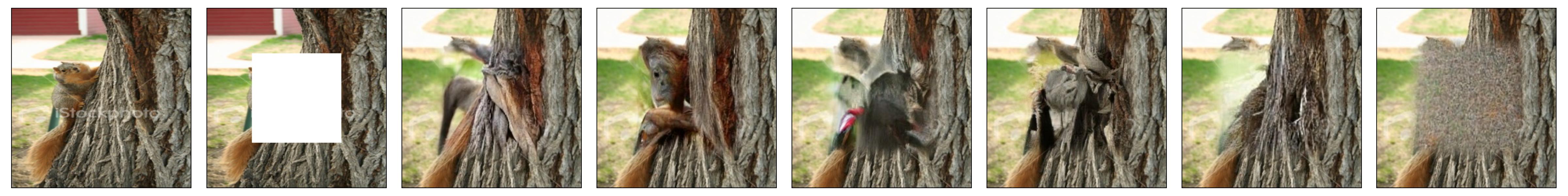}\vspace{-1mm}
     \end{subfigure}
    
     \begin{subfigure}[b]{\textwidth}
     \centering
     \includegraphics[width=\textwidth]{figure/caption_v3.pdf}\vspace{-2mm}
    \caption{{Forget set}}\vspace{1mm}
     \end{subfigure}

     \begin{subfigure}[b]{\textwidth}
     \centering
     \includegraphics[width=\textwidth]{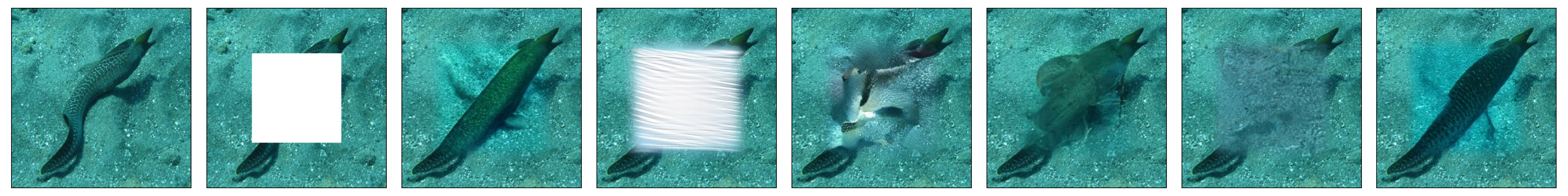}\vspace{-1mm}
     \end{subfigure}
     
     \begin{subfigure}[b]{\textwidth}
     \centering
     \includegraphics[width=\textwidth]{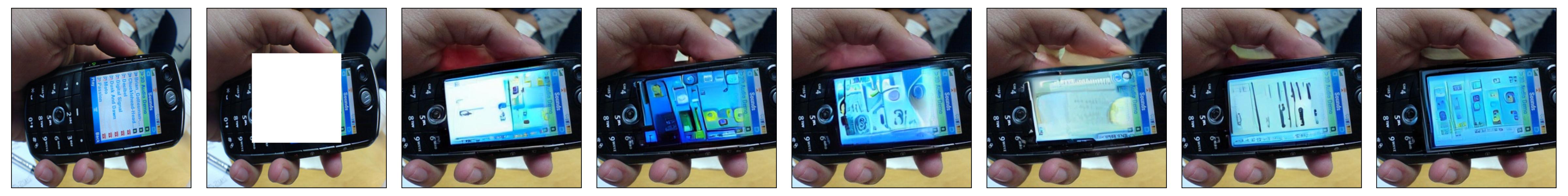}\vspace{-1mm}
     \end{subfigure}
     
     \begin{subfigure}[b]{\textwidth}
     \centering
     \includegraphics[width=\textwidth]{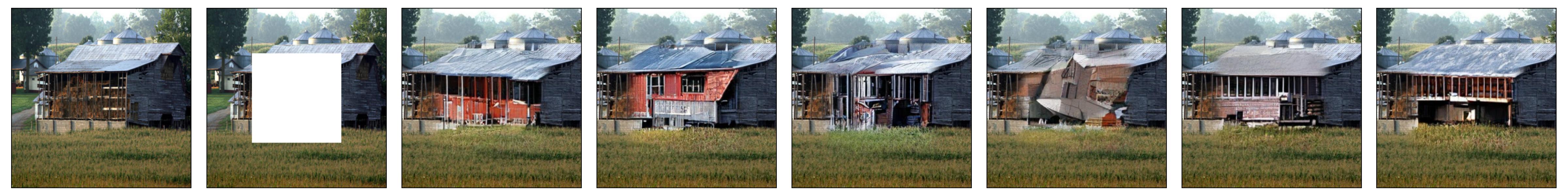}\vspace{-1mm}
     \end{subfigure}
     
     \begin{subfigure}[b]{\textwidth}
     \centering
     \includegraphics[width=\textwidth]{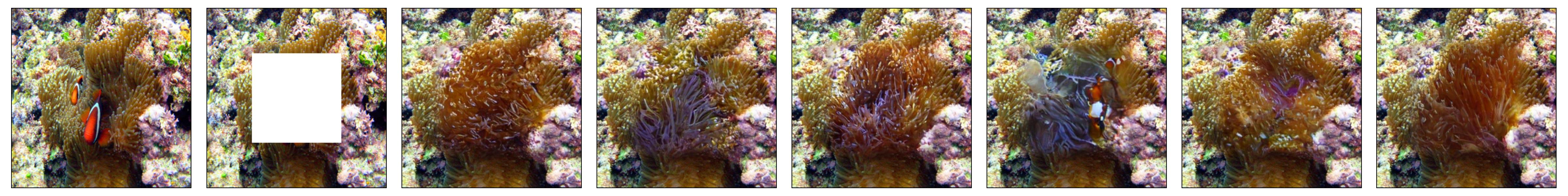}\vspace{-1mm}
     \end{subfigure}
    
     \begin{subfigure}[b]{\textwidth}
     \centering
     \includegraphics[width=\textwidth]{figure/caption_v3.pdf}\vspace{-2mm}
    \caption{{Retain set}}
     \end{subfigure}
 \caption{VQ-GAN: cropping $8\times 8$ patches at the center of the image, where each patch is $ 16 \times 16$ pixels. As shown, our approach has almost identical performance as the original model (\ie, before unlearning); the generated images on forget set are some random noise.   }
 \label{fig:gan_visual_center8by8}
\end{figure}

\subsection{VQ-GAN}
We provide the visualization results under center crop size of $4\times4$ patches in Fig.~\ref{fig:gan_visual_center4by4}.
We also provide more visualization results under center crop size of $8\times8$ patches in Fig.~\ref{fig:gan_visual_center8by8}. As shown in Fig.~\ref{fig:gan_visual_center4by4} and Fig.~\ref{fig:gan_visual_center8by8}, out approach can generate very high quality images on retain set while fill some random noise on the forget set. \textsc{Retain Label} cannot even forget the information on the forget set. 
Moreover, as shown in Fig.~\ref{fig:gan_visual_center4by4}, there are two type of fishes; one is in forget set and another one is in retain set (the fourth and fifth rows). Our approach can handle this subtle difference among difference classes.

\subsection{MAE}
\begin{figure}[htb]
 \centering
  \begin{subfigure}[b]{0.32\textwidth}
 \centering
 \includegraphics[width=\textwidth]{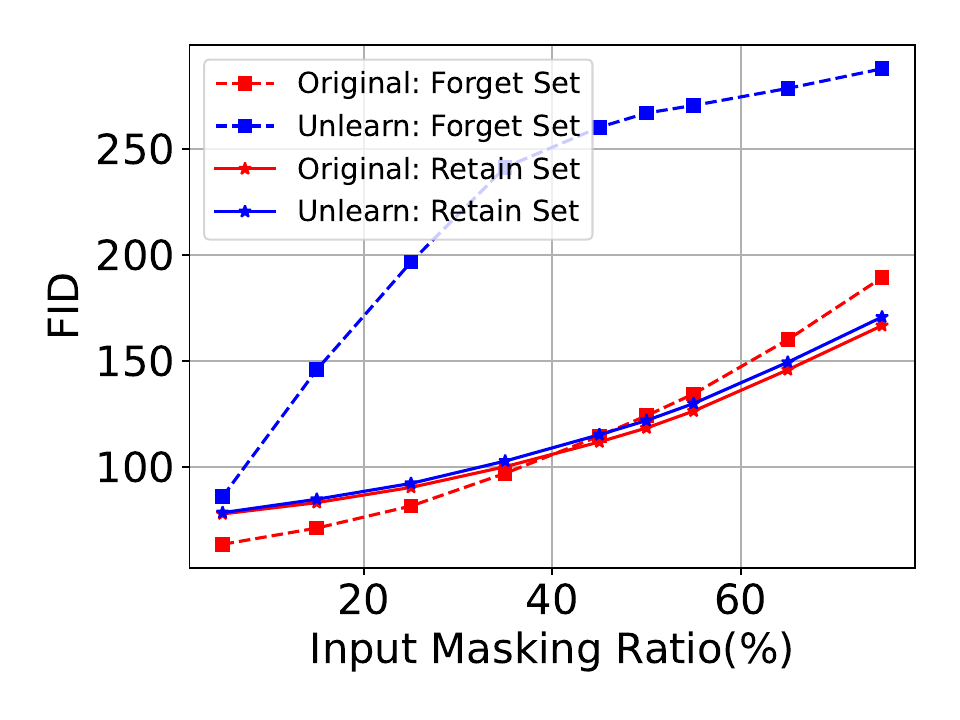}
 \caption{MAE: FID}
 \end{subfigure}
 \hfill
 \begin{subfigure}[b]{0.32\textwidth}
 \centering
 \includegraphics[width=\textwidth]{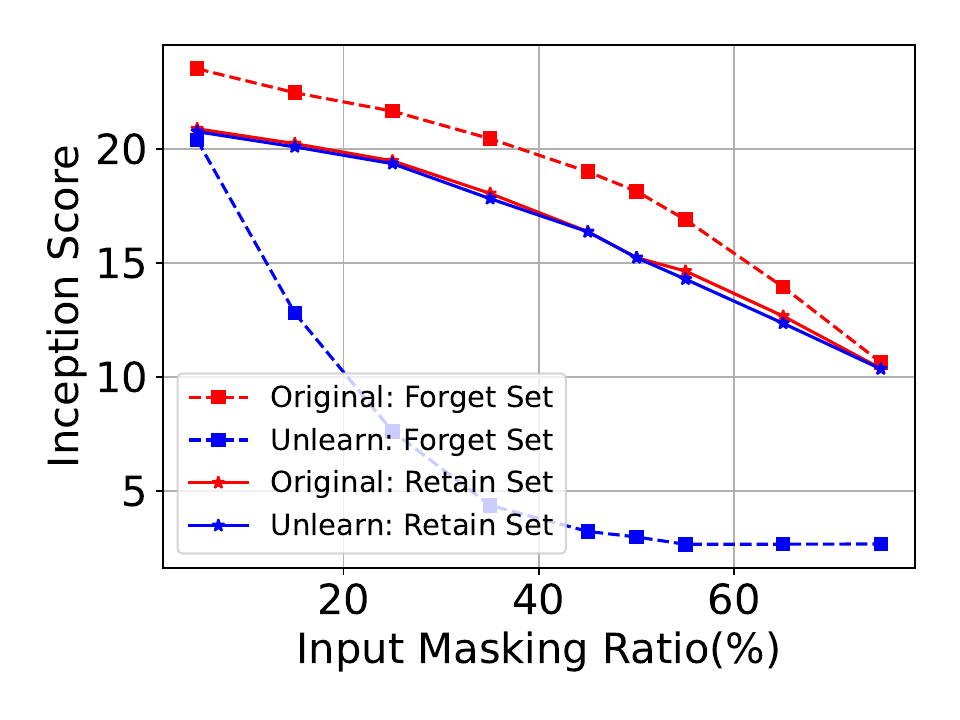}
 \caption{MAE: IS}
 \end{subfigure}
  \hfill
 \begin{subfigure}[b]{0.32\textwidth}
 \centering
 \includegraphics[width=\textwidth]{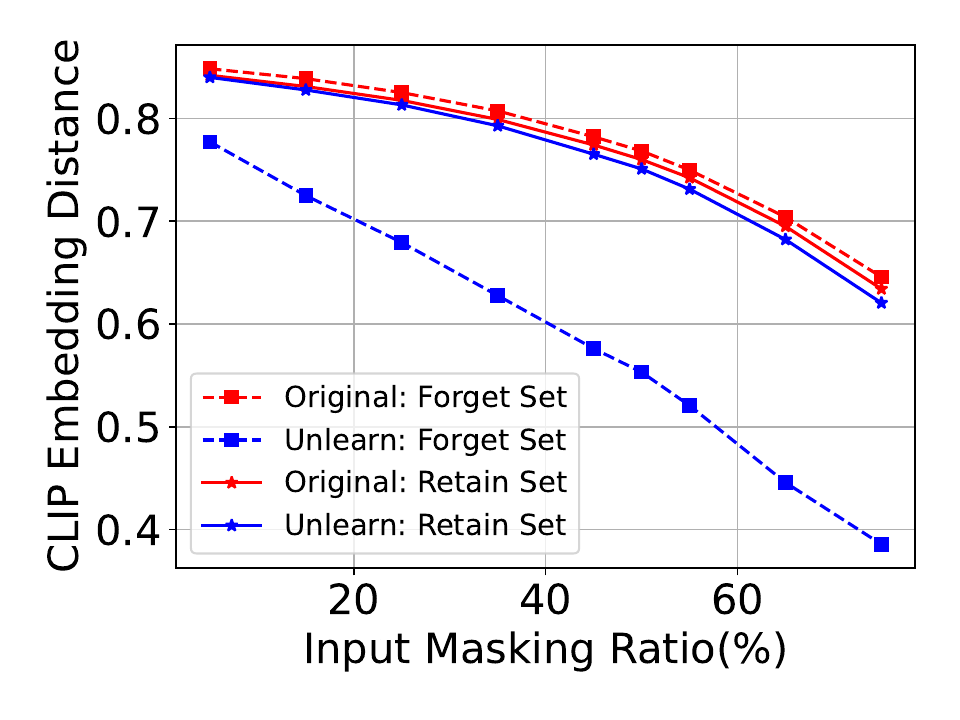}
 \caption{MAE: CLIP}
 \end{subfigure}
 \caption{The quality of reconstructed images given random masked images by MAE under varying random masking ratio; \eg, masking $128$ out of 256 patches means 50\% cropping ratio. We compare the original model and the unlearned model by our approach. As shown, the performance of these two models on the retain set are almost identical, while the unlearned model has a significant performance drop on the forget set. See Fig.~\ref{app:random_masking} for some examples of generated images.}
 \label{fig:mae_random_masking_metric}
\end{figure}

\begin{figure}[htb]\centering
    \begin{subfigure}[b]{0.495\textwidth}
    \centering
    \includegraphics[width=\textwidth]{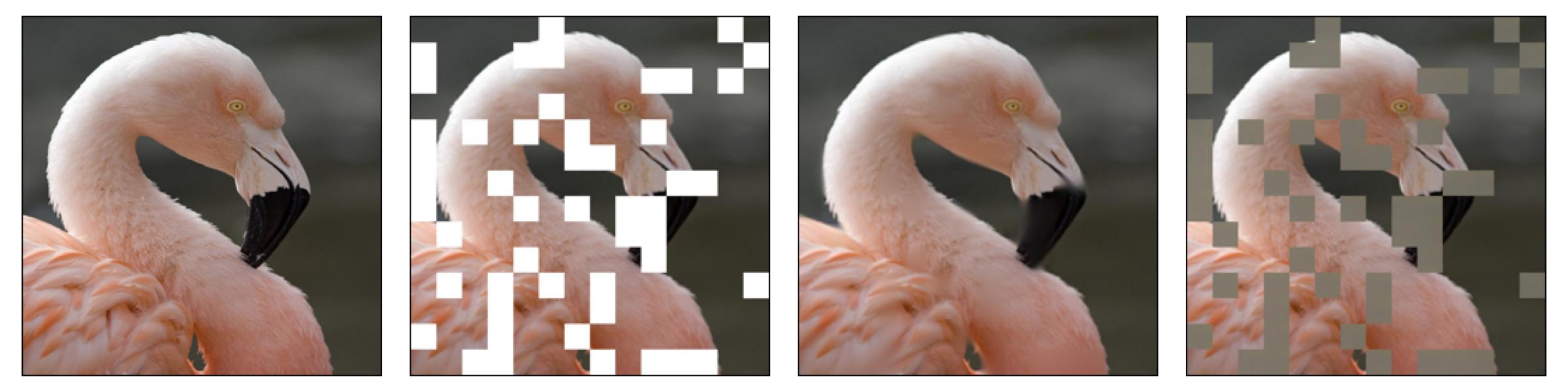}
    \end{subfigure}
    \hfill
    \begin{subfigure}[b]{0.495\textwidth}
    \centering
    \includegraphics[width=\textwidth]{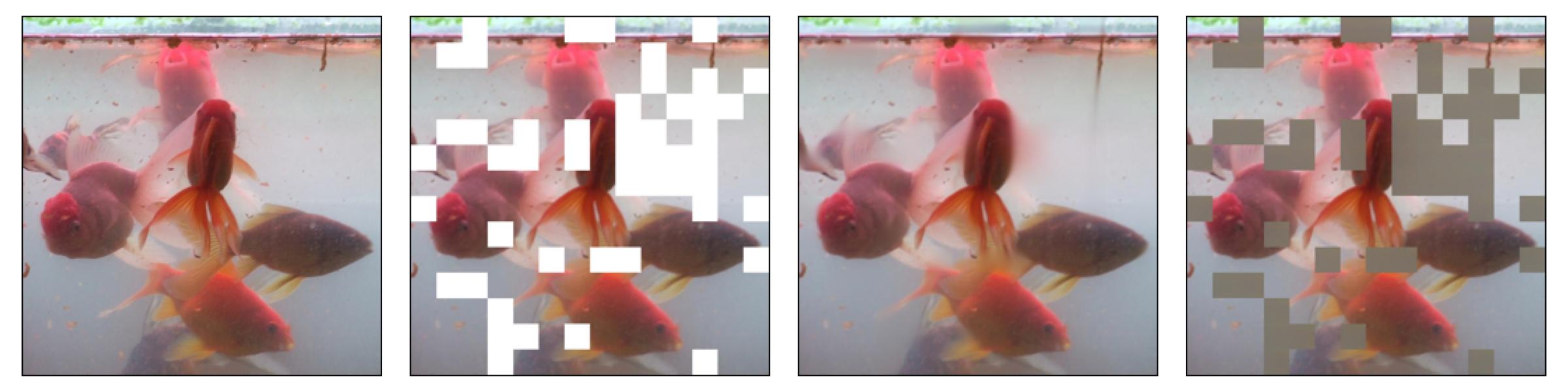}
    \end{subfigure}
    
    \begin{subfigure}[b]{0.495\textwidth}
    \centering
    \includegraphics[width=\textwidth]{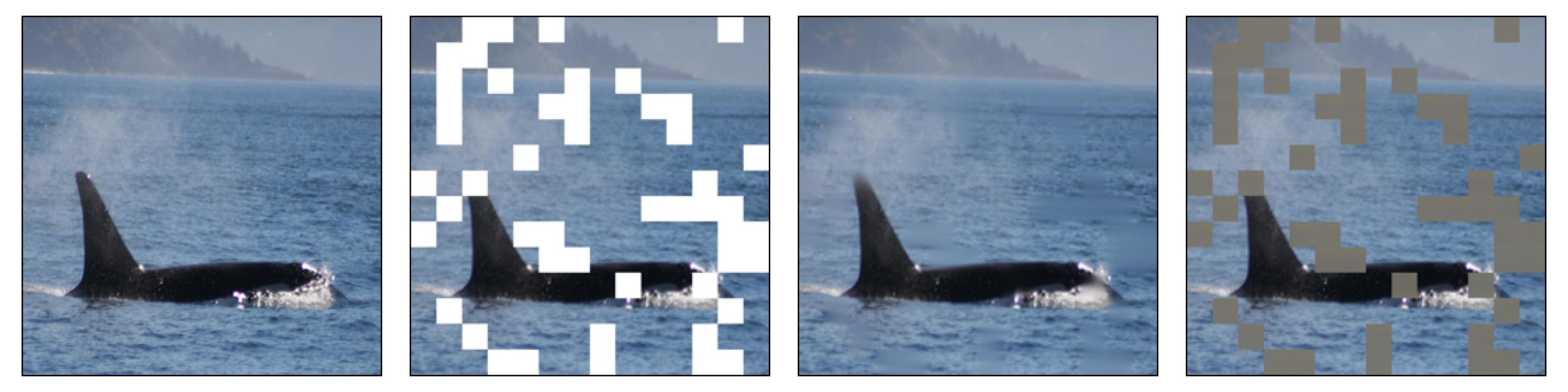}
    \end{subfigure}
    \hfill
    \begin{subfigure}[b]{0.495\textwidth}
    \centering
    \includegraphics[width=\textwidth]{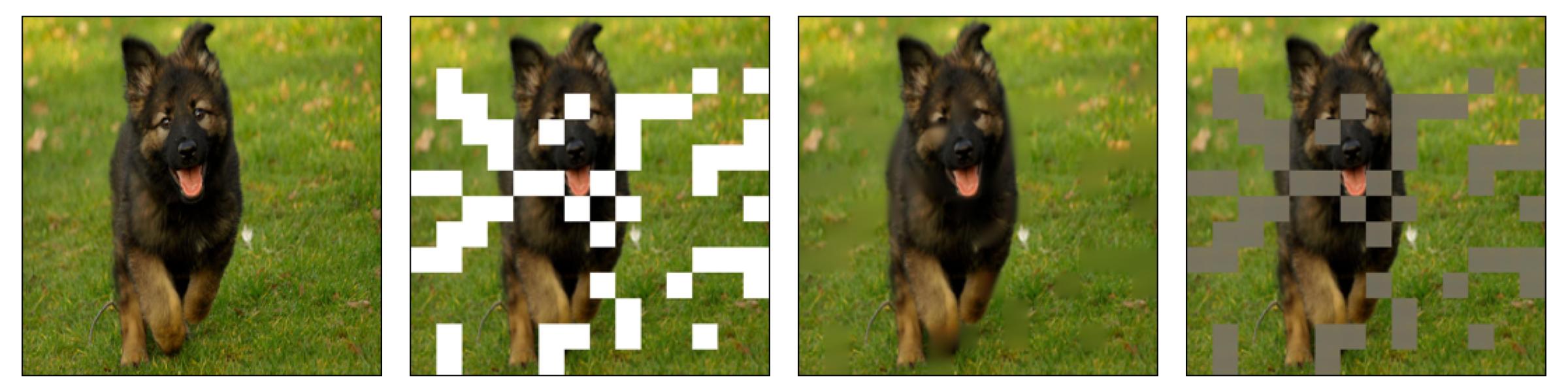}
    \end{subfigure}

    \begin{subfigure}[b]{0.495\textwidth}
    \centering
    \includegraphics[width=\textwidth]{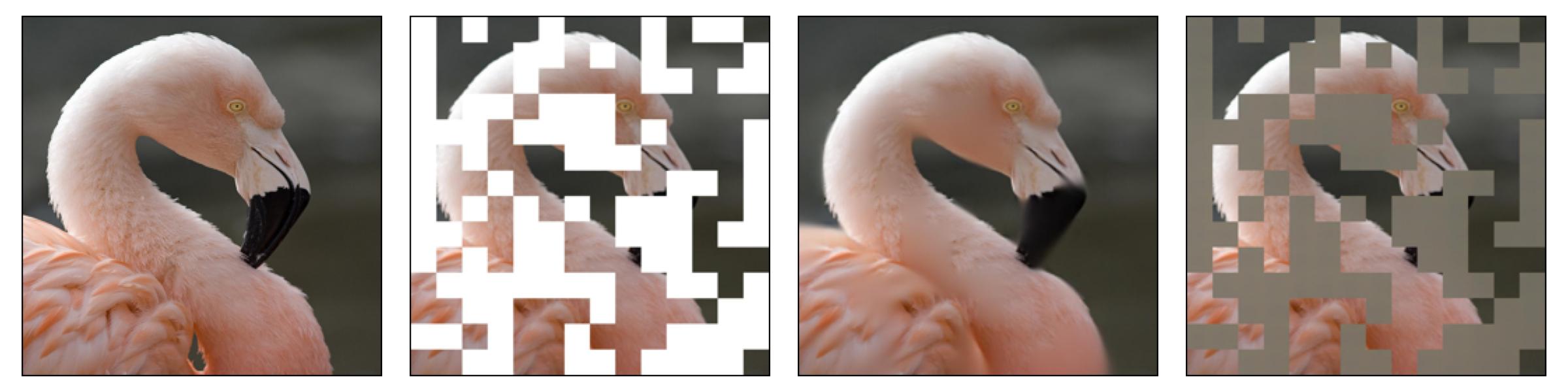}
    \end{subfigure}
    \hfill
    \begin{subfigure}[b]{0.495\textwidth}
    \centering
    \includegraphics[width=\textwidth]{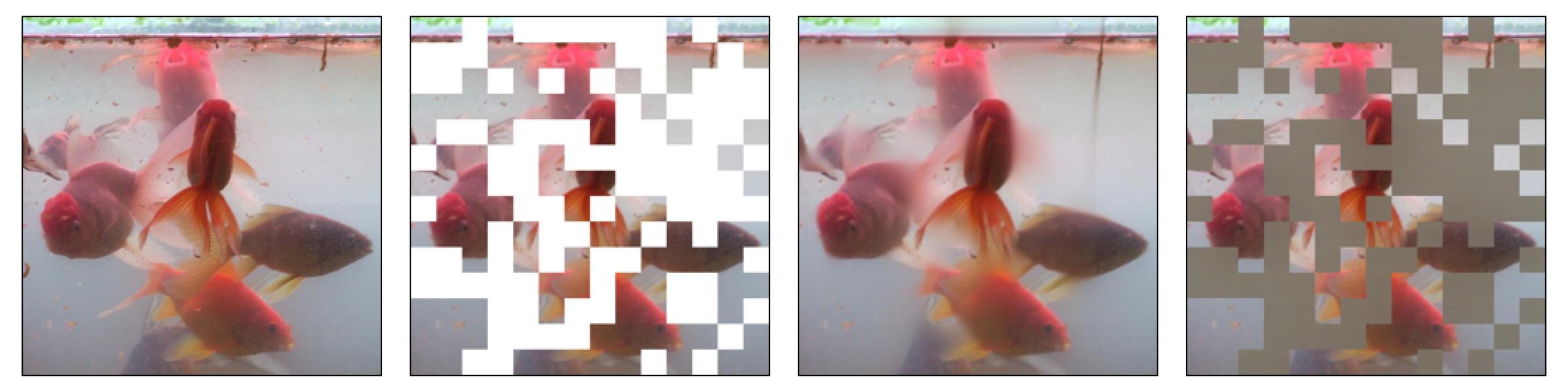}
    \end{subfigure}
    
    \begin{subfigure}[b]{0.495\textwidth}
    \centering
    \includegraphics[width=\textwidth]{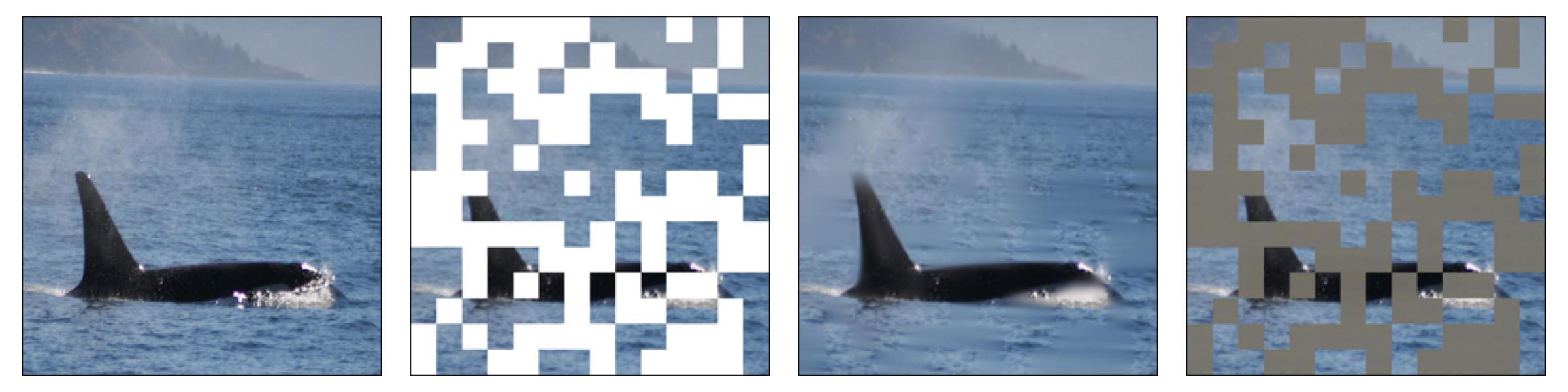}
    \end{subfigure}
    \hfill
    \begin{subfigure}[b]{0.495\textwidth}
    \centering
    \includegraphics[width=\textwidth]{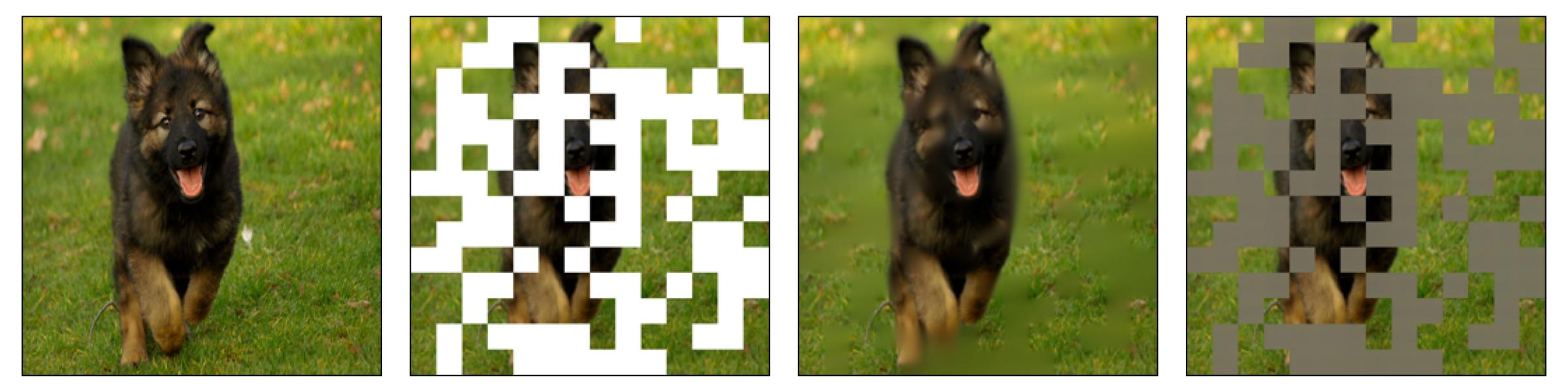}
    \end{subfigure}
    
    \begin{subfigure}[b]{\textwidth}
    \centering
    \includegraphics[width=\textwidth]{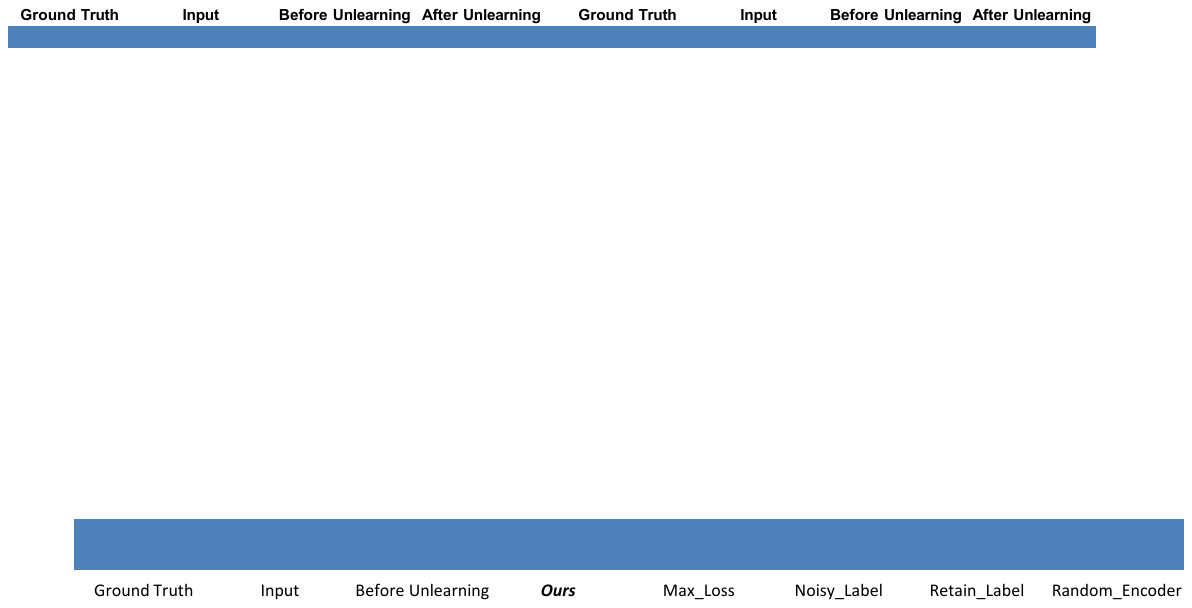}\vspace{-1mm}
    \caption{{Forget set}}
    \end{subfigure}
    \vspace{1mm}

    \begin{subfigure}[b]{0.495\textwidth}
    \centering
    \includegraphics[width=\textwidth]{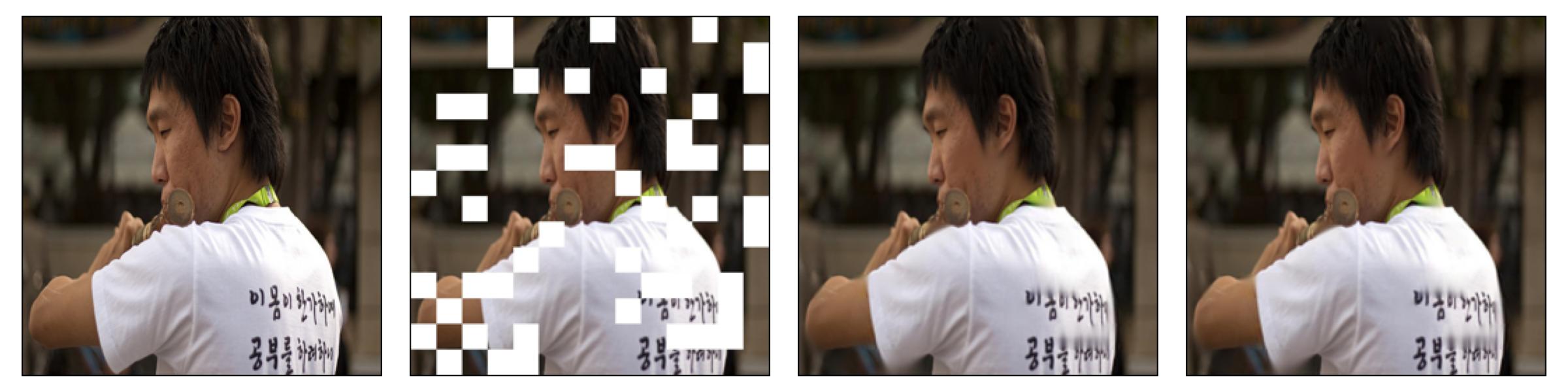}
    \end{subfigure}
    \hfill
    \begin{subfigure}[b]{0.495\textwidth}
    \centering
    \includegraphics[width=\textwidth]{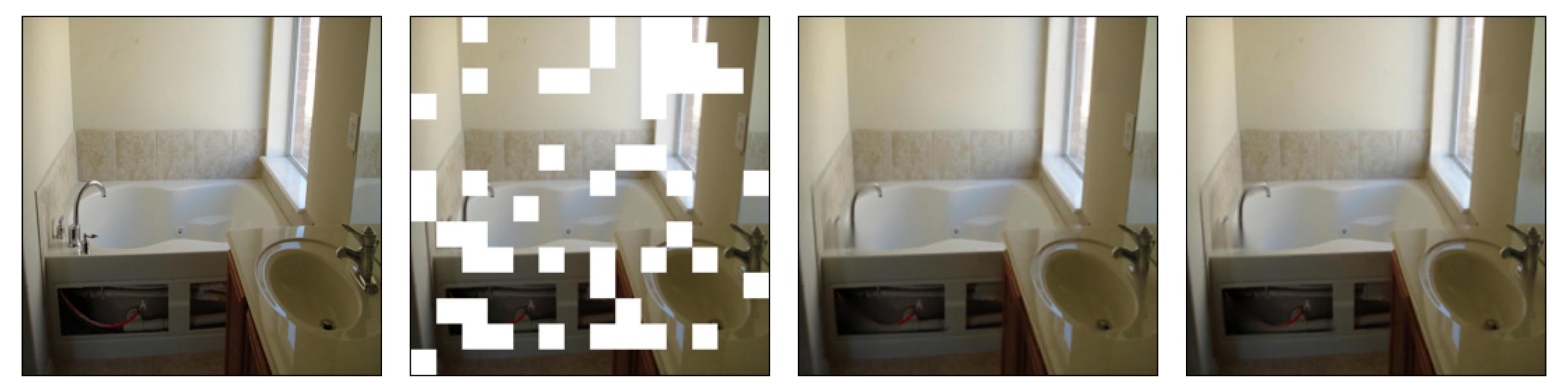}
    \end{subfigure}
    
    \begin{subfigure}[b]{0.495\textwidth}
    \centering
    \includegraphics[width=\textwidth]{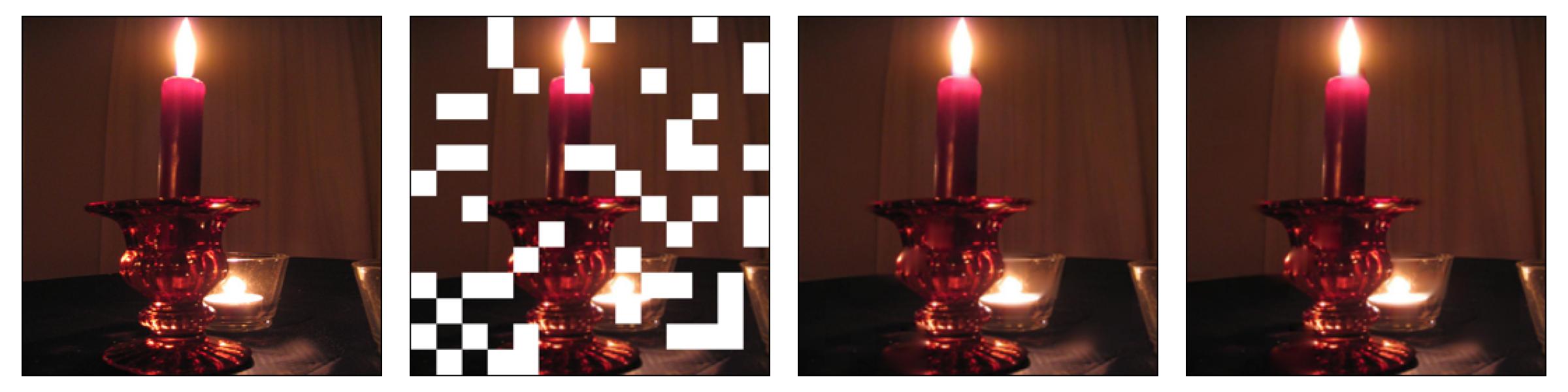}
    \end{subfigure}
    \hfill
    \begin{subfigure}[b]{0.495\textwidth}
    \centering
    \includegraphics[width=\textwidth]{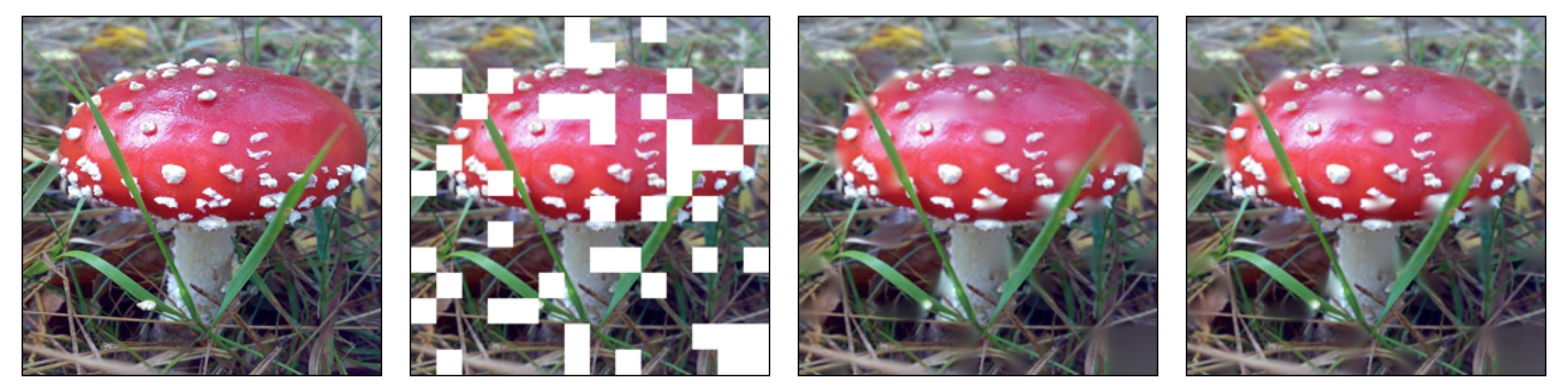}
    \end{subfigure}
    
    \begin{subfigure}[b]{0.495\textwidth}
    \centering
    \includegraphics[width=\textwidth]{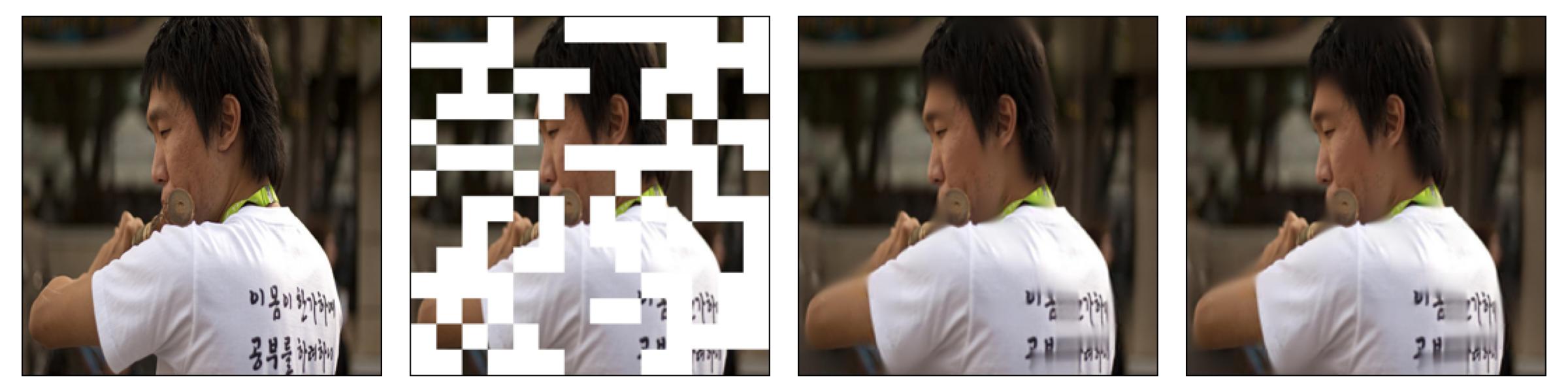}
    \end{subfigure}
    \hfill
    \begin{subfigure}[b]{0.495\textwidth}
    \centering
    \includegraphics[width=\textwidth]{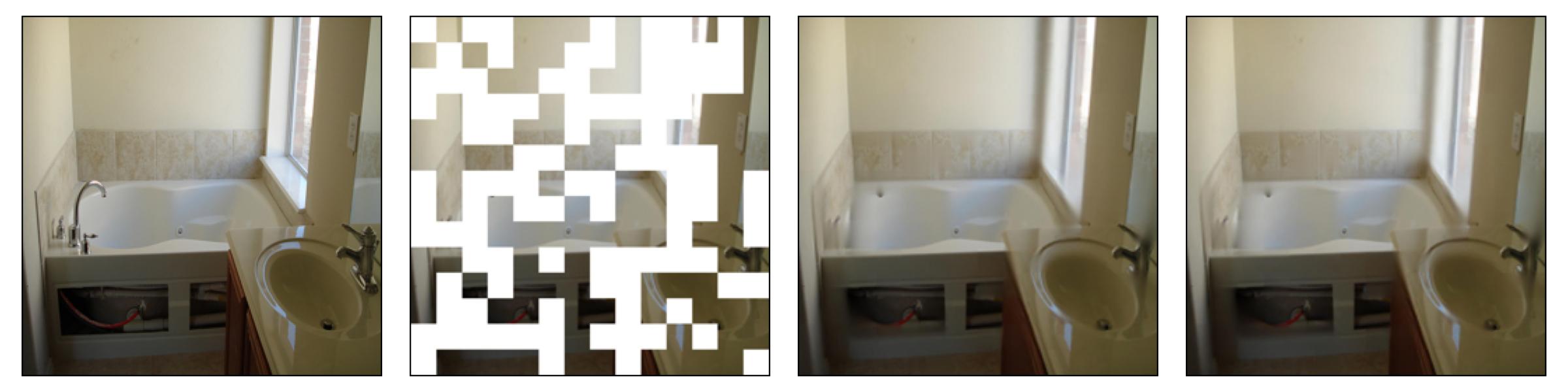}
    \end{subfigure}
    
    \begin{subfigure}[b]{0.495\textwidth}
    \centering
    \includegraphics[width=\textwidth]{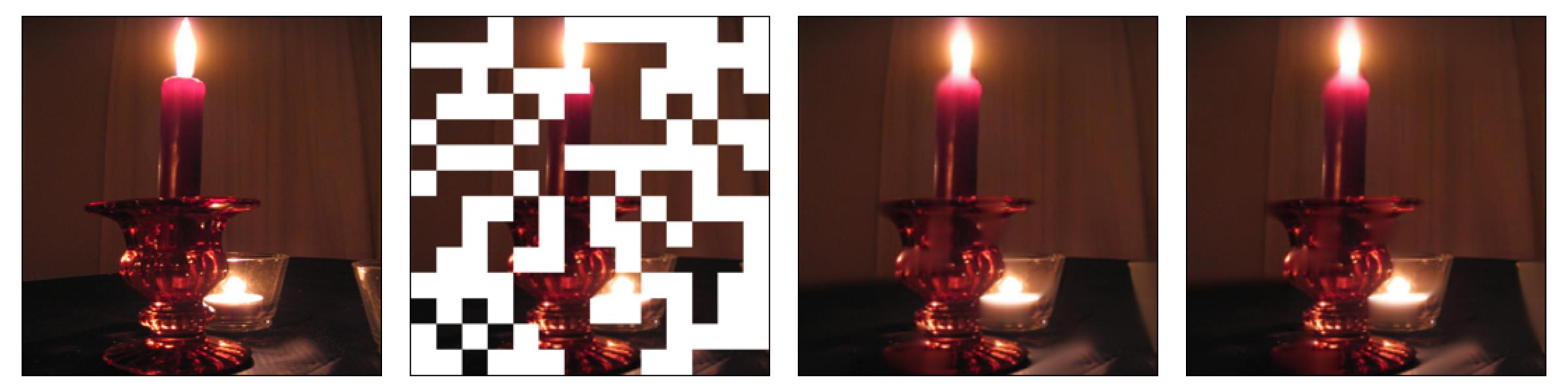}
    \end{subfigure}
    \hfill
    \begin{subfigure}[b]{0.495\textwidth}
    \centering
    \includegraphics[width=\textwidth]{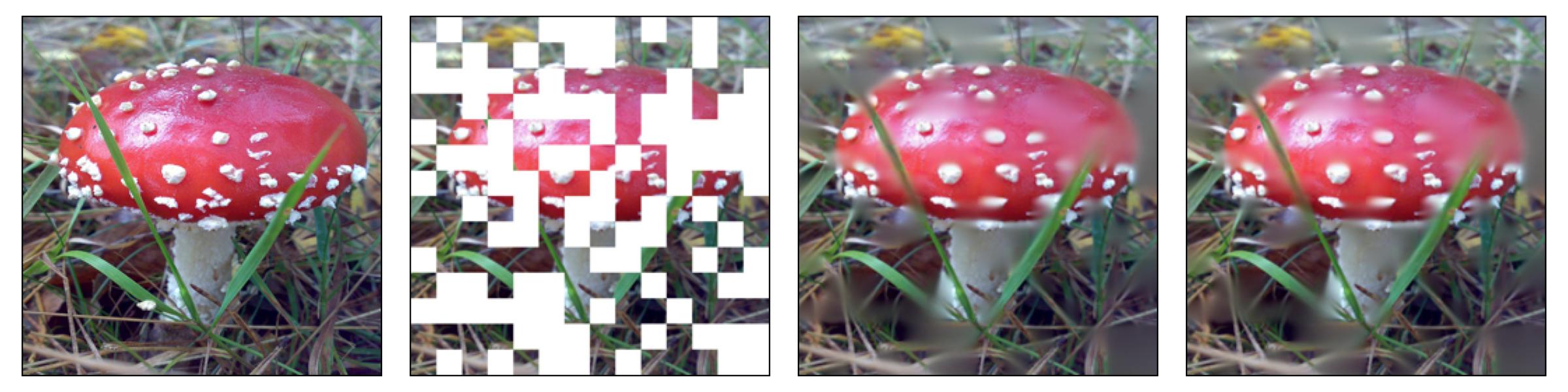}
    \end{subfigure}
    
    \begin{subfigure}[b]{\textwidth}
    \centering
    \includegraphics[width=\textwidth]{figure/caption_four.pdf}\vspace{-1mm}
    \caption{{Retain set}}
    \end{subfigure}
\caption{Reconstruction of random masked images by MAE. For a single image, we test 25\% masking ratio and 50\% masking ratio. }
\label{fig:mae_visual_random}
\end{figure}
\paragraph{Random Masking}\label{app:random_masking}
In the original paper~\citep{mae_masked_he}, MAE primarily focuses on the reconstruction of randomly masked images. Hence, besides the center cropping reported in the main paper, we also report the results under random masking. As shown in Fig.~\ref{fig:mae_random_masking_metric} and Fig.~\ref{fig:mae_visual_random}, the performance on the retain set of the unlearned model is almost identical as original model. In  contrast, there is a significant performance drop on the forget set.

\section{Ablation study}\label{app:ablation_study}
\def\theequation{E.\arabic{equation}}
\def\thelem{E.\arabic{lem}}
\def\thefigure{E.\arabic{figure}}
\def\thetable{E.\arabic{table}}

\begin{figure}[htb]\centering
    \begin{subfigure}[b]{0.495\textwidth}
    \centering
    \includegraphics[width=\textwidth]{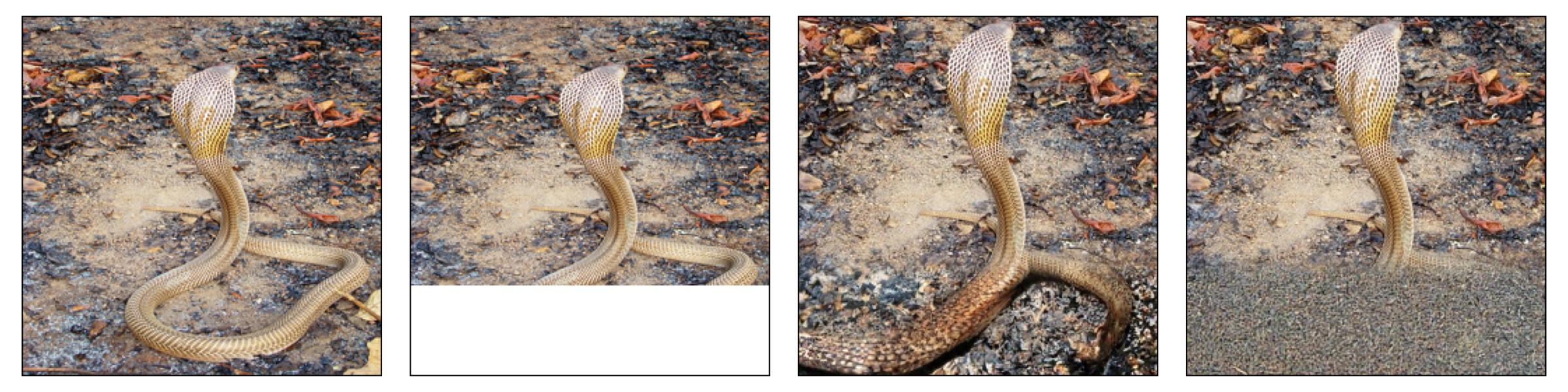}
    \end{subfigure}
    \hfill   
    \begin{subfigure}[b]{0.495\textwidth}
    \centering
    \includegraphics[width=\textwidth]{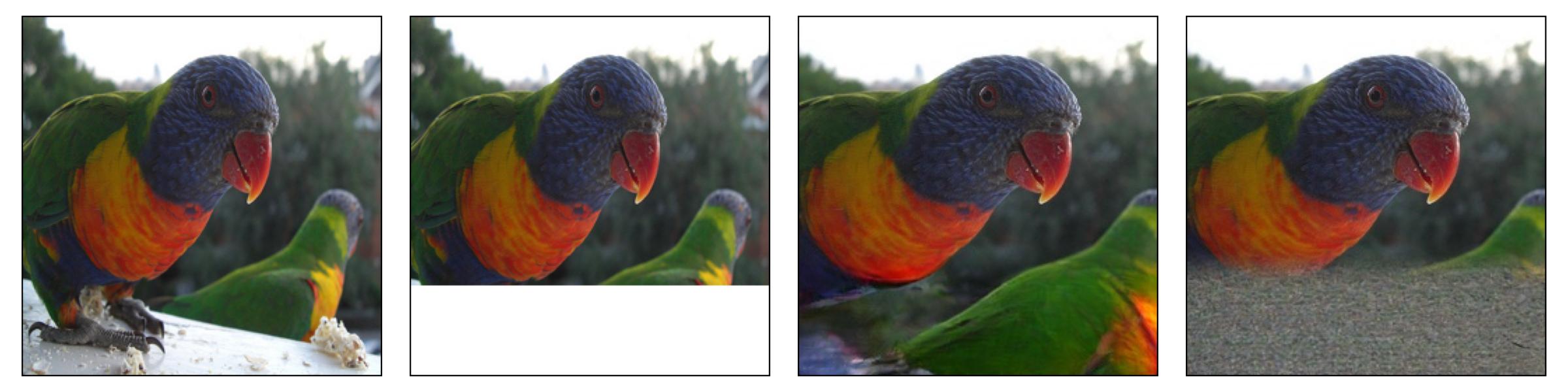}
    \end{subfigure}

    \begin{subfigure}[b]{0.495\textwidth}
    \centering
    \includegraphics[width=\textwidth]{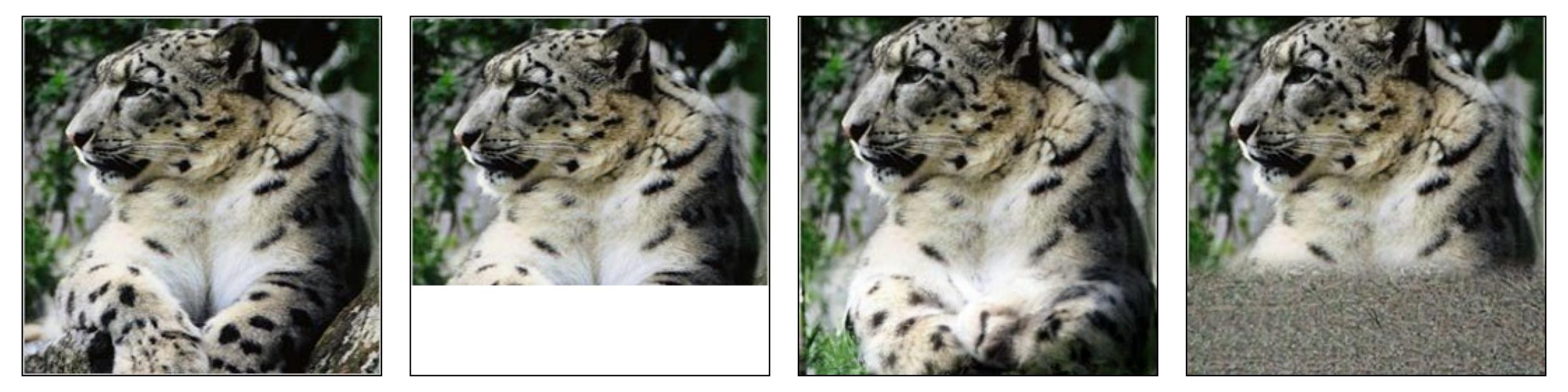}
    \end{subfigure}
    \hfill
    \begin{subfigure}[b]{0.495\textwidth}
    \centering
    \includegraphics[width=\textwidth]{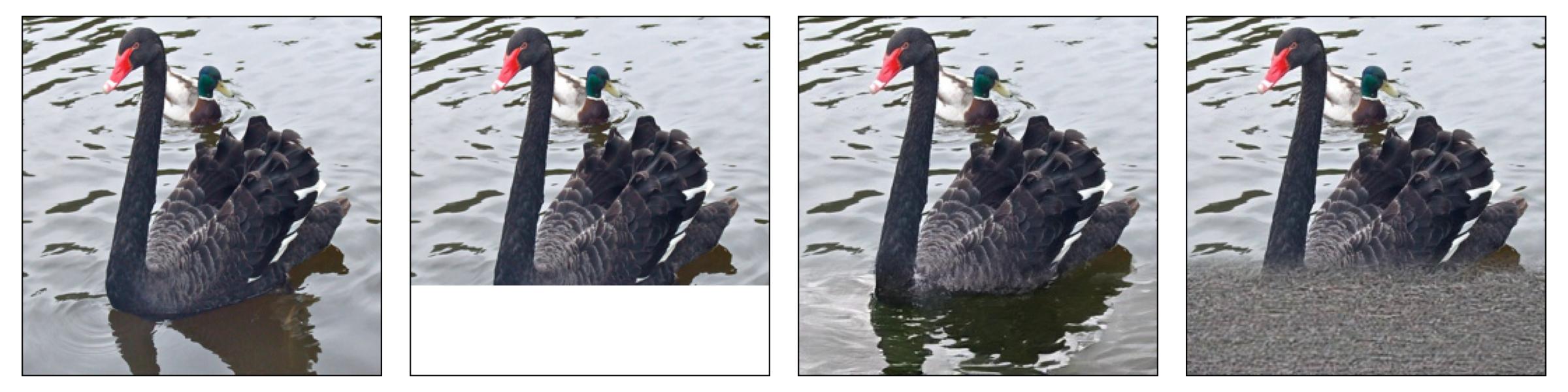}
    \end{subfigure}

    \begin{subfigure}[b]{0.495\textwidth}
    \centering
    \includegraphics[width=\textwidth]{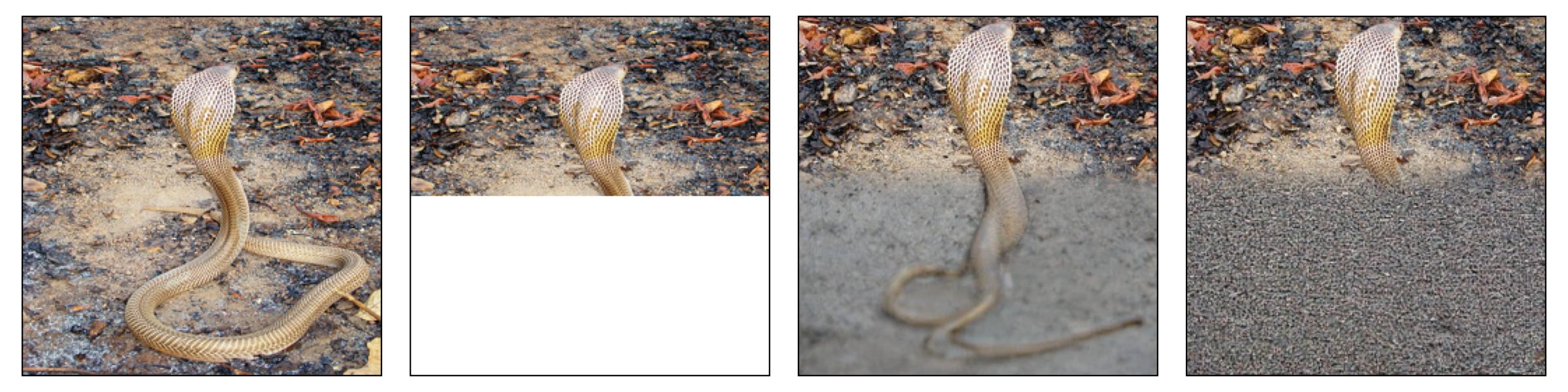}
    \end{subfigure}
    \hfill
    \begin{subfigure}[b]{0.495\textwidth}
    \centering
    \includegraphics[width=\textwidth]{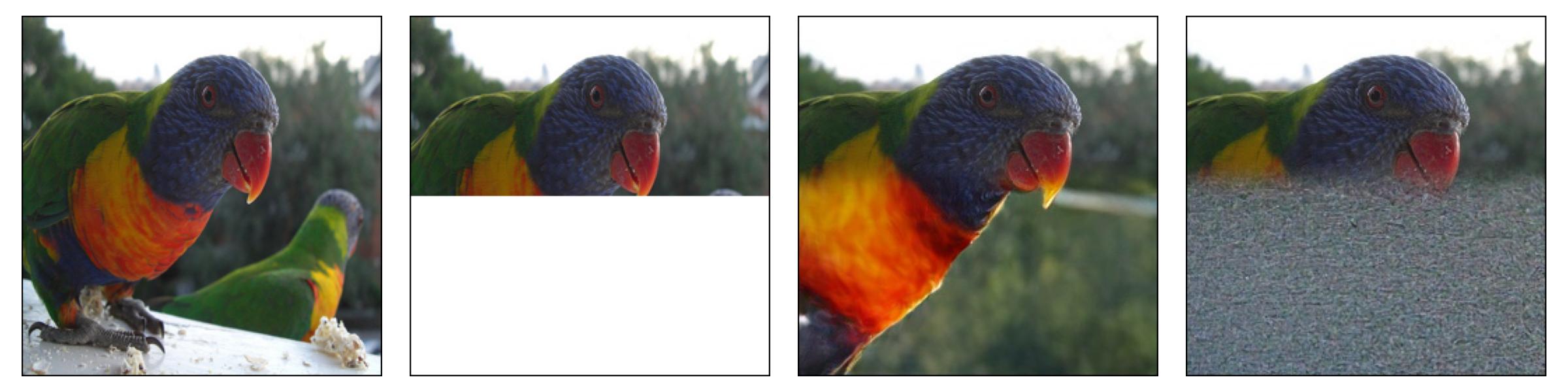}
    \end{subfigure}
    
    \begin{subfigure}[b]{0.495\textwidth}
    \centering
    \includegraphics[width=\textwidth]{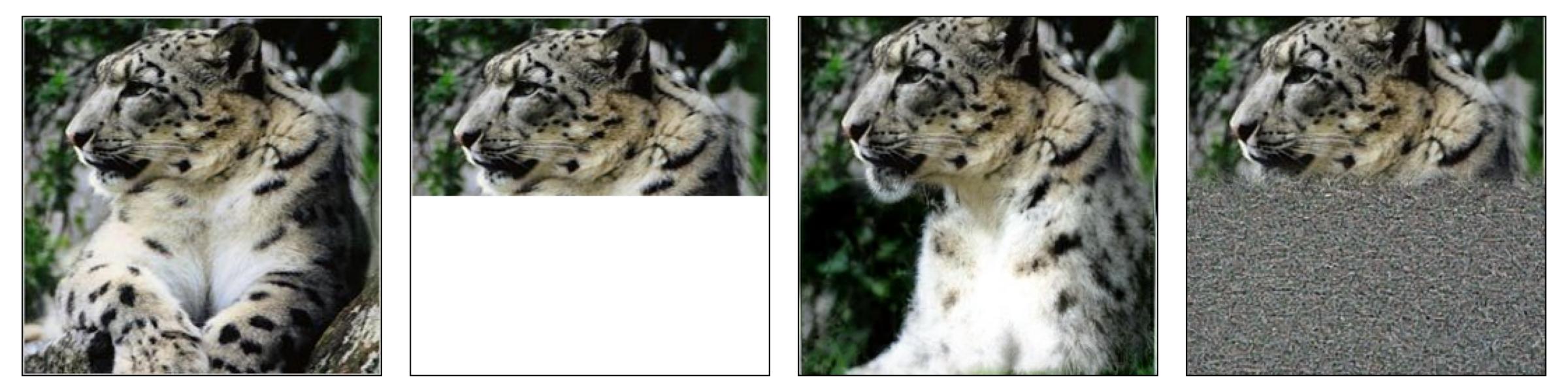}
    \end{subfigure}
    \hfill
    \begin{subfigure}[b]{0.495\textwidth}
    \centering
    \includegraphics[width=\textwidth]{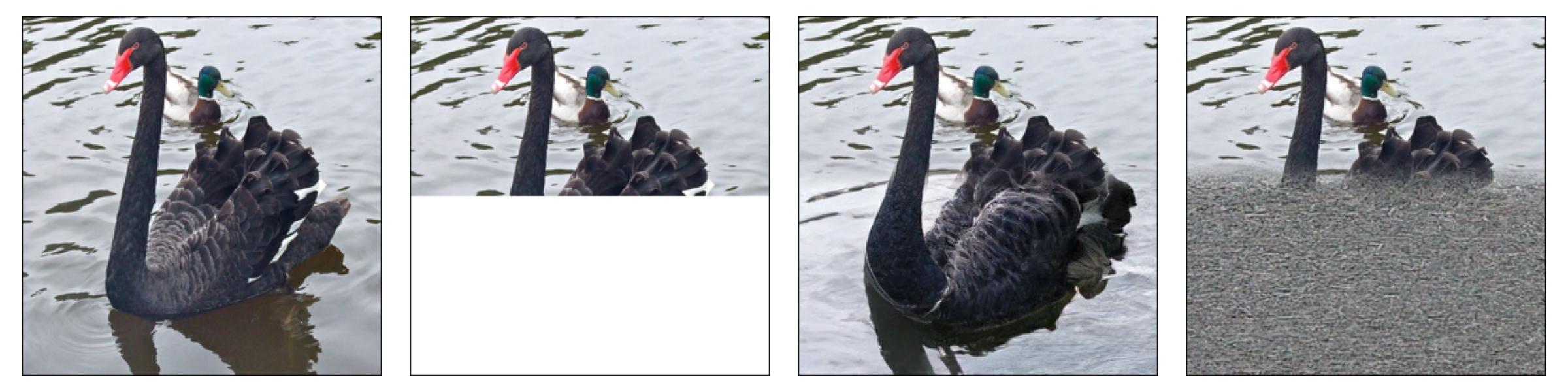}
    \end{subfigure}
    
    \begin{subfigure}[b]{\textwidth}
    \centering
    \includegraphics[width=\textwidth]{figure/caption_four.pdf}\vspace{-1mm}
    \caption{{Forget set}}
    \end{subfigure}
    \vspace{1mm}

    \begin{subfigure}[b]{0.495\textwidth}
    \centering
    \includegraphics[width=\textwidth]{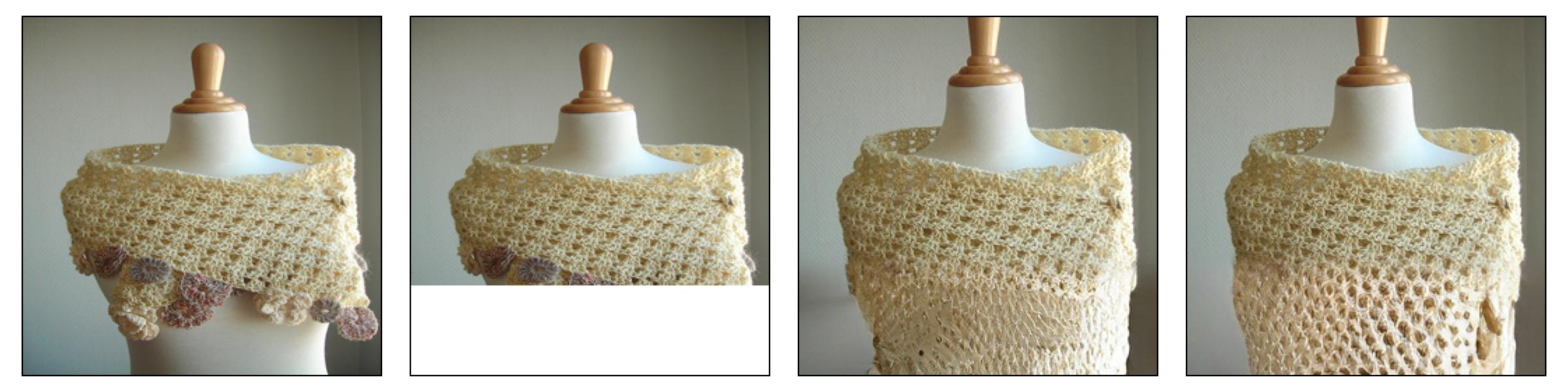}
    \end{subfigure}
    \hfill
    \begin{subfigure}[b]{0.495\textwidth}
    \centering
    \includegraphics[width=\textwidth]{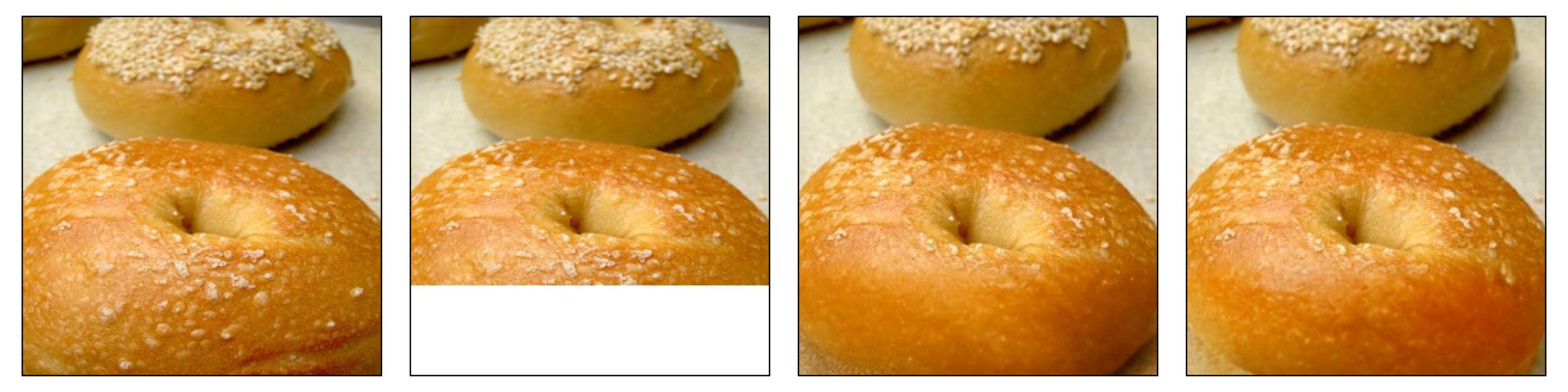}
    \end{subfigure}
 
    \begin{subfigure}[b]{0.495\textwidth}
    \centering
    \includegraphics[width=\textwidth]{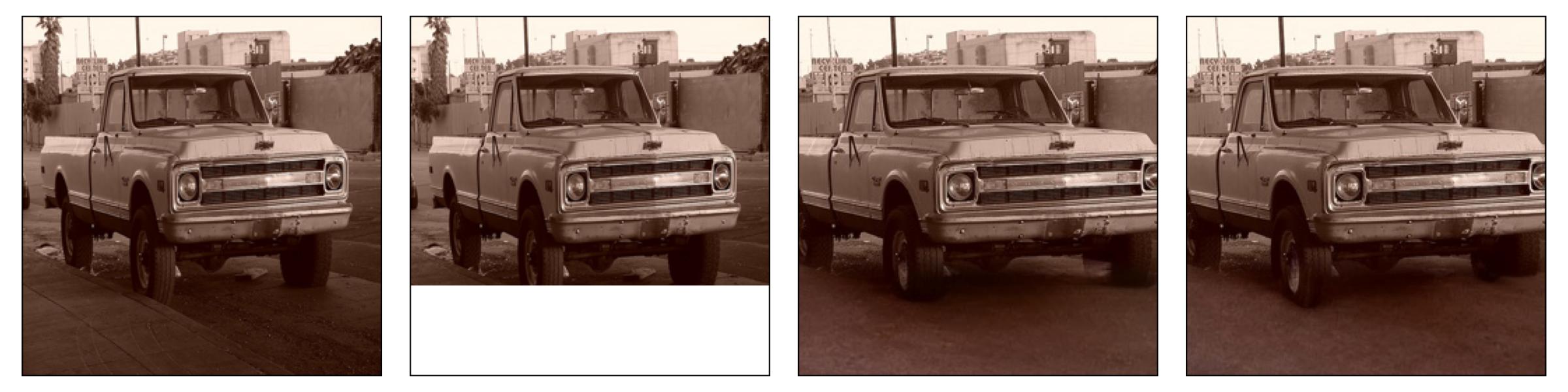}
    \end{subfigure}
    \hfill
    \begin{subfigure}[b]{0.495\textwidth}
    \centering
    \includegraphics[width=\textwidth]{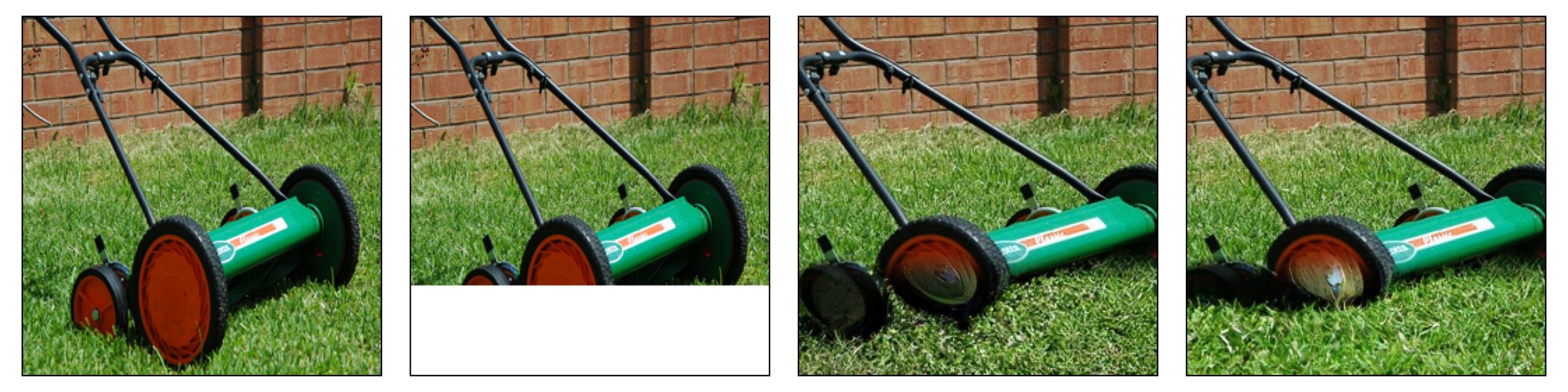}
    \end{subfigure}

    \begin{subfigure}[b]{0.495\textwidth}
    \centering
    \includegraphics[width=\textwidth]{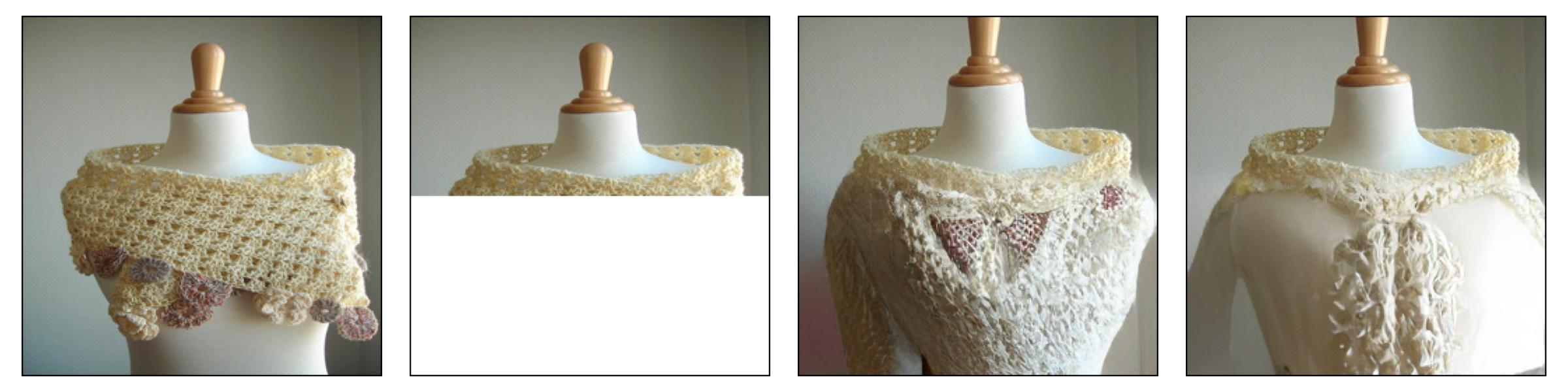}
    \end{subfigure}    
    \hfill
    \begin{subfigure}[b]{0.495\textwidth}
    \centering
    \includegraphics[width=\textwidth]{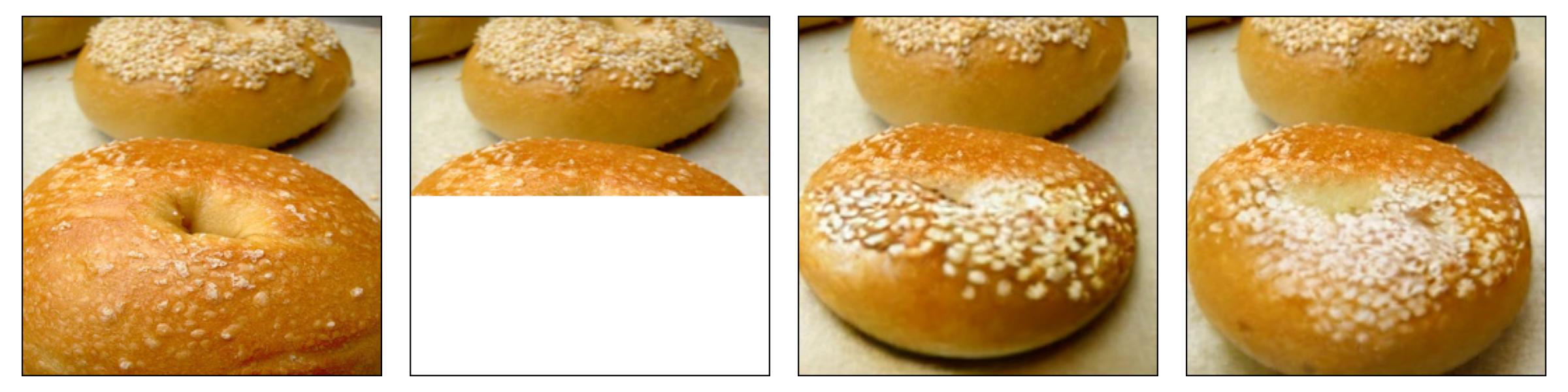}
    \end{subfigure}
    
    \begin{subfigure}[b]{0.495\textwidth}
    \centering
    \includegraphics[width=\textwidth]{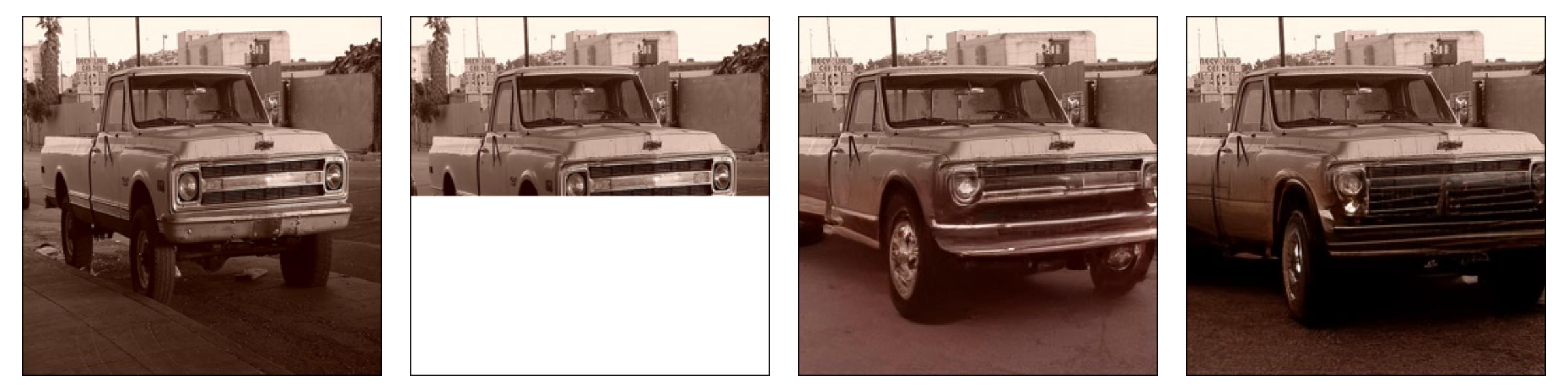}
    \end{subfigure}
    \hfill
    \begin{subfigure}[b]{0.495\textwidth}
    \centering
    \includegraphics[width=\textwidth]{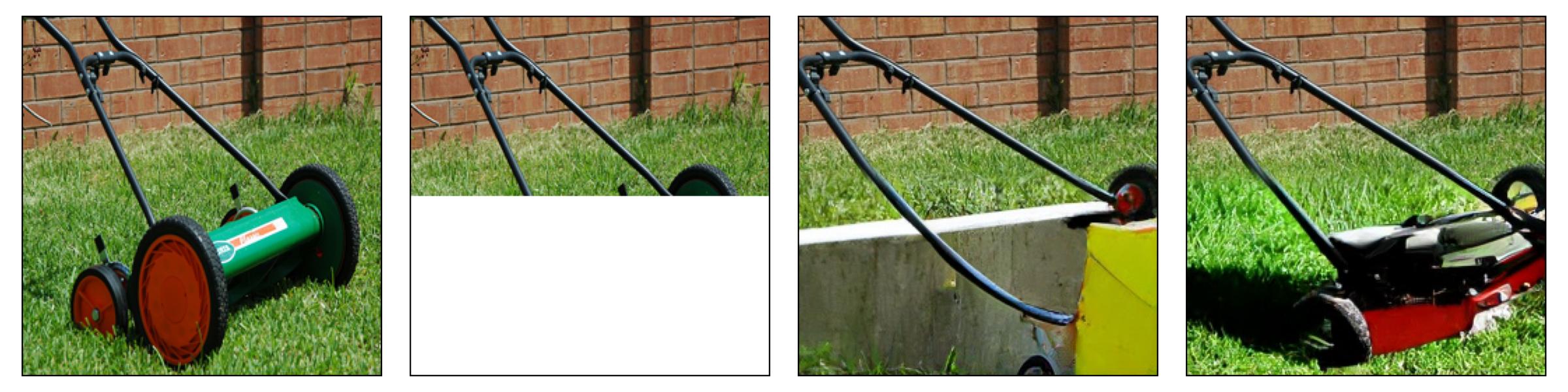}
    \end{subfigure}
    
    \begin{subfigure}[b]{\textwidth}
    \centering
    \includegraphics[width=\textwidth]{figure/caption_four.pdf}\vspace{-1mm}
    \caption{{Retain set}}
    \end{subfigure}
\caption{Ablation study: Downward extension by VQ-GAN. We visualize he performance of the original model (before unlearning) and the obtained model by our approach (after unlearning). We show results on both forget set and retain set. }
\label{fig:gan_down}
\end{figure}

\begin{figure}[htb]\centering
    \begin{subfigure}[b]{0.495\textwidth}
    \centering
    \includegraphics[width=\textwidth]{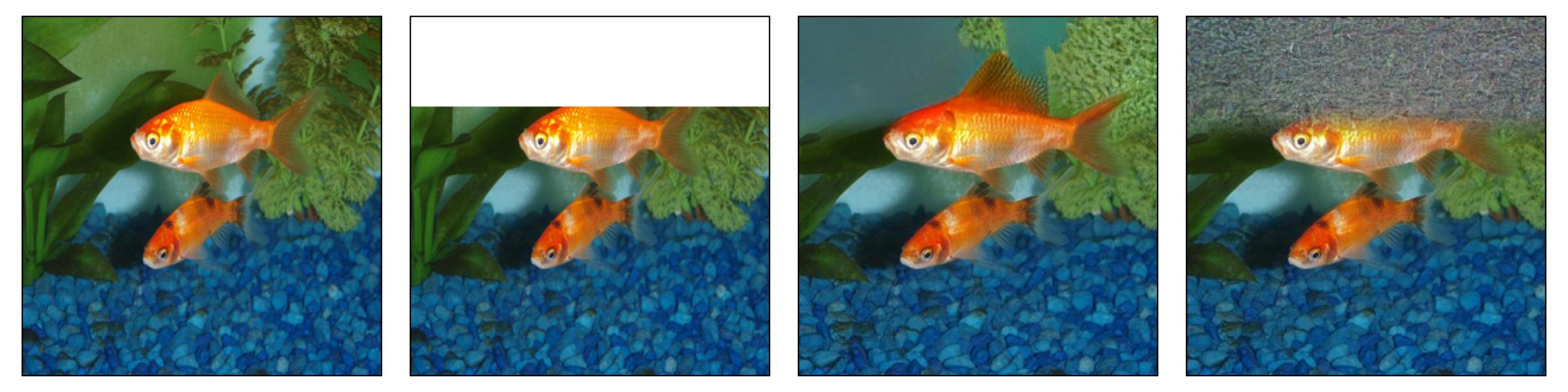}
    \end{subfigure}
    \hfill
    \begin{subfigure}[b]{0.495\textwidth}
    \centering
    \includegraphics[width=\textwidth]{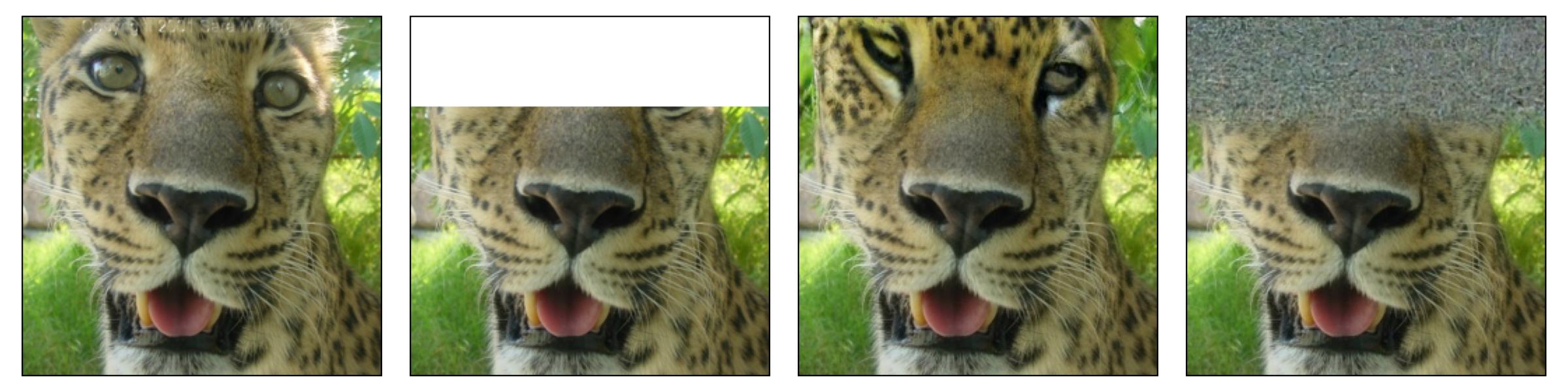}
    \end{subfigure}
    
    \begin{subfigure}[b]{0.495\textwidth}
    \centering
    \includegraphics[width=\textwidth]{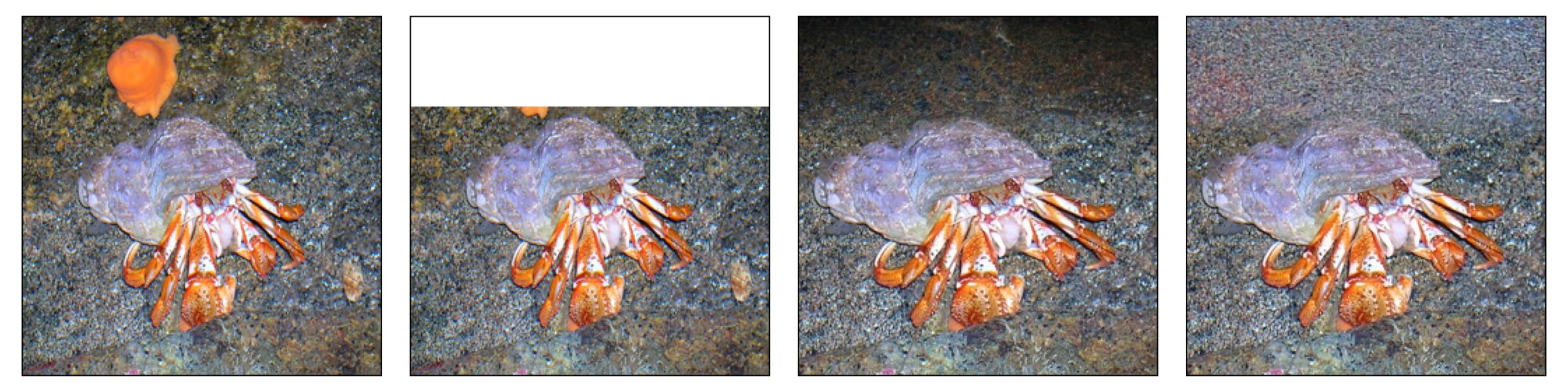}
    \end{subfigure}
    \hfill
    \begin{subfigure}[b]{0.495\textwidth}
    \centering
    \includegraphics[width=\textwidth]{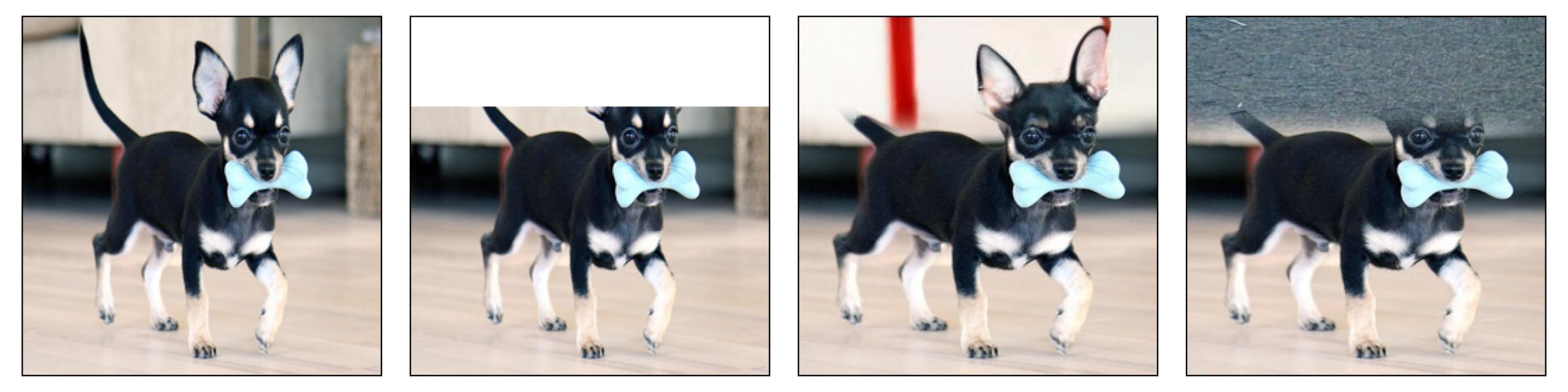}
    \end{subfigure}

    \begin{subfigure}[b]{0.495\textwidth}
    \centering
    \includegraphics[width=\textwidth]{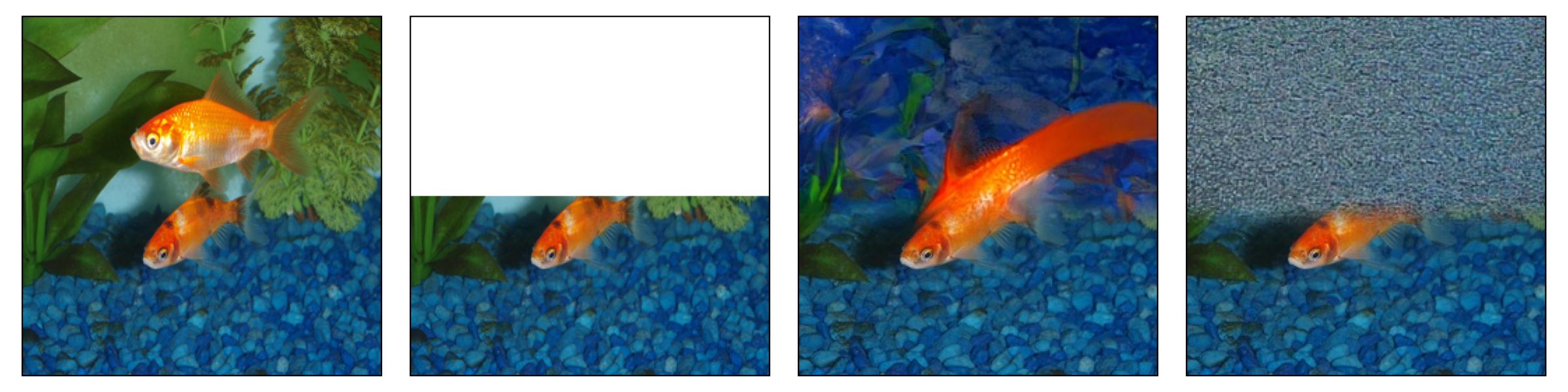}
    \end{subfigure}
    \hfill
    \begin{subfigure}[b]{0.495\textwidth}
    \centering
    \includegraphics[width=\textwidth]{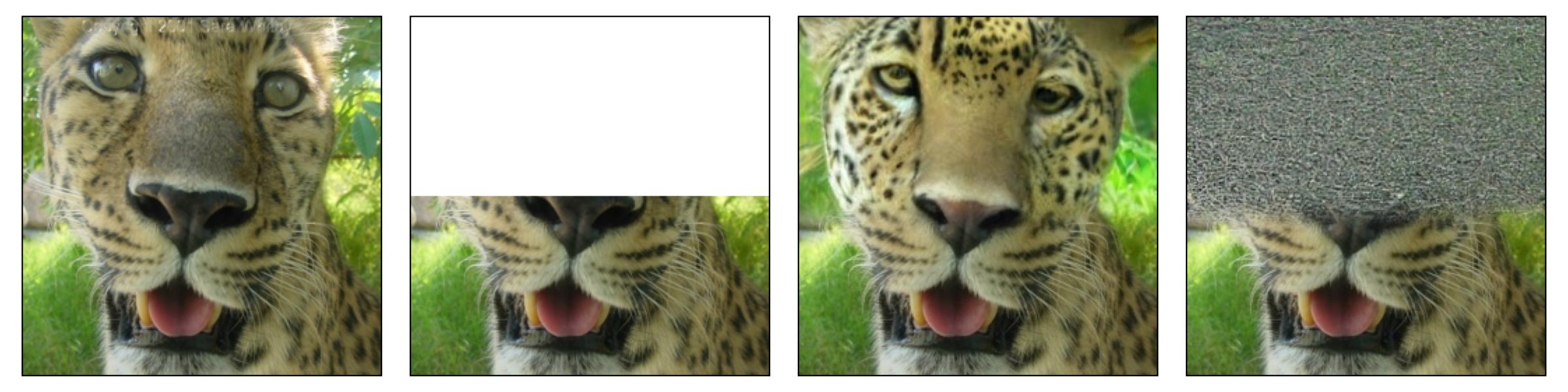}
    \end{subfigure}
    
    \begin{subfigure}[b]{0.495\textwidth}
    \centering
    \includegraphics[width=\textwidth]{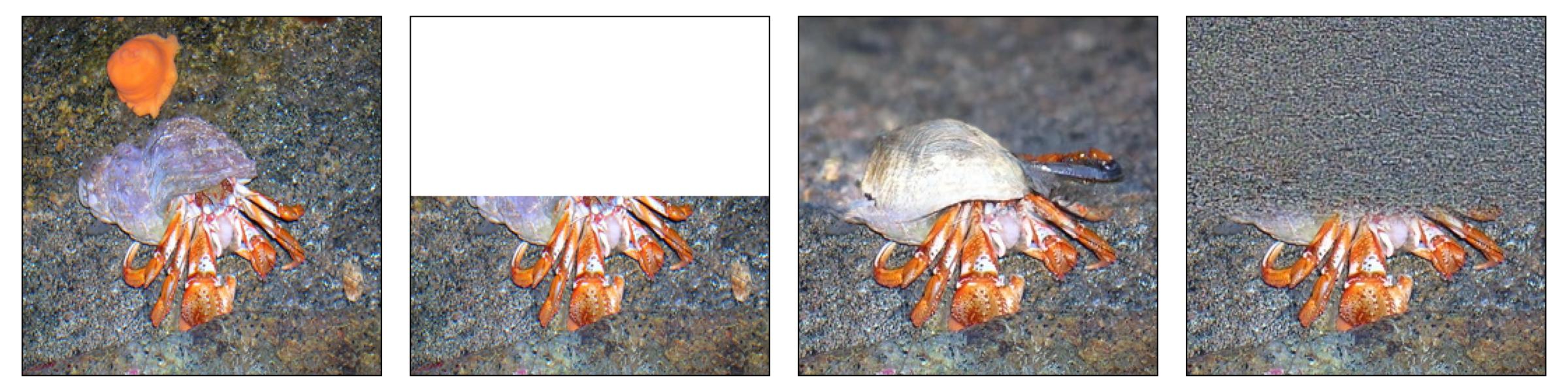}
    \end{subfigure}
    \hfill
    \begin{subfigure}[b]{0.495\textwidth}
    \centering
    \includegraphics[width=\textwidth]{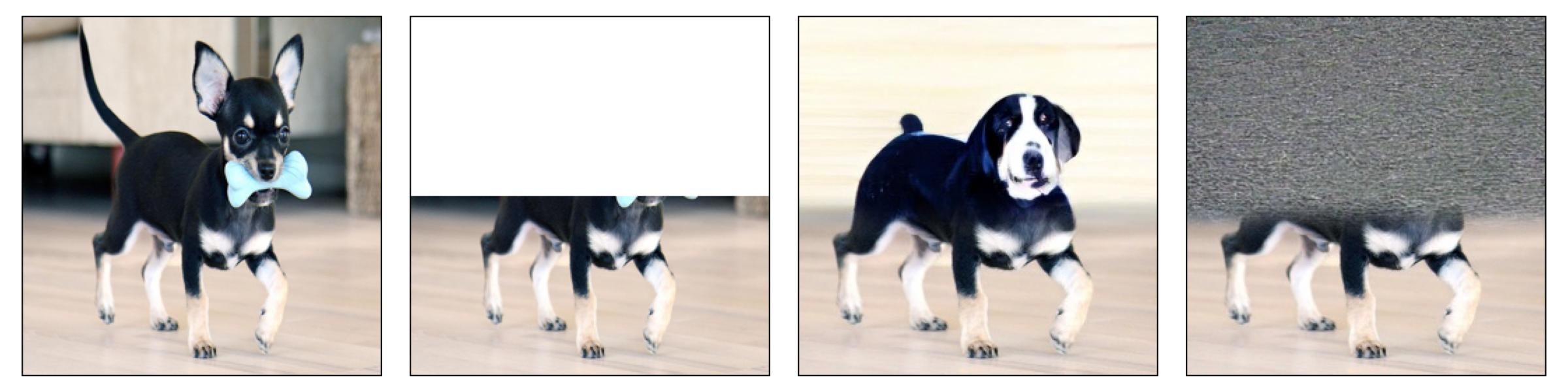}
    \end{subfigure}
    
    \begin{subfigure}[b]{\textwidth}
    \centering
    \includegraphics[width=\textwidth]{figure/caption_four.pdf}\vspace{-1mm}
    \caption{{Forget set}}
    \end{subfigure}
    \vspace{1mm}

    \begin{subfigure}[b]{0.495\textwidth}
    \centering
    \includegraphics[width=\textwidth]{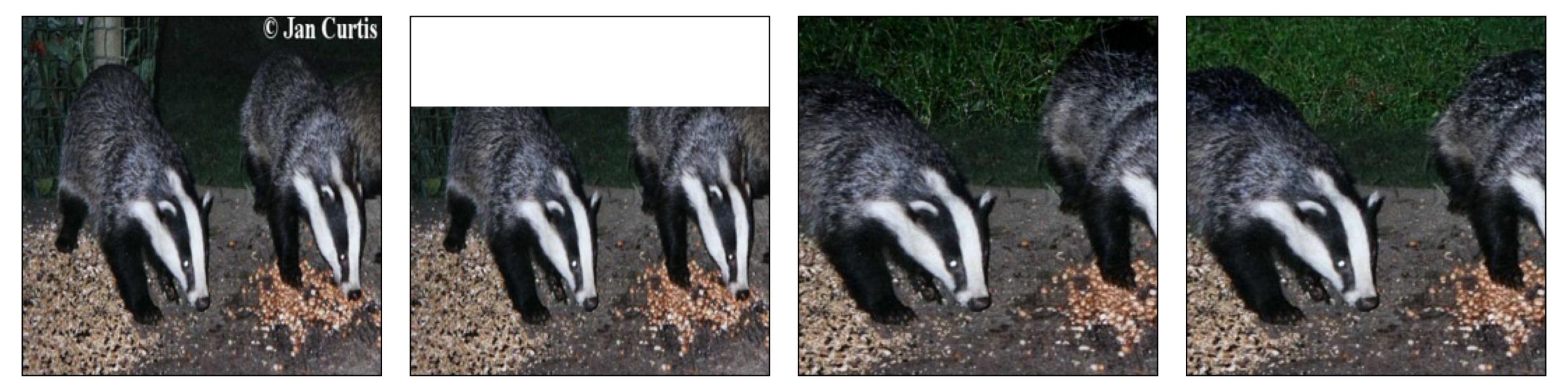}
    \end{subfigure}
    \hfill
    \begin{subfigure}[b]{0.495\textwidth}
    \centering
    \includegraphics[width=\textwidth]{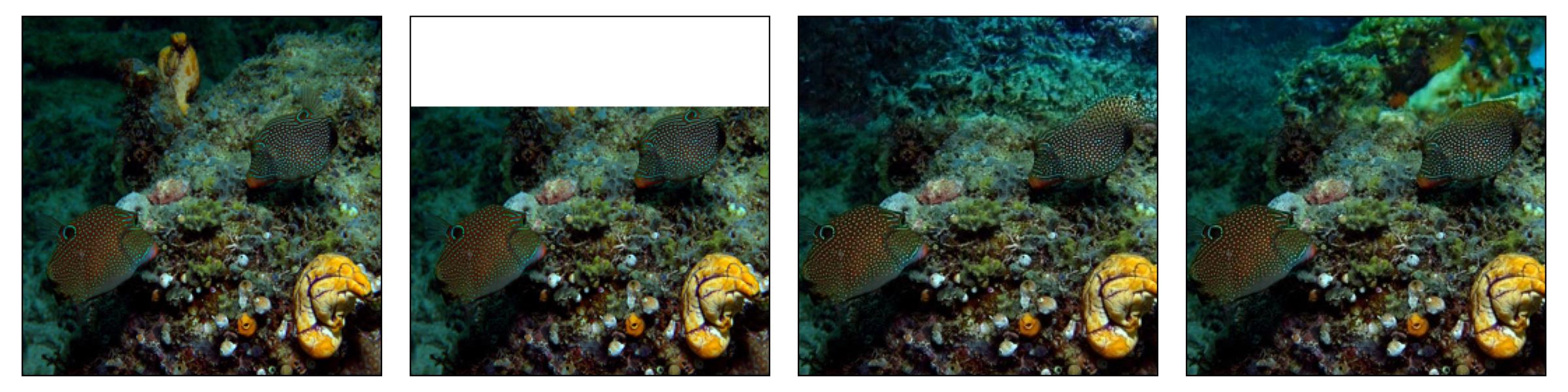}
    \end{subfigure}
    
    \begin{subfigure}[b]{0.495\textwidth}
    \centering
    \includegraphics[width=\textwidth]{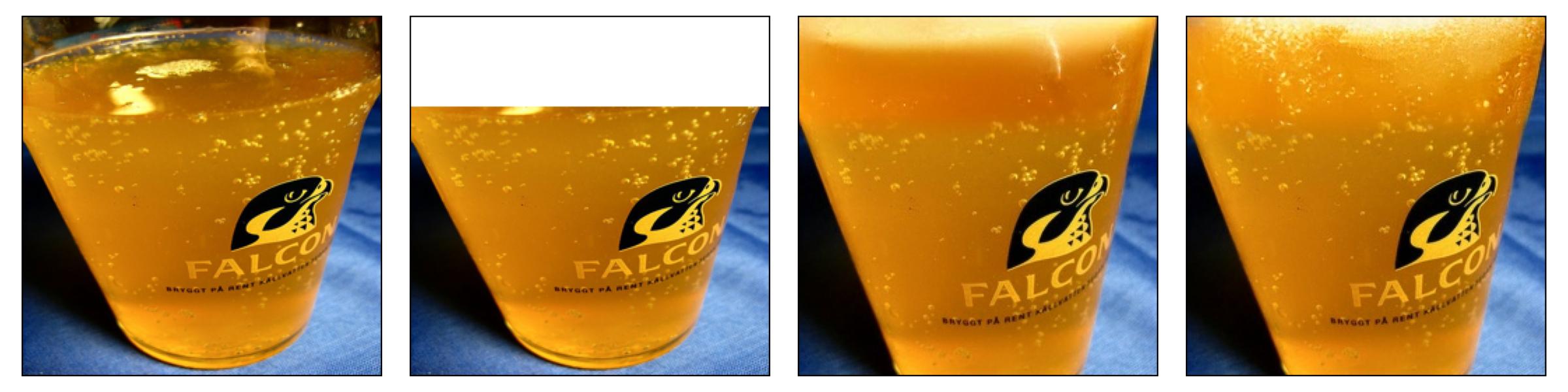}
    \end{subfigure}
    \hfill
    \begin{subfigure}[b]{0.495\textwidth}
    \centering
    \includegraphics[width=\textwidth]{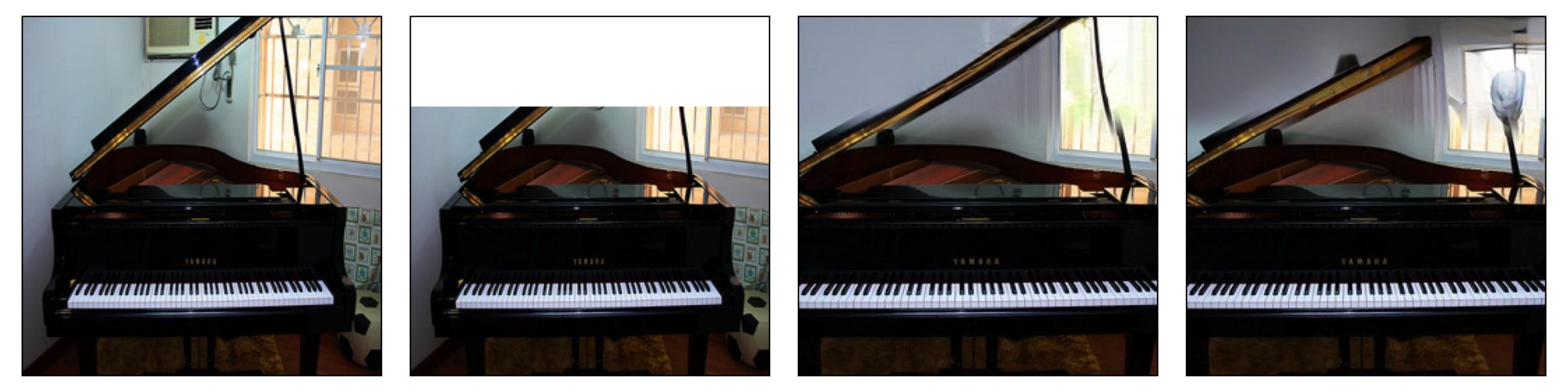}
    \end{subfigure}
    
    \begin{subfigure}[b]{0.495\textwidth}
    \centering
    \includegraphics[width=\textwidth]{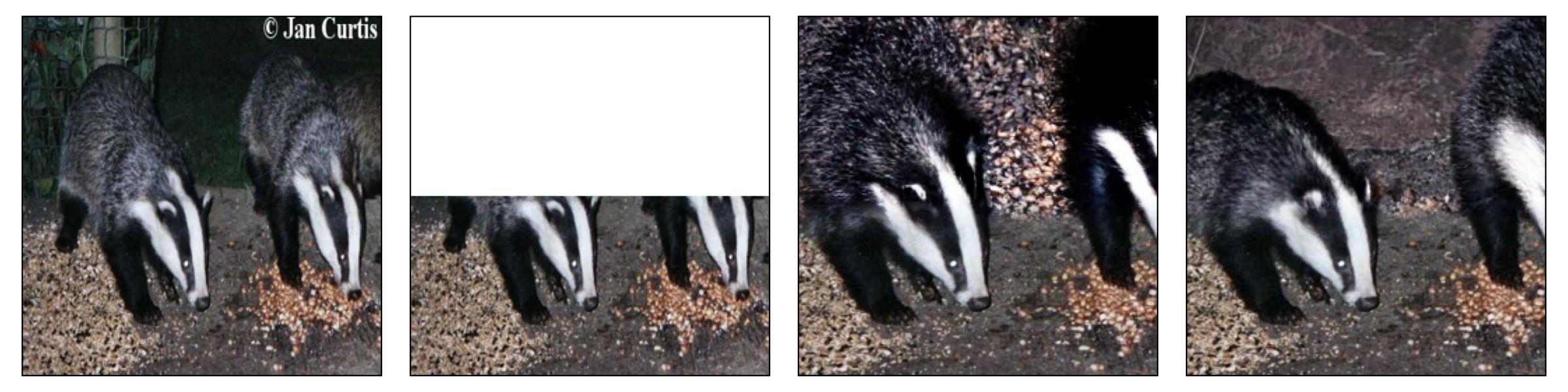}
    \end{subfigure}
    \hfill
    \begin{subfigure}[b]{0.495\textwidth}
    \centering
    \includegraphics[width=\textwidth]{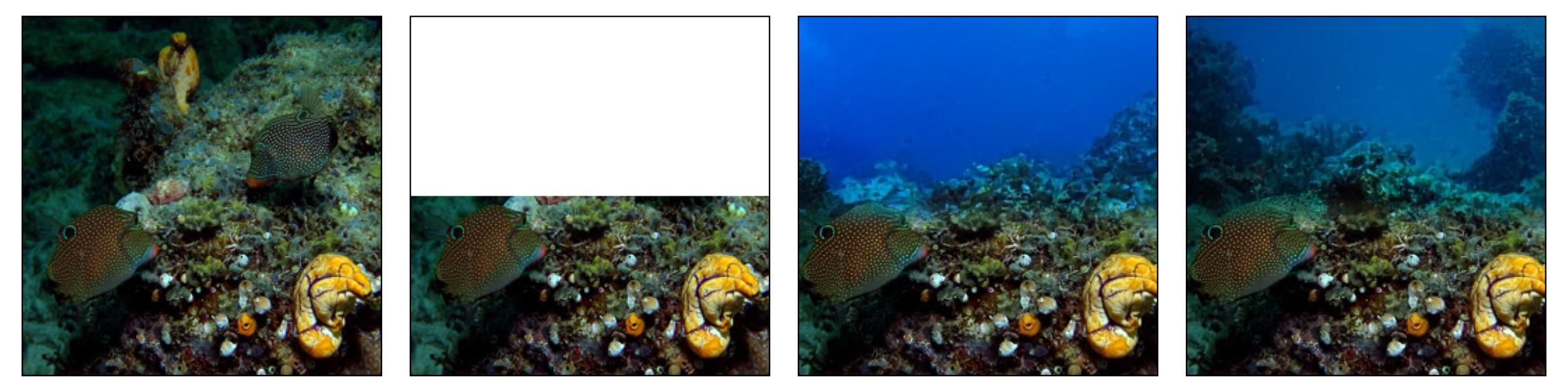}
    \end{subfigure}

    \begin{subfigure}[b]{0.495\textwidth}
    \centering
    \includegraphics[width=\textwidth]{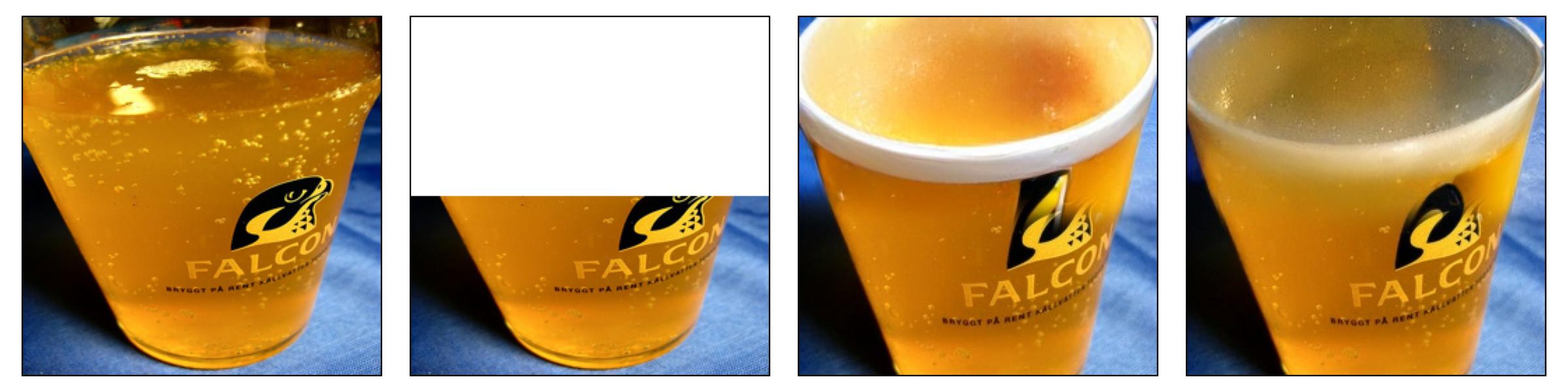}
    \end{subfigure}
    \hfill
    \begin{subfigure}[b]{0.495\textwidth}
    \centering
    \includegraphics[width=\textwidth]{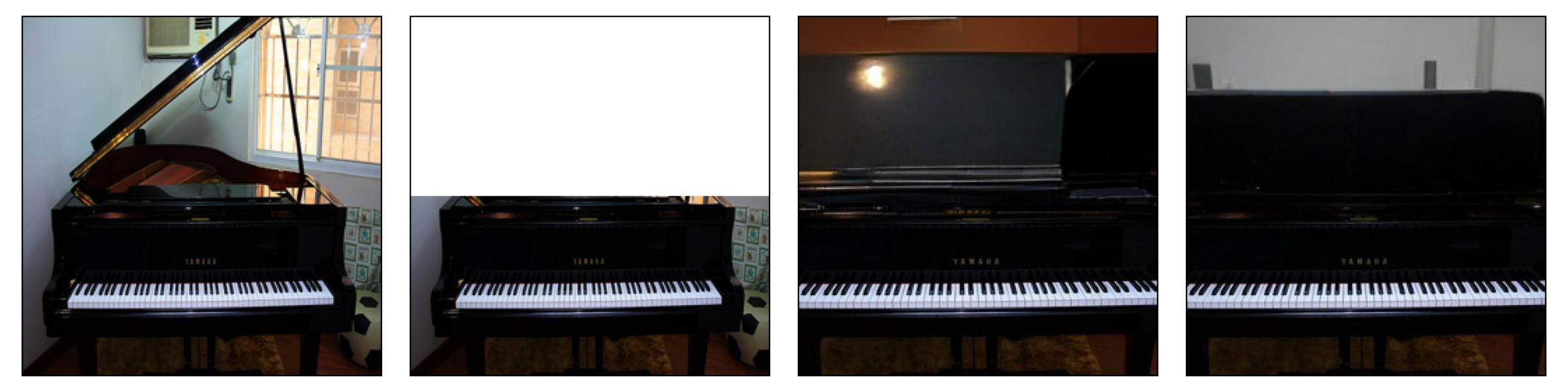}
    \end{subfigure}
    
    \begin{subfigure}[b]{\textwidth}
    \centering
    \includegraphics[width=\textwidth]{figure/caption_four.pdf}\vspace{-1mm}
    \caption{{Retain set}}
    \end{subfigure}
\caption{Ablation study: Upward extension by VQ-GAN. We visualize he performance of the original model (before unlearning) and the obtained model by our approach (after unlearning). We show results on both forget set and retain set. }
\label{fig:gan_up}
\end{figure}

\begin{figure}[htb]\centering
    \begin{subfigure}[b]{0.495\textwidth}
    \centering
    \includegraphics[width=\textwidth]{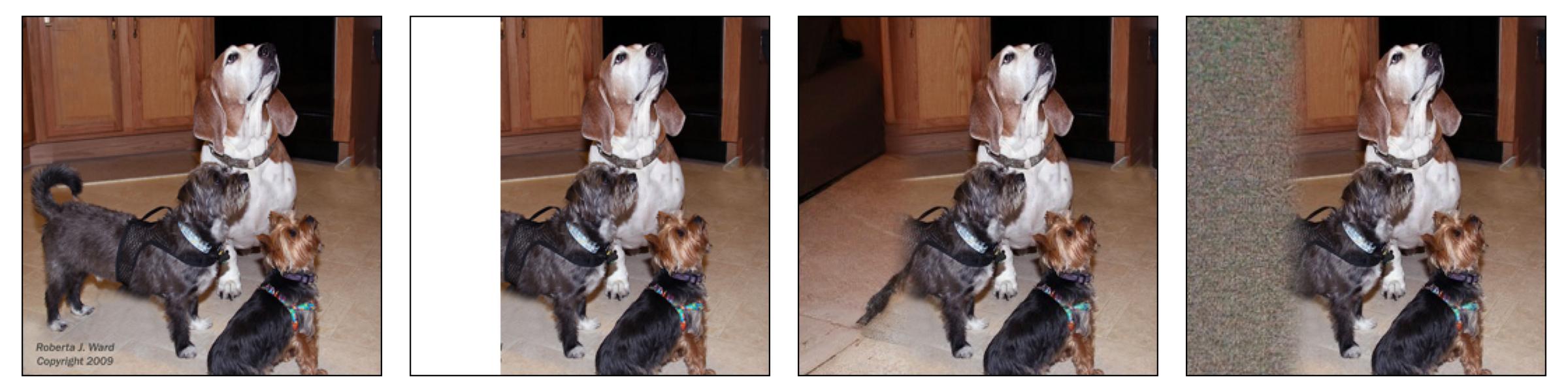}
    \end{subfigure}
    \hfill
    \begin{subfigure}[b]{0.495\textwidth}
    \centering
    \includegraphics[width=\textwidth]{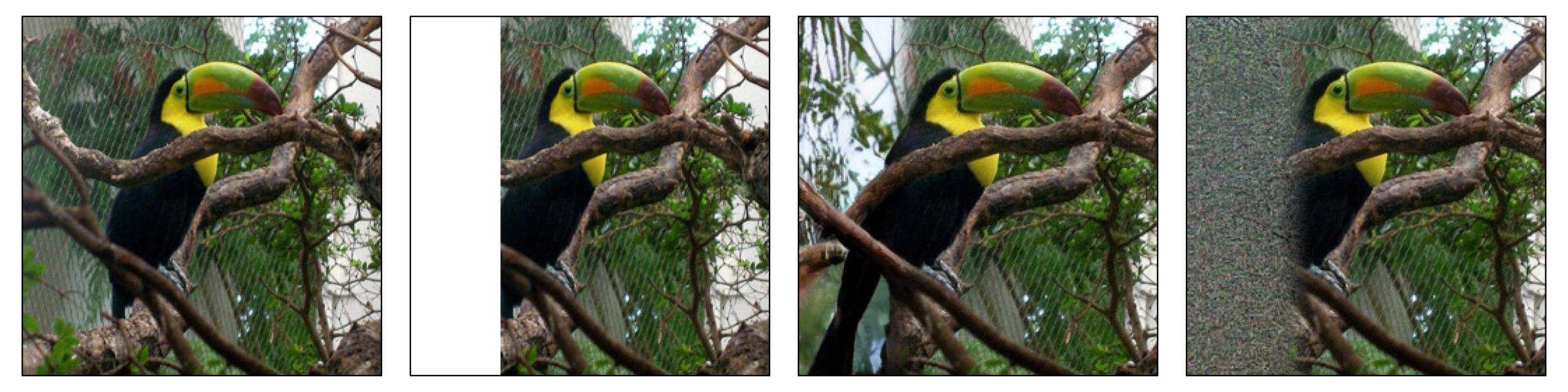}
    \end{subfigure}
    
    \begin{subfigure}[b]{0.495\textwidth}
    \centering
    \includegraphics[width=\textwidth]{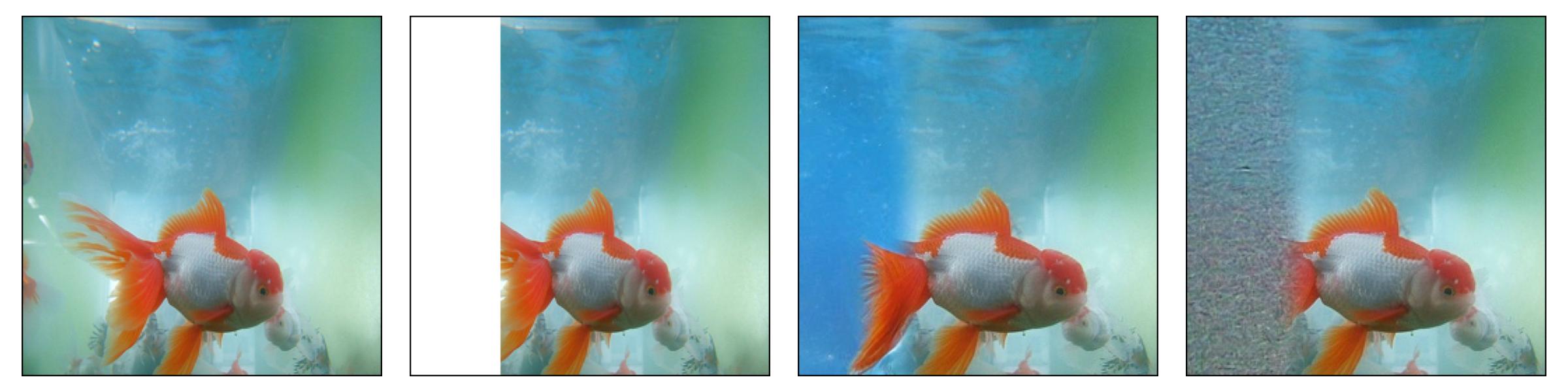}
    \end{subfigure}
    \hfill
    \begin{subfigure}[b]{0.495\textwidth}
    \centering
    \includegraphics[width=\textwidth]{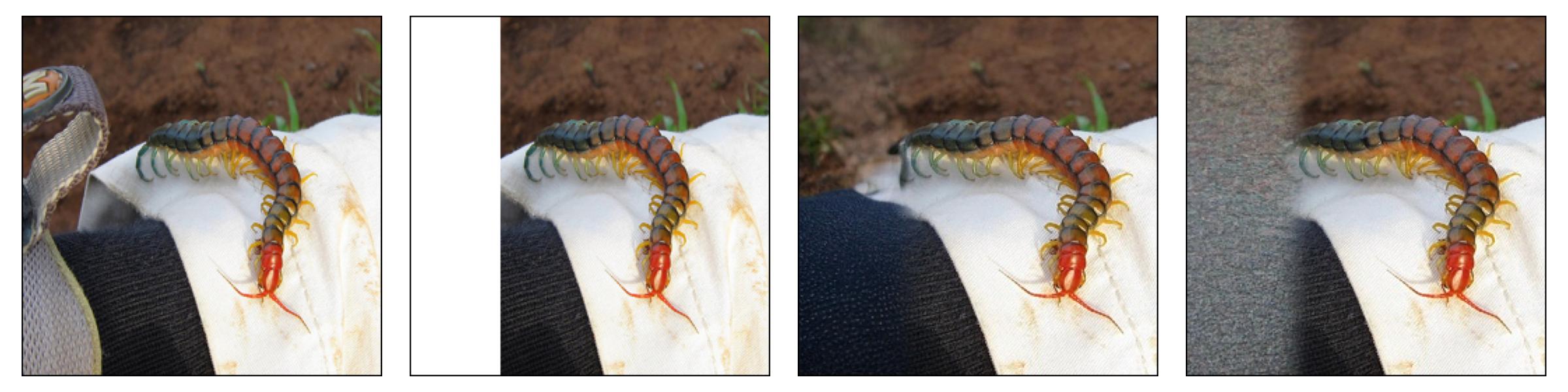}
    \end{subfigure}
        
    \begin{subfigure}[b]{0.495\textwidth}
    \centering
    \includegraphics[width=\textwidth]{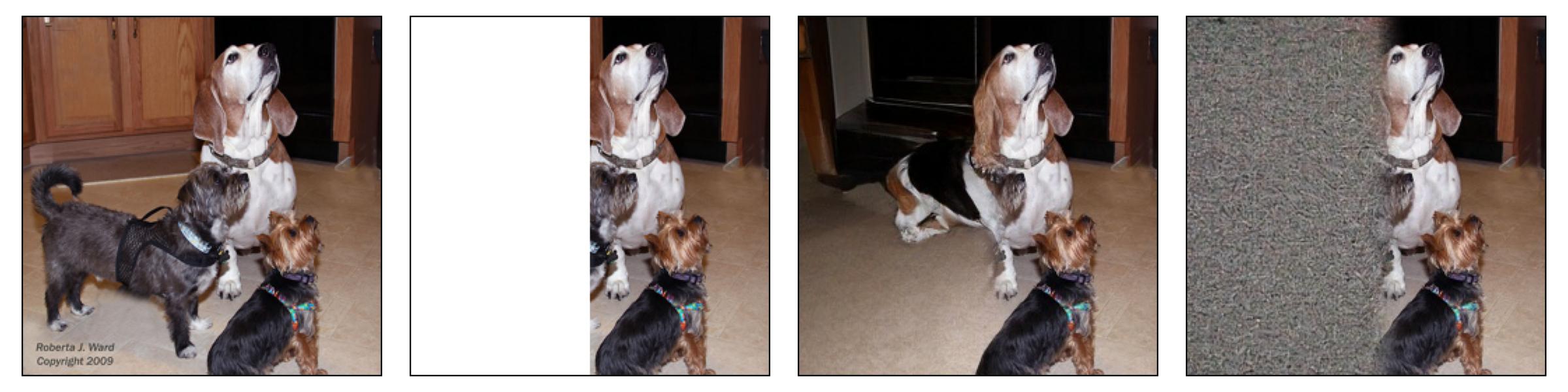}
    \end{subfigure}
    \hfill
    \begin{subfigure}[b]{0.495\textwidth}
    \centering
    \includegraphics[width=\textwidth]{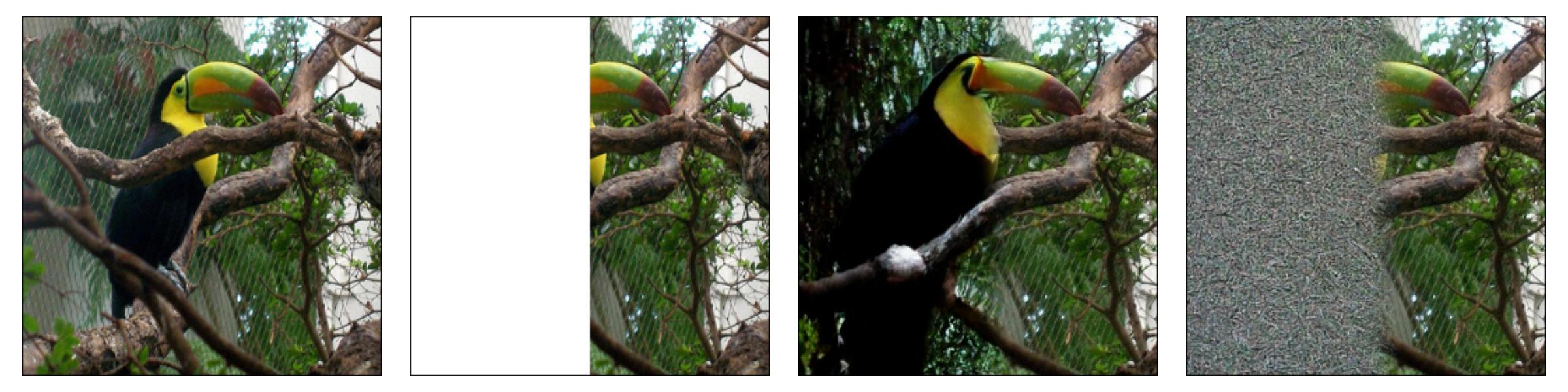}
    \end{subfigure}

    \begin{subfigure}[b]{0.495\textwidth}
    \centering
    \includegraphics[width=\textwidth]{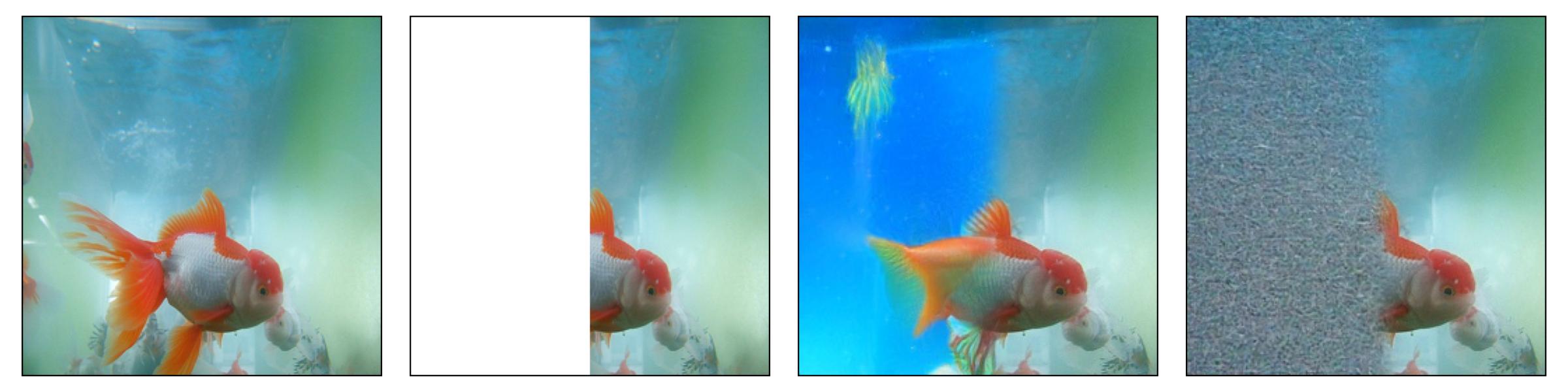}
    \end{subfigure}
    \hfill
    \begin{subfigure}[b]{0.495\textwidth}
    \centering
    \includegraphics[width=\textwidth]{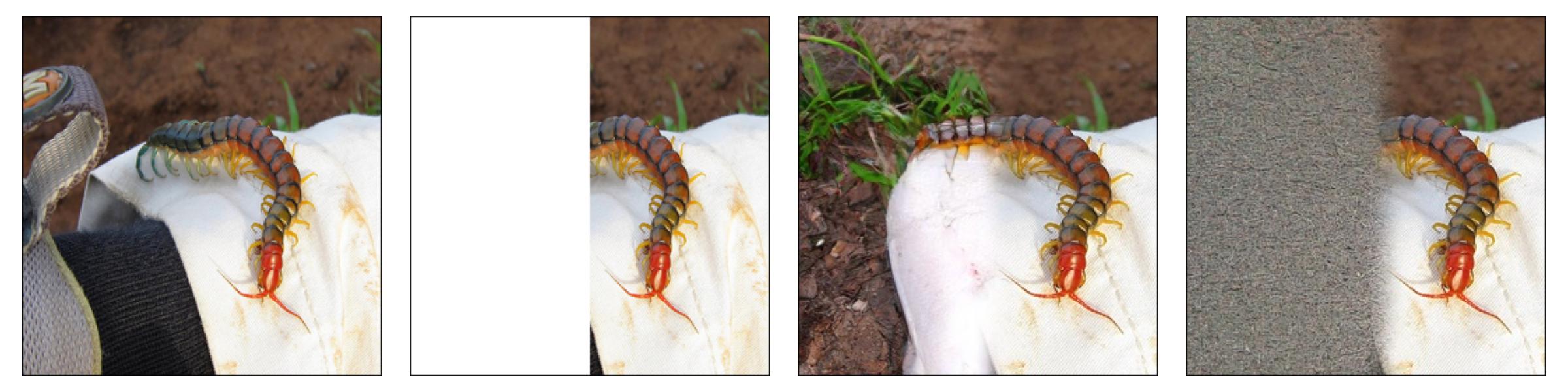}
    \end{subfigure}
    
    \begin{subfigure}[b]{\textwidth}
    \centering
    \includegraphics[width=\textwidth]{figure/caption_four.pdf}\vspace{-1mm}
    \caption{{Forget set}}
    \end{subfigure}
    \vspace{1mm}

    \begin{subfigure}[b]{0.495\textwidth}
    \centering
    \includegraphics[width=\textwidth]{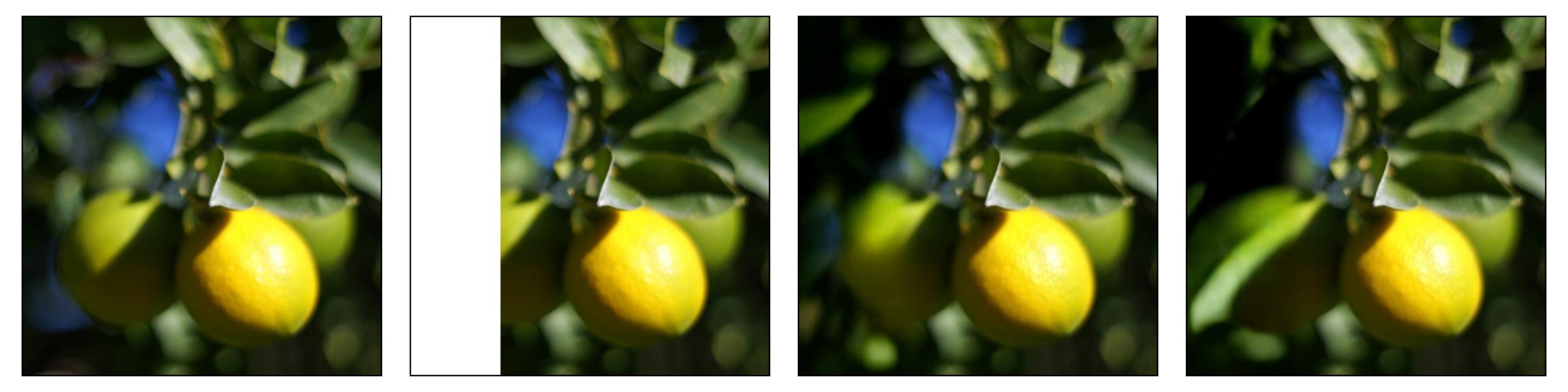}
    \end{subfigure}
    \hfill
    \begin{subfigure}[b]{0.495\textwidth}
    \centering
    \includegraphics[width=\textwidth]{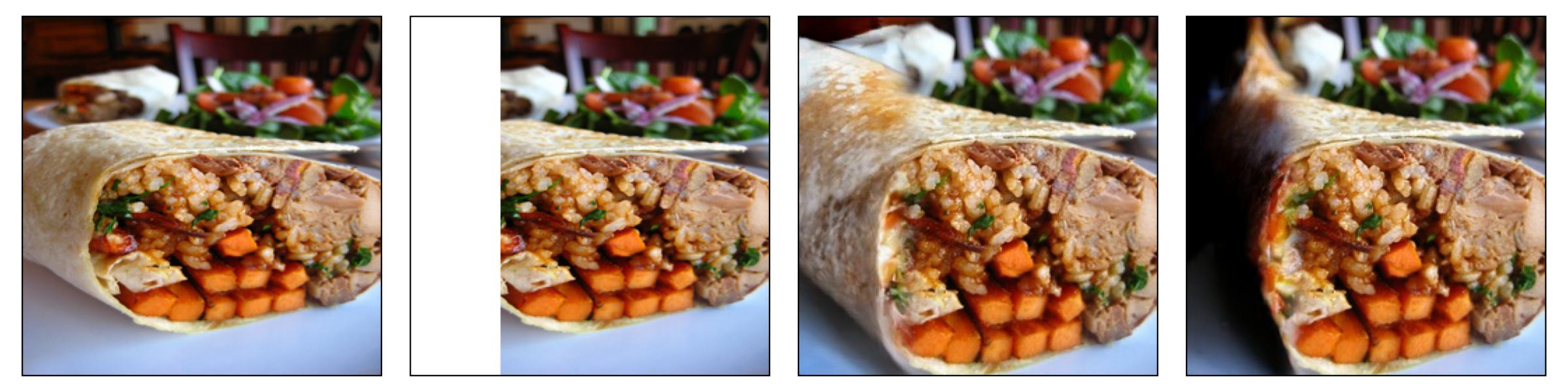}
    \end{subfigure}

    \begin{subfigure}[b]{0.495\textwidth}
    \centering
    \includegraphics[width=\textwidth]{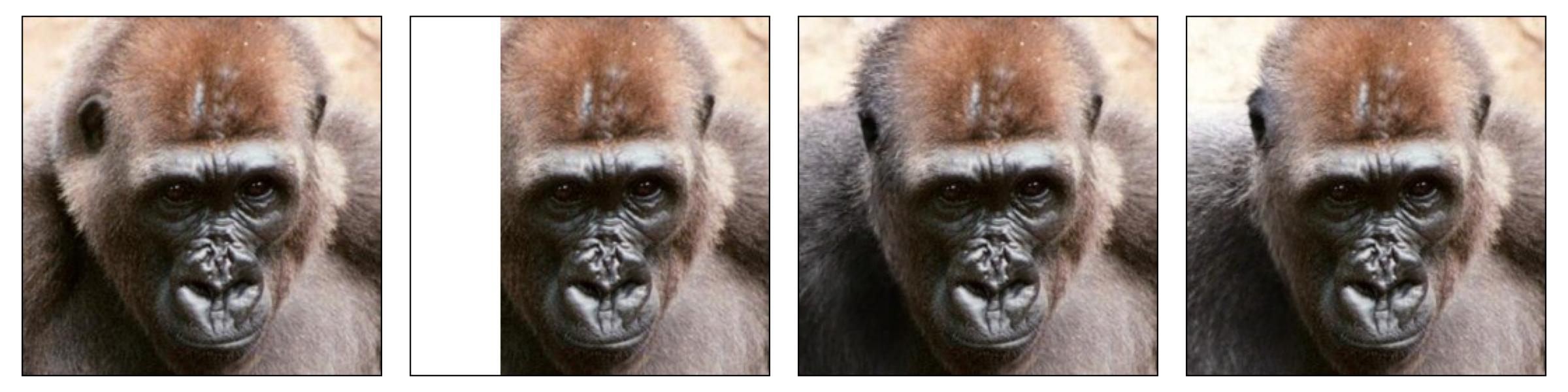}
    \end{subfigure}
    \hfill
    \begin{subfigure}[b]{0.495\textwidth}
    \centering
    \includegraphics[width=\textwidth]{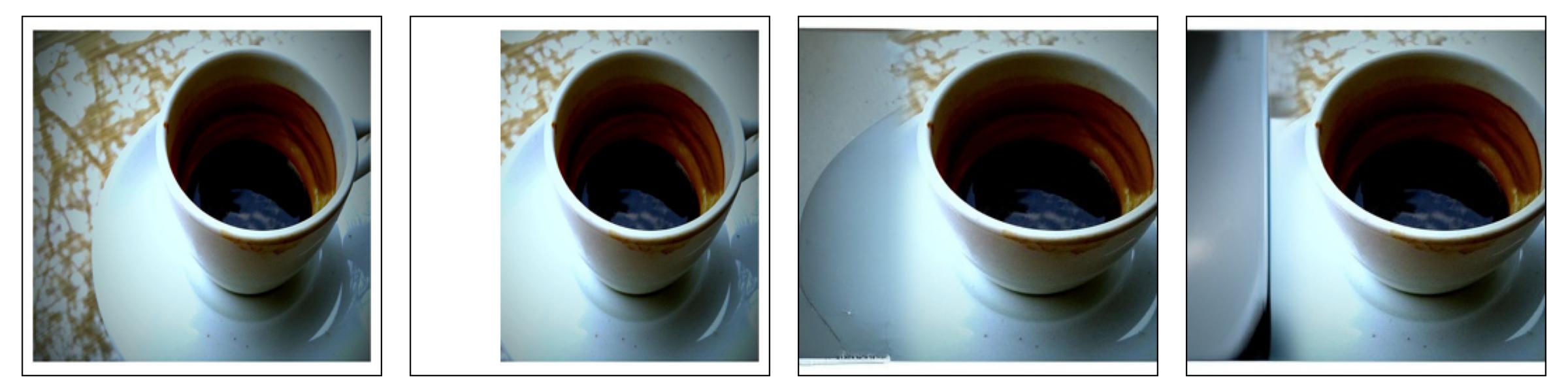}
    \end{subfigure}
    
    \begin{subfigure}[b]{0.495\textwidth}
    \centering
    \includegraphics[width=\textwidth]{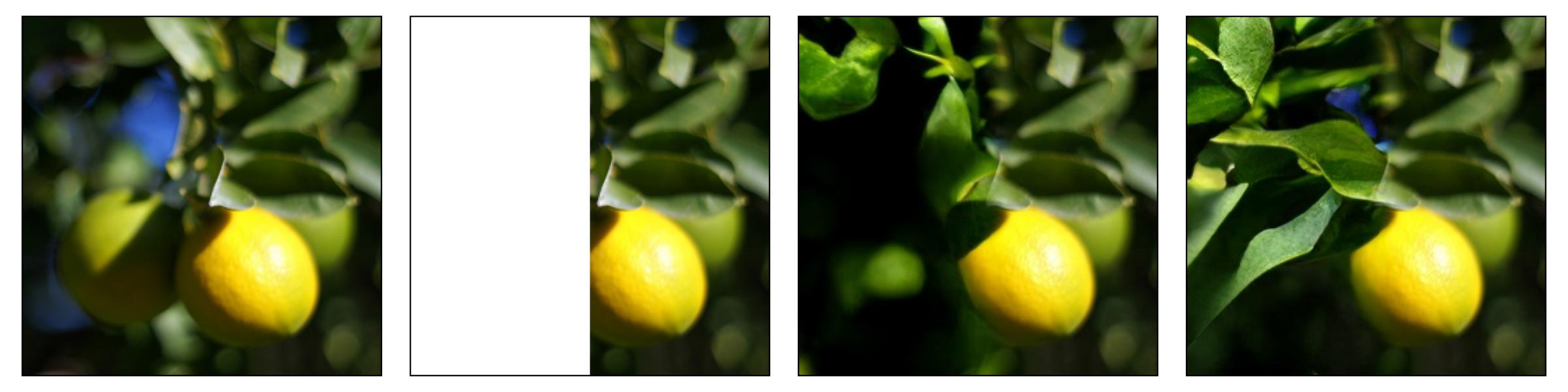}
    \end{subfigure}
    \hfill
    \begin{subfigure}[b]{0.495\textwidth}
    \centering
    \includegraphics[width=\textwidth]{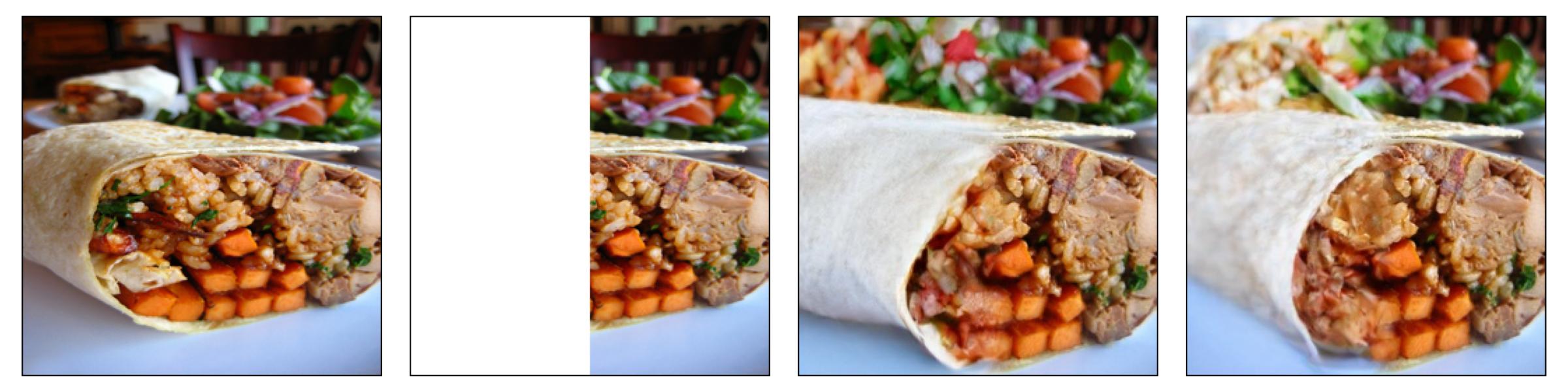}
    \end{subfigure}
    
    \begin{subfigure}[b]{0.495\textwidth}
    \centering
    \includegraphics[width=\textwidth]{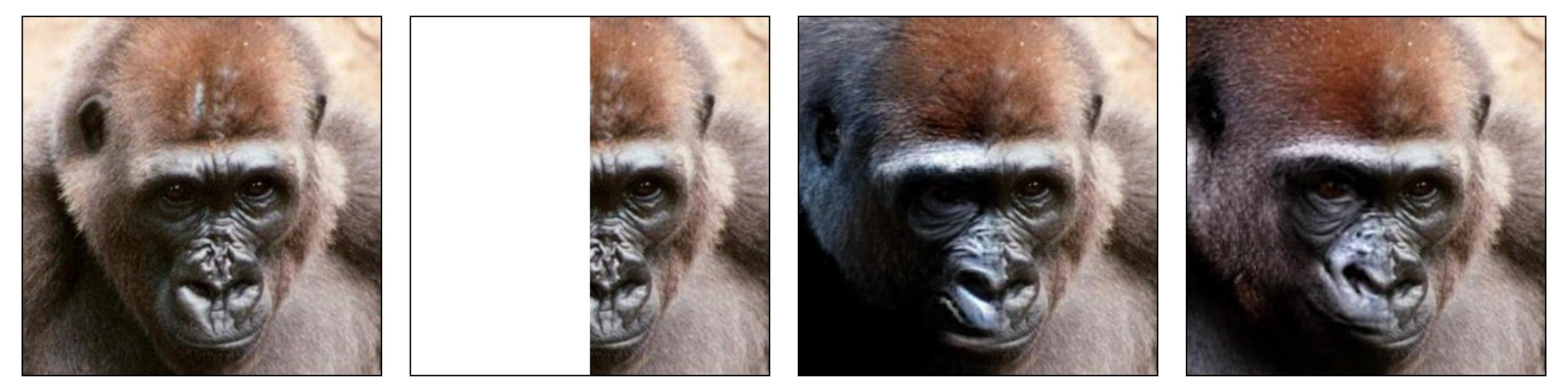}
    \end{subfigure}
    \hfill
    \begin{subfigure}[b]{0.495\textwidth}
    \centering
    \includegraphics[width=\textwidth]{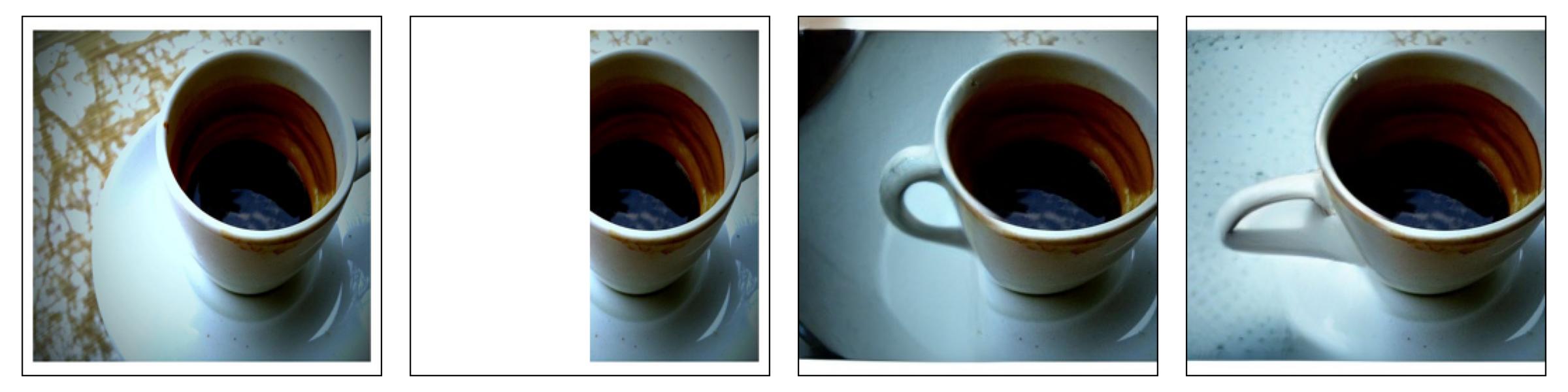}
    \end{subfigure}
    
    \begin{subfigure}[b]{\textwidth}
    \centering
    \includegraphics[width=\textwidth]{figure/caption_four.pdf}\vspace{-1mm}
    \caption{{Retain set}}
    \end{subfigure}
\caption{Ablation study: Leftward extension by VQ-GAN. We visualize he performance of the original model (before unlearning) and the obtained model by our approach (after unlearning). We show results on both forget set and retain set. }
\label{fig:gan_left}
\end{figure}

\begin{figure}[htb]\centering
    \begin{subfigure}[b]{0.495\textwidth}
    \centering
    \includegraphics[width=\textwidth]{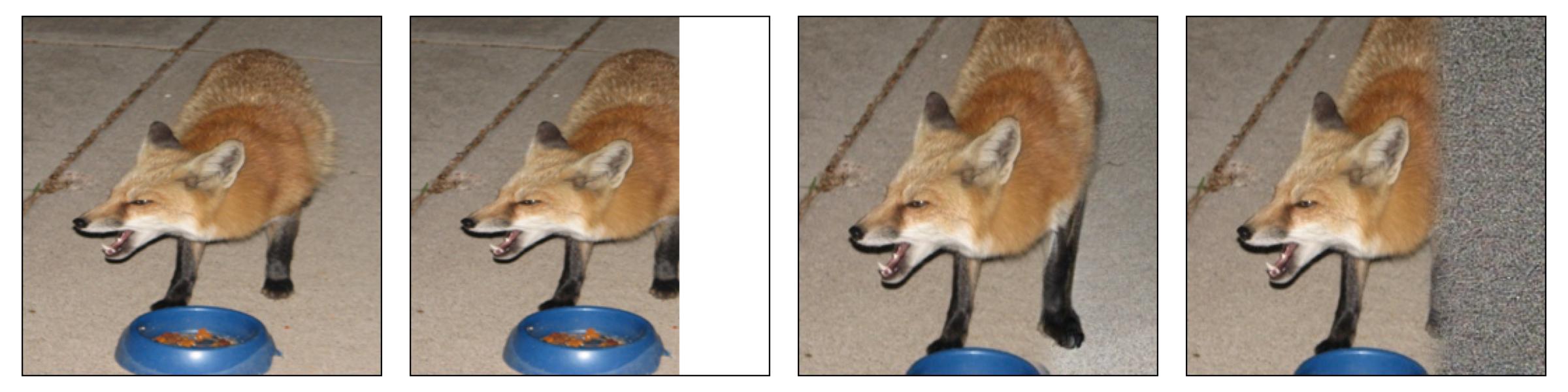}
    \end{subfigure}
    \hfill
    \begin{subfigure}[b]{0.495\textwidth}
    \centering
    \includegraphics[width=\textwidth]{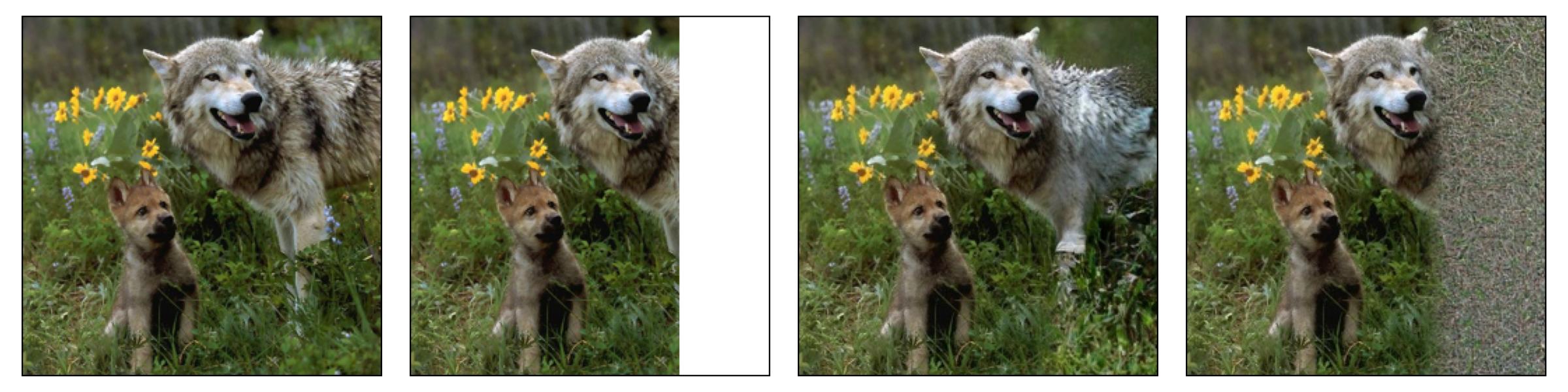}
    \end{subfigure}
    
    \begin{subfigure}[b]{0.495\textwidth}
    \centering
    \includegraphics[width=\textwidth]{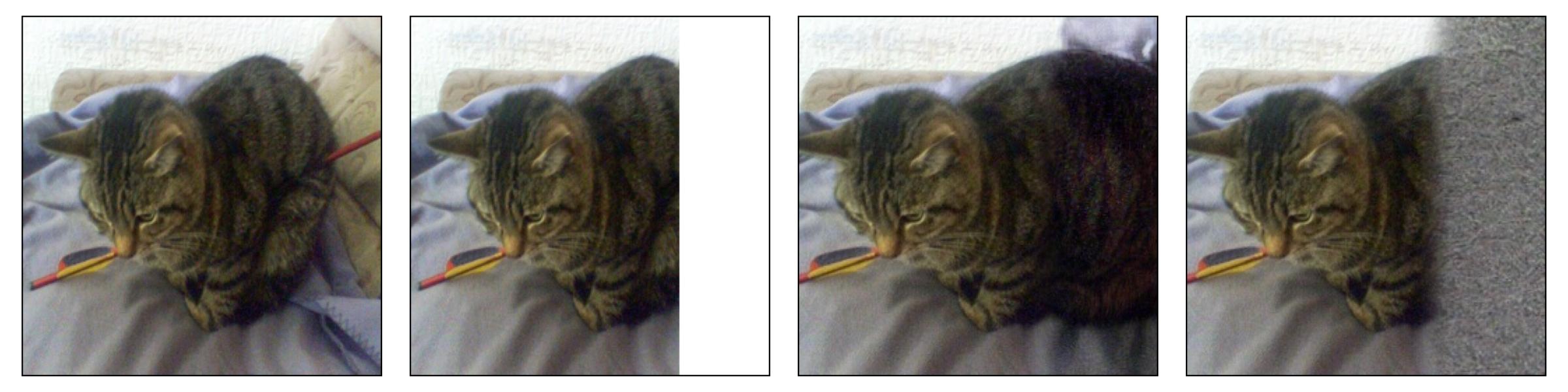}
    \end{subfigure}
    \hfill
    \begin{subfigure}[b]{0.495\textwidth}
    \centering
    \includegraphics[width=\textwidth]{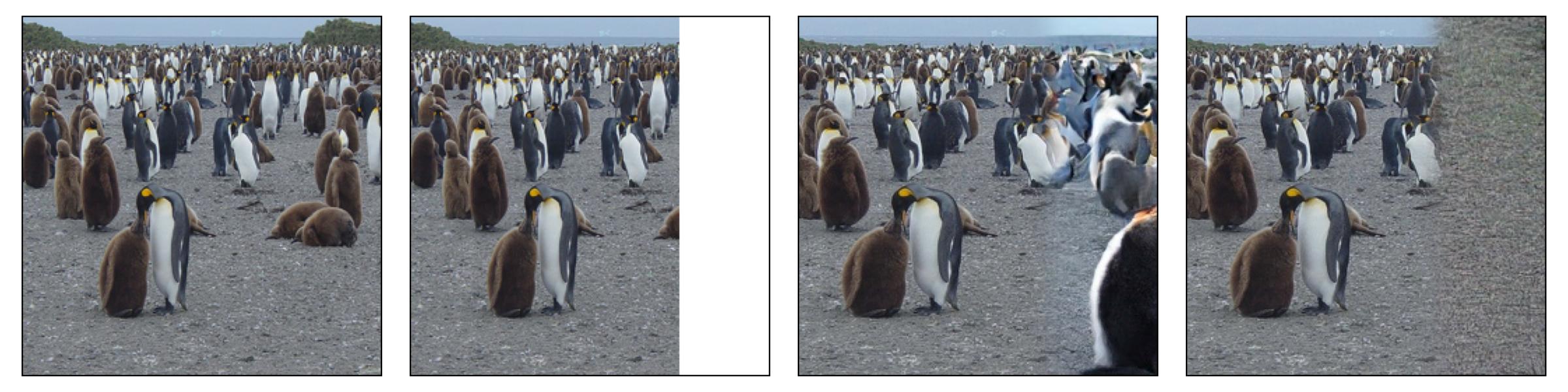}
    \end{subfigure}
    
    \begin{subfigure}[b]{0.495\textwidth}
    \centering
    \includegraphics[width=\textwidth]{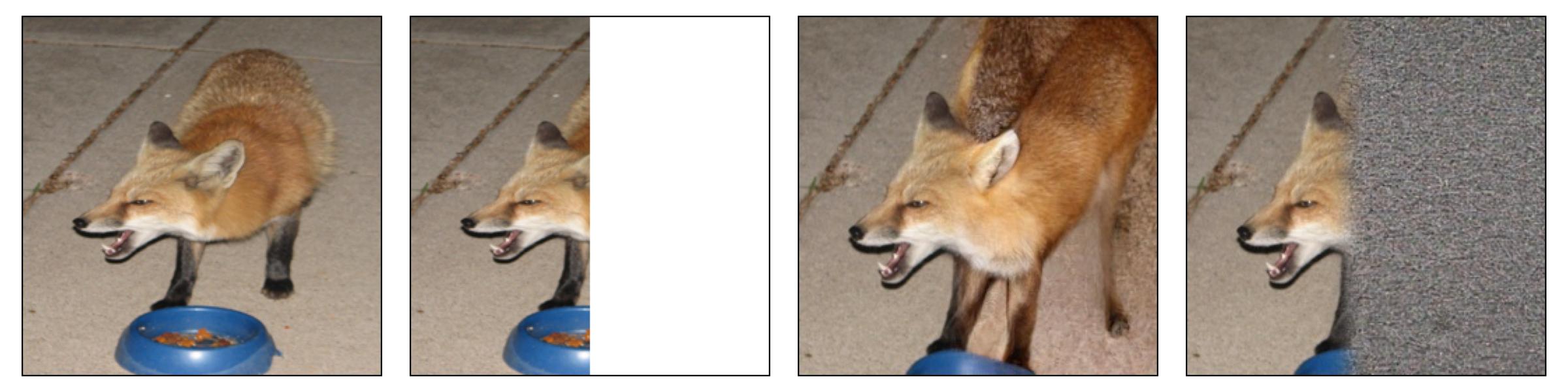}
    \end{subfigure}
    \hfill
    \begin{subfigure}[b]{0.495\textwidth}
    \centering
    \includegraphics[width=\textwidth]{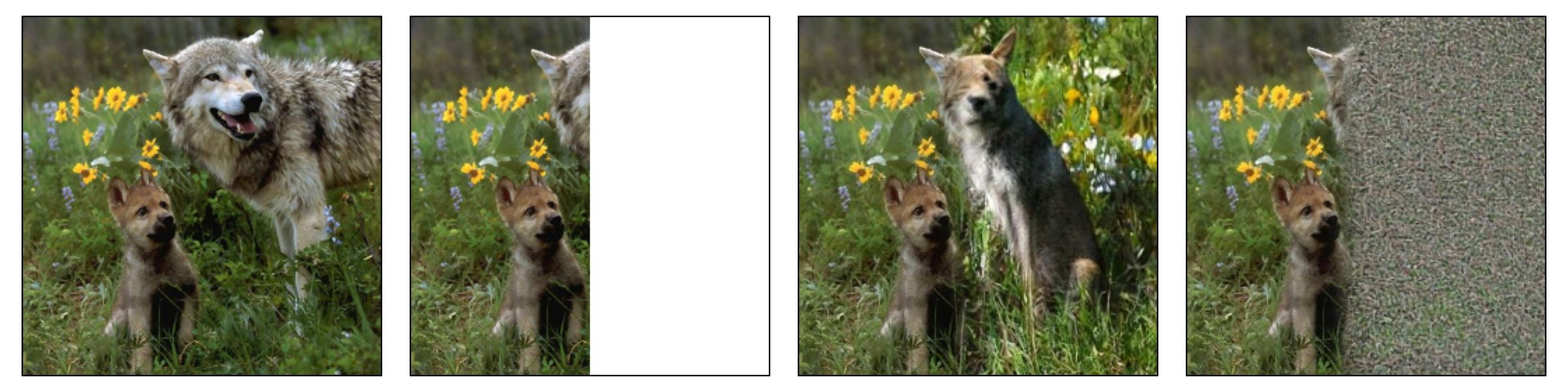}
    \end{subfigure}

    \begin{subfigure}[b]{0.495\textwidth}
    \centering
    \includegraphics[width=\textwidth]{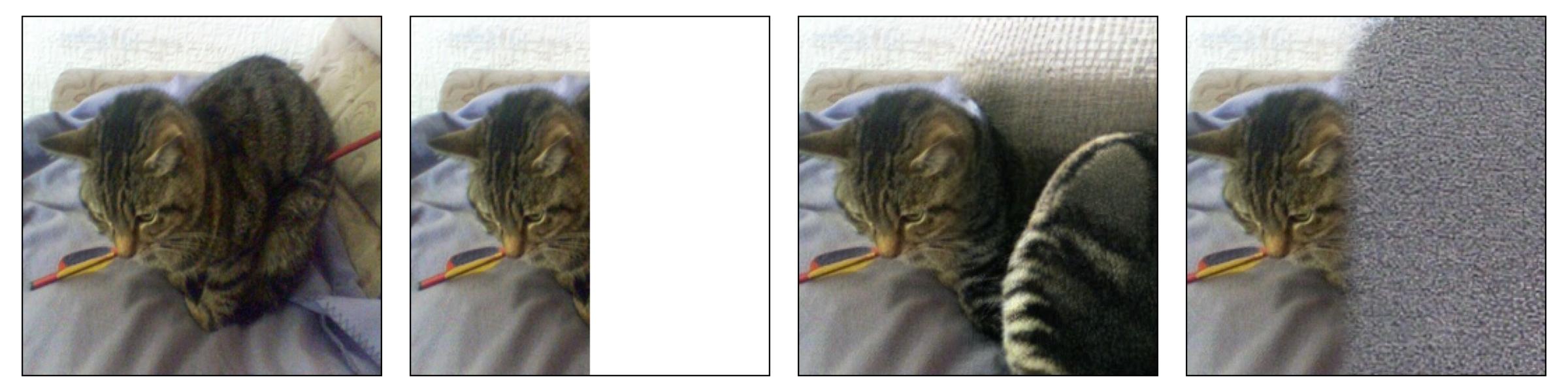}
    \end{subfigure}
    \hfill
    \begin{subfigure}[b]{0.495\textwidth}
    \centering
    \includegraphics[width=\textwidth]{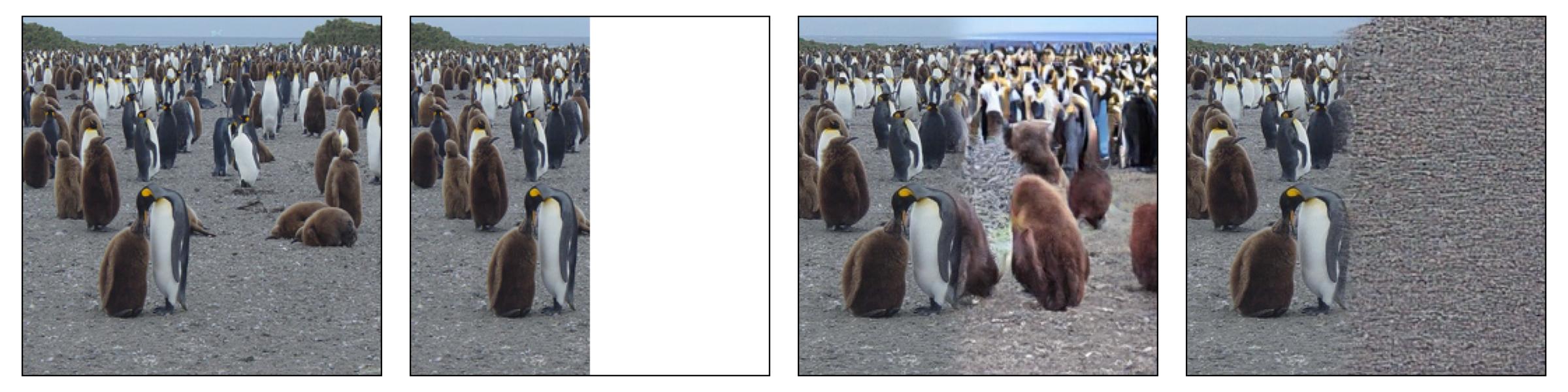}
    \end{subfigure}
    
    \begin{subfigure}[b]{\textwidth}
    \centering
    \includegraphics[width=\textwidth]{figure/caption_four.pdf}\vspace{-1mm}
    \caption{{Forget set}}
    \end{subfigure}
    \vspace{1mm}

    \begin{subfigure}[b]{0.495\textwidth}
    \centering
    \includegraphics[width=\textwidth]{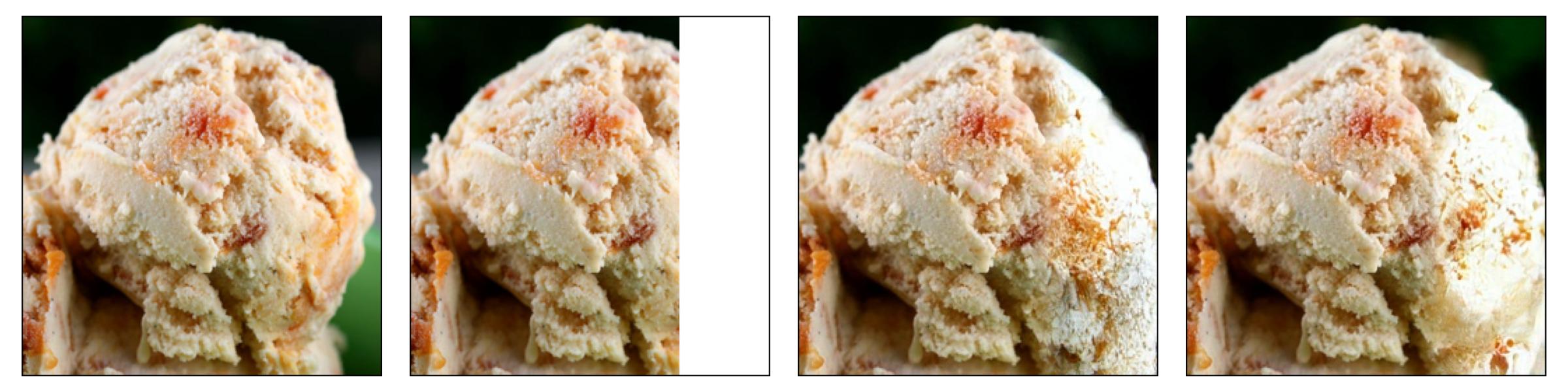}
    \end{subfigure}
    \hfill
    \begin{subfigure}[b]{0.495\textwidth}
    \centering
    \includegraphics[width=\textwidth]{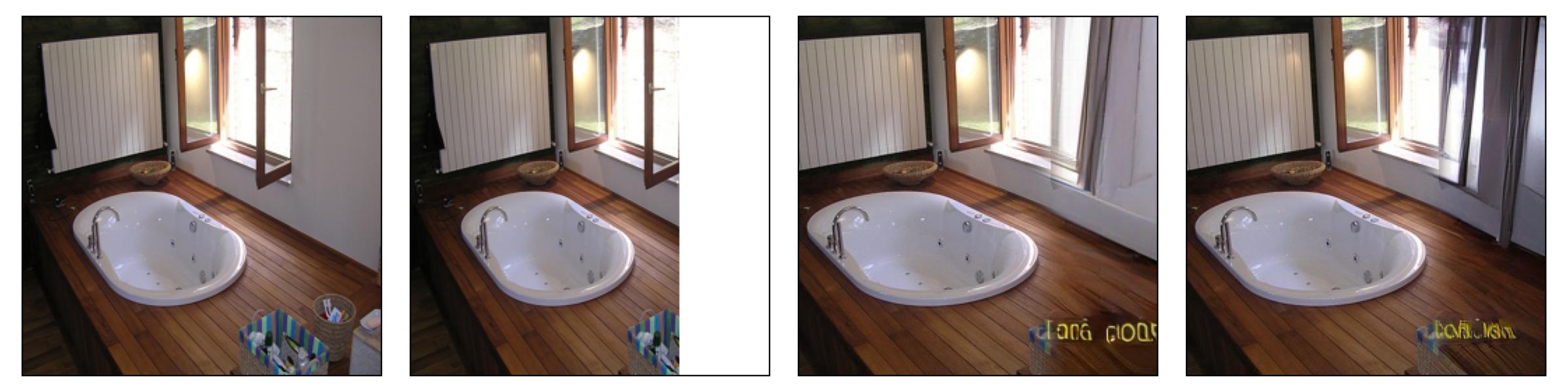}
    \end{subfigure}
    
    \begin{subfigure}[b]{0.495\textwidth}
    \centering
    \includegraphics[width=\textwidth]{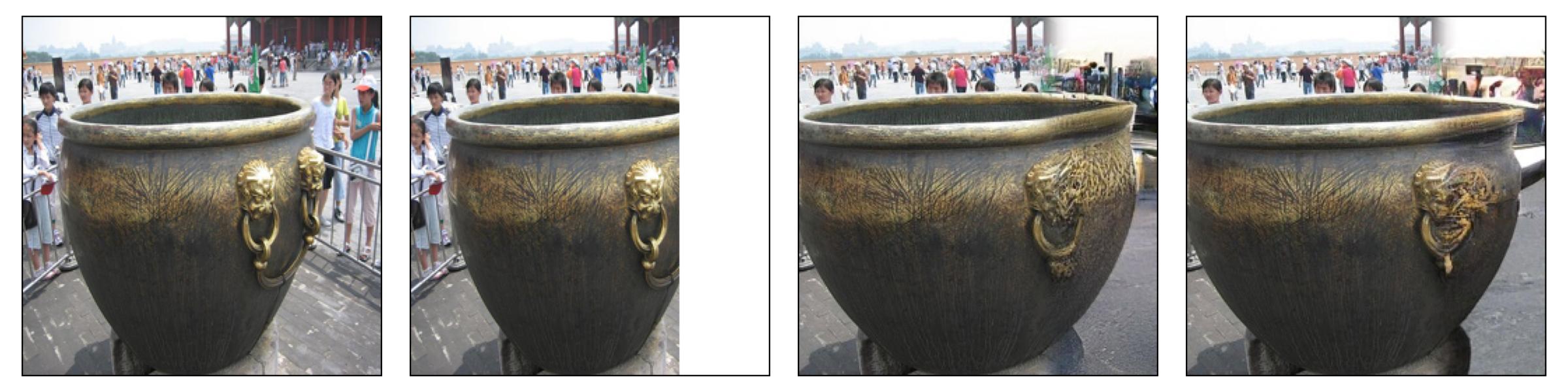}
    \end{subfigure}
    \hfill
    \begin{subfigure}[b]{0.495\textwidth}
    \centering
    \includegraphics[width=\textwidth]{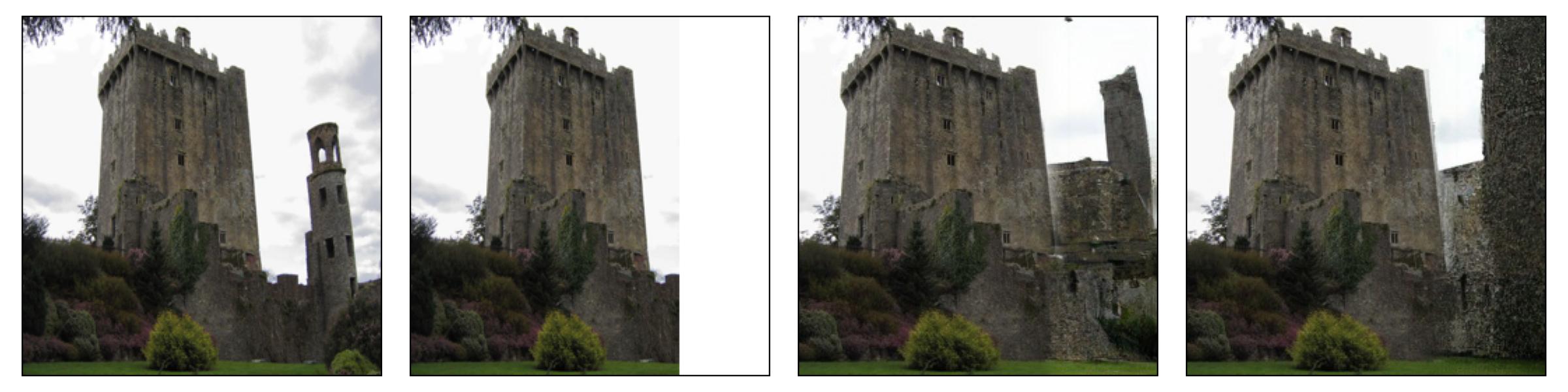}
    \end{subfigure}
    
    \begin{subfigure}[b]{0.495\textwidth}
    \centering
    \includegraphics[width=\textwidth]{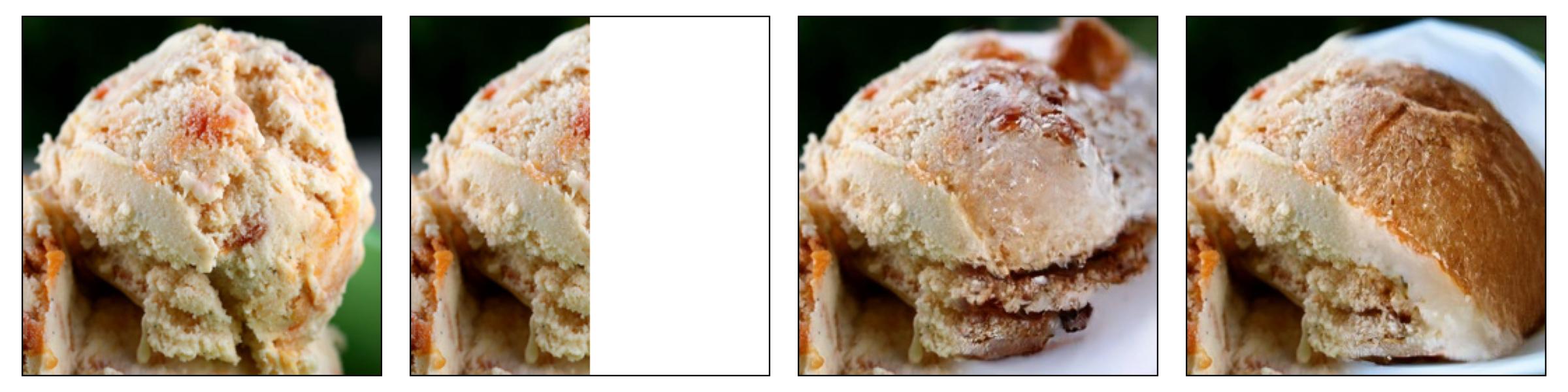}
    \end{subfigure}
    \hfill
    \begin{subfigure}[b]{0.495\textwidth}
    \centering
    \includegraphics[width=\textwidth]{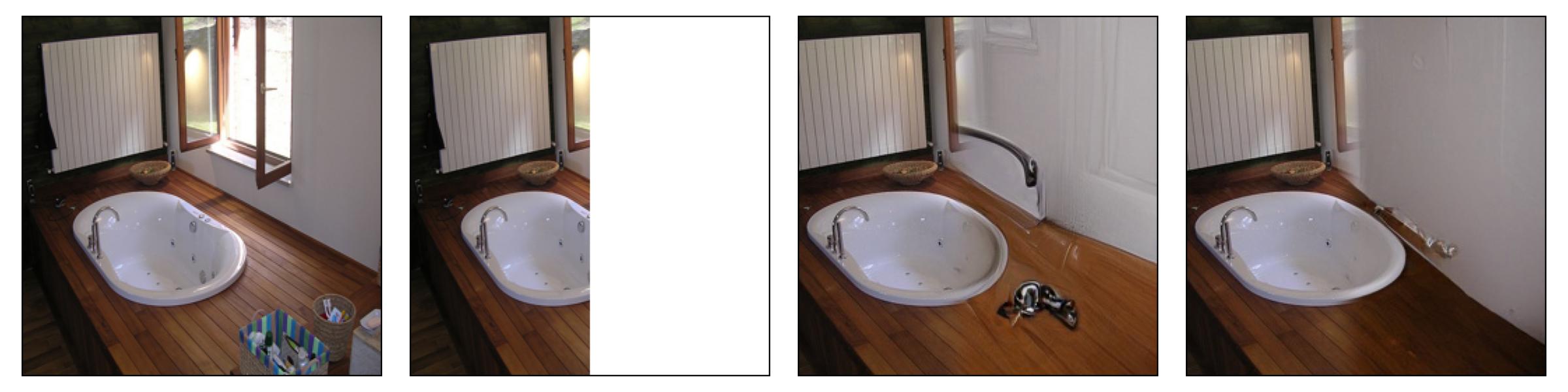}
    \end{subfigure}

    \begin{subfigure}[b]{0.495\textwidth}
    \centering
    \includegraphics[width=\textwidth]{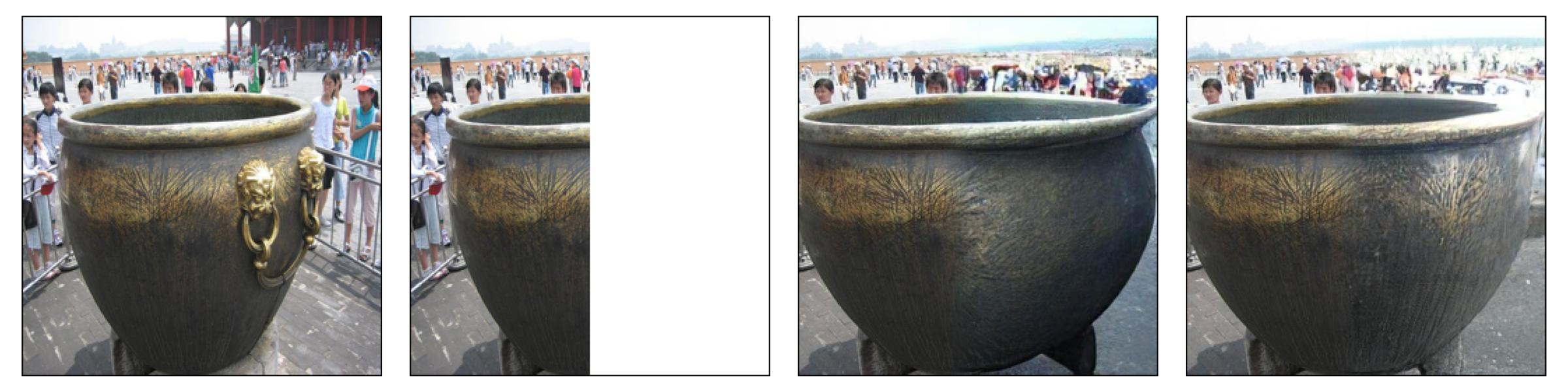}
    \end{subfigure}
    \hfill
    \begin{subfigure}[b]{0.495\textwidth}
    \centering
    \includegraphics[width=\textwidth]{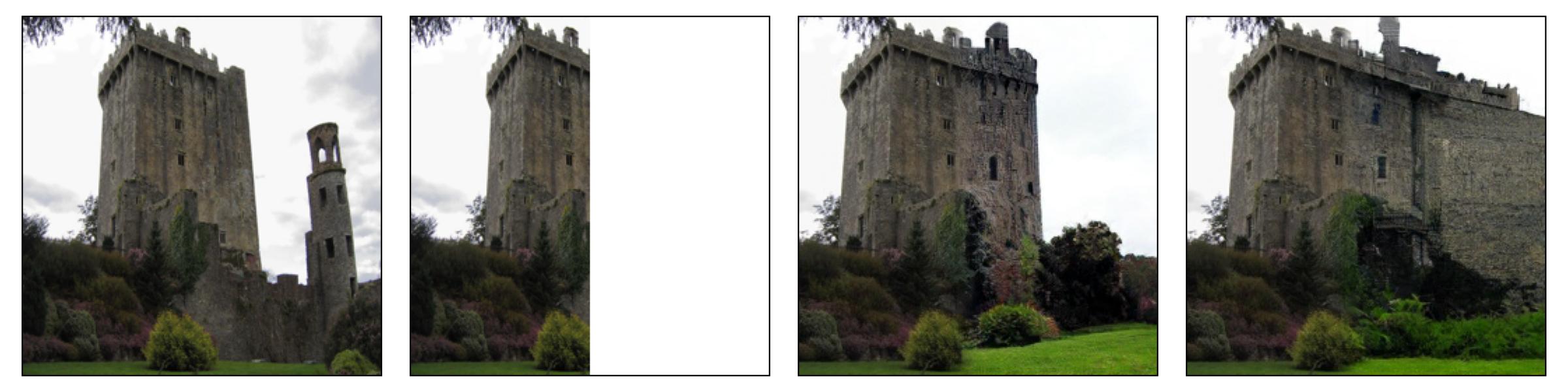}
    \end{subfigure}
    
    \begin{subfigure}[b]{\textwidth}
    \centering
    \includegraphics[width=\textwidth]{figure/caption_four.pdf}\vspace{-1mm}
    \caption{{Retain set}}
    \end{subfigure}
\caption{Ablation study: Rightward extension by VQ-GAN. We visualize he performance of the original model (before unlearning) and the obtained model by our approach (after unlearning). We show results on both forget set and retain set. }
\label{fig:gan_right}
\end{figure}

\begin{figure}[htb]\centering
    \begin{subfigure}[b]{0.495\textwidth}
    \centering
    \includegraphics[width=\textwidth]{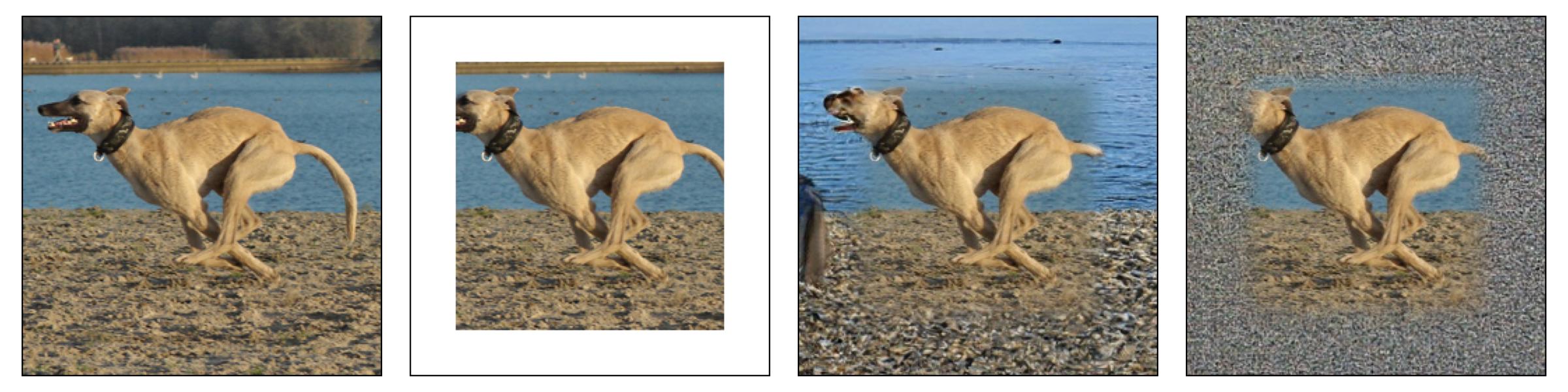}
    \end{subfigure}
    \hfill
    \begin{subfigure}[b]{0.495\textwidth}
    \centering
    \includegraphics[width=\textwidth]{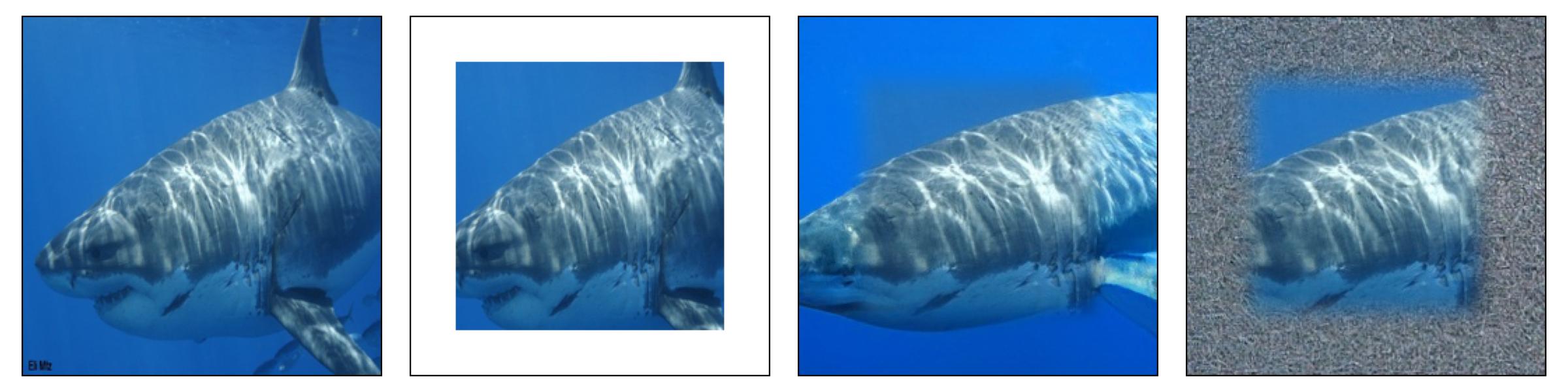}
    \end{subfigure}
    
    \begin{subfigure}[b]{0.495\textwidth}
    \centering
    \includegraphics[width=\textwidth]{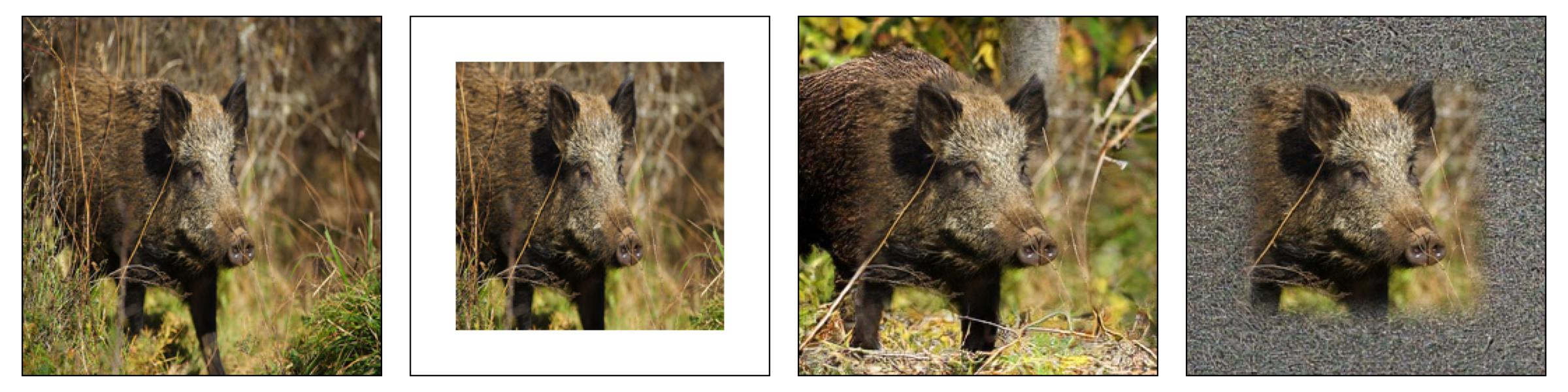}
    \end{subfigure}
    \hfill
    \begin{subfigure}[b]{0.495\textwidth}
    \centering
    \includegraphics[width=\textwidth]{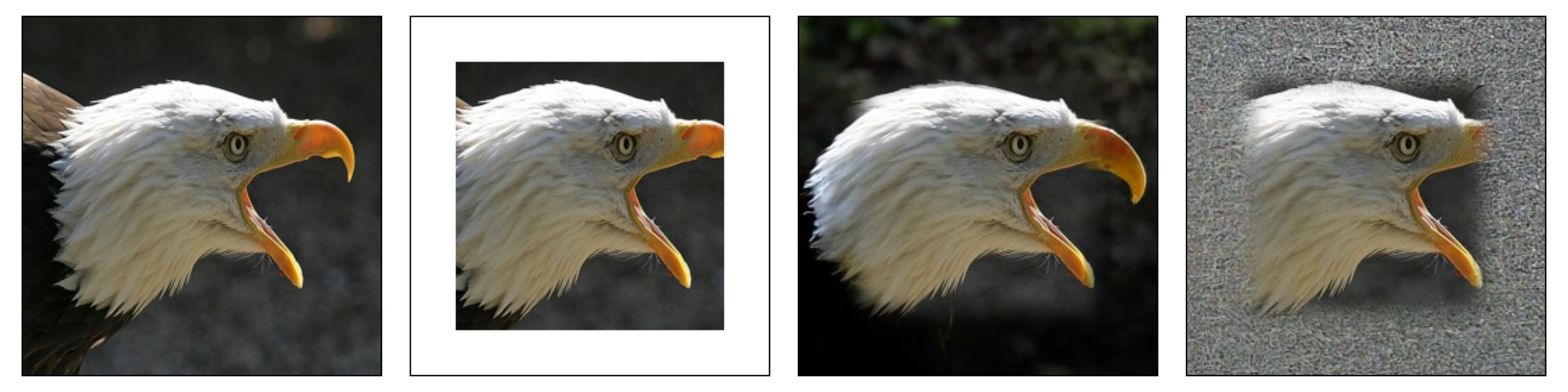}
    \end{subfigure}
    
    \begin{subfigure}[b]{0.495\textwidth}
    \centering
    \includegraphics[width=\textwidth]{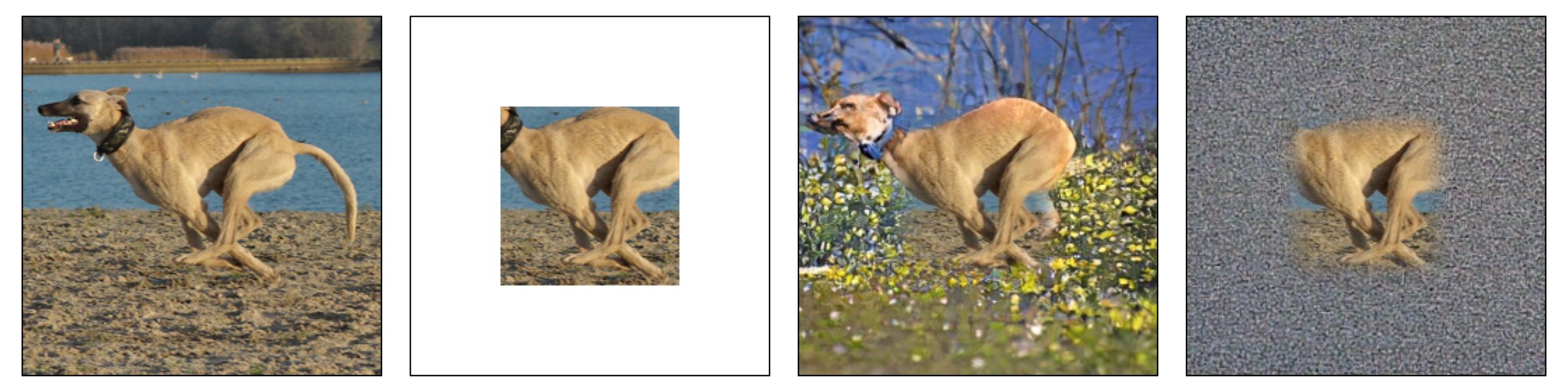}
    \end{subfigure}
    \hfill
    \begin{subfigure}[b]{0.495\textwidth}
    \centering
    \includegraphics[width=\textwidth]{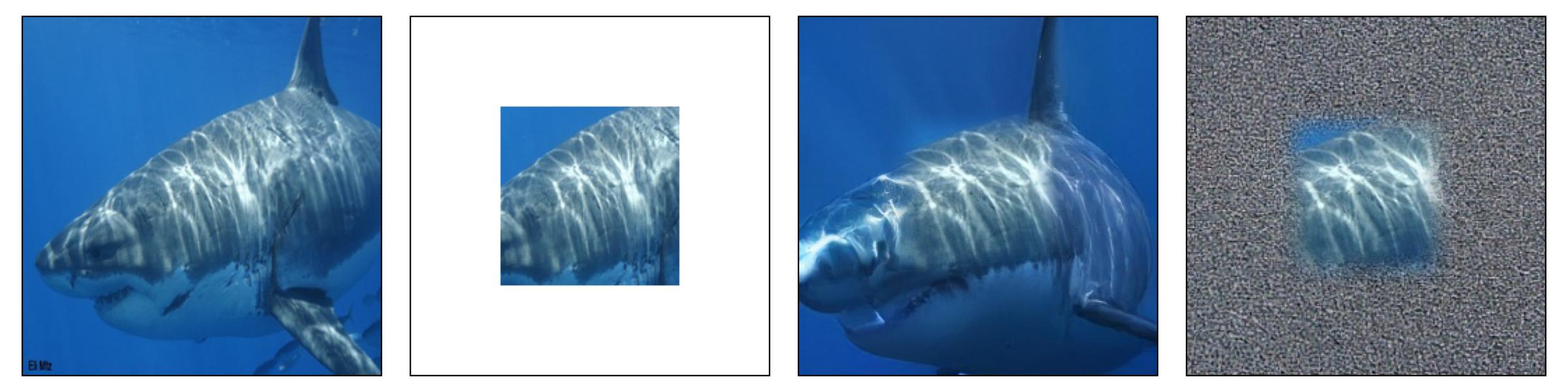}
    \end{subfigure}

    \begin{subfigure}[b]{0.495\textwidth}
    \centering
    \includegraphics[width=\textwidth]{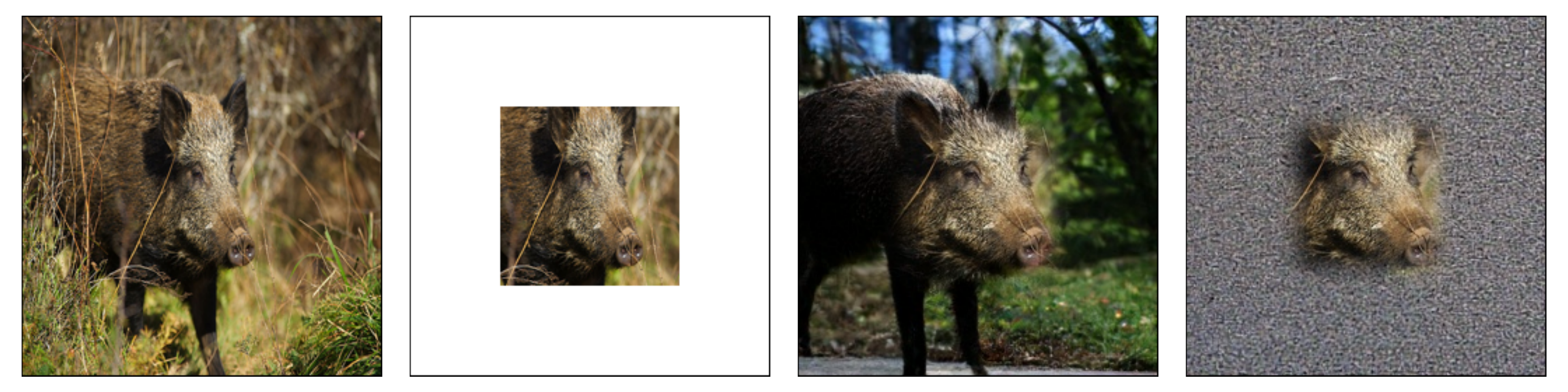}
    \end{subfigure}
    \hfill
    \begin{subfigure}[b]{0.495\textwidth}
    \centering
    \includegraphics[width=\textwidth]{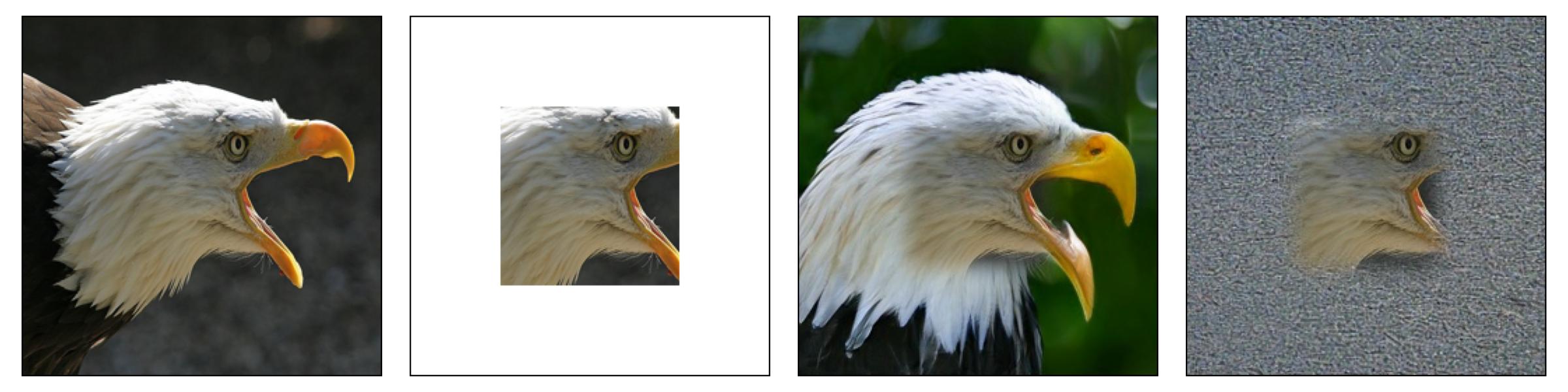}
    \end{subfigure}
    
    \begin{subfigure}[b]{\textwidth}
    \centering
    \includegraphics[width=\textwidth]{figure/caption_four.pdf}\vspace{-1mm}
    \caption{{Forget set}}
    \end{subfigure}
    \vspace{1mm}

    \begin{subfigure}[b]{0.495\textwidth}
    \centering
    \includegraphics[width=\textwidth]{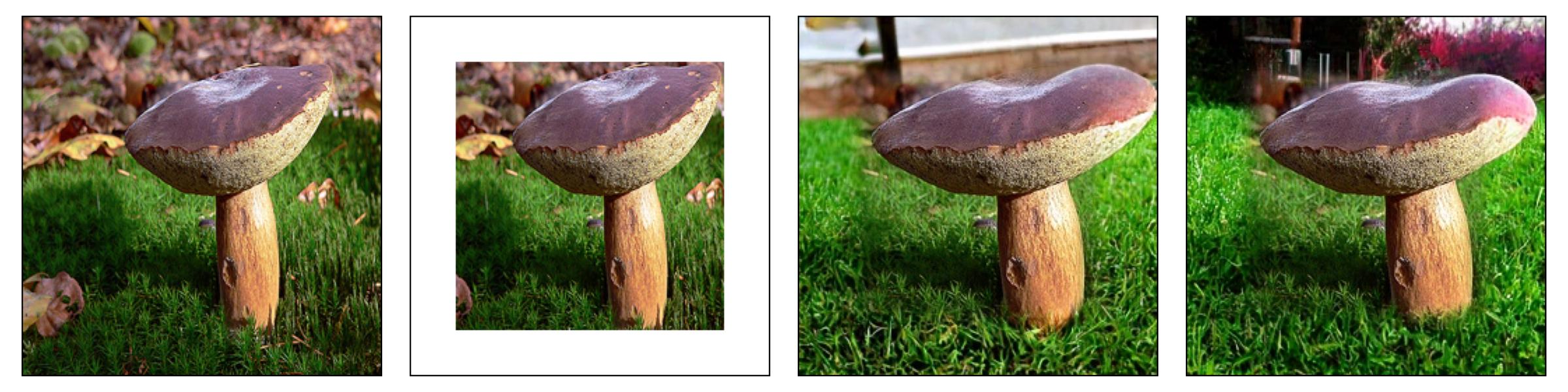}
    \end{subfigure}
    \hfill
    \begin{subfigure}[b]{0.495\textwidth}
    \centering
    \includegraphics[width=\textwidth]{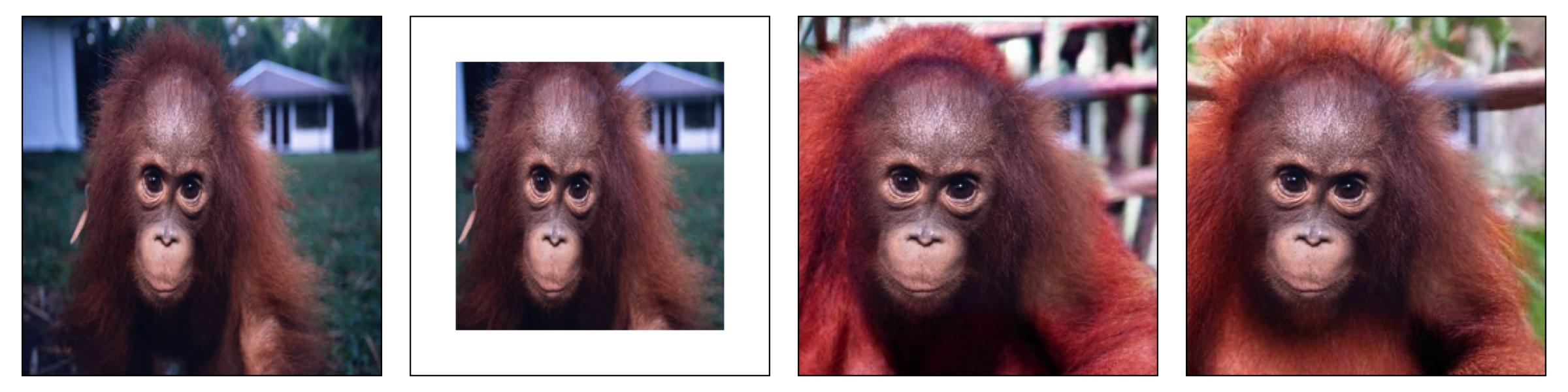}
    \end{subfigure}
    
    \begin{subfigure}[b]{0.495\textwidth}
    \centering
    \includegraphics[width=\textwidth]{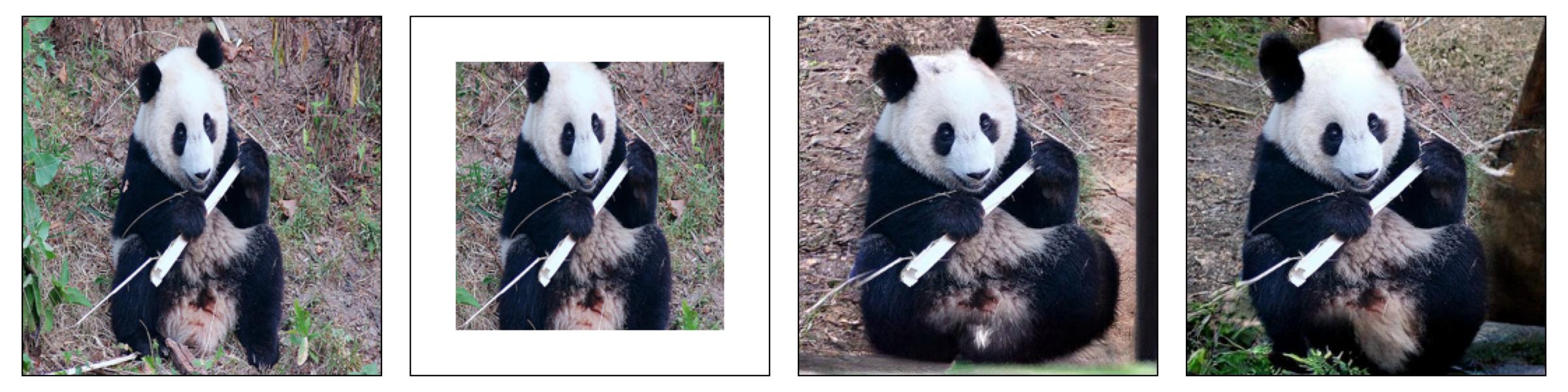}
    \end{subfigure}
    \hfill
    \begin{subfigure}[b]{0.495\textwidth}
    \centering
    \includegraphics[width=\textwidth]{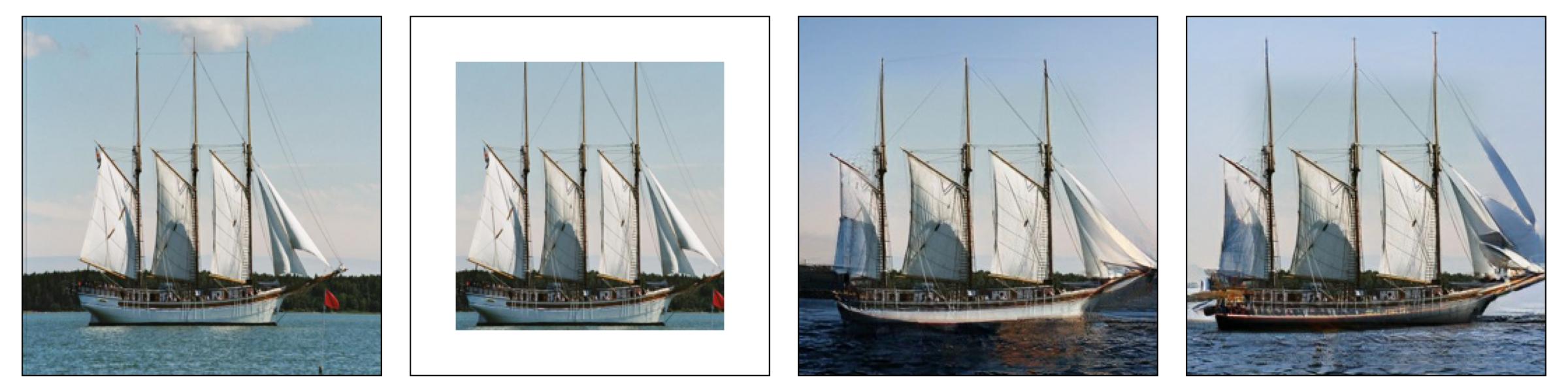}
    \end{subfigure}
    
    \begin{subfigure}[b]{0.495\textwidth}
    \centering
    \includegraphics[width=\textwidth]{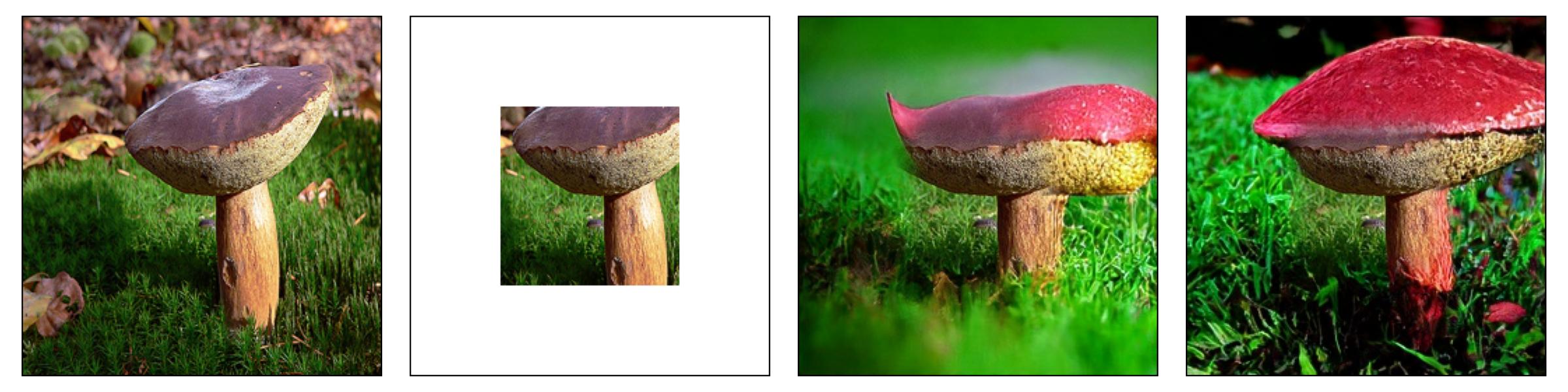}
    \end{subfigure}
    \hfill
    \begin{subfigure}[b]{0.495\textwidth}
    \centering
    \includegraphics[width=\textwidth]{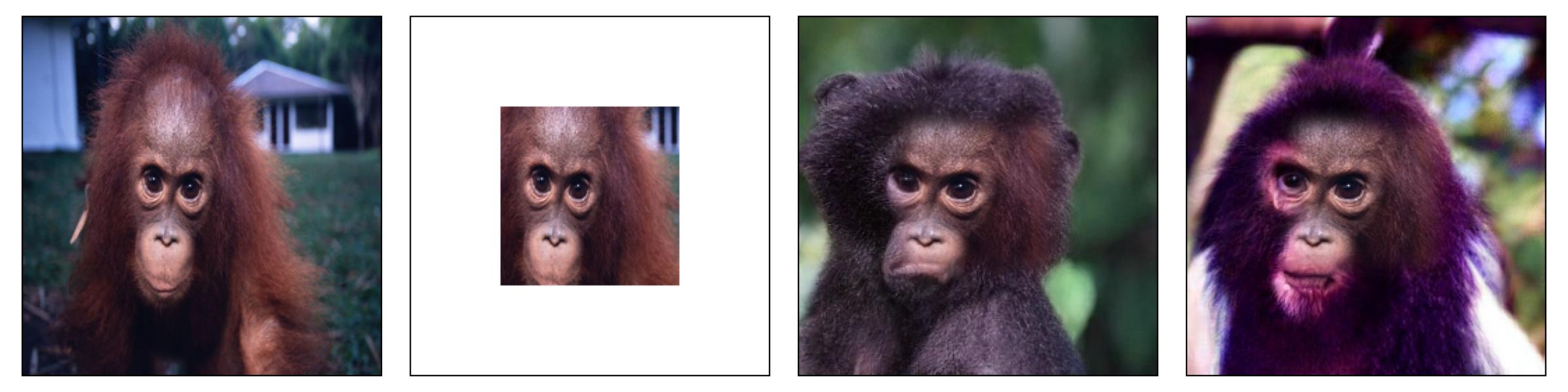}
    \end{subfigure}
    
    \begin{subfigure}[b]{0.495\textwidth}
    \centering
    \includegraphics[width=\textwidth]{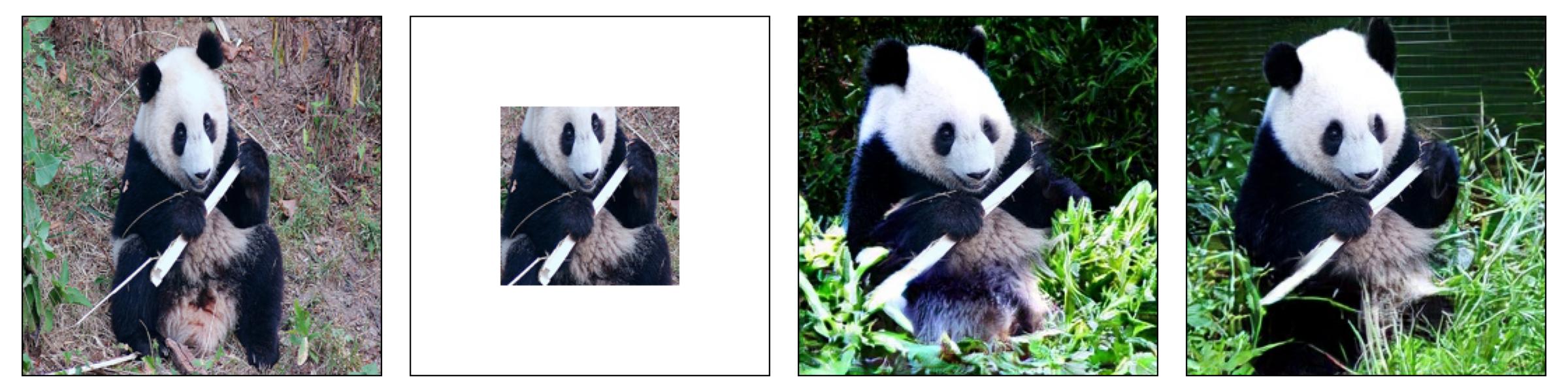}
    \end{subfigure}
    \hfill
    \begin{subfigure}[b]{0.495\textwidth}
    \centering
    \includegraphics[width=\textwidth]{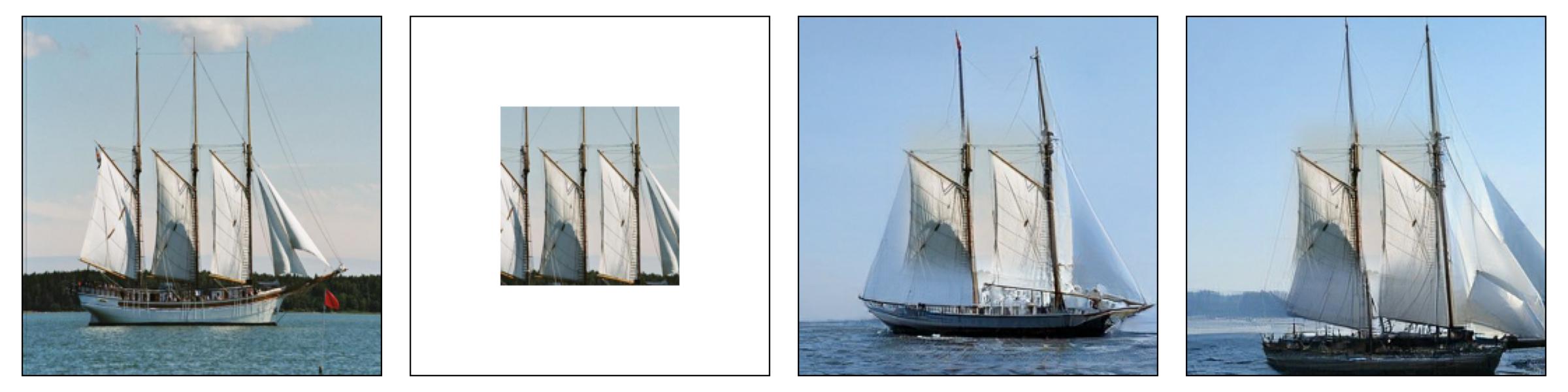}
    \end{subfigure}
    
    \begin{subfigure}[b]{\textwidth}
    \centering
    \includegraphics[width=\textwidth]{figure/caption_four.pdf}\vspace{-1mm}
    \caption{{Retain set}}
    \end{subfigure}
\caption{Ablation study: Outpaint by VQ-GAN. We visualize he performance of the original model (before unlearning) and the obtained model by our approach (after unlearning). We show results on both forget set and retain set. }
\label{fig:gan_outpaint}
\end{figure}

\begin{figure}[htb]\centering
    \begin{subfigure}[b]{0.32\textwidth}
    \centering
    \includegraphics[width=\textwidth]{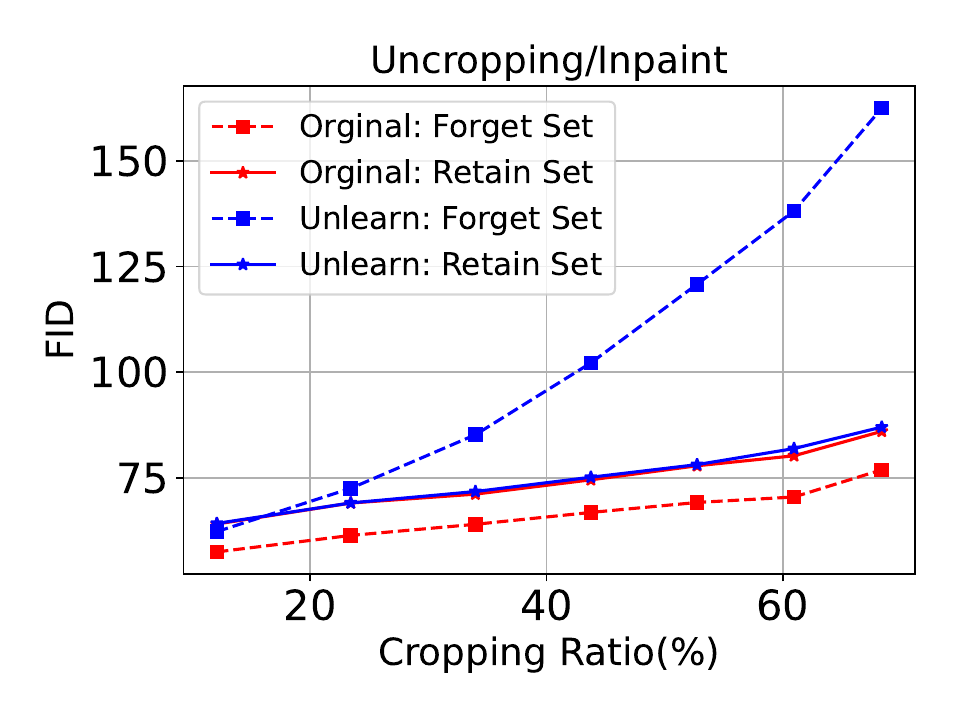}
    \end{subfigure}
    \hfill
    \begin{subfigure}[b]{0.32\textwidth}
    \centering
    \includegraphics[width=\textwidth]{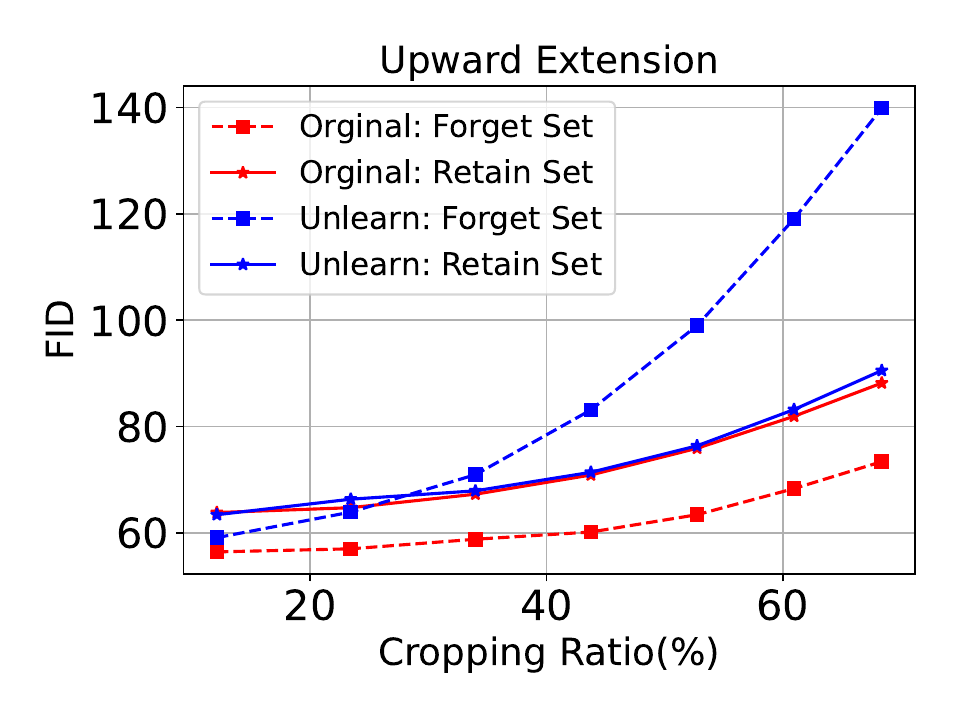}
    \end{subfigure}
    \hfill
    \begin{subfigure}[b]{0.32\textwidth}
    \centering
    \includegraphics[width=\textwidth]{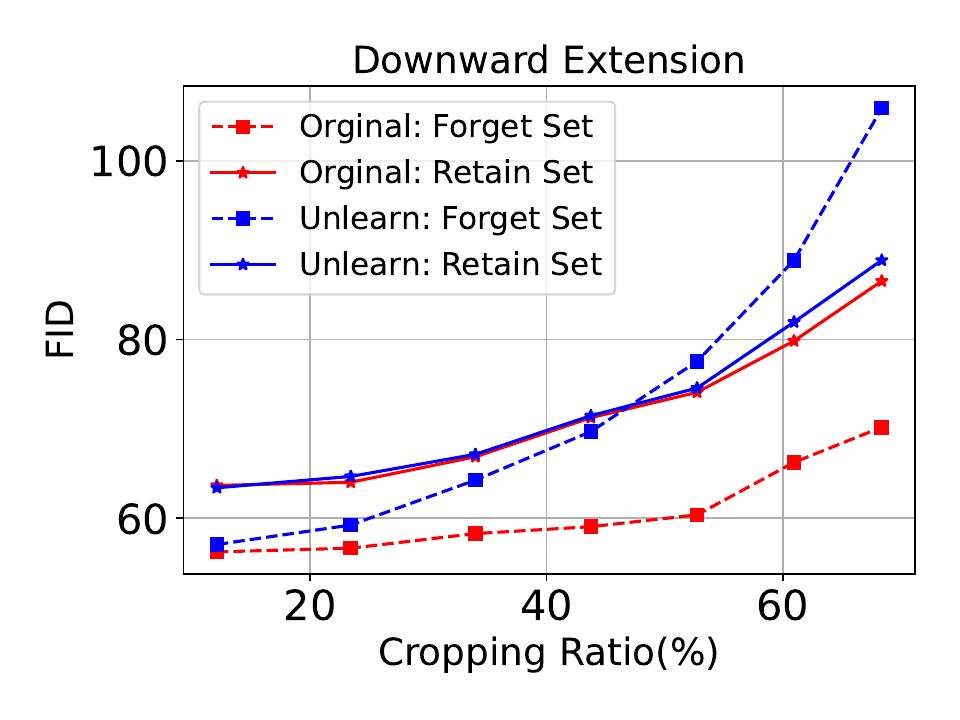}
    \end{subfigure}

    \begin{subfigure}[b]{0.32\textwidth}
    \centering
    \includegraphics[width=\textwidth]{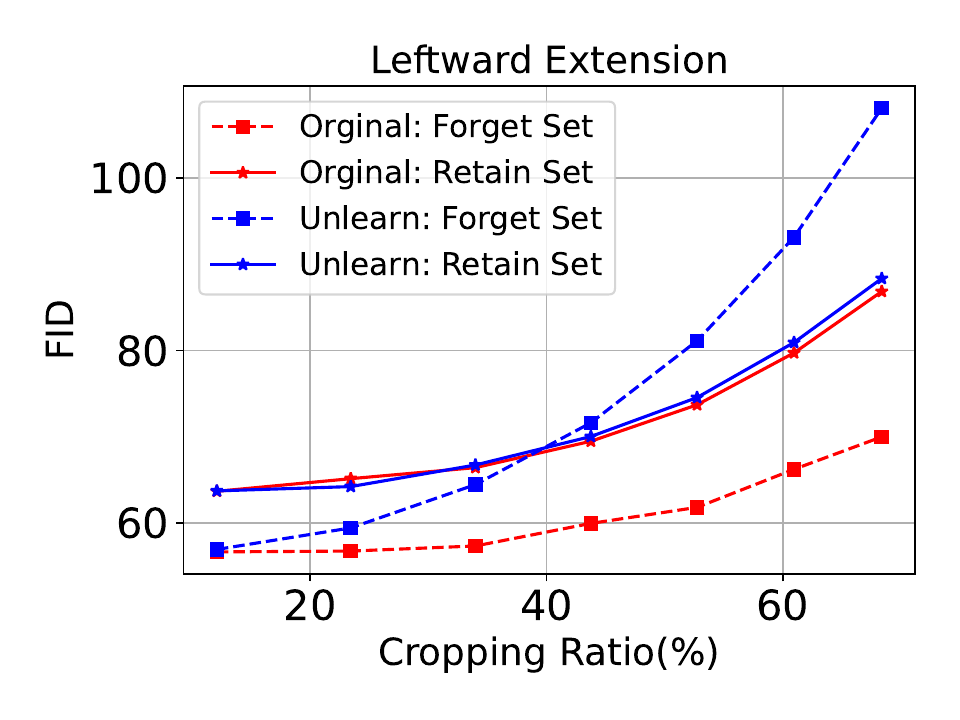}
    \end{subfigure}
    \hfill
    \begin{subfigure}[b]{0.32\textwidth}
    \centering
    \includegraphics[width=\textwidth]{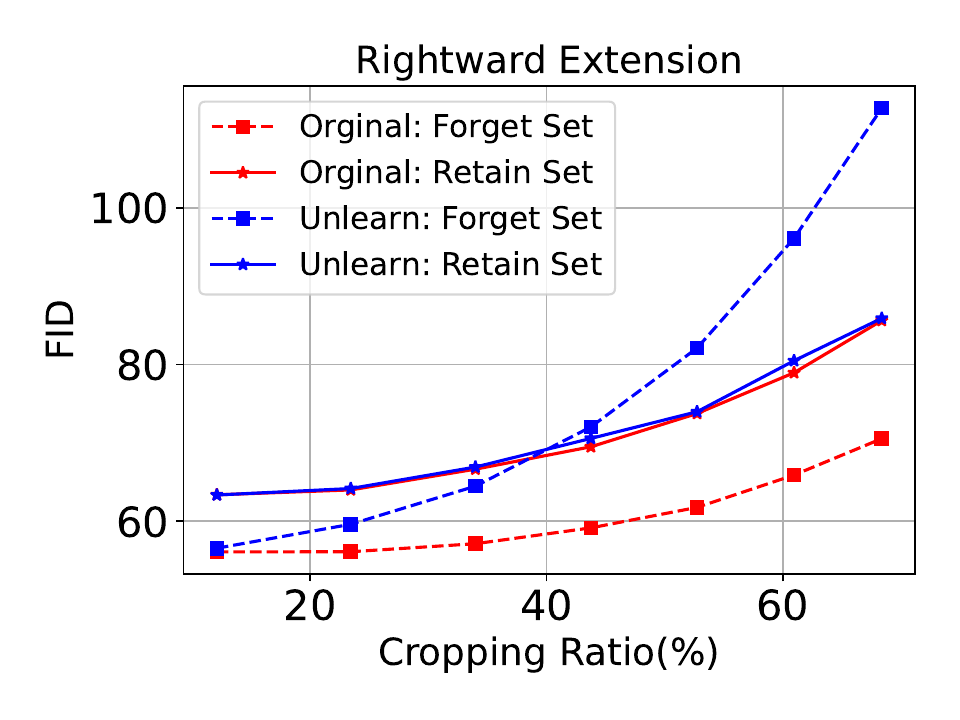}
    \end{subfigure}
    \hfill
    \begin{subfigure}[b]{0.32\textwidth}
    \centering
    \includegraphics[width=\textwidth]{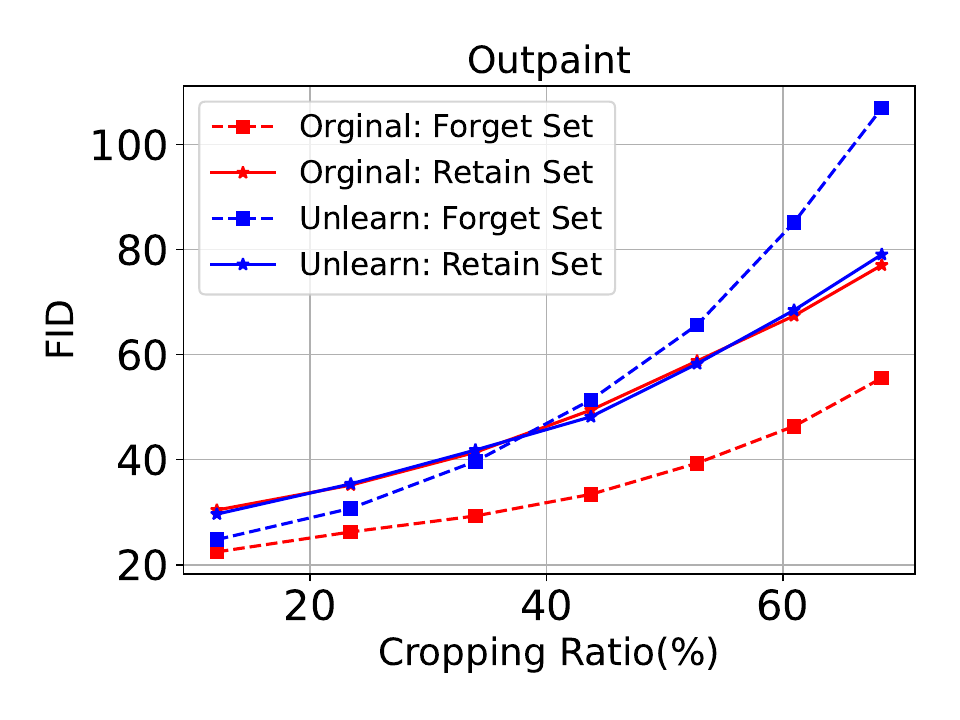}
    \end{subfigure}
\caption{Ablation study: FID \textit{vs.} input cropping ratios under different I2I generation tasks. }
\label{fig:gan_ratios_FID}
\end{figure}    
 
\begin{figure}[htb]\centering
    \begin{subfigure}[b]{0.32\textwidth}
    \centering
    \includegraphics[width=\textwidth]{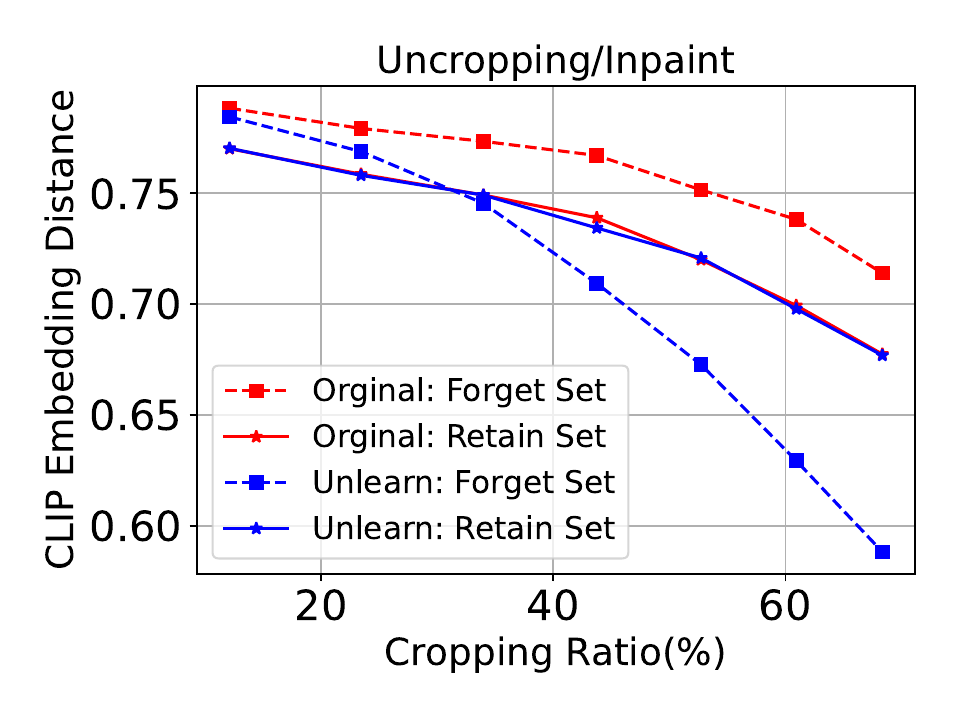}
    \end{subfigure}
    \hfill
    \begin{subfigure}[b]{0.32\textwidth}
    \centering
    \includegraphics[width=\textwidth]{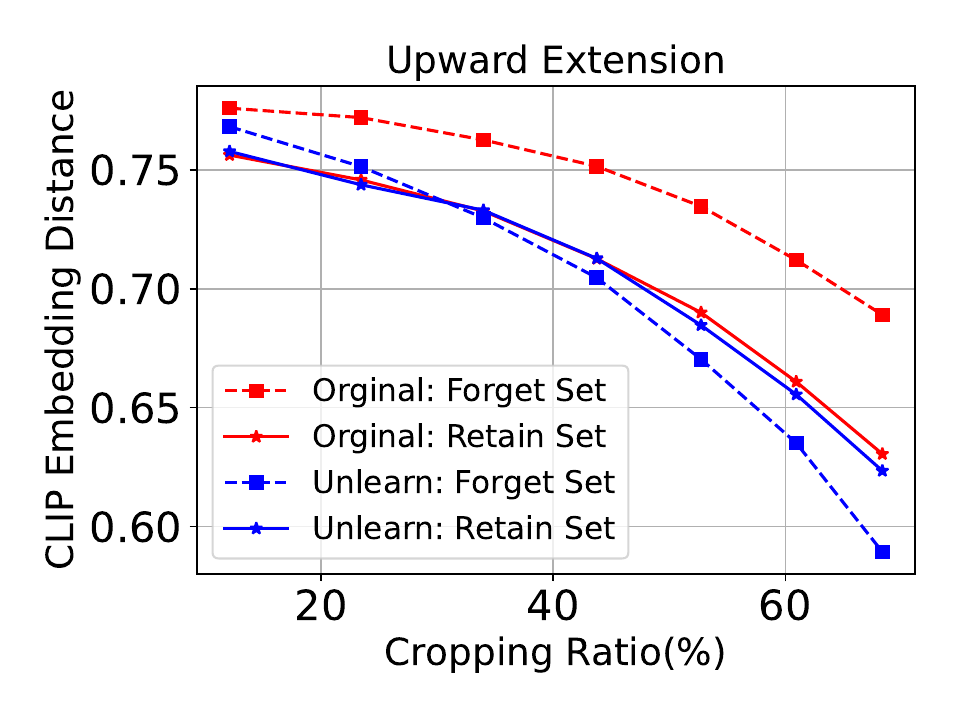}
    \end{subfigure}
    \hfill
    \begin{subfigure}[b]{0.32\textwidth}
    \centering
    \includegraphics[width=\textwidth]{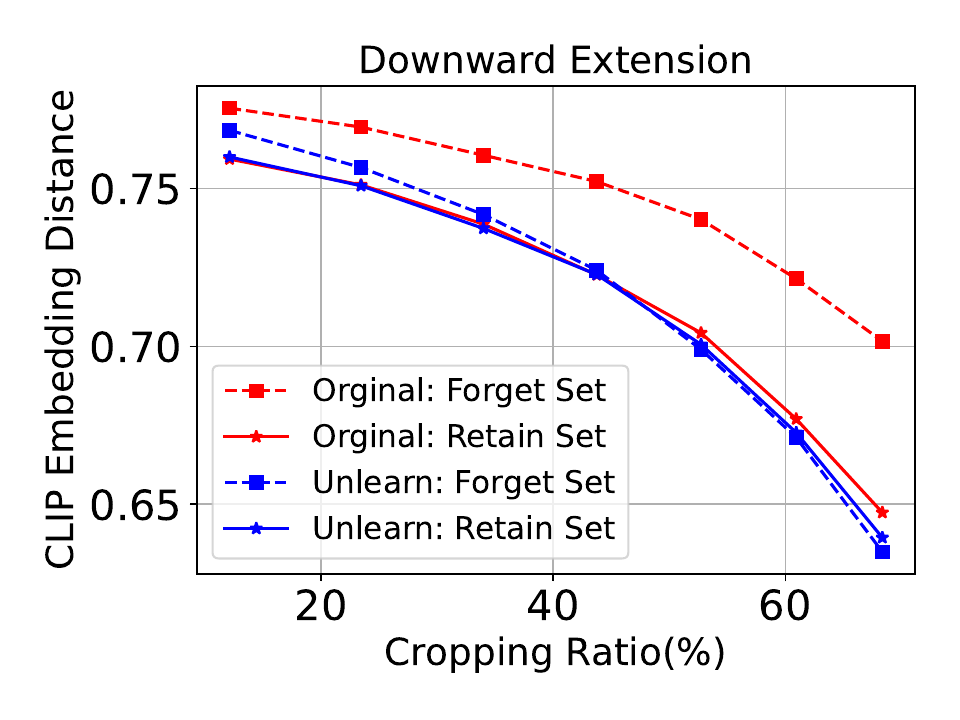}
    \end{subfigure}

    \begin{subfigure}[b]{0.32\textwidth}
    \centering
    \includegraphics[width=\textwidth]{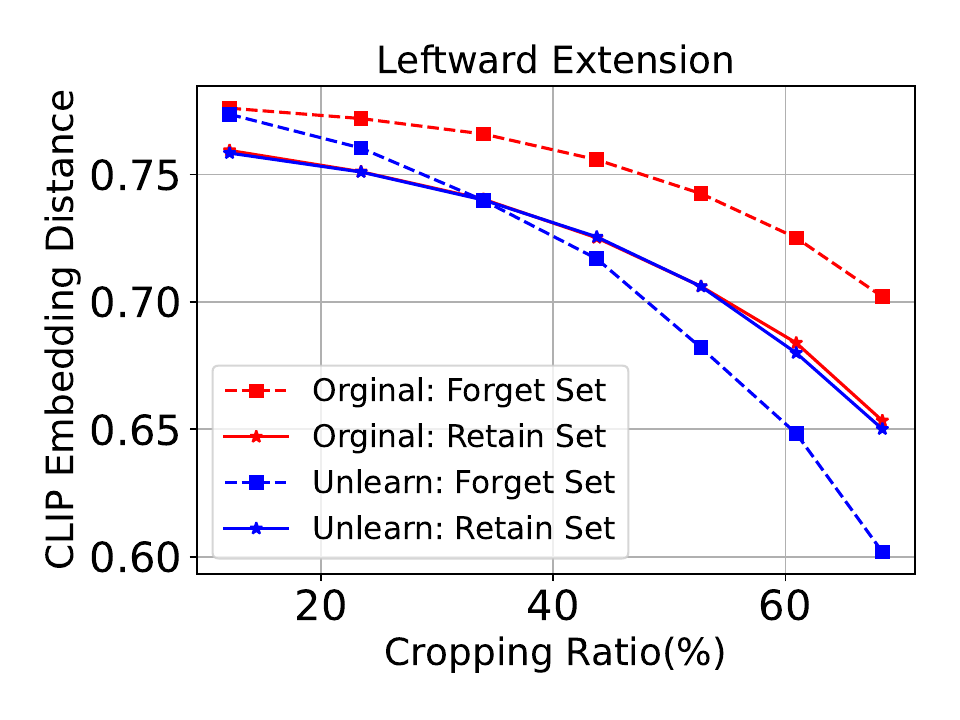}
    \end{subfigure}
    \hfill
    \begin{subfigure}[b]{0.32\textwidth}
    \centering
    \includegraphics[width=\textwidth]{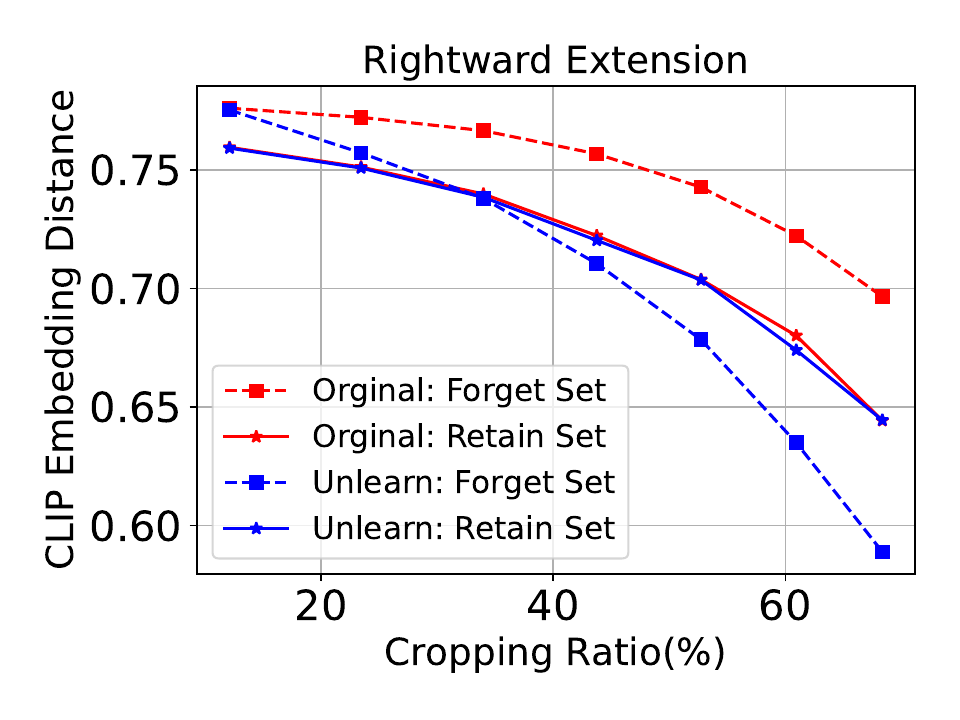}
    \end{subfigure}
    \hfill
    \begin{subfigure}[b]{0.32\textwidth}
    \centering
    \includegraphics[width=\textwidth]{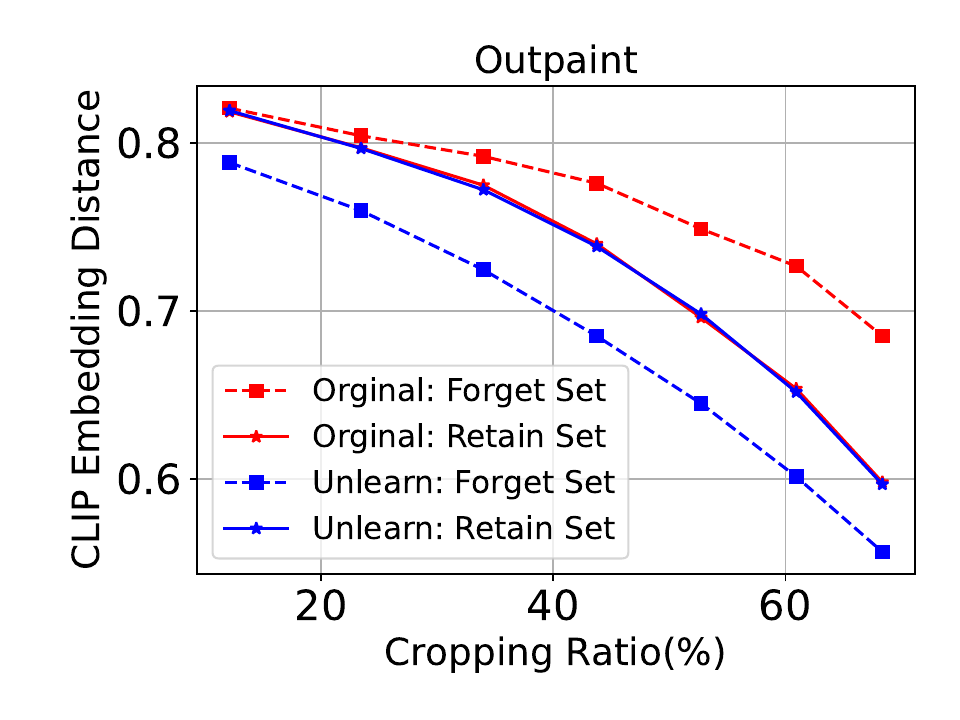}
    \end{subfigure}    

\caption{Ablation study: CLIP embedding distance \textit{vs.} input cropping ratios under different  I2I generation tasks. }
\label{fig:gan_ratios_clip}
\end{figure}

\subsection{Varying Cropping Pattern and Ratio}\label{app:varyingmaskingtype_ratio}
In the main paper, we primarily show the results of uncropping/inpainting tasks. We vary the cropping patterns and use the VQ-GAN to generate the images under these different patterns. We shown the generated images with downward extension in Fig.~\ref{fig:gan_down}, upward extension in Fig.~\ref{fig:gan_up}, leftward extension in Fig.~\ref{fig:gan_left}, rightward extension in Fig.~\ref{fig:gan_right}, and outpaint in Fig.~\ref{fig:gan_outpaint}, respectively. 
As shown in these images, our method is  robust to different cropping types. 

As shown in Fig.~\ref{fig:gan_ratios_FID} and Fig.~\ref{fig:gan_ratios_clip}, we also vary the input cropping ratio and report the FID and CLIP embedding distance. Clearly, our approach is robust to different cropping ratios.

\begin{table}[h]
\centering
\caption{The comparison of cross validation in Appendix~\ref{app:cross_val}. We first use different models to do the inpainting tasks, i.e., reconstructing the central cropped patches (\textsc{First} reconstructions). Given the \textsc{First} reconstructions as input, we then use the original model (before unlearning) to do the outpainting tasks and get the \textsc{Second} reconstructions, i.e., re-generated the outer patches based on these reconstructed central patches from \textsc{First} reconstructions. We then evaluate the quality of these \textsc{Second} reconstructed images under various metrics. Lower FID, higher IS, and higher CLIP values indicate higher image quality.}\label{tab:cross_val}

    \begin{tabular}{ccccccc}
    \toprule
    & \multicolumn{2}{c}{FID}            & \multicolumn{2}{c}{IS}            & \multicolumn{2}{c}{CLIP}          \\ \cmidrule(lr){2-3}\cmidrule(lr){4-5}\cmidrule(lr){6-7}
    & \multicolumn{1}{c}{$D_R$} & $D_F$  & \multicolumn{1}{c}{$D_R$} & $D_F$ & \multicolumn{1}{c}{$D_R$} & $D_F$ \\ \toprule
    Original Model    & \multicolumn{1}{c}{36.14} & 28.96  & \multicolumn{1}{c}{27.28} & 27.95 & \multicolumn{1}{c}{0.37}  & 0.49  \\ \midrule
    Unlearned   Model & \multicolumn{1}{c}{36.74} & 156.68 & \multicolumn{1}{c}{26.62} & 7.67  & \multicolumn{1}{c}{0.37}  & 0.28  \\ \bottomrule
    \end{tabular}
\end{table}

\begin{figure}[htb]
     \centering
      \begin{subfigure}[b]{0.32\textwidth}
     \centering
     \includegraphics[width=\textwidth]{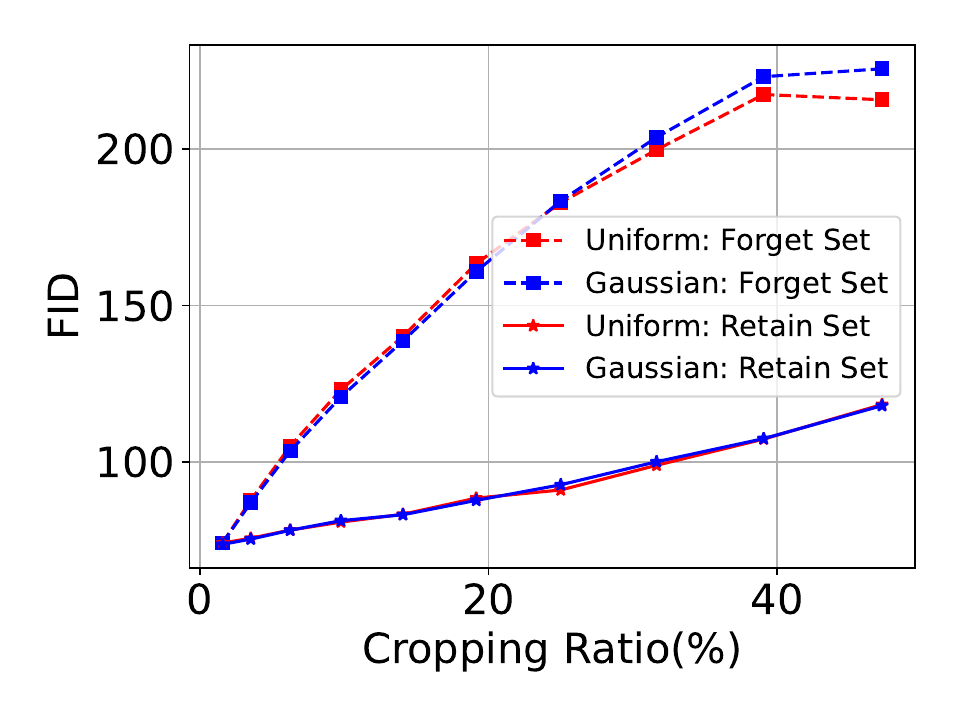}\vspace{-1mm}
     \caption{VQ-GAN: FID}
     \end{subfigure}
     \hfill
     \begin{subfigure}[b]{0.32\textwidth}
     \centering
     \includegraphics[width=\textwidth]{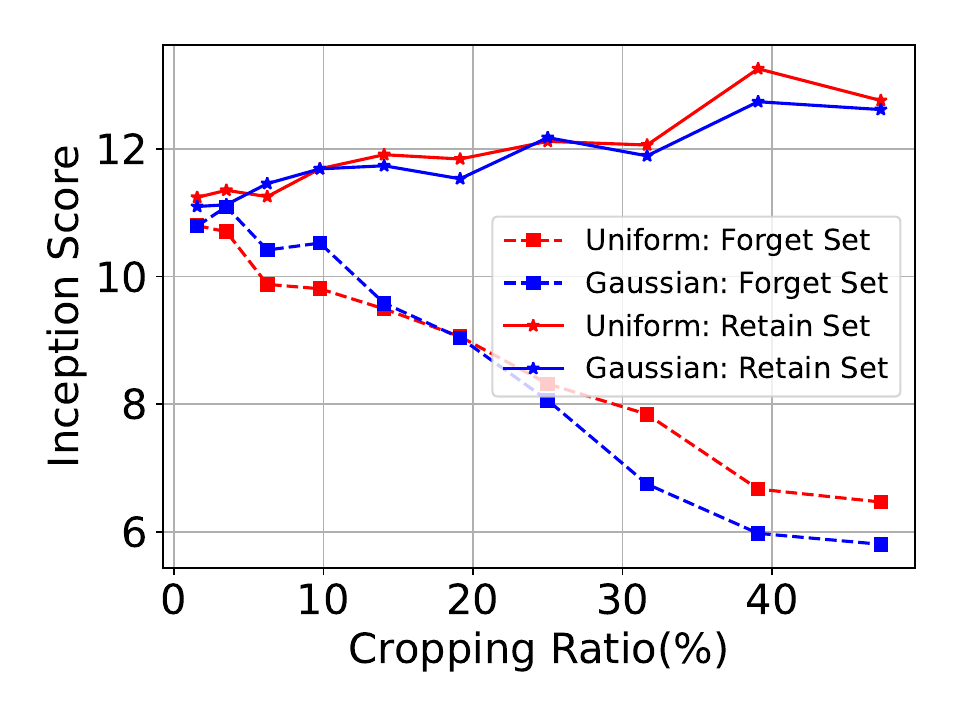}\vspace{-1mm}
     \caption{VQ-GAN: IS}
     \end{subfigure}
      \hfill
     \begin{subfigure}[b]{0.32\textwidth}
     \centering
     \includegraphics[width=\textwidth]{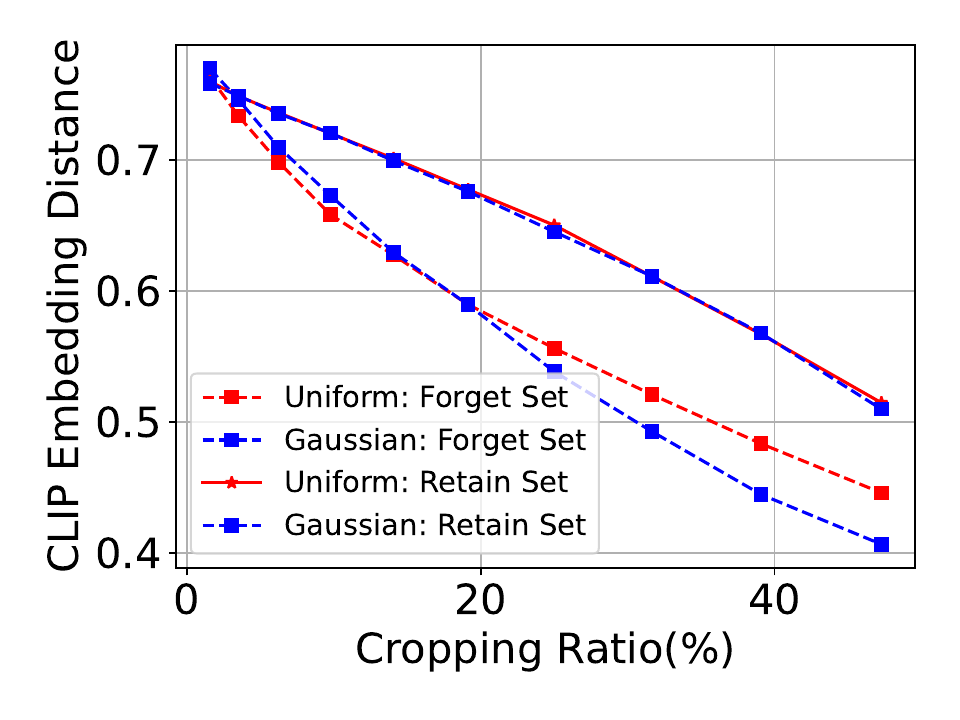}\vspace{-1mm}
     \caption{VQ-GAN: CLIP}
     \end{subfigure}
     
     \begin{subfigure}[b]{0.32\textwidth}
     \centering
     \includegraphics[width=\textwidth]{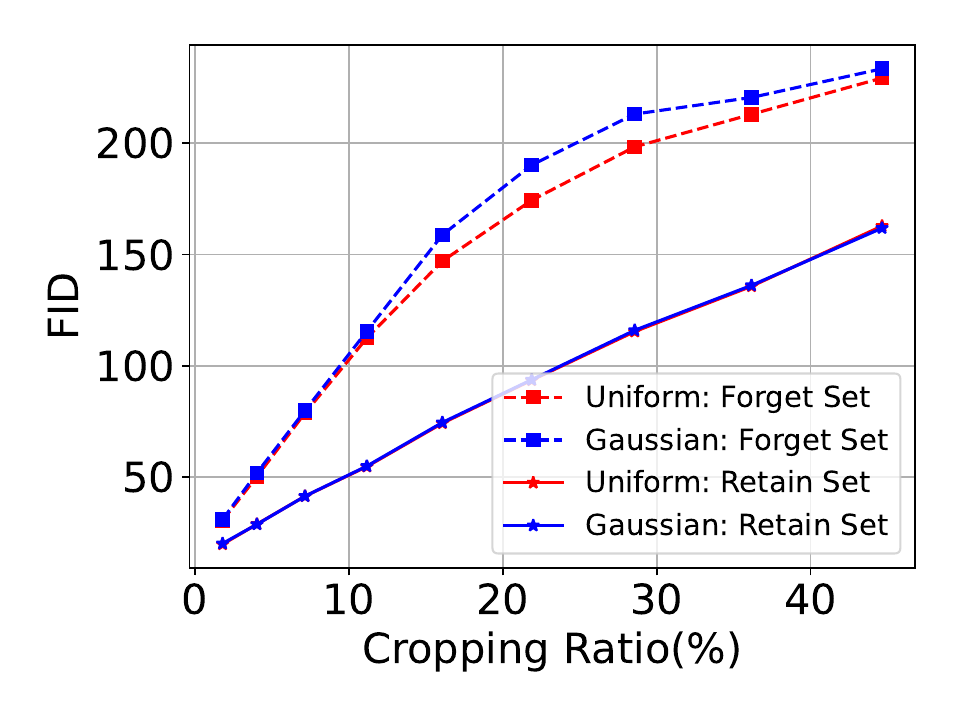}\vspace{-1mm}
     \caption{MAE: FID}
     \end{subfigure}
     \hfill
     \begin{subfigure}[b]{0.32\textwidth}
     \centering
     \includegraphics[width=\textwidth]{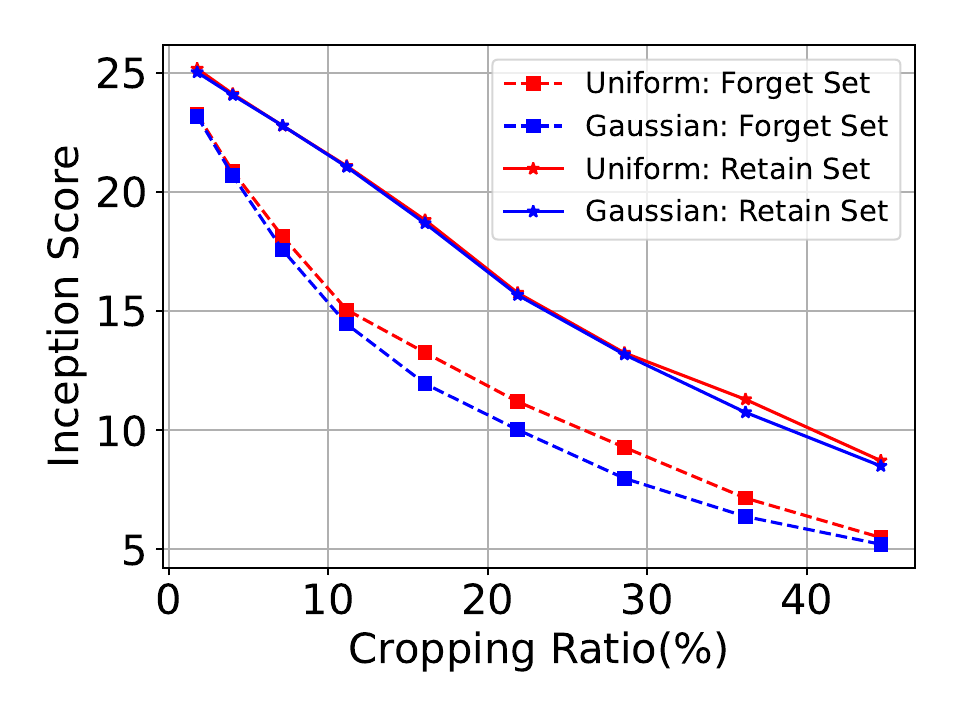}\vspace{-1mm}
     \caption{MAE: IS}
     \end{subfigure}
      \hfill
     \begin{subfigure}[b]{0.32\textwidth}
     \centering
     \includegraphics[width=\textwidth]{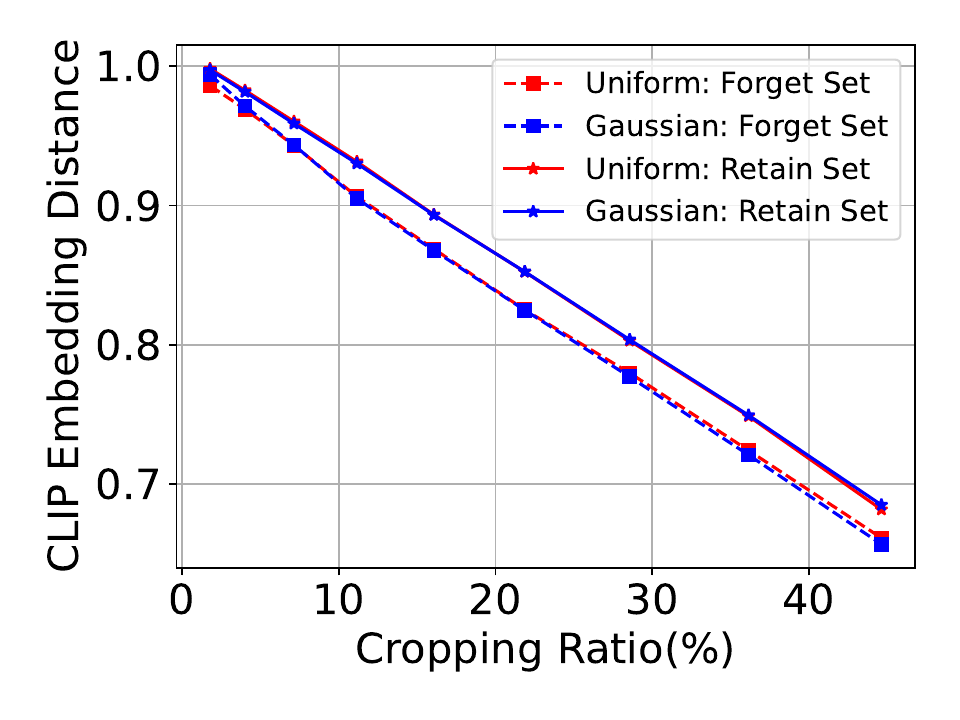}\vspace{-1mm}
     \caption{MAE: CLIP}
     \end{subfigure}
\caption{Ablation study of noise type. We test the performance under varying ratios of central cropping; \eg, cropping $8\times8$ out of 256 patches means a 25\% cropping ratio. For VQ-GAN and MAE, Gaussian noise achieves better forgetting in the forget set in general (\ie, it achieves higher FID, lower IS and lower CLIP on forget set).}\vspace{-4mm}
\label{fig:noise_type}
\end{figure}

\paragraph{Noise type.} We compare the performance of Gaussian Noise used in our method with uniform noise for VQ-GAN and MAE. Specifically, we set the noise in Eq.~\eref{eq:final_optimization} as $n\sim\mathcal{U}[-1,1]$ then conduct the unlearning under the same setup.
We use the obtained models to generate 500 images for both retain set and forget set. We then compute  multiple metrics for different noise types. As shown in Fig.~\ref{fig:noise_type}, Gaussian noise achieves better forgetting in the forget set in general (\ie, higher FID, lower IS and lower CLIP on forget sets). These results empirically verify the benefits of using Gaussian Noise, which coincides with our analysis (cf.~Lemma~\ref{lemma:optimal_gaussian}).

\vspace{-2mm}

\subsection{Cross Validation of Generated Images}\label{app:cross_val}
To further verify the generated images do not have the information from forget set, we conducted the following experiment for VQ-GAN:
\begin{itemize}
    \item We first conduct the unlearning for VQ-GAN for ImageNet to obtain the \textit{unlearned model}, under the same setup mentioned in Section~\ref{sec:exp}.
    \item Second, given the center-cropped input images (cropping central $8\times 8$ patches), we then use this \textit{unlearned model} to generate the images on both forget set and retain set (i.e., image uncropping/inpainting). Here we call them \textsc{First} generated images.
    \item Given these \textsc{First} generated images, we only keep the reconstructed central $8\times 8$ patches (cropping the outer ones) as the new input to the original VQ-GAN (i.e., the model before unlearning) and get the newly generated images (here we call them \textsc{Second} generated images). We then evaluate the quality of \textsc{Second} generated images for both forget set and retain set and report the results. 
    \item We also conduct the above process for the \textit{original model} (i.e., before unlearning) as the baseline/reference. The results are given below:
\end{itemize}

\begin{figure}[htb]\centering
    \begin{subfigure}[b]{0.48\textwidth}
    \centering
    \includegraphics[width=\textwidth]{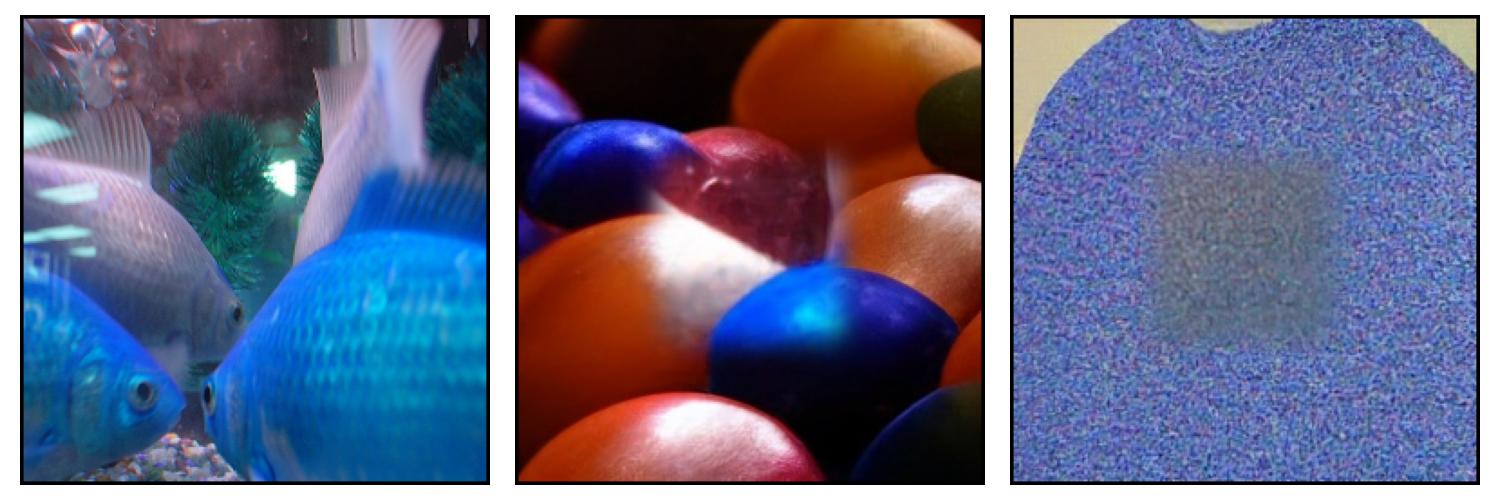}
    \end{subfigure}
    \hfill
    \begin{subfigure}[b]{0.48\textwidth}
    \centering
    \includegraphics[width=\textwidth]{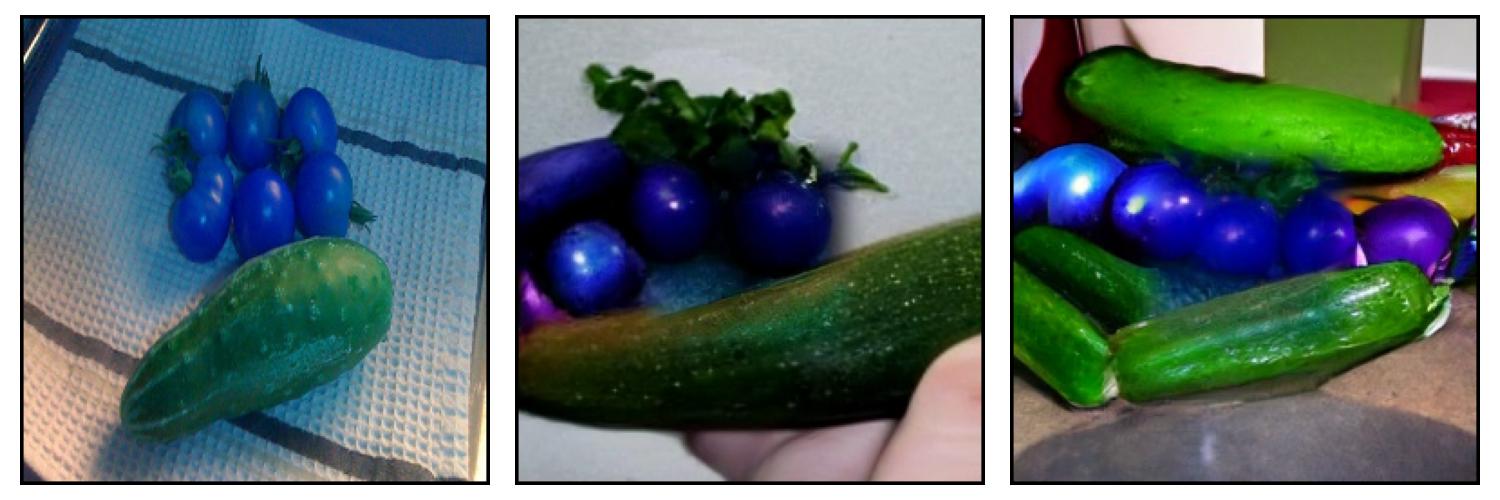}
    \end{subfigure}
    
    \begin{subfigure}[b]{0.48\textwidth}
    \centering
    \includegraphics[width=\textwidth]{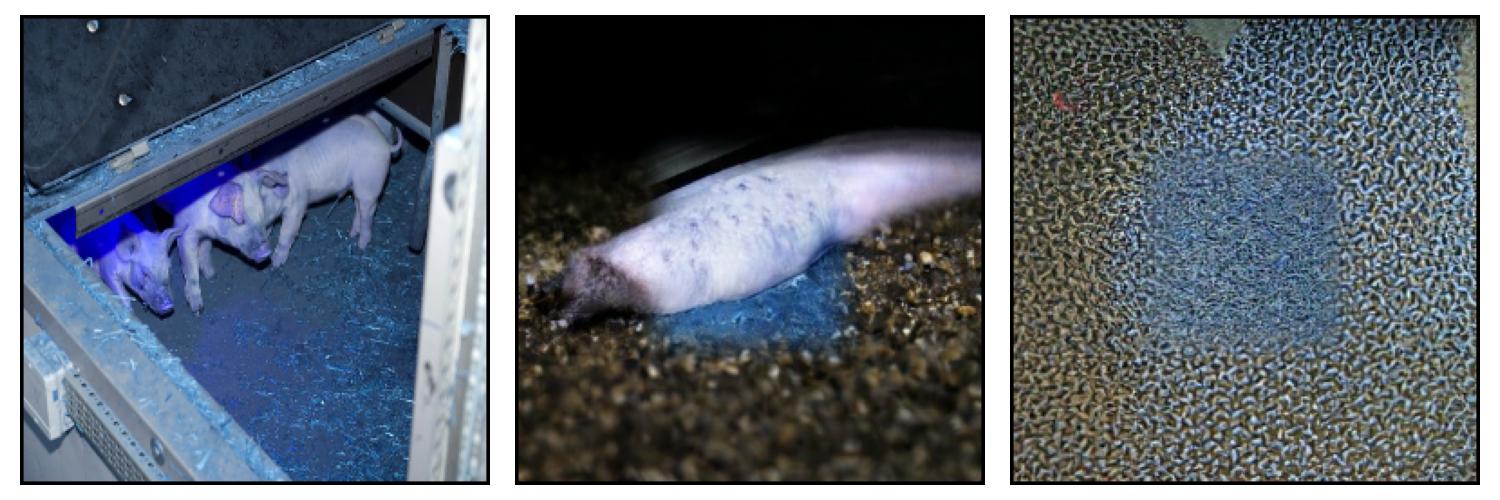}
    \end{subfigure}
    \hfill
    \begin{subfigure}[b]{0.48\textwidth}
    \centering
    \includegraphics[width=\textwidth]{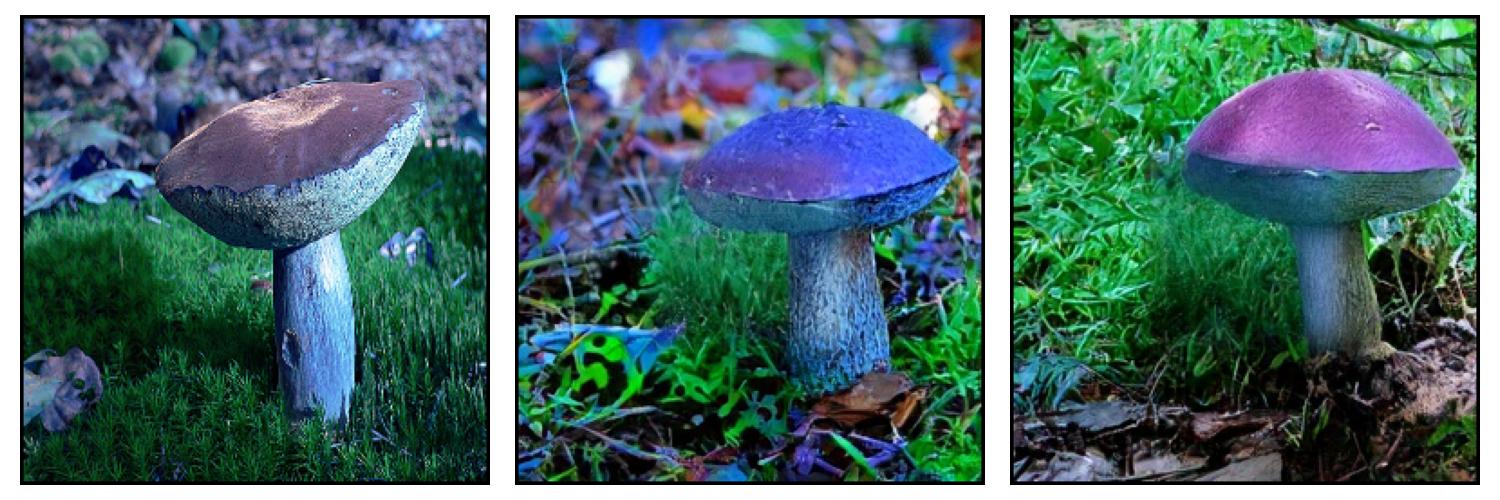}
    \end{subfigure}
    
    \begin{subfigure}[b]{0.48\textwidth}
    \centering
    \includegraphics[width=\textwidth]{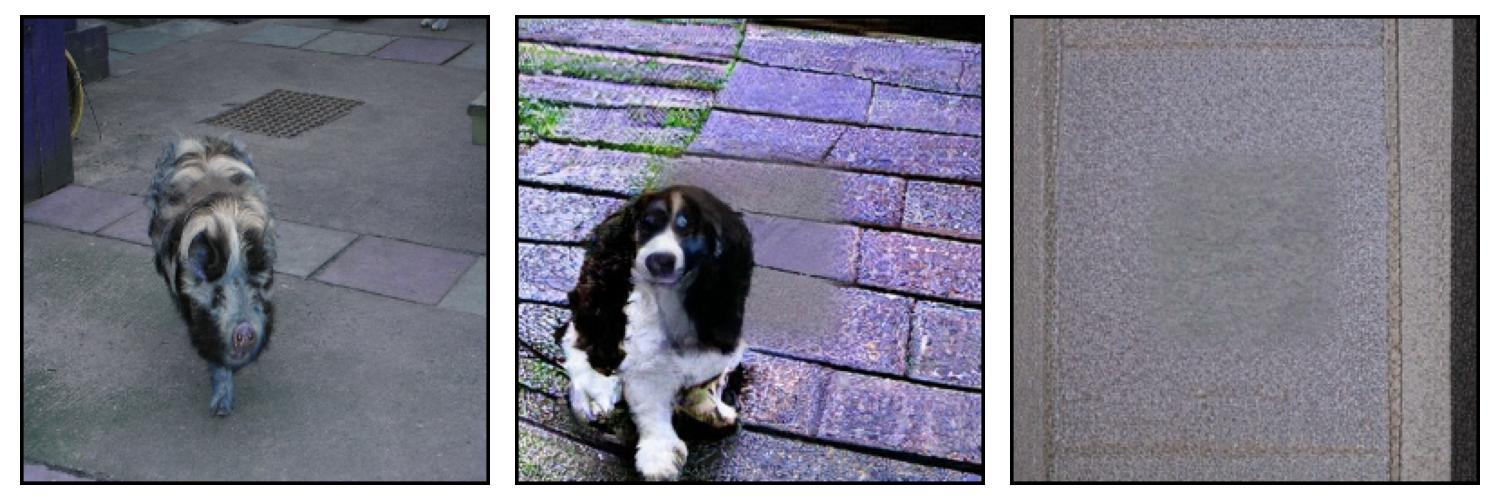}
    \end{subfigure}
    \hfill
    \begin{subfigure}[b]{0.48\textwidth}
    \centering
    \includegraphics[width=\textwidth]{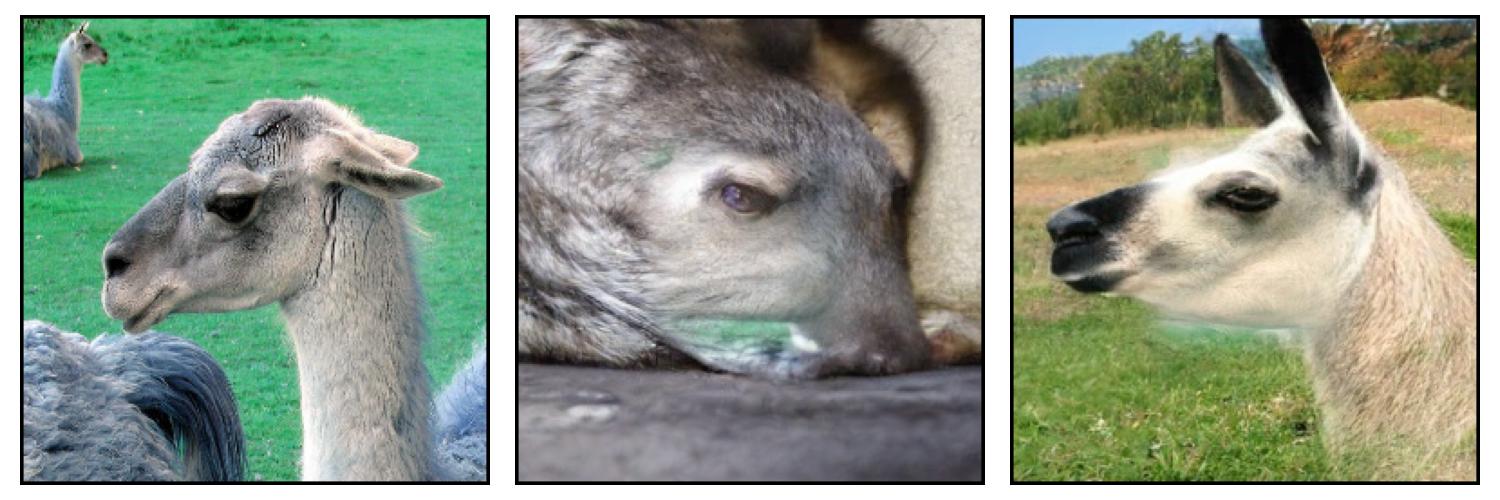}
    \end{subfigure}

    \begin{subfigure}[b]{0.48\textwidth}
    \centering
    \includegraphics[width=\textwidth]{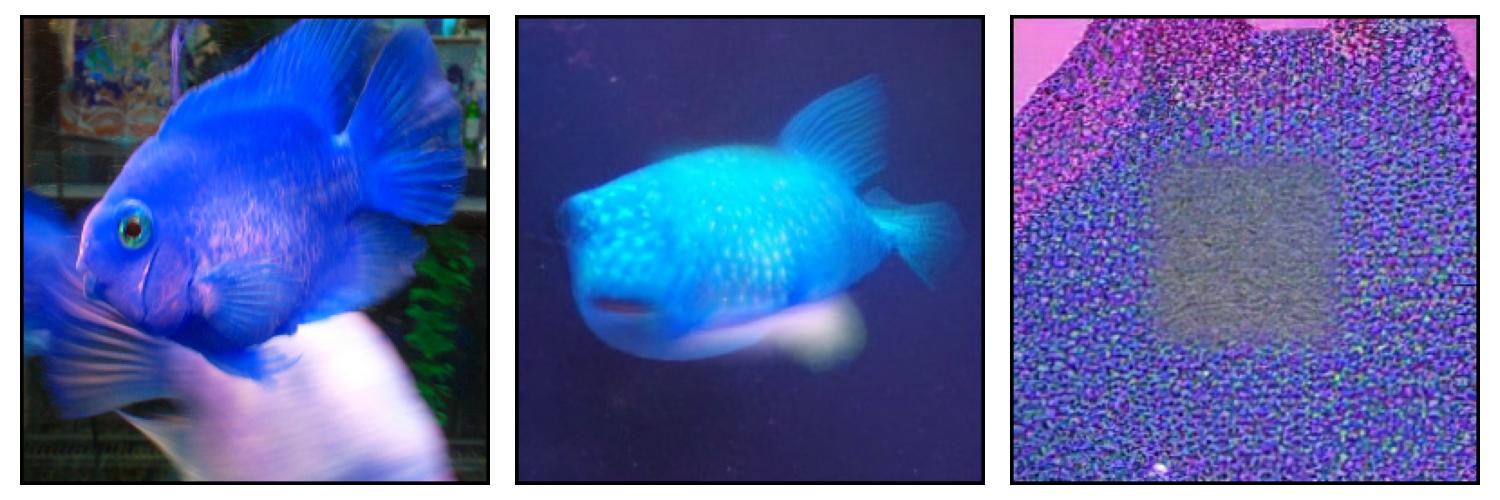}
    \end{subfigure}
    \hfill
    \begin{subfigure}[b]{0.48\textwidth}
    \centering
    \includegraphics[width=\textwidth]{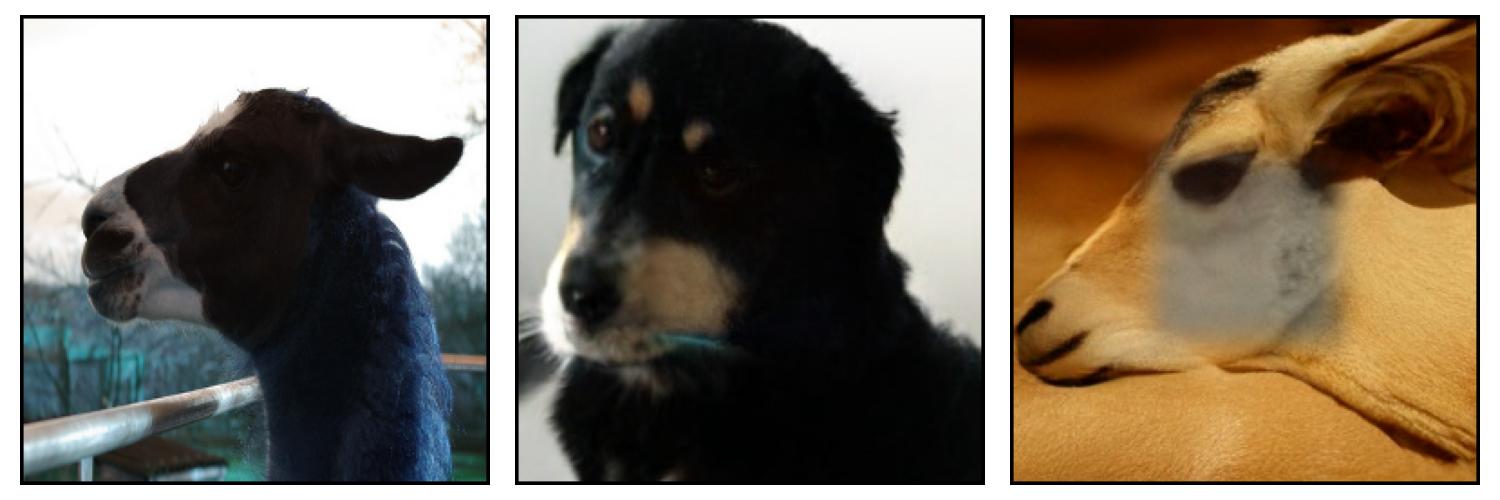}
    \end{subfigure}

    \begin{subfigure}[b]{0.48\textwidth}
    \centering
    \includegraphics[width=\textwidth]{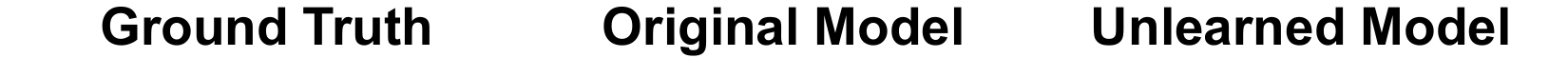}
    \caption{Forget Set}
    \end{subfigure}
    \hfill
    \begin{subfigure}[b]{0.48\textwidth}
    \centering
    \includegraphics[width=\textwidth]{figure/rebuttal/caption_rebuttal.pdf}
    \caption{Retain Set}
    \end{subfigure}

\caption{Visualization of cross validation in Appendix~\ref{app:cross_val}. We first use different models to do the inpainting tasks, i.e., reconstructing the central cropped patches (\textsc{First} reconstructions). Given the \textsc{First} reconstructions as input, we then use the original model (before unlearning) to do the outpainting tasks and get the \textsc{Second} reconstructions, i.e., re-generated the outer patches based on these reconstructed central patches from \textsc{First} reconstructions. We then evaluate the quality of these \textsc{Second} reconstructed images under various metrics. }
\label{fig:cross_Val}
\end{figure}

As shown in Table~\ref{tab:cross_val} and Figure~\ref{fig:cross_Val}, compared to the original model, given the \textit{noise} (i.e., \textsc{First} generated images on the forget set) from the unlearned model as the input, the \textsc{Second} generated images has very low quality in terms of all these three metrics. This means that the \textit{noise} indeed are not correlated with the real forget images.

In contrast, the performance on the retain set is almost the same before and after unlearning. This indicates that our approach indeed preserves the knowledge on the retain set well.

\end{document}